%% file: iclr2023_conference.tex
\documentclass{article} 
\usepackage{iclr2023_preprint,times}

\input{math_commands.tex}

\usepackage{hyperref}
\usepackage{url}
\usepackage{subfigure}
\usepackage{graphicx}
\usepackage{color}

\usepackage{enumitem}

\title{ The Onset of Variance-Limited Behavior for Networks in the Lazy and Rich Regimes
}

\author{Alexander Atanasov\thanks{These authors contributed equally.}\text{  \,}
$^{\S}$$^{\ddagger}$
, \text{ } Blake Bordelon$^{*}$
$^{\dagger}$$^{\ddagger}$
, \text{ } Sabarish Sainathan
$^{\dagger}$$^{\ddagger}$
\& Cengiz Pehlevan
$^{\dagger}$$^{\ddagger}$ 
\\
 \S Department of Physics\\
$^{\dagger}$John A. Paulson School of Engineering and Applied Sciences\\
$^{\ddagger}$Center for Brain Science\\
\, Harvard University\\
\, Cambridge, MA 02138, USA \\
\texttt{\{atanasov,blake\_bordelon,cpehlevan\}@g.harvard.edu} \\
}

\newcommand{\ntk}{NTK$_\infty$\,}
\newcommand{\entk}{eNTK$_0$\,}
\newcommand{\entkf}{eNTK$_f$\,}

\begin{document}

\maketitle

\begin{abstract}
For small training set sizes $P$, the generalization error of wide neural networks is well-approximated by the error of an infinite width neural network (NN), either in the kernel or mean-field/feature-learning regime. However, after a critical sample size $P^*$, we empirically find the finite-width network generalization becomes worse than that of the infinite width network.
In this work, we empirically study the transition from infinite-width behavior to this \textit{variance-limited} regime  as a function of sample size $P$ and network width $N$. We find that finite-size effects can become relevant for very small dataset sizes on the order of $P^* \sim \sqrt{N}$ for polynomial regression with ReLU networks. We discuss the source of these effects using an argument based on the variance of the NN's final neural tangent kernel (NTK). This transition can be pushed to larger $P$ by enhancing feature learning or by ensemble averaging the networks. 
We find that the learning curve for regression with the final NTK is an accurate approximation of the NN learning curve. Using this, we provide  a toy model which also exhibits $P^* \sim \sqrt{N}$ scaling and has $P$-dependent benefits from feature learning.

\end{abstract}

\section{Introduction}



Deep learning systems are achieving state of the art performance on a variety of  tasks \citep{tan2019efficientnet,hoffmann2022training}. Exactly how their generalization is controlled by network architecture, training procedure, and task structure is still not fully understood. One promising direction for deep learning theory in recent years is the infinite-width limit. Under a certain parameterization, infinite-width networks yield a kernel method known as the neural tangent kernel (NTK) \citep{Jacot2018NeuralTK, Lee2019WideNN}. Kernel methods are easier to analyze, allowing for accurate prediction of the generalization performance of wide networks in this regime \citep{bordelon_icml_learning_curve,Canatar2021SpectralBA, bahri2021scaling, simon2021neural}. Infinite-width networks can also operate in the \textit{mean-field} regime if network outputs are rescaled by a small parameter $\alpha$ that enhances feature learning \citep{mei2018mean, Chizat2019OnLT, geiger2020disentangling, Yang2020FeatureLI, bordelon2022self}. 

While infinite-width networks provide useful limiting cases for deep learning theory, real networks have finite width. Analysis at finite width is more difficult, since predictions are dependent on the initialization of parameters. While several works have attempted to analyze feature evolution and kernel statistics at large but finite width \citep{Dyer2020Asymptotics, roberts2021principles}, the implications of finite width on generalization are not entirely clear. Specifically, it is unknown at what value of the training set size $P$ the effects of finite width become relevant, what impact this critical $P$ has on the learning curve, and how it is affected by feature learning.


To identify the effects of finite width and feature learning on the deviation from infinite width learning curves, we empirically study neural networks trained across a wide range of output scales $\alpha$, widths $N$, and training set sizes $P$ on the simple task of polynomial regression with a ReLU neural network. Concretely, our experiments show the following:

\begin{itemize}
    \item Learning curves for polynomial regression transition exhibit significant finite-width effects very early, around $P \sim \sqrt{N}$. Finite-width NNs at large $\alpha$ are always outperformed by their infinite-width counterparts. We show this gap is driven primarily by variance of the predictor over initializations \citep{geiger2020scaling}. Following prior work \citep{bahri2021scaling}, we refer to this as the \textit{variance-limited regime}. We compare three distinct ensembling methods to reduce error in this regime. 
    \item Feature-learning NNs show improved generalization both before and after the transition to the variance limited regime. Feature learning can be enhanced through re-scaling the output of the network by a small scalar $\alpha$ or by training on a more complex task (a higher-degree polynomial). We show that alignment between the final NTK and the target function on test data improves with feature learning and sample size. 
    \item We demonstrate that the learning curve for the NN is well-captured by the learning curve for kernel regression with the \textit{final} empirical NTK, \entkf, as has been observed in other works \citep{vyas2022limitations, geiger2020disentangling, atanasov2021neural, wei2022more}. 
    \item Using this correspondence between the NN and the final NTK, we provide a cursory account of how fluctuations in the final NTK over random initializations are suppressed at large width $N$ and large feature learning strength. In a toy model, we reproduce several scaling phenomena, including the $P \sim \sqrt{N}$ transition and the improvements due to feature learning through an alignment effect.  
\end{itemize}

We validate that these effects qualitatively persist in the realistic setting of wide ResNets \cite{zagoruyko2017wide} trained on CIFAR in appendix \ref{sec:resnet}.

Overall, our results indicate that the onset of finite-width corrections to generalization in neural networks become relevant when the scale of the variance of kernel fluctuations becomes comparable to the bias component of the generalization error in the bias-variance decomposition. The variance contribution to generalization error can be reduced both through ensemble averaging and through feature learning, which we show promotes higher alignment between the final kernel and the task. We construct a model of noisy random features which reproduces the essential aspects of our observations.

\vspace{-10pt}
\subsection{Related Works}

\citet{geiger2020scaling} analyzed the scaling of network generalization with the number of model parameters. Since the NTK fluctuates with variance $O(N^{-1})$ for a width $N$ network \citep{Dyer2020Asymptotics, roberts2021principles}, they find that finite width networks in the lazy regime generically perform worse than their infinite width counterparts. 



The scaling laws of networks over varying $N$ and $P$ were also studied, both empirically and theoretically by \citet{bahri2021scaling}. They consider two types of learning curve scalings. First, they describe \textit{resolution-limited} scaling, where either training set size or width are effectively infinite and the scaling behavior of generalization error with the other quantity is studied. There, the scaling laws can been obtained by the theory in \citet{bordelon_icml_learning_curve}. Second, they analyze \textit{variance-limited} scaling where width or training set size are fixed to a finite value and the other parameter is taken to infinity.  While that work showed for any fixed $P$ that the learning curve converges to the infinite width curve as $O(N^{-1})$, these asymptotics do not predict, for fixed $N$, at which value of $P$ the NN learning curve begins to deviate from the infinite width theory. This is the focus of our work.


The contrast between rich and lazy networks has been empirically studied in several prior works. Depending on the structure of the task, the lazy regime can have either worse \citep{fort2020deep} or better \citep{ortiz2021can, geiger2020disentangling} performance than the feature learning regime. For our setting, where the signal depends on only a small number of relevant input directions, we expect representation learning to be useful, as discussed in \citep{ghorbani_outperform,paccolat2021geometric}. Consequently, we posit and verify that the rich network will outperform the lazy one.

Our toy model is inspired by the literature on random feature models. Analysis of generalization for two layer networks at initialization in the limit of high dimensional data have been carried out using techniques from random matrix theory \citep{mei2022generalization, hu2020universality,adlam2020neural,dhifallah2020precise,adlam2020understanding} and statistical mechanics \citep{gerace2020generalisation, d2020double, d2020triple}. Several of these works have identified that when $N$ is comparable to $P$, the network generalization error has a contribution from variance over initial parameters. Further, they provide a theoretical explanation of the benefit of ensembling predictions of many networks trained with different initial parameters. Recently, \citet{ba2022high} studied regression with the hidden features of a two layer network after taking one step of gradient descent, finding significant improvements to the learning curve due to feature learning. \citet{zavatone2022contrasting} analyzed linear regression for Bayesian deep linear networks with width $N$ comparable to sample size $P$ and demonstrated the advantage of training multiple layers compared to only training the only last layer, finding that feature learning advantage has leading correction of scale $(P/N)^2$ at small $P/N$.

\section{Problem Setup and Notation}

We consider a supervised task with a dataset $\mathcal D = \{\bm x^\mu, y^\mu \}_{\mu=1}^P$ of size $P$. The pairs of data points are drawn from a population distribution $p(\bm x, y)$. Our experiments will focus on training networks to interpolate degree $k$ polynomials on the sphere (full details in Appendix \ref{sec:expt_details}). For this task, the infinite width network learning curves can be found analytically. In particular at large $P$ the generalization error scales as $1/P^2$ \citep{bordelon_icml_learning_curve}. We take a single output feed-forward NN $\tilde f_{\theta}: \mathbb R^D \to \mathbb R$ with hidden width $N$ for each layer. We let $\theta$ denote all trainable parameters of the network. Using NTK parameterization \citep{Jacot2018NeuralTK}, the activations for an input $\bm x$ are given by
\begin{equation}\label{eq:network_defn}
\begin{aligned}
    h^{(\ell)}_i = \frac{\sigma}{\sqrt{N}} \sum_{j=1}^{N} W_{ij}^{(\ell)} \varphi(h_{j}^{(\ell-1)}), \quad \ell = 2, \dots L, \quad h^{(1)}_i = \frac{\sigma}{\sqrt{D}} \sum_{j=1}^D W^{(1)}_{ij} x_j.
\end{aligned}
\end{equation}
Here, the output of the network is $\tilde f_\theta = h_1^{(L)}$. We take $\varphi$ to be a positively homogenous function, in our case a ReLU nonlinearity, but this is not strictly necessary (Appendix \ref{app:alpha_no_weight_rescale}). At initialization we have $W_{ij} \sim \mathcal N(0, 1)$. Consequently, the scale of the output at initialization is $O(\sigma^L)$. As a consequence of the positive homogeneity of the network, the scale of the output is given by $\alpha = \sigma^L$. $\alpha$ controls the feature learning strength of a given NN. Large $\alpha$ corresponds to a \textit{lazy} network while small $\alpha$ yields a \textit{rich} network with feature movement. More details on how $\alpha$ controls feature learning are given in Appendix \ref{sec:why_alpha} and \ref{app:alpha_no_weight_rescale}.

In what follows, we will denote the infinite width NTK limit of this network by NTK$_\infty$. We will denote its finite width linearization by $\mathrm{eNTK}_0(\bm x, \bm x') := \sum_{\theta} \partial_{\theta} f(\bm x) \partial_{\theta} f(\bm x') |_{\theta=\theta_0}$, and we will denote its linearization around its final parameters $\theta_f$ by $\mathrm{eNTK}_f(\bm x, \bm x') := \sum_{\theta} \partial_{\theta} f(\bm x) \partial_{\theta} f(\bm x') |_{\theta=\theta_f}$. Following other authors \citep{Chizat2019OnLT, adlam2020neural}, we will take the output to be $f_{\theta}(\bm x) := \tilde f_{\theta}(\bm x) - \tilde f_{\theta_0} (\bm x)$. Thus, at initialization the function output is $0$. We explain this choice further in Appendix \ref{sec:expt_details}. The parameters are then trained with full-batch gradient descent on a mean squared error loss. 
We denote the final network function starting from initialization $\theta_0$ on a dataset $\mathcal D$ by $f^*_{\theta_0, \mathcal D}(\bm x)$ or $f^*$ for short. The generalization error is calculated using a held-out test set and approximates the population risk $E_g(f) := \left< (f(\bm x) - y)^2 \right>_{\bm x, y \sim p(\bm x, y)}$.


\section{Empirical Results}


In this section, we will study learning curves for ReLU NNs trained on polynomial regression tasks of varying degrees. We take our task to be learning $y = Q_k(\bm \beta \cdot \bm x)$ where $\bm \beta$ is random vector of norm $1/D$ and $Q_k$ is the $k$th gegenbauer polynomial. We will establish the following key observations, which we will set out to theoretically explain in Section 4.

\begin{enumerate}
    \item Both \entk and sufficiently lazy networks perform strictly worse than \ntk, but the ensembled predictors approach the \ntk test error. 
    \item NNs in the feature learning regime of small $\alpha$ can outperform \ntk for an intermediate range of $P$. Over this range, the effect of ensembling is less notable.
    \item Even richly trained finite width NNs eventually perform worse than NTK$_\infty$ at sufficiently large $P$. However, these small $\alpha$ feature-learning networks become variance-limited at larger $P$ than lazy networks. Once in the variance-limited regime, all networks benefit from ensembling over initializations. 
    \item For all networks, the transition to the variance-limited regime begins at a $P^*$ that scales sub-linearly with $N$. For polynomial regression, we find $P^* \sim \sqrt N$.
\end{enumerate}
These findings support our hypothesis that finite width introduces variance in \entk over initializations, which ultimately leads to variance in the learned predictor and higher generalization error. Although we primarily focus on polynomial interpolation tasks in this paper, in Appendix \ref{sec:more_plots} we provide results for wide ResNets trained on CIFAR and observe that rich networks also outperform lazy ones, and that lazy ones benefit more significantly from ensembling. 

\subsection{Finite width effects cause the onset of a variance limited regime}



\begin{figure}[h]
    \centering
    \subfigure[$k=2$ generalization error]{\includegraphics[width=0.4\linewidth]{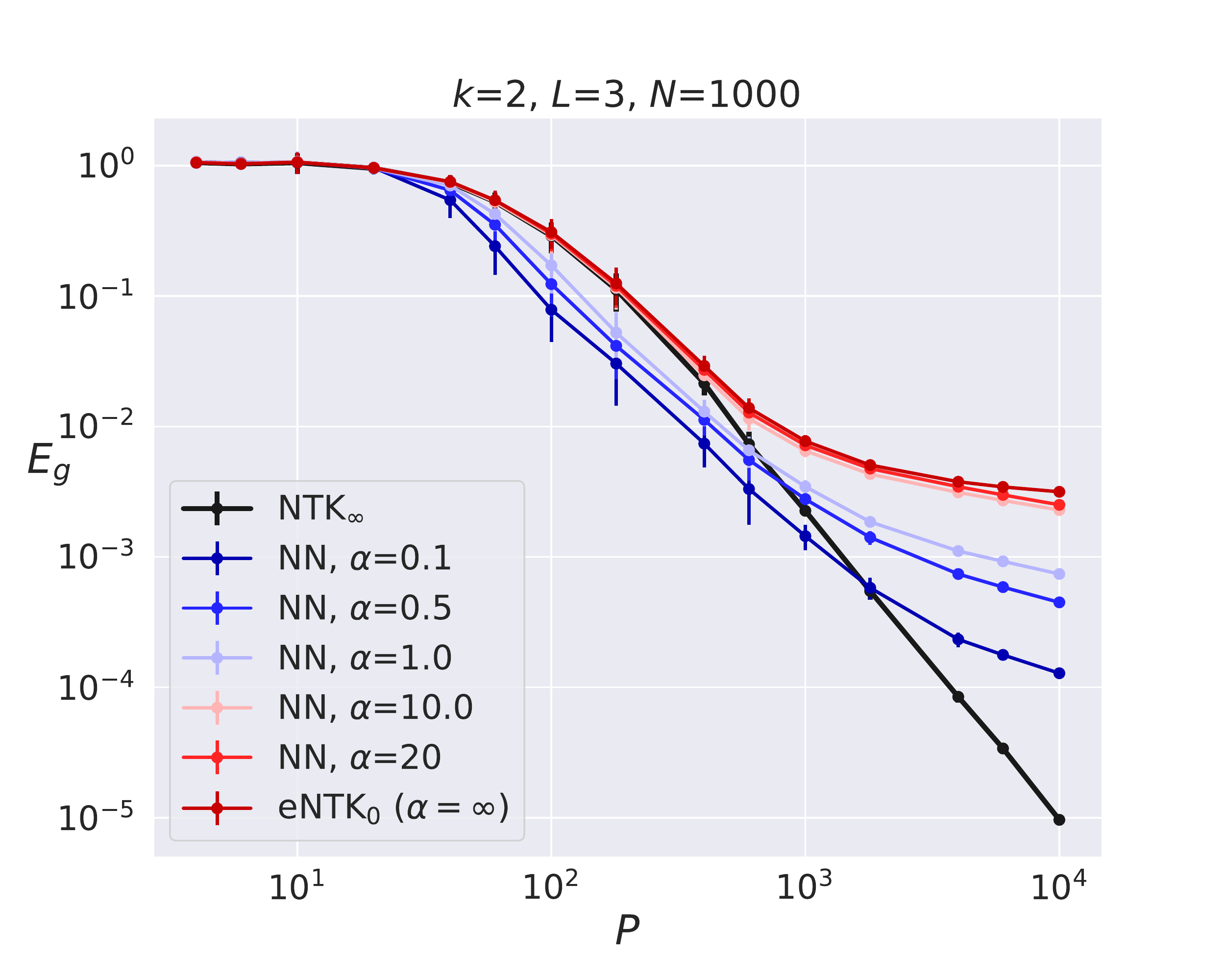}}
    \subfigure[$k=2$ $20$-fold ensemble error]{\includegraphics[width=0.4\linewidth]{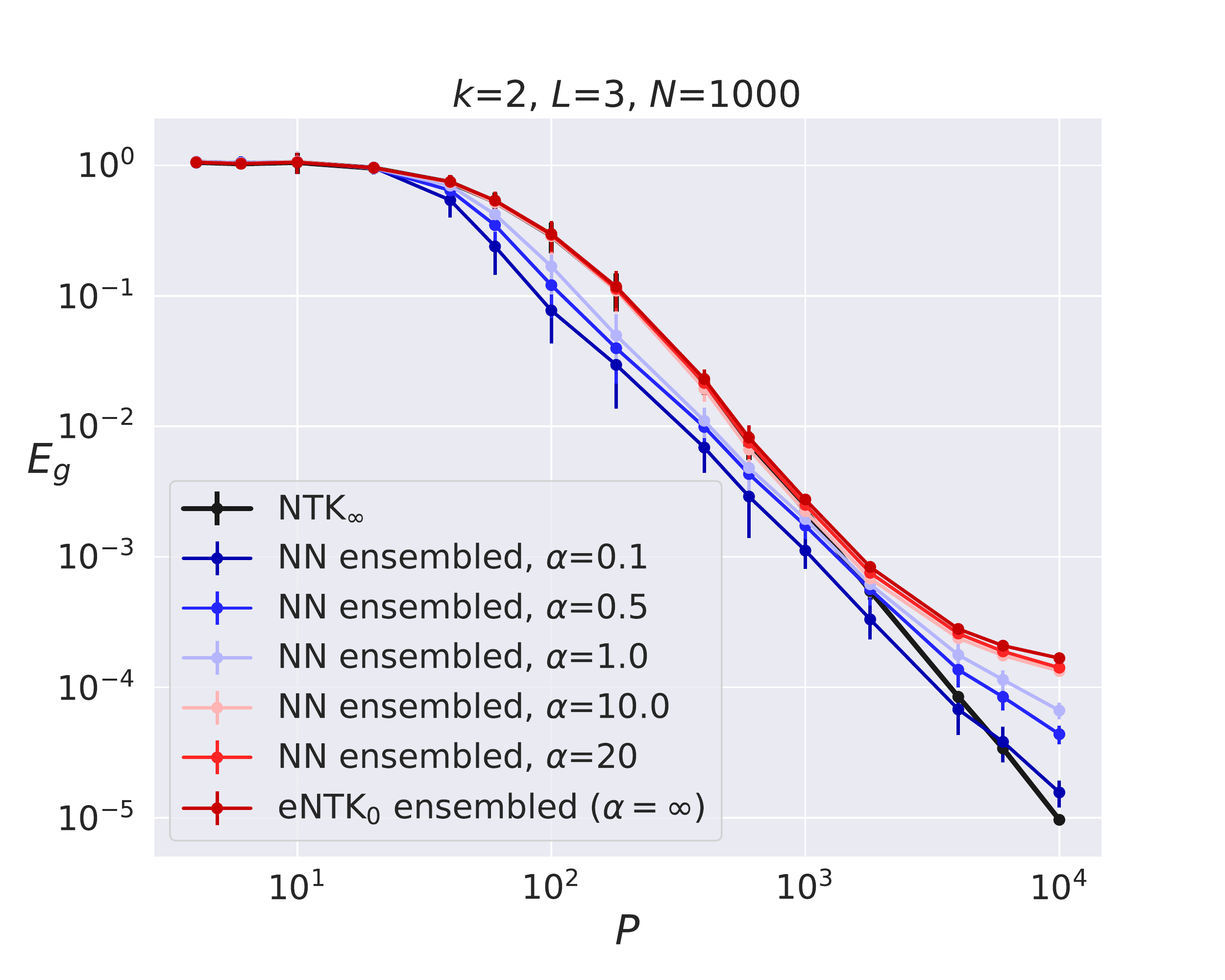}}
    \subfigure[$k=4$ generalization error]{\includegraphics[width=0.4\linewidth]{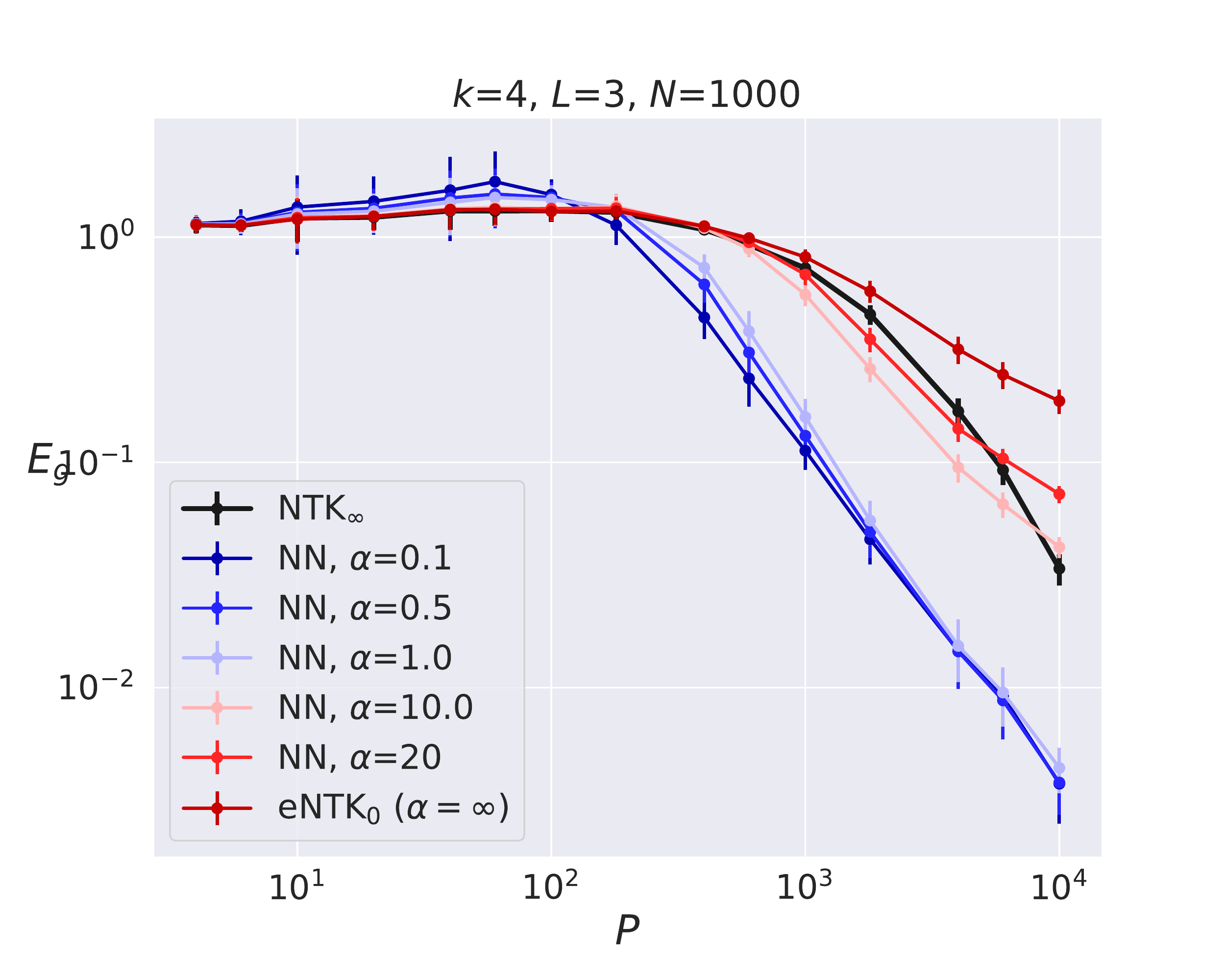}}
    \subfigure[$k=4$ $20$-fold ensemble error]{\includegraphics[width=0.4\linewidth]{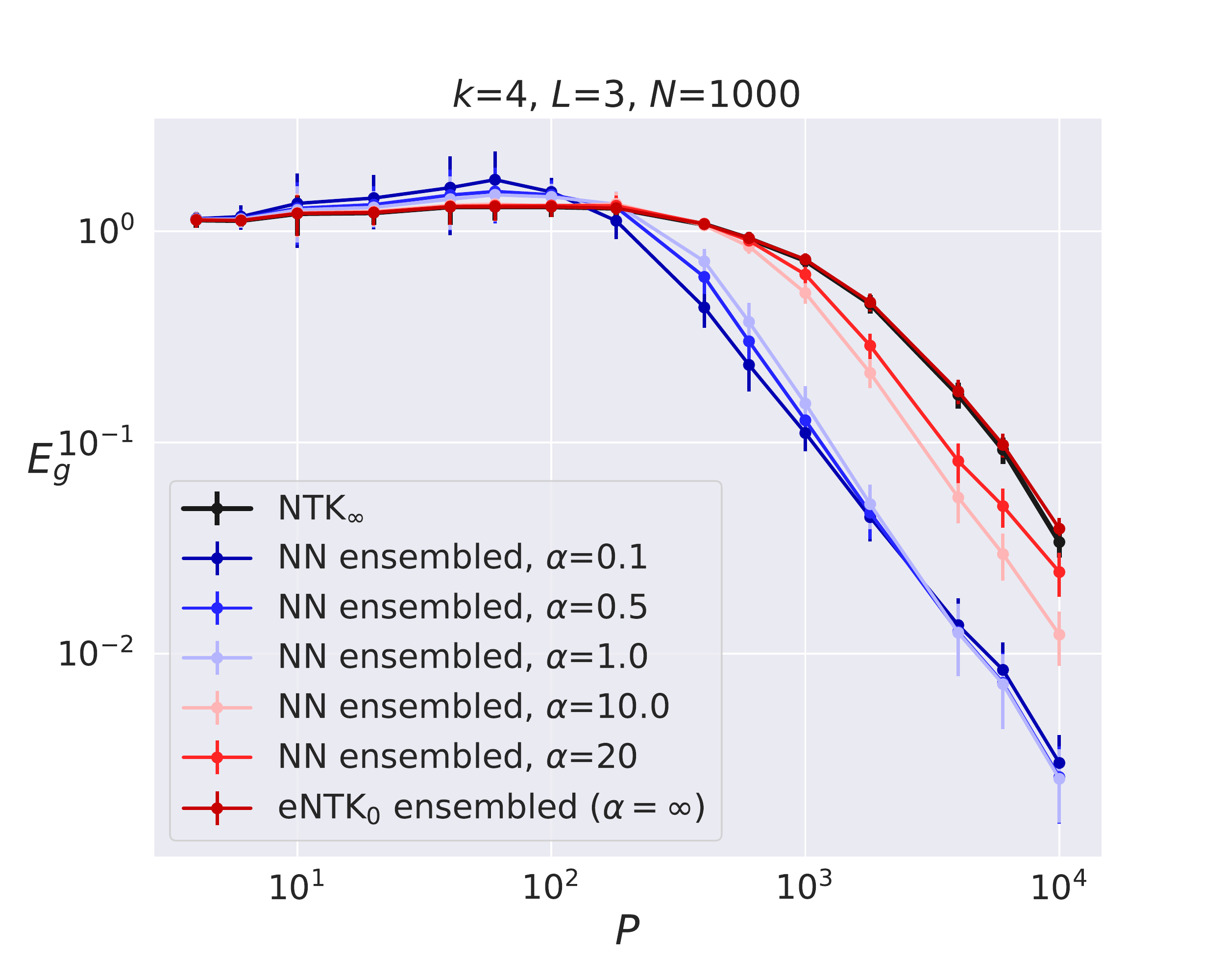}}
    \caption{Generalization errors of depth $L=3$ neural networks across a range of $\alpha$ values compared to \ntk. The regression for \ntk was calculated using the Neural Tangents package \citep{neuraltangents2020}. The exact scaling of \ntk is known to go asymptotically as $P^{-2}$ for this task. a) Lazy networks perform strictly worse than \ntk while rich networks can outperform it for an intermediate range of $P$ before their performance is also limited. b) Ensembling 20 networks substantially improves lazy network and \entk generalization, as well as asymptotic rich network generalization. This indicates that at sufficiently large $P$, these neural networks become limited by variance due to initialization. The error bars in a) and c)  denote the variance due to both both training set and initialization. The error bars in b), d) denote the variance due to the train set. 
    }
    \label{fig:relu_demo_variance} 

\end{figure}

In this section, we first investigate how finite width NN learning curves differ from infinite width NTK regression. In Figure \ref{fig:relu_demo_variance} we show the generalization error $E_g(f^*_{\theta_0, \mathcal D})$ for a depth 3 network with width $N=1000$ trained on a quadratic $k=2$ and quartic $k=4$ polynomial regression task. Additional plots for other degree polynomials are provided in Appendix \ref{sec:more_plots}. We sweep over $P$ to show the effect of more data on generalization, which is the main relationship we are interested in studying. For each training set size we sweep over a grid of 20 random draws of the train set and 20 random network initializations. This for 400 trained networks in total at each choice of $P, k, N, \alpha$. We see that a discrepancy arises at large enough $P$ where the neural networks begin to perform worse than \ntk \!.

We probe the source of the discrepancy between finite width NNs and \ntk by ensemble averaging network predictions $\bar f_{\mathcal D}(\bm x) := \langle f_{\theta_0, \mathcal D}^* (\bm x) \rangle_{\theta_0}$ over $E=20$ initializations $\theta_0$. In Figures \ref{fig:relu_demo_variance}b and \ref{fig:relu_demo_variance}d, we calculate the error of $\bar f_{\mathcal D}(\bm x)$, each trained on the same dataset. We then plot $E_g(\bar f_{\mathcal D})$. This ensembled error approximates the bias in a bias-variance decomposition (Appendix \ref{sec:bias_variance}). Thus, any gap between \ref{fig:relu_demo_variance} (a) and \ref{fig:relu_demo_variance} (b) is driven by variance of $f_{\theta, \mathcal D}$ over $\theta$.

\begin{figure}[t]
    \centering
    \subfigure[$k=3$ generalization error]{\includegraphics[width=0.32\linewidth]{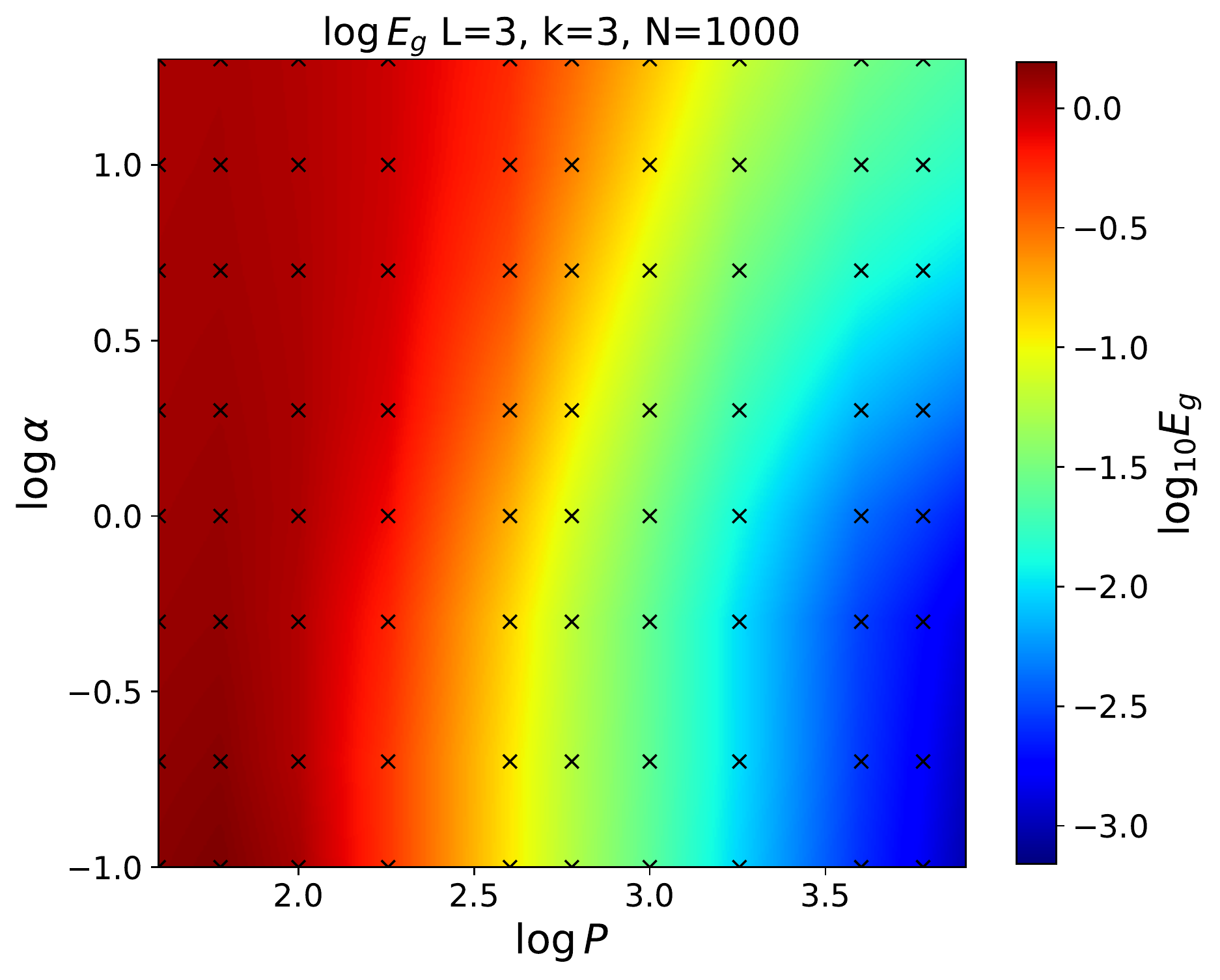}}
    \subfigure[$k=3$ variance fraction]{\includegraphics[width=0.32\linewidth]{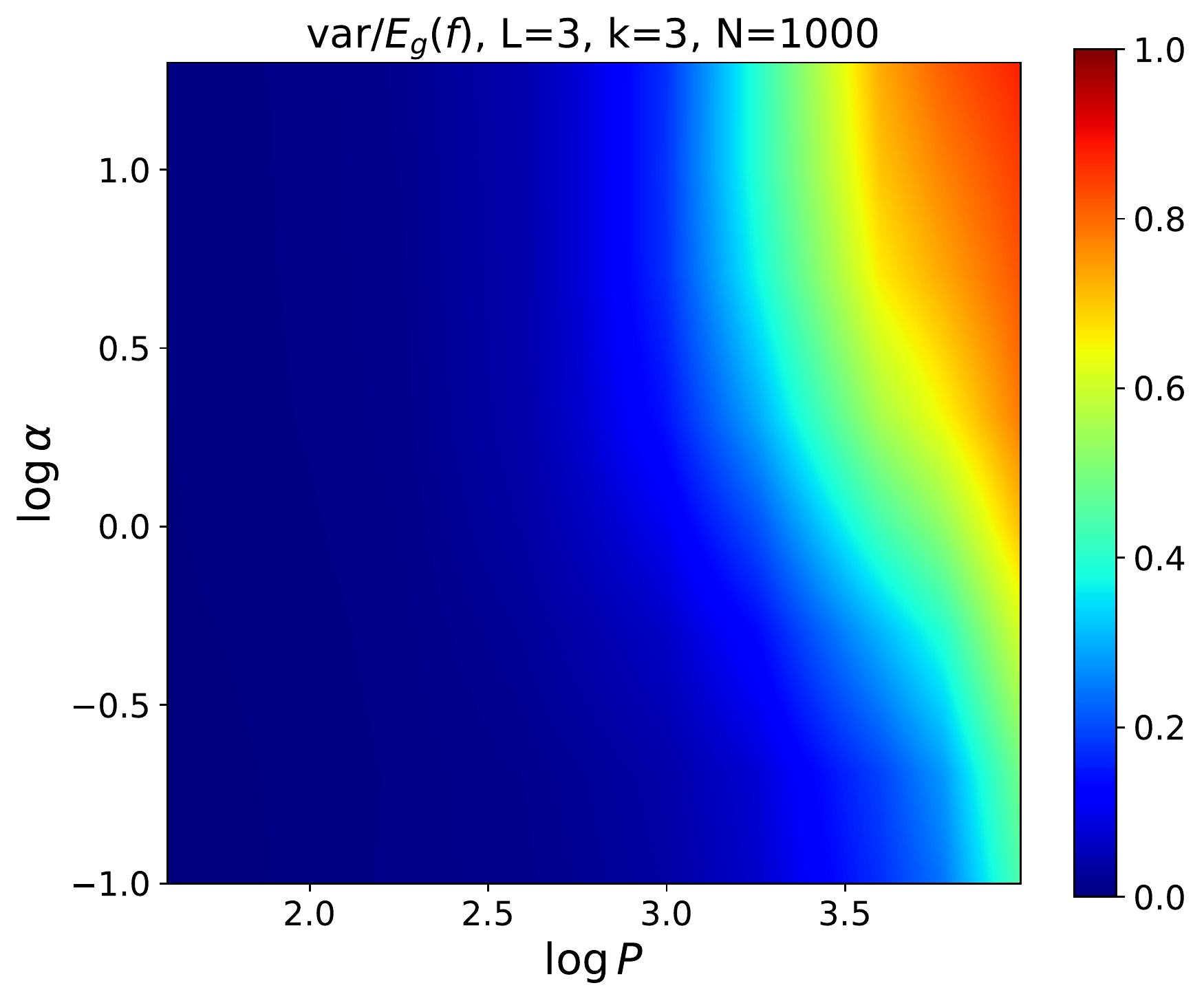}}
    \subfigure[$k=3$ alignment]{\includegraphics[width=0.32\linewidth]{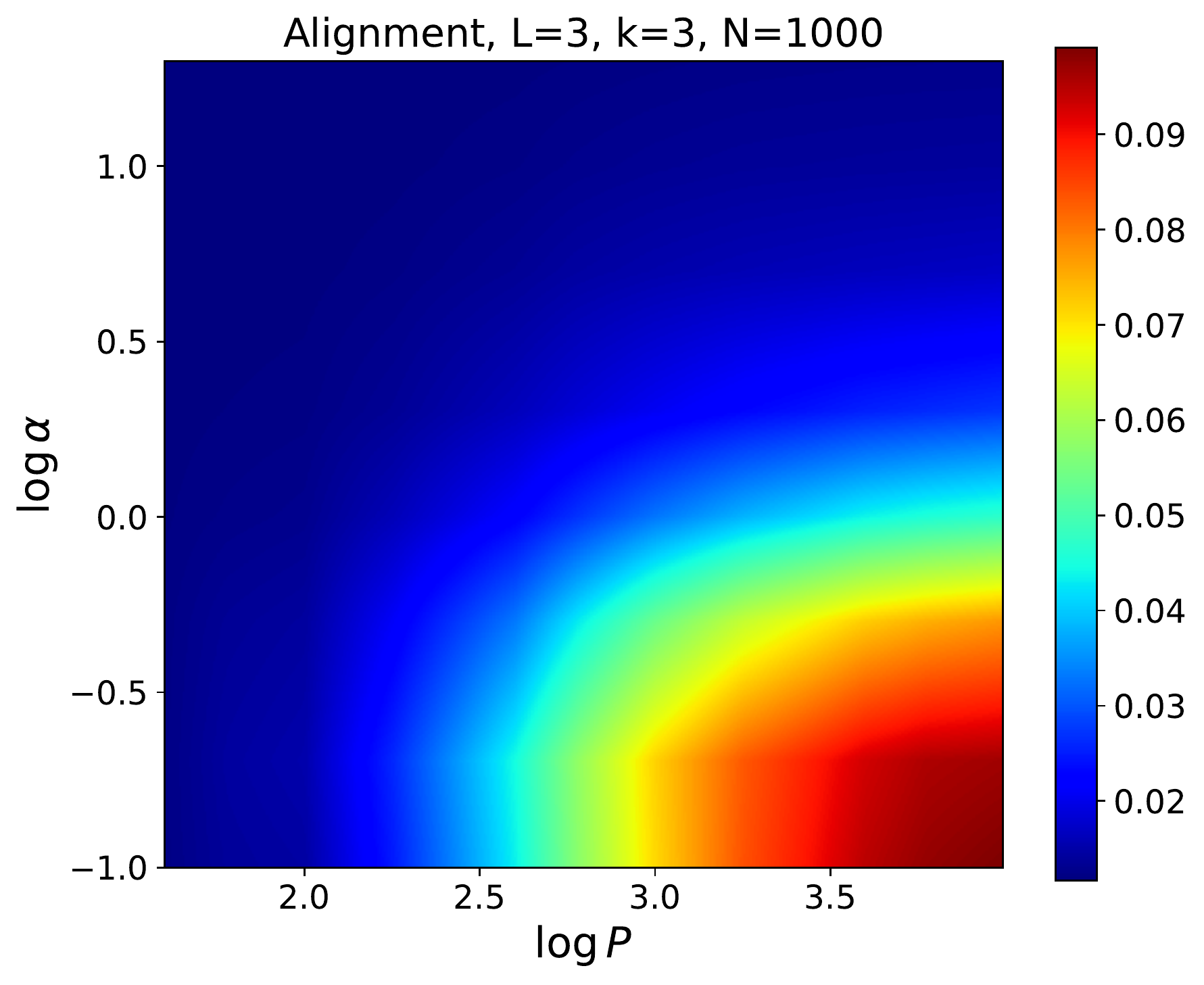}}
    \caption{Phase plots in the $P, \alpha$ plane of a) The log generalization error $\log_{10} E_g(f^\star)$, b) The fraction of generalization error removed by ensembling $1-E_g(\bar f^\star)/E_g(f^\star)$, c) Kernel-task alignment measured by $\frac{\bm{y}^T K_f \bm{y}}{|\bm y|^2 \mathrm{Tr} K_f }$ where $\bm y$ and $K_f$ are evaluated on test data. We have plotted `x' markers in a) to show the points where the NNs were trained. }
    \label{fig:phase_plot}
\end{figure}

We sharpen these observations with phase plots of NN generalization, variance and kernel alignment over $P, \alpha$, as shown in Figure \ref{fig:phase_plot}. In Figure \ref{fig:phase_plot}a, generalization for NNs in the rich regime (small $\alpha$) have lower final $E_g$ than lazy networks. As the dataset grows, the fraction of $E_g$ due to initialization variance (that is, the fraction removed by ensembling) strictly increases (\ref{fig:phase_plot} (b)). We will show why this effect occurs in section \ref{sec:variance}. Figure \ref{fig:phase_plot}b shows that, at any fixed $P$, the variance is lower for small $\alpha$. To measure the impact of feature learning on the \entkf, we plot its alignment with the target function, measured as $\frac{\vy^\top \mK \vy}{|\vy|^2 \text{Tr}\mK}$ for a test set of targets $[\bm y]_\mu$ and kernel $[\bm K]_{\mu \nu} = \mathrm{eNTK}_f(\vx_\mu,\vx_\nu)$. Alignment of the kernel with the target function is known to be related to good generalization \citep{Canatar2021SpectralBA}. In Section \ref{sec:random_kernel}, we revisit these effects in a simple model which relates kernel alignment and variance reduction.

In addition to initialization variance, variance over dataset $\mathcal D$ contributes to the total generalization error. Following \citep{adlam2020understanding}, we discuss a symmetric decomposition of the variance in Appendix \ref{sec:bias_variance}, showing the contribution from dataset variance and the effects of bagging. We find that most of the variance in our experiments is due to initialization.  

We show several other plots of the results of these studies in the appendix. We show the effect of bagging (Figure \ref{fig:fine_grained_variance}), phase plots of different degree target functions (Figures \ref{fig:fine_grained_gen_curves}, \ref{fig:gen_errs_over_k}), phase plots over $N, \alpha$ (Figure \ref{fig:N_alpha_plots}) and a comparison of network predictions against the initial and final kernel regressors (Figures \ref{fig:pred_comp_k=3}, \ref{fig:pred_comp_k=1}).

\subsection{Final NTK Variance leads to Generalization Plateau}\label{sec:variance}

In this section, we show how the variance over initialization can be interpreted as kernel variance in both the rich and lazy regimes. We also show how this implies a plateau for the generalization error.

To begin, we demonstrate empirically that all networks have the same generalization error as kernel regression solutions with their \textit{final} eNTKs. At large $\alpha$, the initial and the final kernel are already close, so this follows from earlier results of \citet{Chizat2019OnLT}. In the rich regime, the properties of the \entkf have been studied in several prior works. Several have empirically demonstrated that the \entkf is a good match to the final network predictor for a trained network \citep{long2021after, vyas2022limitations,wei2022more} while others have given conditions under which such an effect would hold true \citep{atanasov2021neural, bordelon2022self}. We comment on this in appendix \ref{sec:entkf}.
We show in Figure \ref{fig:eNTK_f} how the final network generalization error matches the generalization error of \entkf. As a consequence, we can use  \entkf to study the observed generalization behavior. 


\begin{figure}
    \centering
    \subfigure[$E_g^{NN} = E_g^{NTK_f}$]{\includegraphics[width=0.4\linewidth]{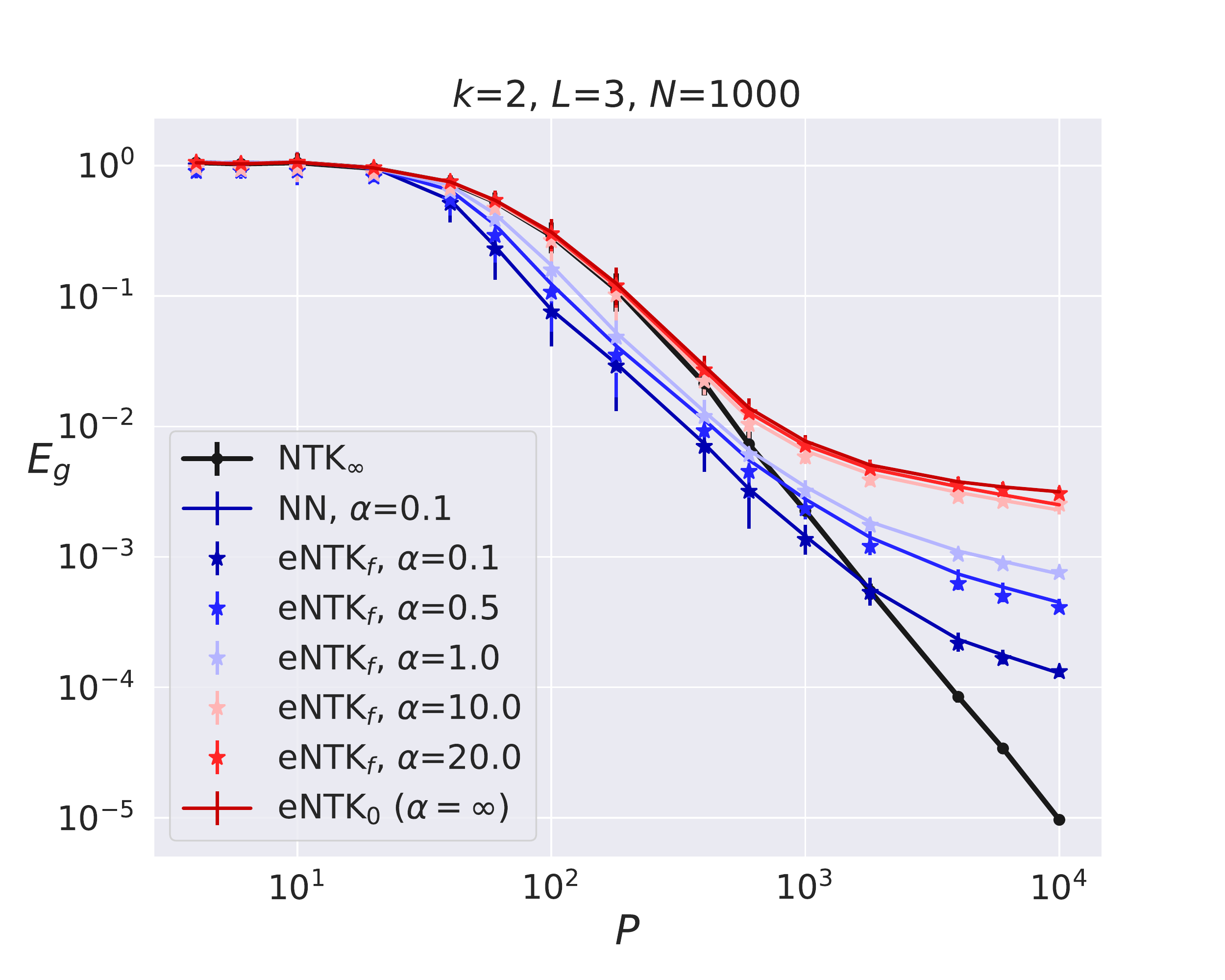}}
    \subfigure[$E_g^{NN} = E_g^{NTK_f}$ across $N, \alpha$]{\includegraphics[width=0.4\linewidth]{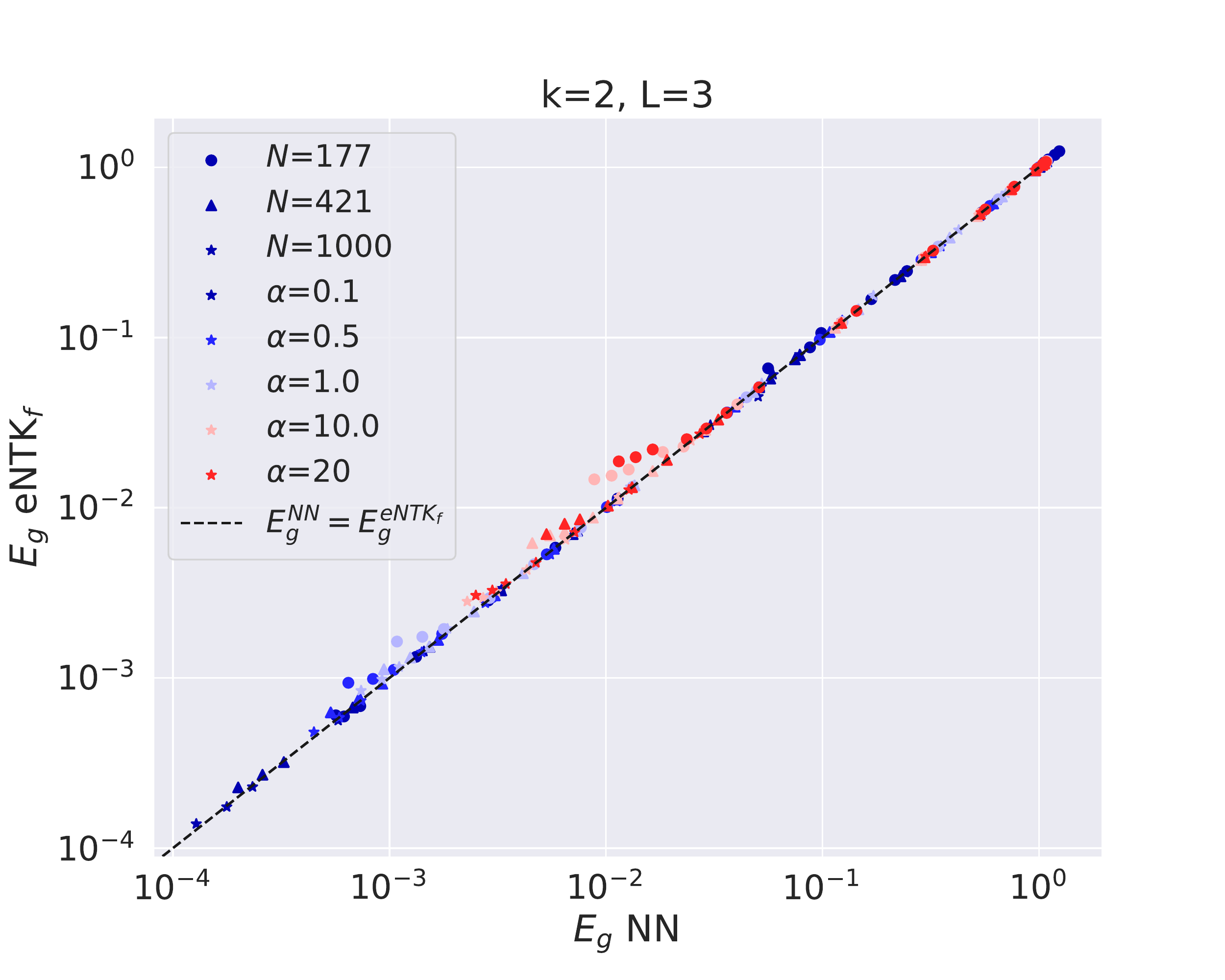}}
    
    \caption{Kernel regression with \entkf reproduces the learning curves of the NN with high fidelity. (a) Learning curves across different laziness settings $\alpha$ in a width $1000$ network. The solid black curve is the infinite width network. Colored curves are the NN generalizations. Stars represent the \entkf \!s, and lie on top of the corresponding NN learning curves.  (b) The agreement of generalizations between NNs and \entkf\!s across different $N$ and $\alpha$. Here the colors denote different $\alpha$ values while the dot, triangle and star markers denote networks of $N=\{177, 421,1000\}$ respectively.}
    \label{fig:eNTK_f}
\end{figure}

Next, we relate the variance of the final predictor $f_{\theta_0, \mathcal D}^*$ to the corresponding infinite width network $f^\infty_{\mathcal D}$.
The finite size fluctuations of the kernel at initialization have been studied in \citep{Dyer2020Asymptotics, hanin2019finite, roberts2021principles}. The variance of the kernel elements has been shown to scale as $1/N$. 
We perform the following bias-variance decomposition: Take $f_{\theta_0, \mathcal D}$ to be the \entk predictor, or a sufficiently lazy network trained to interpolation on a dataset $\mathcal D$. Then,
\begin{equation}\label{eq:init_bias_variance}
\begin{aligned}
\langle (f_{\theta_0, \mathcal D}^*(\bm x) - y)^2 \rangle_{\theta_0, \mathcal D, \bm x, y} 
&= \langle ( f^\infty_{\mathcal D} (\bm x) - y)^2 \rangle_{\mathcal D, \bm x, y} + O(1/N).
\end{aligned}
\end{equation}
We demonstrate this equality using a relationship between the infinite-width network and an infinite ensemble of finite-width networks derived in Appendix \ref{sec:bias_variance}.  There we also show that the $O(1/N)$ term is strictly positive for sufficiently large $N$. Thus, for lazy networks of sufficiently large $N$, finite width effects lead to strictly worse generalization error. The decomposition in Equation \ref{eq:init_bias_variance} continues to hold for rich networks at small $\alpha$ if $f^{\infty}$ is interpreted as the infinite-width mean field limit. In this case one can show that ensembles of rich networks are approximating an infinite width limit in the mean-field regime. See Appendix \ref{sec:bias_variance} for details. 

\subsection{Feature Learning delays variance limited transition}\label{sec:feature_learning}

\begin{figure}
    \centering
    \subfigure[Scaling of $P_{1/2}$ with $\alpha$]{\includegraphics[width=0.35\linewidth]{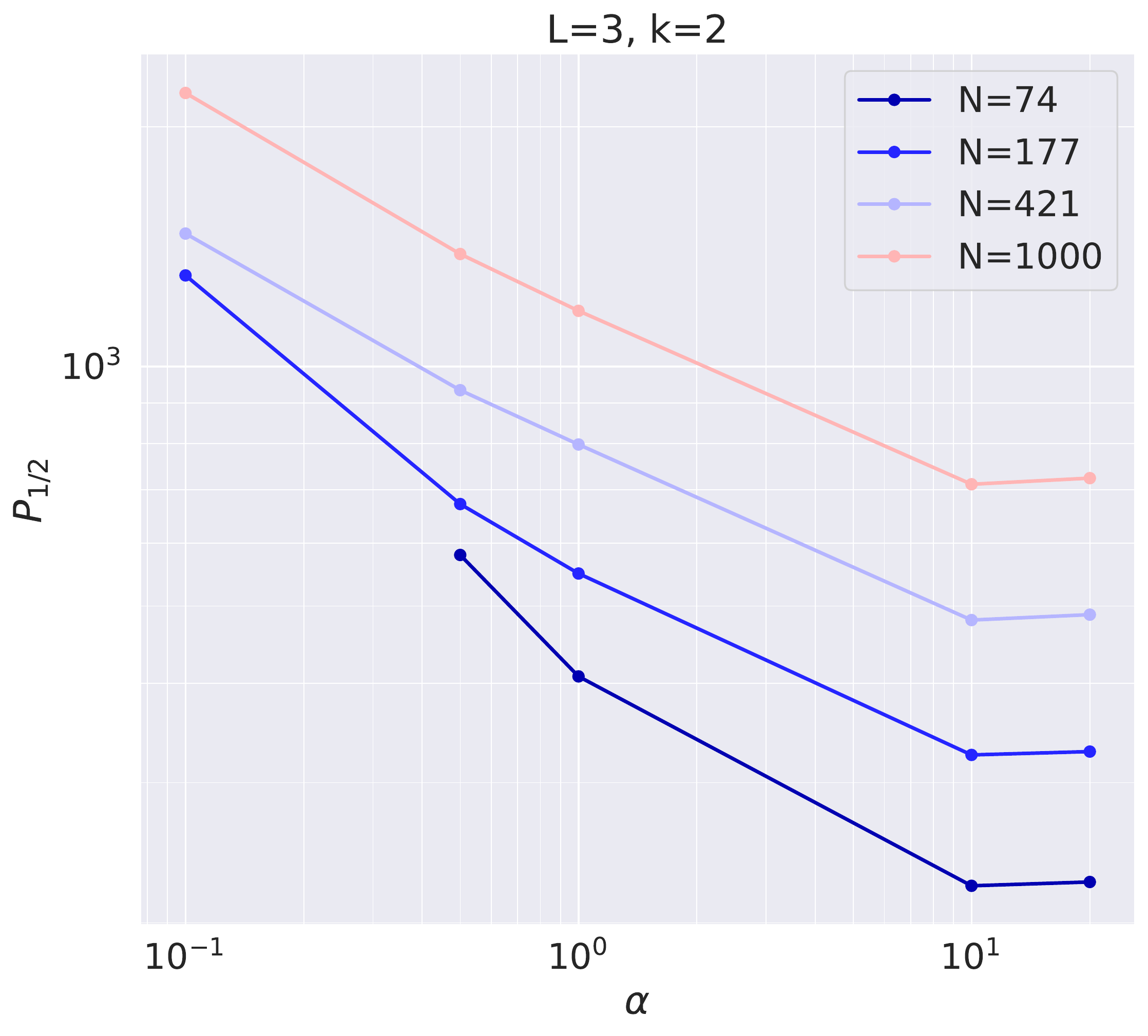}} \hspace{.3in}
    \subfigure[Scaling of $P_{1/2}$ with $N$]{\includegraphics[width=0.35\linewidth]{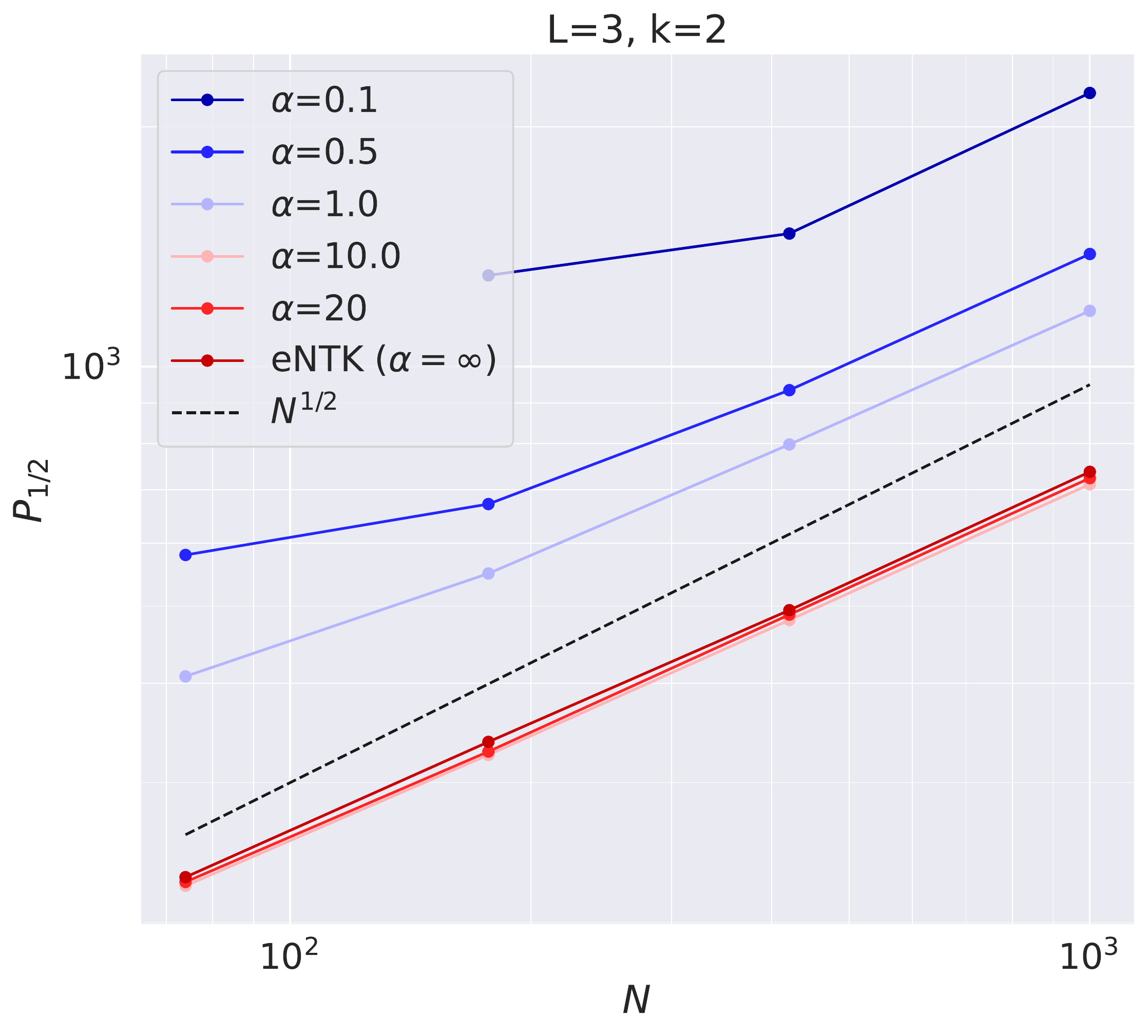}}
    
    \caption{Critical sample size $P_{1/2}$ measures the onset of the variance limited regime as a function of $\alpha$ at fixed $N$. (a) More feature learning (small $\alpha$) delays the transition to the variance limited regime. (b) $P_{1/2}$ as a function of $N$ for fixed $\alpha$ has roughly $P_{1/2} \sim \sqrt{N}$ scaling. }
    \label{fig:variance_onset}
\end{figure}

We now consider how feature learning alters the onset of the variance limited regime, and how this onset scales with $\alpha, N$. We define the onset of the variance limited regime to take place at the value $P^* = P_{1/2}$ where over half of the generalization error is due to variance over initializations. Equivalently we have $E_g(\bar f^*)/E_g(f^*) = 1/2$. By using an interpolation method together with bisection, we solve for $P_{1/2}$ and plot it in Figure \ref{fig:variance_onset}. 

Figure \ref{fig:variance_onset}b shows that $P_{1/2}$ scales as $\sqrt{N}$ for this task. In the next section, we shall show that this scaling is governed by the fact that $P_{1/2}$ is close to the value where the infinite width network generalization curve $E_g^\infty$ is equal to the variance of \entkf. In this case the quantities to compare are $E_g^\infty \approx P^{-2}$ and $\mathrm{Var} \, \mathrm{eNTK}_f \approx N^{-1}$.

We can understand the delay of the variance limited transition, as well as the lower value of the final plateau using a mechanistic picture similar to the effect observed in \citet{atanasov2021neural}. In that setting, under small initialization, the kernel follows a deterministic trajectory, picking up a low rank component in the direction of the train set targets $\bm y \bm y^\top$, and then changing only in scale as the network weights grow to interpolate the dataset. In their case, for initial output scale $\sigma^L$, \entkf is deterministic up to a variance of $O(\sigma)$. In our case, the kernel variance at initialization scales as $\sigma^{2L}/N$. As $\sigma \to 0$ the kernel's trajectory becomes deterministic up to a variance term scaling with $\sigma$ as $O(\sigma)$, which implies that the final predictor also has a variance scaling as $O(\sigma)$.





\section{Signal plus noise correlated feature model}\label{sec:random_kernel}

\begin{figure}
    \centering
    \subfigure[Ensembling Methods]{\includegraphics[width=0.3\linewidth]{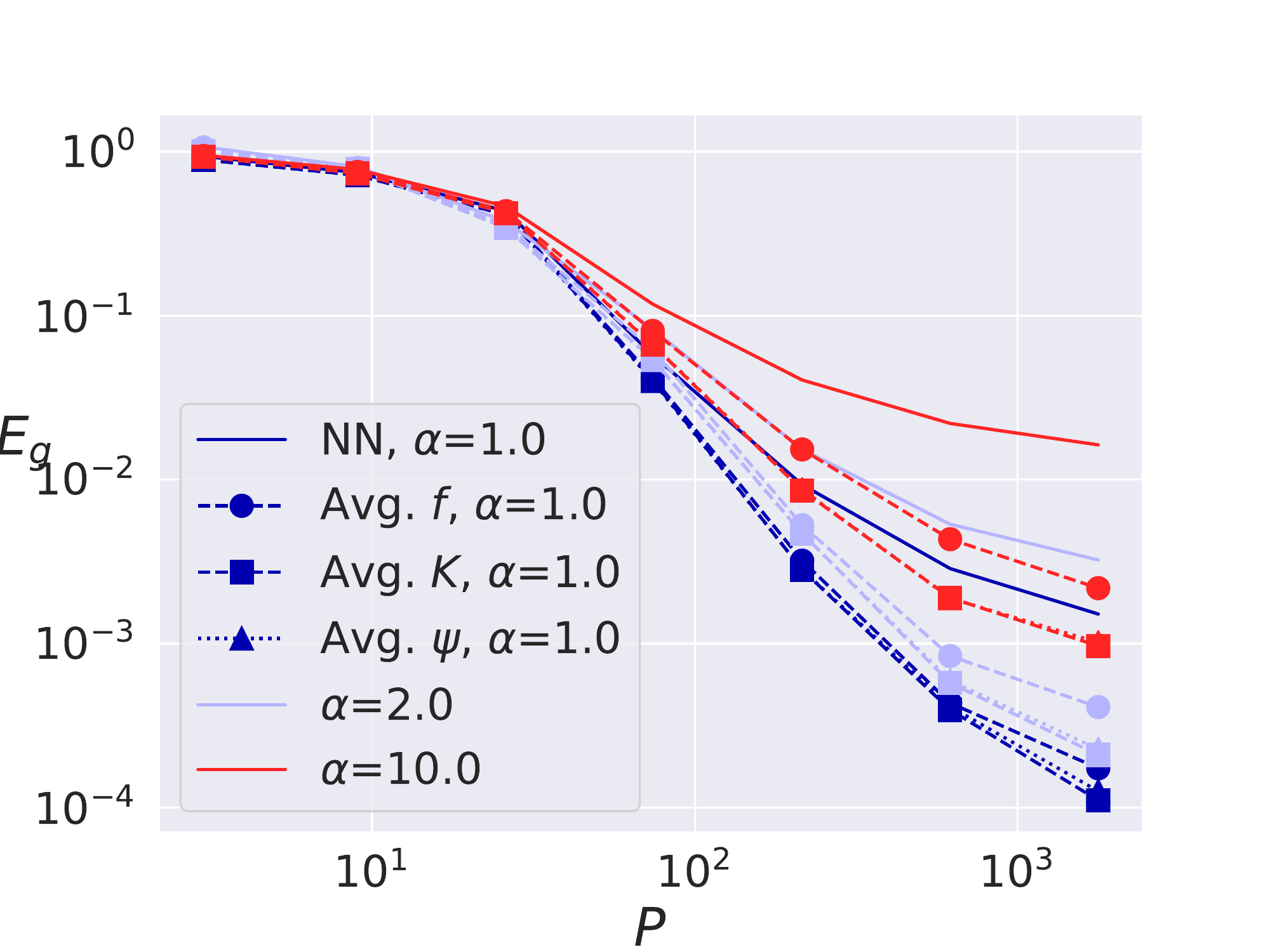}}
    \subfigure[Reduction in $E_g$ for Each Ensembling Technique]{\includegraphics[width=0.68\linewidth]{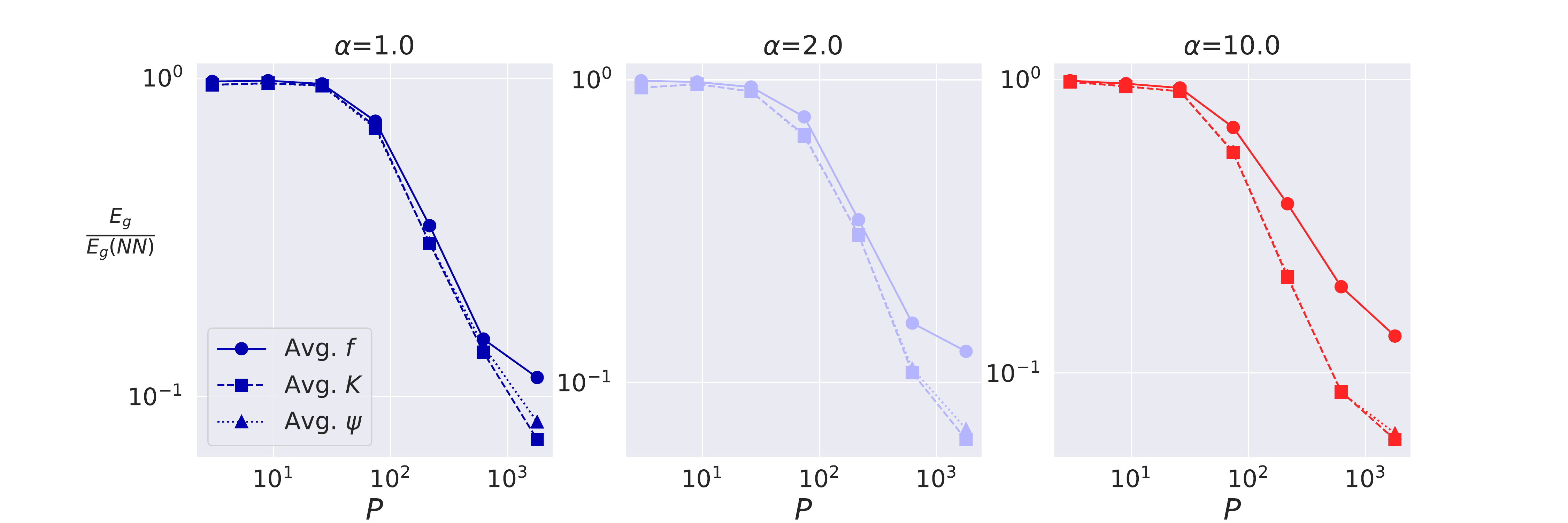}}
    \caption{The random feature model suggests three possible types of ensembling: averaging the output function $f(\vx,\theta)$, averaging \entkf $K(\vx,\vx';\theta)$, and averaging the induced features $\bm\psi(\vx,\theta)$. We analyze these ensembling methods for a $k=1$ task with a width $N=100$ ReLU network. (a) While all ensembling methods improve generalization, averaging either the kernel $\left< K \right>$ or features $\left< \psi \right>$ gives a better improvement to generalization than averaging the output function $\left< f \right>$. Computing final kernels for many richly trained networks and performing regression with this averaged kernel gives the best performance. (b) We plot the relative error of each ensembling method against the single init neural network. The gap between ensembling and the single init NN becomes evident for sufficiently large $P \sim P_{1/2}$. For small $\alpha$, all ensembling methods perform comparably, while for large $\alpha$ ensembling the kernel or features gives much lower $E_g$ than averaging the predictors. }
    \label{fig:ensembling_comparison}
\end{figure}

In Section \ref{sec:variance} we have shown that in both the rich and lazy regimes, the generalization error of the NN is well approximated by the generalization of a kernel regression solution with \entkf\!. This finding motivates an analysis of the generalization of kernel machines which depend on network initialization $\theta_0$. Unlike many analyses of random feature models which specialize to two layer networks and focus on high dimensional Gaussian random data \citep{mei2022generalization, adlam2020neural, gerace2020generalisation, ba2022high}, we propose to analyze regression with the \entkf for more general feature structures. This work builds on the kernel generalization theory for kernels developed with statistical mechanics \citep{bordelon_icml_learning_curve,Canatar2021SpectralBA, simon2021neural, loureiro_lenka_feature_maps}. We will attempt to derive approximate learning curves in terms of the \entkf\!'s signal and noise components, which provide some phenomenological explanations of the onset of the variance limited regime and the benefits of feature learning. Starting with the final NTK $K_{\theta_0}(\vx,\vx')$ which depends on the random initial parameters $\theta_0$, we project its square root $K^{1/2}_{\theta_0}(\vx,\vx')$ (as defined in equation \ref{eq:K_sqrt}) on a fixed basis $\{b_k(x)\}_{k=1}^\infty$ orthonormal with respect to $p(\vx)$. This defines a feature map
\begin{align}
    \psi_k(\vx,\theta_0) = \int d\vx' p(\vx')  K^{1/2}_{\theta_0}(\vx,\vx') b_k(\vx') \ , \ k \in \{1,...,\infty\} .
\end{align}
The kernel can be reconstructed from these features $K_{\theta_0}(\vx,\vx')= \sum_k \psi_k(\vx,\theta_0) \psi_k(\vx',\theta_0)$. The kernel interpolation problem can be solved by performing linear regression with features $\bm\psi(\vx, \theta_0)$. Here, $\vw(\theta_0) = \lim_{\lambda \to 0}\text{argmin}_{\vw} \sum_{\mu=1}^P [\vw \cdot \bm\psi(\vx_\mu,\theta_0) - y_\mu]^2 + \lambda |\vw|^2$. The learned function $f(\vx,\theta_0)= \vw(\theta_0) \cdot \bm\psi(\vx,\theta_0)$ is the minimum norm interpolator for the kernel $K(\vx,\vx';\theta_0)$ and matches the neural network learning curve as seen in Section \ref{sec:variance}. In general, since the rank of $K$ is finite for a finite size network, the $\psi_k(\vx,\bm\theta_0)$ have correlation matrix of finite rank $N_{\mathcal H}$. Since the target function $y$ does not depend on the initialization $\theta_0$, we decompose it in terms of a fixed set of features $\bm\psi_M(\vx) \in \mathbb{R}^{M}$ (for example, the first $M$ basis functions $\{b_k\}_{k=1}^M$). 
In this random feature model, one can interpret the initialization-dependent fluctuations in $K(\vx,\vx';\theta_0)$ as generating fluctuations in the features $\bm\psi(\vx,\theta_0)$ which induce fluctuations in the learned network predictor $f(\vx,\theta_0)$. To illustrate the relative improvements to generalization from denoising these three different objects, in Figure \ref{fig:ensembling_comparison}, we compare averaging the final kernel $K$, averaging the induced features $\psi$, and averaging network predictions $f$ directly. For all $\alpha$, all ensembling methods provide improvements over training a single NN. However, we find that averaging the kernel directly and performing regression with this kernel exhibits the largest reduction in generalization error. Averaging features performs comparably. However, ensemble averaging network predictors does not perform as well as either of these other two methods. The gap between ensembling methods is more significant in the lazy regime (large $\alpha$) and is negligible in the rich regime (small $\alpha$).

\subsection{Toy Models and Approximate Learning Curves}

To gain insight into the role of feature noise, we characterize the test error associated with a Gaussian covariate model in a high dimensional limit $P,M,N_{\mathcal H} \to \infty$ with $\alpha = P/M , \eta = N_{\mathcal H}/M$. 
\begin{align}
    y = \frac{1}{\sqrt M} \bm\psi_M \cdot \vw^*  ,  f = \frac{1}{\sqrt M} \bm\psi \cdot \vw  , \ \bm\psi = \mA(\bm\theta_0) \bm\psi_M + \bm\epsilon , \ \begin{bmatrix}
    \bm\psi_M
    \\
    \bm\epsilon
    \end{bmatrix} \sim \mathcal{N}\left( 0 , \begin{bmatrix} \bm\Sigma_M & 0
    \\
    0 & \bm\Sigma_{\epsilon} \end{bmatrix} \right)
\end{align}
This model was also studied by \citet{loureiro_lenka_feature_maps} and subsumes the classic two layer random feature models of prior works \citep{hu2020universality, adlam2020neural, mei2022generalization}. The expected generalization error for any distribution of $\mA(\bm\theta_0)$ has the form
\begin{align}
     &\mathbb{E}_{\bm\theta_0} E_g(\bm\theta_0) = \mathbb{E}_{\mA} \frac{1}{1-\gamma} \frac{1}{M} \vw^* \bm\Sigma_M^{1/2} \left[\mI - \hat{q} {\bm\Sigma}_s^{1/2} \mA^\top \mG \mA {\bm\Sigma}_s^{1/2}  - \hat{q}  {\bm\Sigma}_s^{1/2}  \mA^\top \mG^2 \mA  {\bm\Sigma}_s^{1/2} \right] \bm\Sigma_M^{1/2} \vw^* \nonumber
     \\
     &\mG = \left( \mI + \hat{q} \mA \bm\Sigma_M \mA^\top + \hat{q} \bm\Sigma_\epsilon \right)^{-1} \ , \ \hat{q} = \frac{\alpha}{\lambda + q} \ , \ q = \text{Tr}\mG[\mA\bm\Sigma_M\mA^\top + \bm\Sigma_{\epsilon}] ,
\end{align}
where $\alpha = P/M$ and $\gamma = \frac{\alpha}{(\lambda+q)^2} \text{Tr} \mG^2 [\mA\bm\Sigma_M\mA^\top + \bm\Sigma_{\epsilon}]^2$. Details of the calculation can be found in Appendix \ref{sec:replica_calculation}. We also provide experiments showing the predictive accuracy of the theory in Figure \ref{fig:theory_qualitative_match}. In general, we do not know the induced distribution of $\mA(\theta_0)$ over disorder $\theta_0$. In Appendix \ref{app:gauss_A_computation}, we compute explicit learning curves for a simple toy model where $\mA(\bm\theta_0)'s$ entries as i.i.d. Gaussian over the random initialization $\bm\theta_0$. A similar random feature model was recently analyzed with diagrammatic techniques by \citet{roberts_random_feature}. In the high dimensional limit $M,P,N_{\mathcal H} \to \infty$ with $P/M= \alpha, N_{\mathcal H}/M =\eta$, our replica calculation demonstrates that test error is self-averaging (the same for every random instance of $\mA$) which we describe in Appendix \ref{app:gauss_A_computation} and Figure \ref{fig:gauss_covariate_verify}.


\subsection{Explaining Feature Learning Benefits and Error Plateaus}

\begin{figure}[t]
    \centering
     \subfigure[$E_g^{NN}$ for different $N$]{\includegraphics[width=0.32\linewidth]{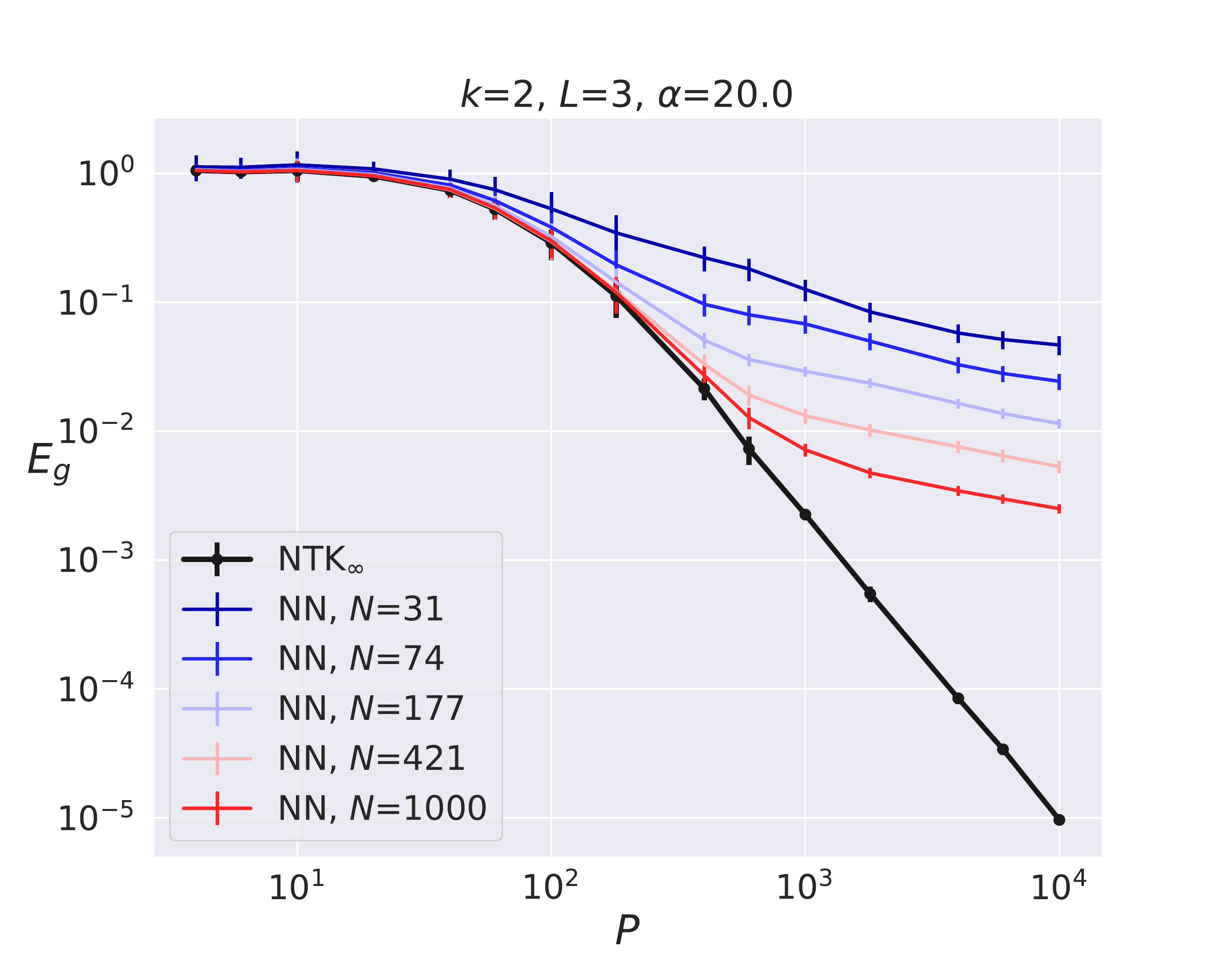}}
    \subfigure[Small $N$ $\approx$ Large $\sigma^2$]{\includegraphics[width=0.32\linewidth]{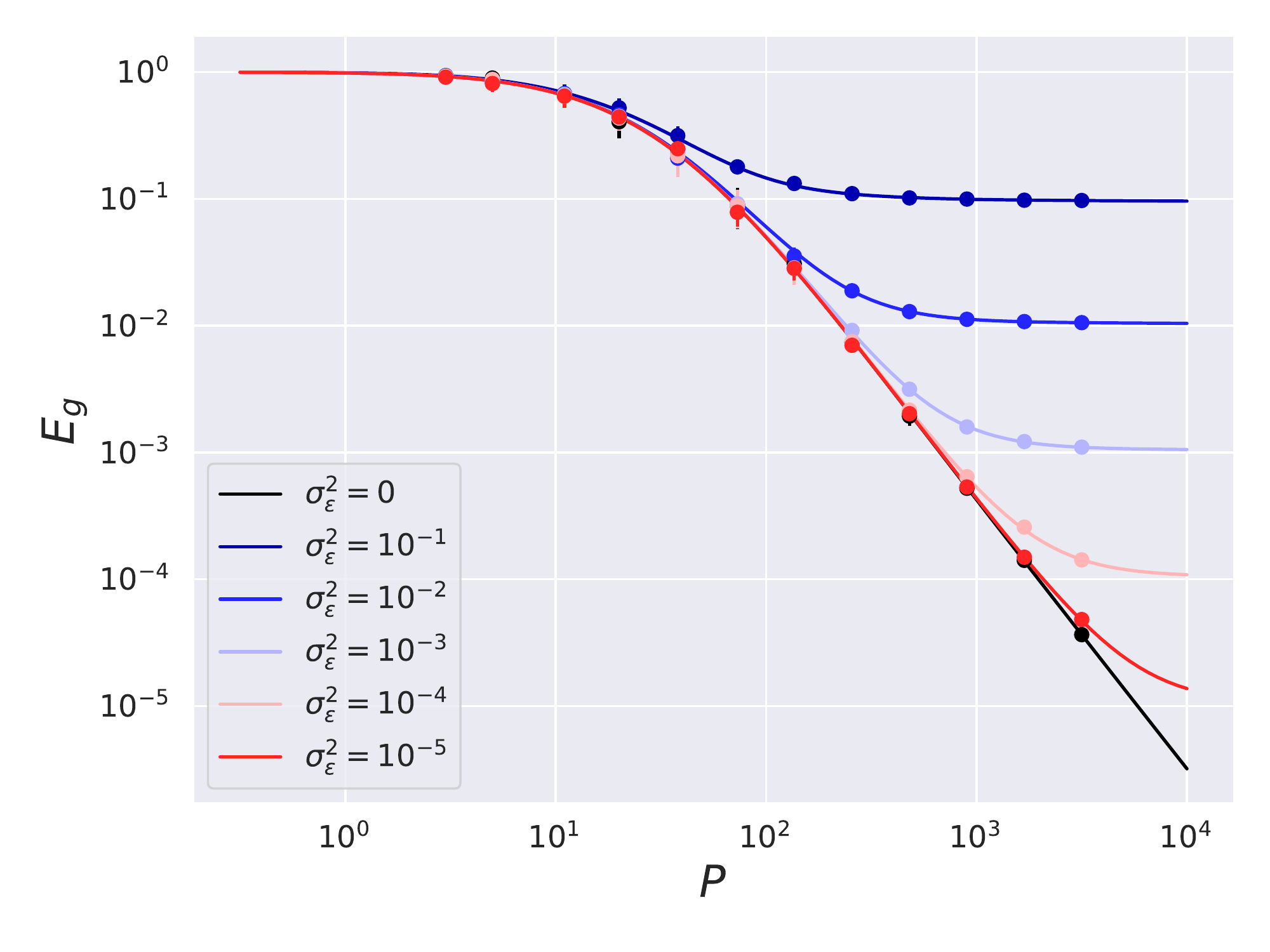}}
    \subfigure[Variance Limited Transition]{\includegraphics[width=0.32\linewidth]{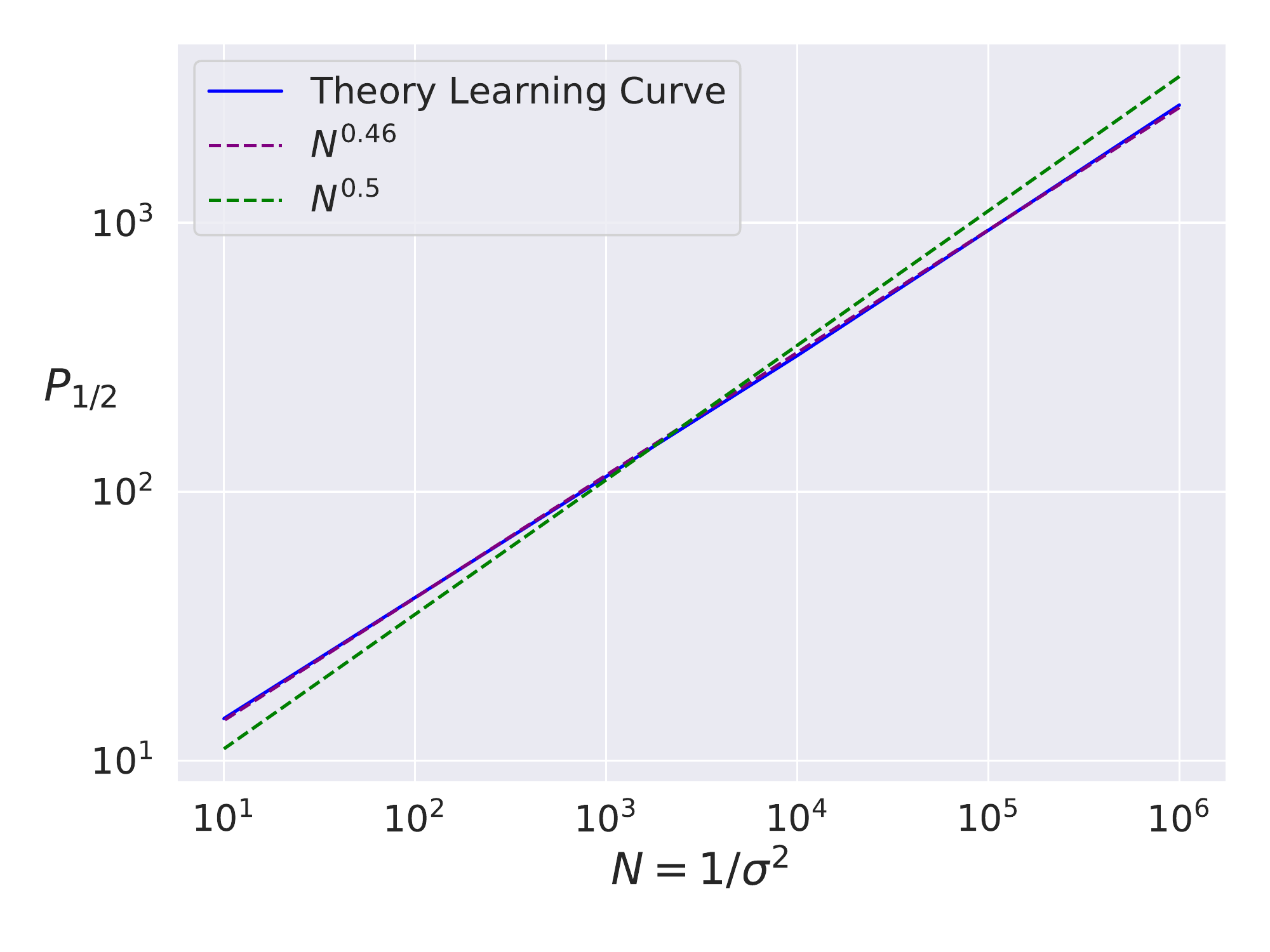}}
    \subfigure[Feature Scalings]{\includegraphics[width=0.32\linewidth]{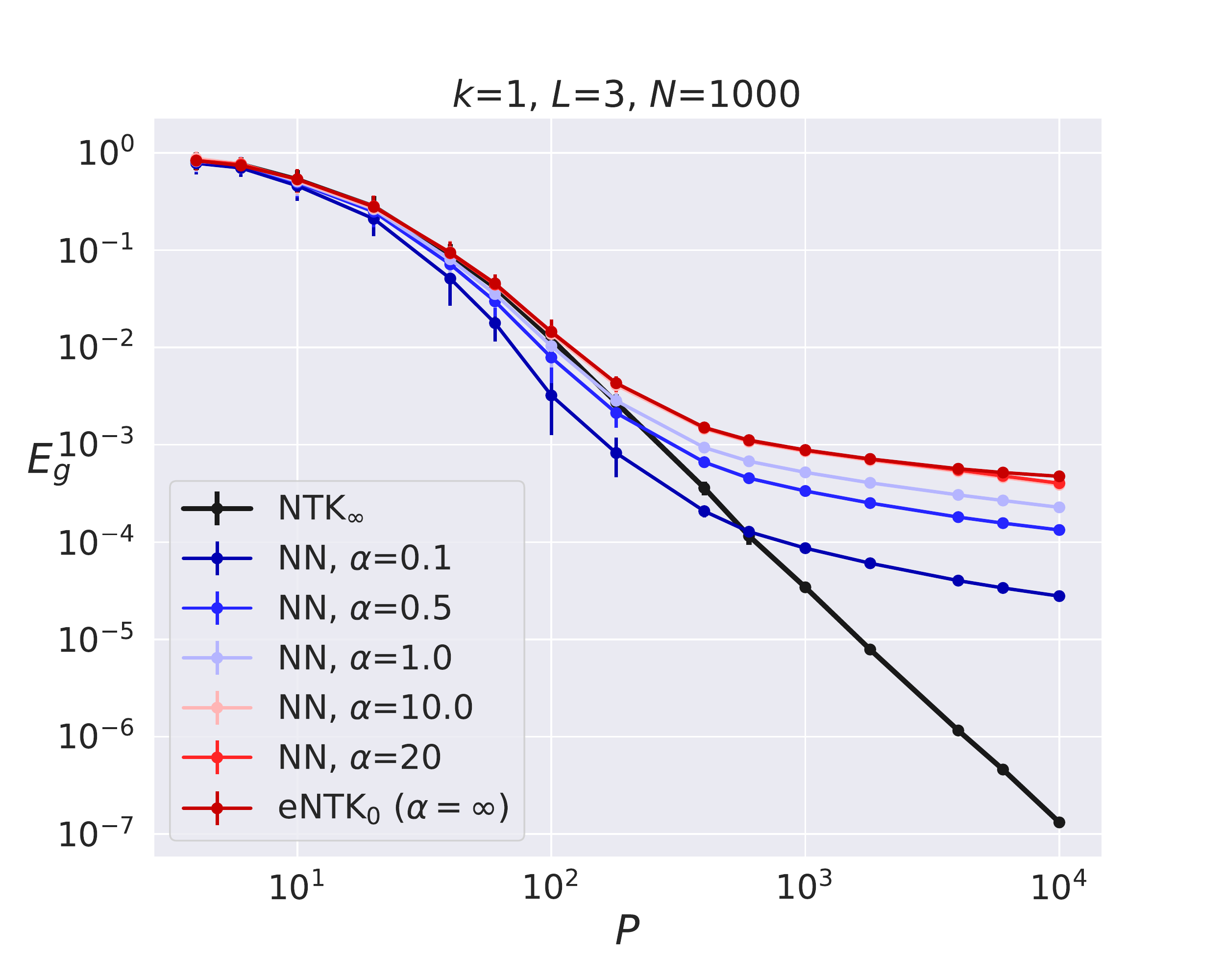}} 
    \subfigure[Richness $\approx$ Amplified $\bm\Sigma_M$]{\includegraphics[width=0.32\linewidth]{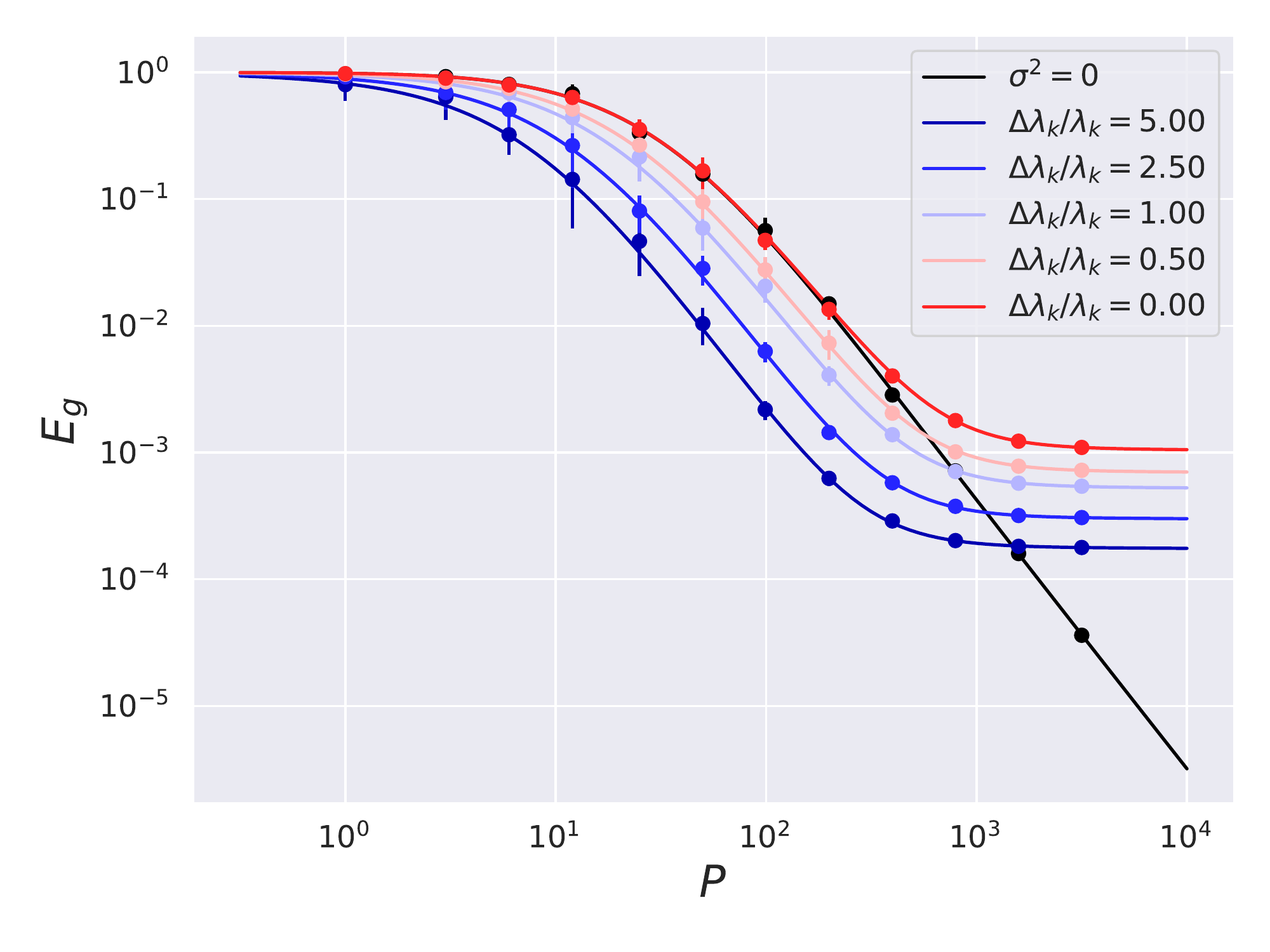}} 
    \subfigure[$P$-dependent Amplification]{\includegraphics[width=0.32\linewidth]{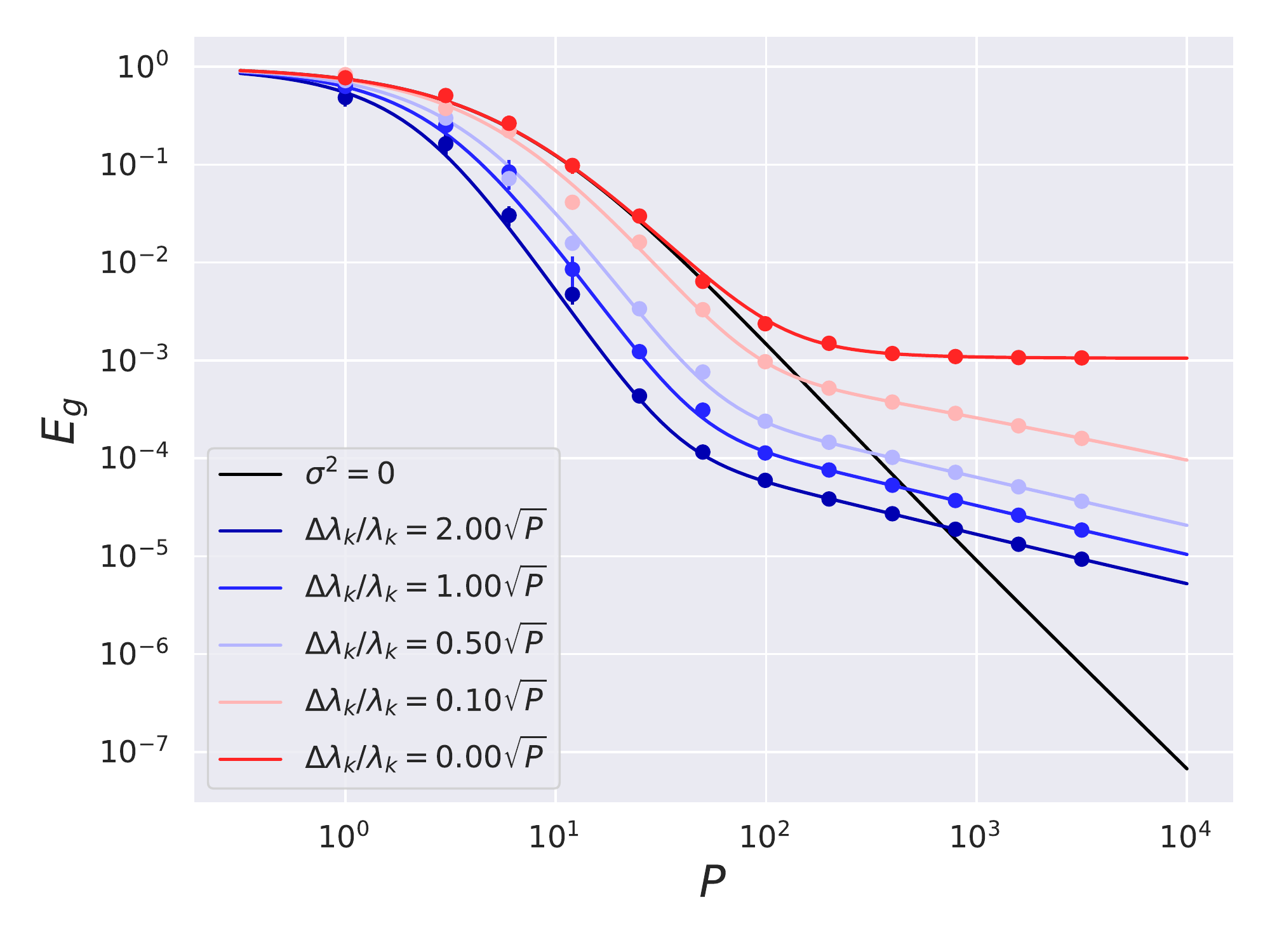}}
    \caption{ A toy model of noisy features reproduces qualitative dependence of learning curves on kernel fluctuations and feature learning. (a) The empirical learning curves for networks of varying width $N$ at large $\alpha$. (b) Noisy kernel regression learning curve with noise $\bm\Sigma_{\epsilon} = \sigma_{\epsilon}^2 \bm\Sigma_M$ and $\mA$ is a projection matrix preserving $20$-k top eigenmodes of $\bm\Sigma_M$, which was computed from the \ntk for a depth $3$ ReLU network. 
    (c) This toy model reproduces the approximate scaling of the transition sample size $P_{1/2} \sim N^{1/2}$ if $\sigma^2_{\epsilon} \sim N^{-1}$. (d) NNs trained with varying richness $\alpha$. Small $\alpha$ improves the early learning curve and asymptotic behavior. (e) Theory curves for a kernel with amplified eigenvalue $\lambda_k \to \lambda_k + \Delta \lambda_k$ for the target eigenfunction. This amplification mimics the effect of enhanced kernel alignment in the low $\alpha$ regime. Large amplification improves generalization performance. (f) $P$-dependent alignment where $\Delta \lambda_k \sim \sqrt{P}$ gives a better qualitative match to (d). }
    \label{fig:theory_qualitative_match}
\end{figure}

Using this theory, we can attempt to explain some of the observed phenomena associated with the onset of the variance limited regime. First, we note that the kernels exhibit fluctuations over initialization with variance $O(1/N)$, either in the lazy or rich regime. In Figure \ref{fig:theory_qualitative_match} (a), we show learning curves for networks of different widths in the lazy regime. Small width networks enter the variance limited regime earlier and have higher error. Similarly, if we alter the scale of the noise $\bm\Sigma_{\epsilon}= \sigma^2_{\epsilon} \bm\Sigma_M$ in our toy model, the corresponding transition time $P_{1/2}$ is smaller and the asymptotic error is higher. In Figure \ref{fig:theory_qualitative_match} (c), we show that our theory also predicts the onset of the variance limited regime at $P_{1/2} \sim \sqrt{N}$ if $\sigma^2_{\epsilon} \sim N^{-1}$. We stress that this scaling is a consequence of the structure of the task. Since the target function is an eigenfunction of the kernel, the infinite width error goes as $1/P^2$ \citep{bordelon_icml_learning_curve}. Since variance scales as $1/N$, bias and variance become comparable at $P \sim \sqrt{N}$. Often, realistic tasks exhibit power law decays where $E_g^{N=\infty} = P^{-\beta}$ with $\beta < 2$ \citep{spigler2020asymptotic,bahri2021scaling}, where we'd expect a transition around $P_{1/2} \sim N^{1/\beta}$. 

Using our model, we can also approximate the role of feature learning as enhancement in the signal correlation along task-relevant eigenfunctions. In Figure \ref{fig:theory_qualitative_match} (d) we plot the learning curves for networks trained with different levels of feature learning, controlled by $\alpha$. We see that feature learning leads to improvements in the learning curve both before and after onset of variance limits. In Figure \ref{fig:theory_qualitative_match} (e)-(f), we plot the theoretical generalization for kernels with enhanced signal eigenvalue for the task eigenfunction $y(\vx) = \phi_k(\vx)$. This enhancement, based on the intuition of kernel alignment, leads to lower bias and lower asymptotic variance. However, this model does not capture the fact that feature learning advantages are small at small $P$ and that the slopes of the learning curves are different at different $\alpha$. Following the observation of \citet{Paccolat_2021} that kernel alignment can occur with scale $\sqrt{P}$, we plot the learning curves for signal enhancements that scale as $\sqrt{P}$. Though this toy model reproduces the onset of the variance limited regime $P_{1/2}$ and the reduction in variance due to feature learning, our current result is not the complete story. A more refined future theory could use the structure of neural architecture to constrain the structure of the $\mA$ distribution. 

\section{Conclusion}

We performed an extensive empirical study for deep ReLU NNs learning a fairly simple polynomial regression problems. For sufficiently large dataset size $P$, all neural networks under-perform the infinite width limit, and we demonstrated that this worse performance is driven by initialization variance. We show that the onset of the variance limited regime can occur early in the learning curve with $P_{1/2} \sim \sqrt{N}$, but this can be delayed by enhancing feature learning. Finally, we studied a simple random-feature model to attempt to explain these effects and qualitatively reproduce the observed behavior, as well as quantitatively reproducing the relevant scaling relationship for $P_{1/2}$. This work takes a step towards understanding scaling laws in regimes where finite-size networks undergo feature learning. This has implications for how the choice of initialization scale,  neural architecture, and number networks in an ensemble can be tuned to achieve optimal performance under a fixed compute and data budget.



\bibliography{iclr2023_conference}
\bibliographystyle{iclr2023_conference}

\newpage

\appendix
\section{Details on Experiments}\label{sec:expt_details}


We generated the dataset $\mathcal D = \{\bm x^\mu, y^\mu \}_{\mu=1}^P$ by sampling $\bm x^\mu$ uniformly on $\mathbb S^{D-1}$, the unit sphere in $\mathbb R^D$. $\tilde y$ was then generated as a Gegenbauer polynomial of degree $k$ of a 1D projection of $\bm x$, $\tilde y = Q_k(\bm \beta \cdot \bm x)$. Because the scale of the output of the neural network relative to the target is a central quantity in this work, it is especially important to make sure the target is appropriately scaled to unit norm. We did this by defining the  target to be $y = \tilde y/ \sqrt{\langle Q_k(\bm \beta \cdot \bm x)^2 \rangle_{\bm x \sim \mathbb S^{D-1}}}$. The denominator can be easily and accurately approximated by Monte Carlo sampling.

We used JAX \citep{bradbury2018jax} for all neural network training. We built multi-layer perceptrons (MLPs) of depth 2 and 3. Most of the results are reported for depth 3 perceptrons, where there is a separation between the width of the network $N$ and the number of parameters $N^2$. Sweeping over more depths and architectures is possible, but because of the extensive dimensionality of the hyperparameter search space, we have not yet experimented with deeper networks.

We considered MLPs with no bias terms. Since the Gegenbauer polynomials are mean zero, we do not need biases to fit the training set and generalize well. We have also verified that adding trainable biases does not change the final results in any substantial way.

As mentioned in the main text, we consider the final output function to be the initial network output minus the output at initialization:
\begin{equation}
    f_{\theta}(\bm x) = \tilde f_{\theta} (\bm x) - \tilde f_{\theta_0}(\bm x).
\end{equation}
Here, only $\theta$ is differentiated through, while $\theta_0$ is held fixed. The rationale for this choice is that without this subtraction, in the lazy limit the trained neural network output can be written as 
\begin{equation}
    \tilde f_{\theta}^*(\bm x) = \tilde f_{\theta_0}(\bm x) + \sum_{\mu \nu} \bm k_\mu(\bm x) [\bm K^{-1}]_{\mu \nu} (y^\nu -  \tilde f_{\theta_0}(\bm x)).
\end{equation}
This is the same as doing \entk regression on the shifted targets $y^\mu - \tilde f_{\theta_0}(\bm x)$. At large initialization the shift $\tilde f_{\theta_0}(\bm x)$ amounts to adding random, initialization-dependent noise to the targets. By instead performing the subtraction, the lazy limit can be interpreted as a kernel regression on the targets themselves, which is preferable.

We trained this network with full batch gradient descent with a learning rate $\eta$ so that
\begin{equation}
\begin{aligned}
    \Delta \theta &= - \eta \nabla_\theta \mathcal L(\mathcal D, \theta),\\
    \mathcal L(\mathcal D, \theta) &:= \frac{1}{P} \sum_{\mu =1}^P |f_\theta(\bm x^\mu) - y^\mu|^2.
\end{aligned}
\end{equation}
Each network was trained to an interpolation threshold of $10^{-6}$. If a network could not reach this threshold in under 30k steps, we checked if the training error was less than $10$ times the generalization error. If this was not satisfied, then that run of the network was discarded.

For each fixed $P, k$, we generated 20 independent datasets. For each fixed $N, \alpha$ we generated 20 independent neural network initializations. This $20 \times 20$ table yields a total of 400 neural networks trained on every combination of initialization and dataset choice.

The infinite width network predictions were calculated using the Neural Tangents package \citep{neuraltangents2020}. The finite width \entk \!s were also calculated using the empirical methods in Neural Tangents. They were trained to interpolation using the \texttt{gradient\_descent\_mse} method. This is substantially faster than training the linearized model using standard full-batch gradient descent, which we have found to take a very long time for most networks. We use the same strategy for the \entkf \!s.

For the experiments in the main text, we have taken the input dimension to be $D = 10$ and sweep over $k=1,2,3,4$. We swept over 15 values $P$ in logspace from size $30$ to size 10k, and over 6 values of $N$ in logspace from size $30$ to size $2150$. We then swept over alpha values $0.1, 0.5, 1.0, 10.0, 20.0$. Depending on $\alpha, N$, we tuned the learning rate $\eta$ of the network small enough to stay close to the gradient flow limit, but allow for the interpolation threshold to be feasibly reached. 

For each of the 1800 settings of $P, N, \alpha, k$ and each of the 400 networks, 400 \entk \!s, 400 \entkf \!s, and  20 \ntk \!s, the generalization error was saved, as well as a vector of $\hat y$ predictions on a test set of 2000 points. In addition, for the neural networks we saved both initial and final parameters. All are saved as lists of numpy arrays in a directory of about 1TB. We plan to make the results of our experiments publicly accessible, alongside the code to generate them. 

\subsection{CIFAR Experiments}\label{sec:CIFAR_expt}

We apply the same methodology of centering the network and allowing $\alpha$ to control the degree of laziness 
by redefining 
\begin{equation}
    f_\theta(\bm x) = \alpha ( \tilde f_{\theta} (\bm x) - \tilde f_{\theta_0}(\bm x) ).
\end{equation}
We consider the task of binary classification for CIFAR-10. In order to allow $P$ to become large we divide the data into two classes: animate and inanimate objects. We choose to subsample eight classes and superclass them into two: (cat, deer, dog, horse) vs (airplane, automobile, ship, truck). Each superclass consists of 20,000 training examples and 4,000 test examples retrieved from the CIFAR-10 dataset.

On subsets of this dataset, we train wide residual networks (ResNets) \cite{zagoruyko2017wide} of width $64$ and block size $1$ with the NTK parameterization \cite{Jacot2018NeuralTK} on this task using mini-batch gradient descent with batch size of 256 and MSE loss. Step sizes are governed by the Adam optimizer \cite{kingma2014adam} with initial learning rate $\eta_0 = 10^{-3}.$ Every network is trained for 24,000 steps, such that under nearly all settings of $\alpha$ and dataset size the network has attained infinitesimal train loss.

We sweep $\alpha$ from $10^{-3}$ to $10^{0}$ and $P$ from $2^9$ to $2^{15}$. For each value of $P$, we randomly sample five training datasets of size $P$ and compute ensembles of size 20. For each network in an ensemble the initialization and the order of the training data is randomly chosen independently of those for the other networks.

\section{Fine-grained bias-variance decomposition}\label{sec:bias_variance}

\subsection{Fine Grained Decomposition of Generalization Error}
Let $\mathcal D$ be a dataset of $(\bm x^\mu, y^\mu)_{\mu=1}^P \sim p(\bm x, y)$ viewed as a random variable. Let $\theta_0$ represent the initial parameters of a neural network, viewed as a random variable.  In the case of no label noise, as in section 2.2.1 of \cite{adlam2020understanding}, we derive the symmetric decomposition of the generalization error in terms of the variance due to initialization and the variance due to the dataset. We have
\begin{equation}
\begin{aligned}
    E_g(f^*_{\theta_0, \mathcal D}) &= \langle  (f^*_{\theta_0, \mathcal D}(\bm x) - y)^2 \rangle_{\bm x, y}  =  \langle  ( \langle f^*_{\theta_0, \mathcal D}(\bm x) \rangle_{\theta_0, \mathcal D} - y)^2 \rangle_{\bm x, y} + \mathbb E_{\bm x} \mathrm{Var}_{\theta_0, \mathcal D} f^*_{\theta_0, \mathcal D}(y) \\
    &= \mathrm{Bias}^2 + V_{\mathcal D} + V_{\theta_0} + V_{\mathcal D, \theta_0}.
\end{aligned}
\end{equation}
Here we have defined
\begin{align}
    \mathrm{Bias}^2 &= \langle  ( \langle f^*_{\theta_0, \mathcal D}(\bm x) \rangle_{\theta_0, \mathcal D} - y)^2 \rangle_{\bm x, y},\\
    V_{\mathcal D} &= \mathbb E_{\bm x}\, \mathrm{Var}_{\mathcal D} \, \mathbb E_{\theta_0} [f^*_{\theta_0, \mathcal D}(\bm x) | \mathcal D] = \mathbb E_{\bm x}\, \mathrm{Var}_{\mathcal D} \, \bar f_{\mathcal D}^*(\bm x),\\
    V_{\theta_0} &=  \mathbb E_{\bm x}\, \mathrm{Var}_{\theta_0} \, \mathbb E_{\mathcal D} [f^*_{\theta_0, \mathcal D}(\bm x) | \theta_0],\\
    V_{\mathcal D, \theta_0} &=  \mathbb E_{\bm x} \mathrm{Var}_{\theta_0, \mathcal D} f^*_{\theta_0, \mathcal D}(y)  - V_{\theta_0} - V_{\mathcal D}.
\end{align}
$V_{\mathcal D}$ and $V_{\theta_0}$ give the components of the variance explained by variance in $\mathcal D, \theta_0$ respectively. $V_{\mathcal D, \theta_0}$ is the remaining part of the variance not explained by either of these two sources. As in the main text, $\bar f^*_{\mathcal D} (\bm x)$ is the ensemble average of the trained predictors over initializations. $E_{\mathcal D} [f^*_{\theta_0, \mathcal D}(\bm x) | \theta_0]$ is commonly referred to as the bagged predictor. In the next subsection we study these terms empirically.

\subsection{Empirical Study of Dataset Variance}

\begin{figure}[h]
    \centering
    \subfigure[Initialization variance $k=2$]{\includegraphics[width=0.32\linewidth]{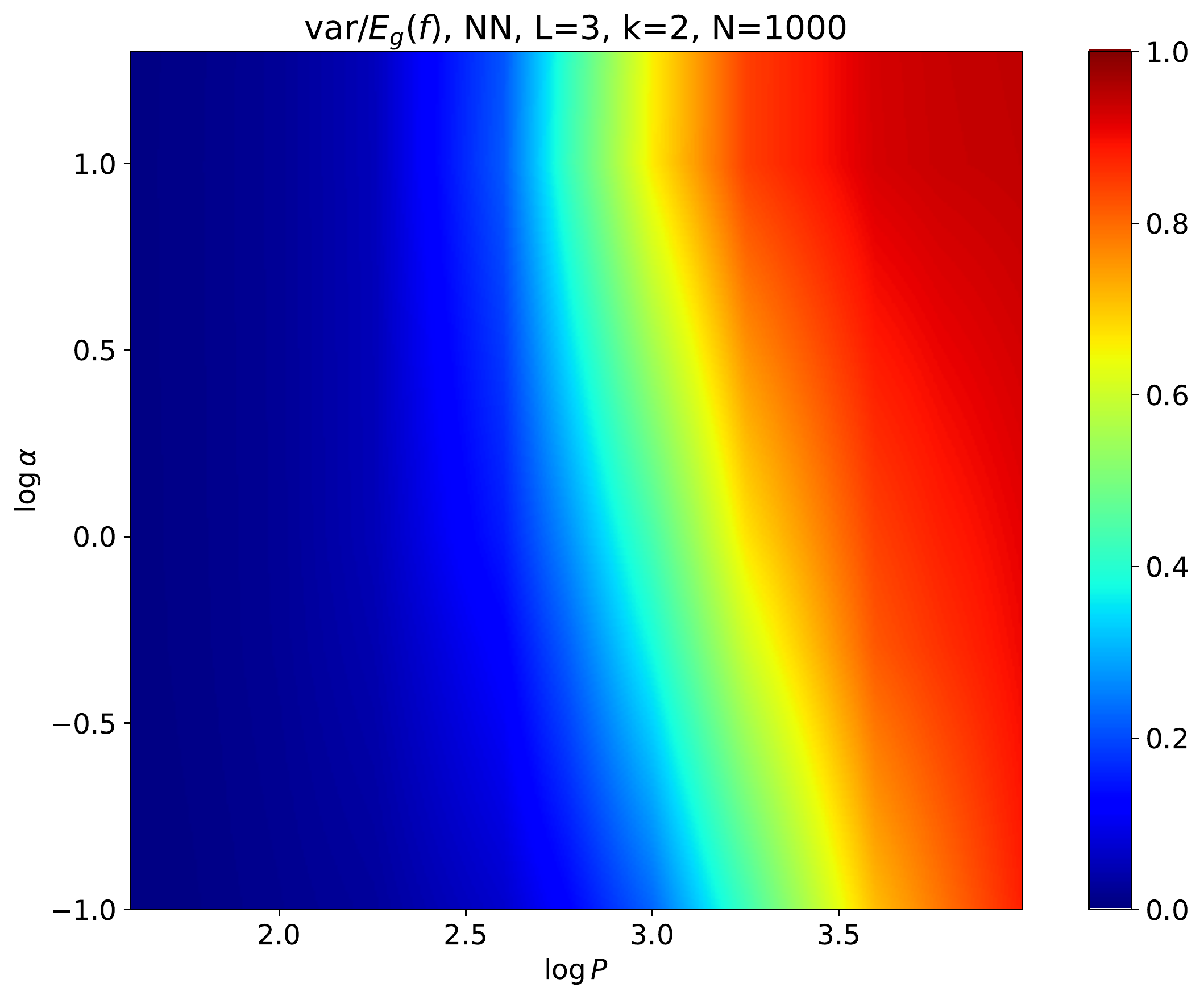}}
    \subfigure[Initialization variance $k=3$]{\includegraphics[width=0.32\linewidth]{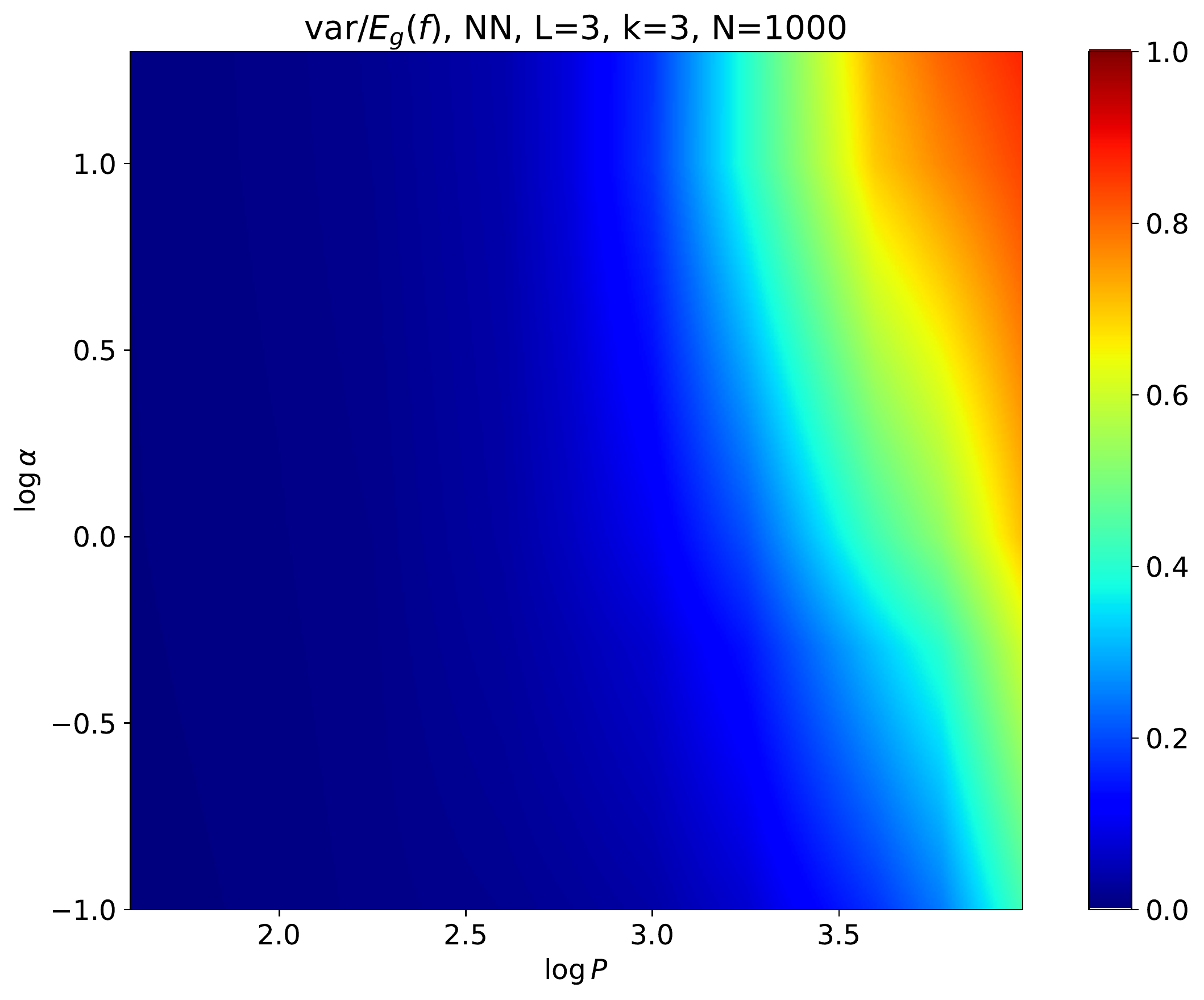}}
    \subfigure[Initialization variance $k=4$]{\includegraphics[width=0.32\linewidth]{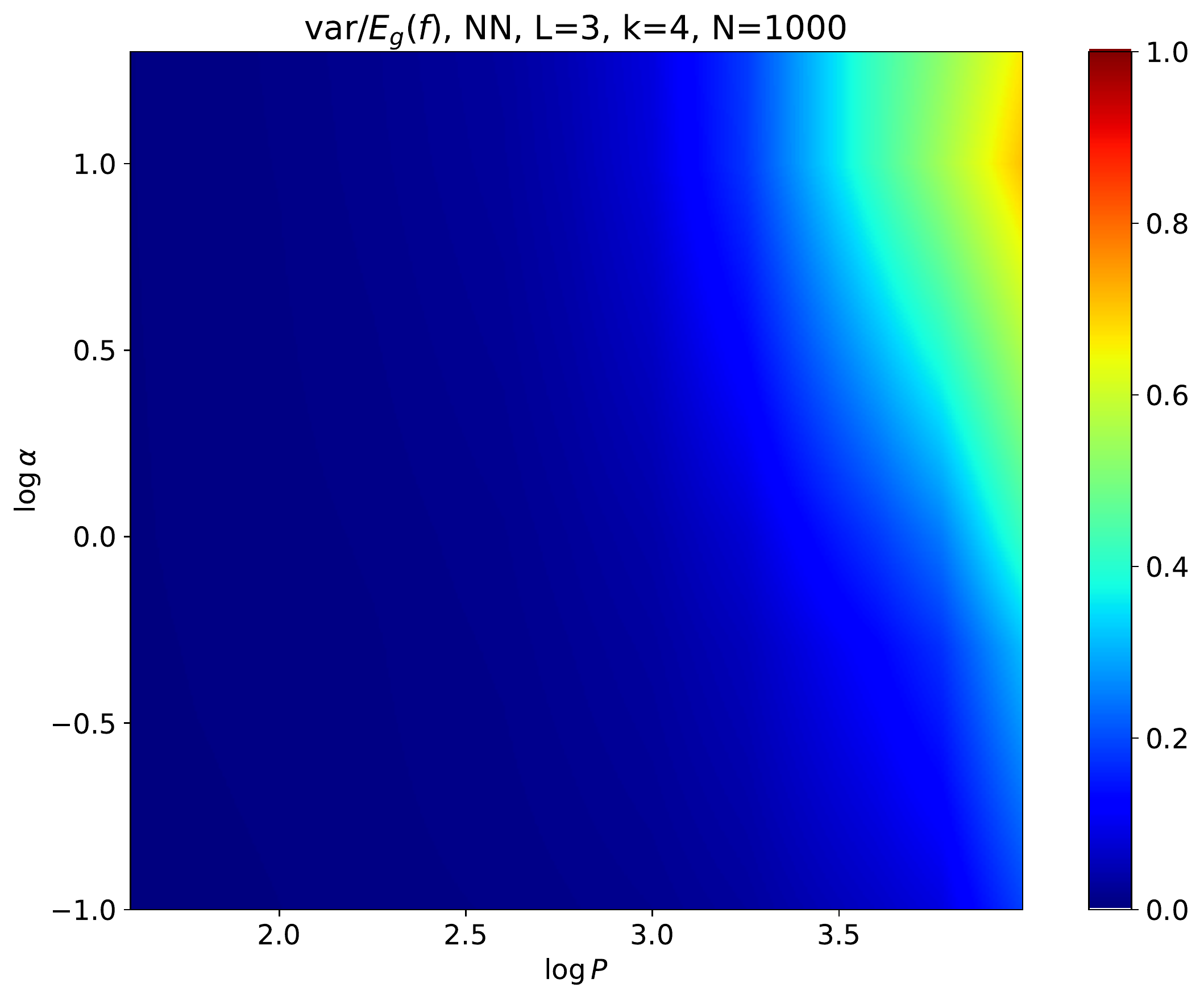}}
    \subfigure[Dataset variance $k=2$]{\includegraphics[width=0.32\linewidth]{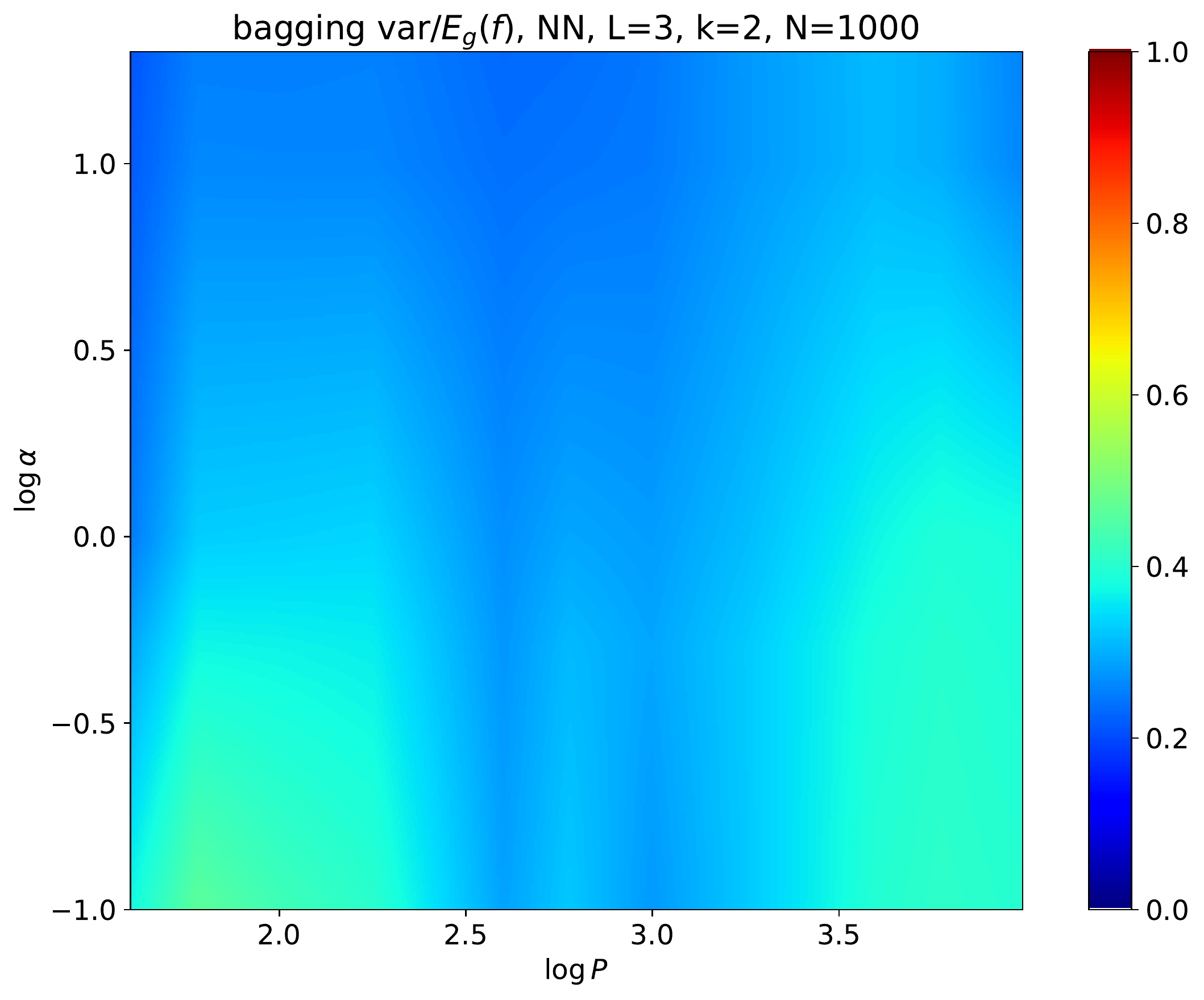}}
    \subfigure[Dataset variance $k=3$]{\includegraphics[width=0.32\linewidth]{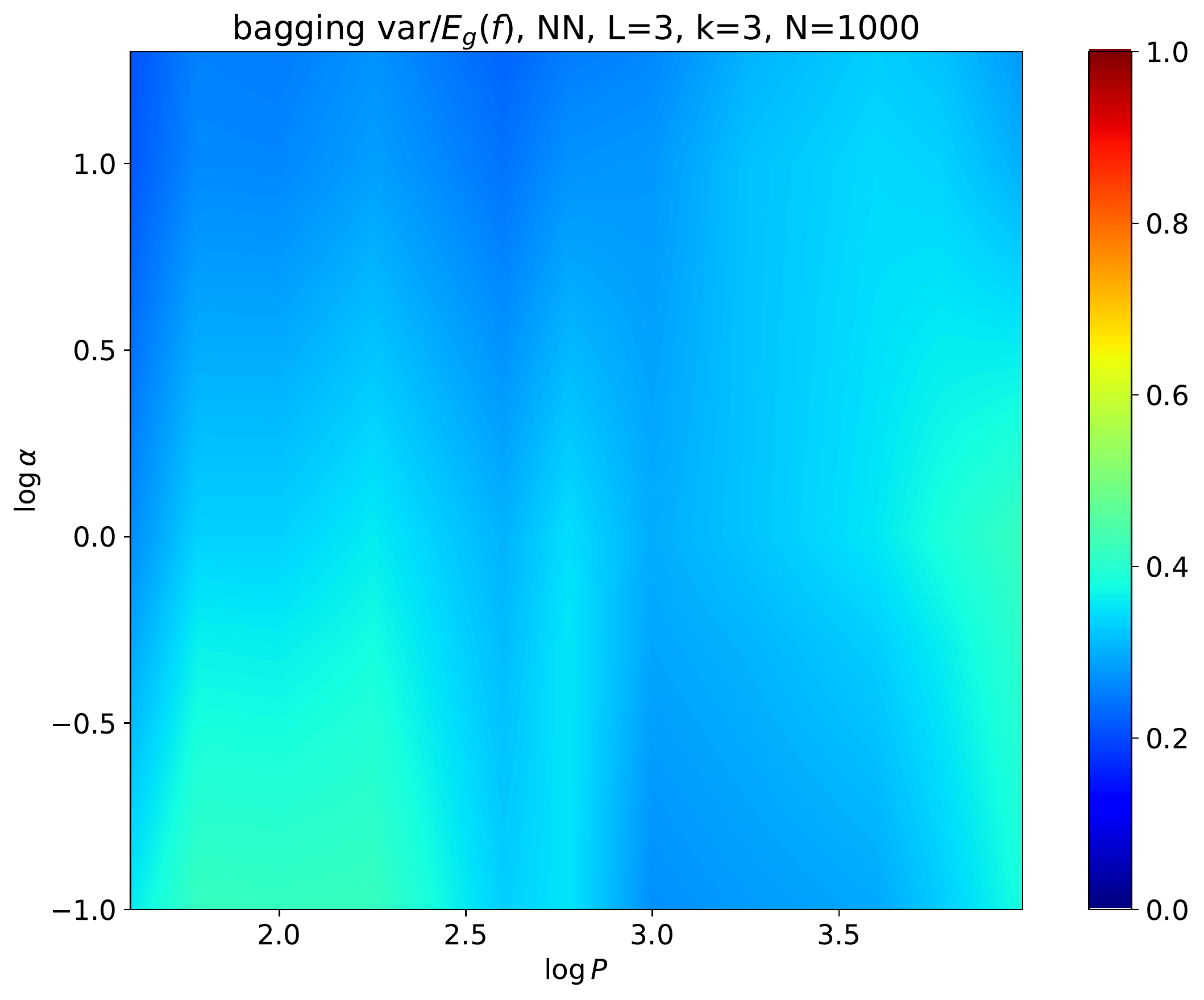}}
    \subfigure[Dataset variance $k=4$]{\includegraphics[width=0.32\linewidth]{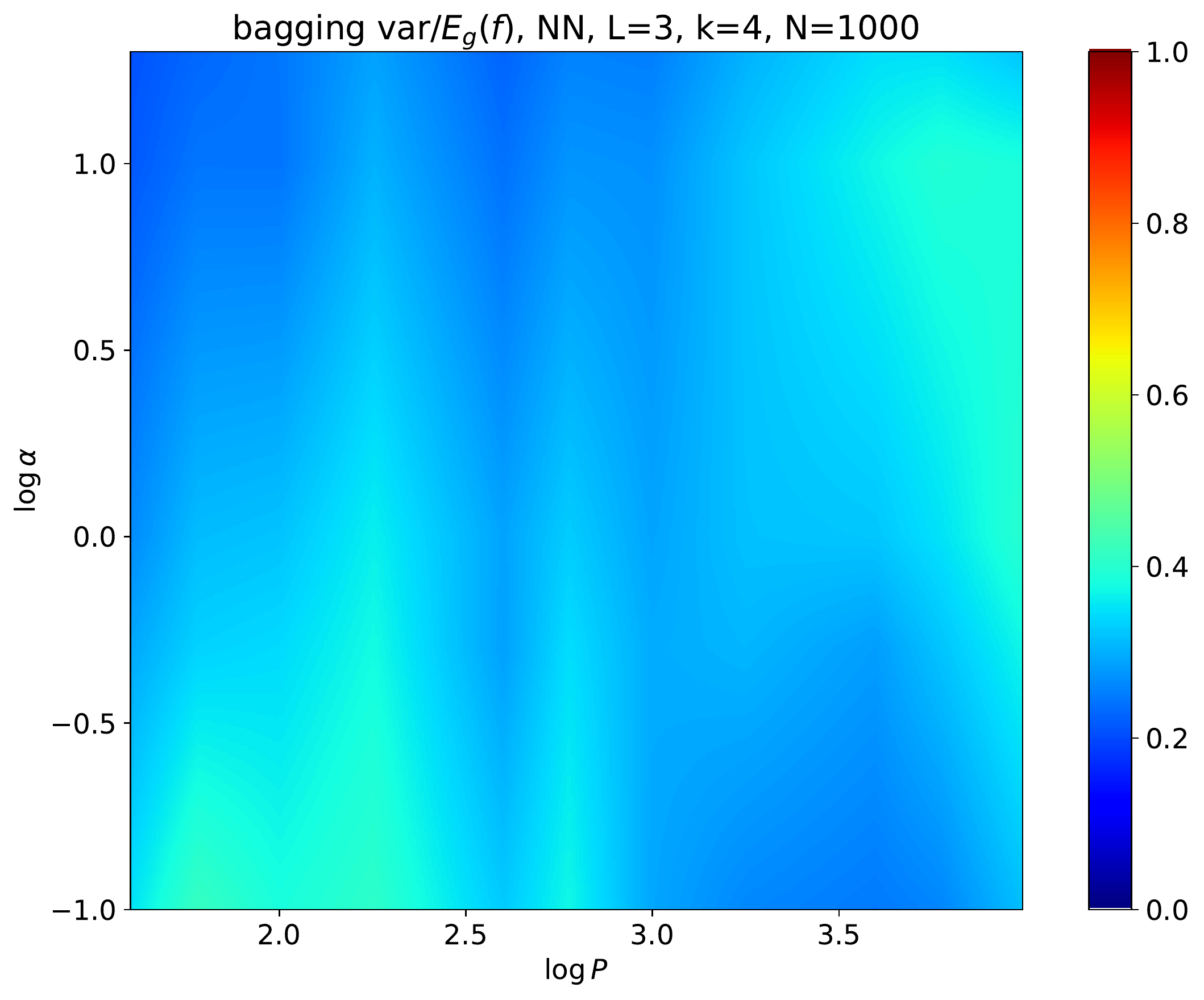}}
    \subfigure[Total variance $k=2$]{\includegraphics[width=0.32\linewidth]{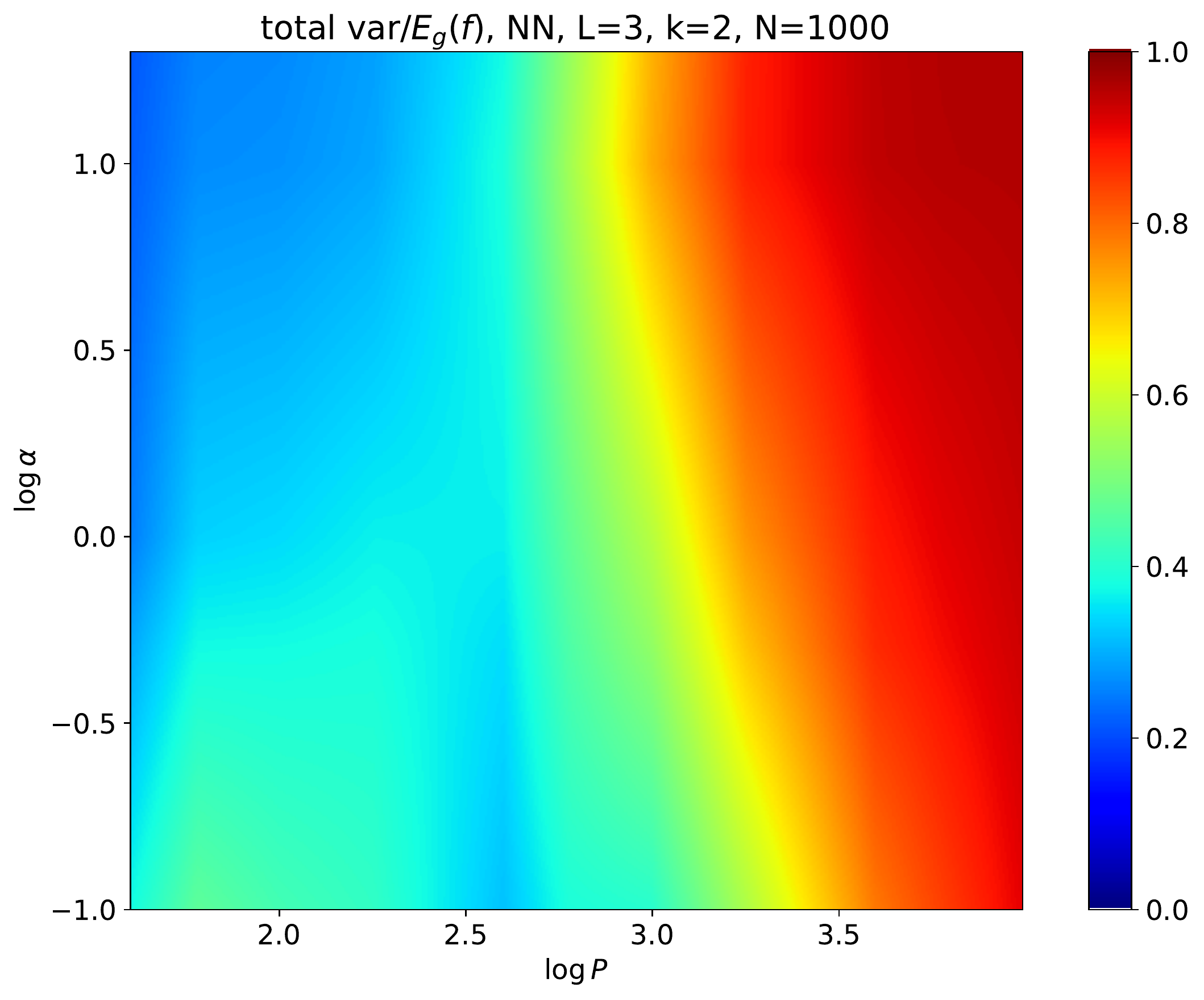}}
    \subfigure[Total variance $k=3$]{\includegraphics[width=0.32\linewidth]{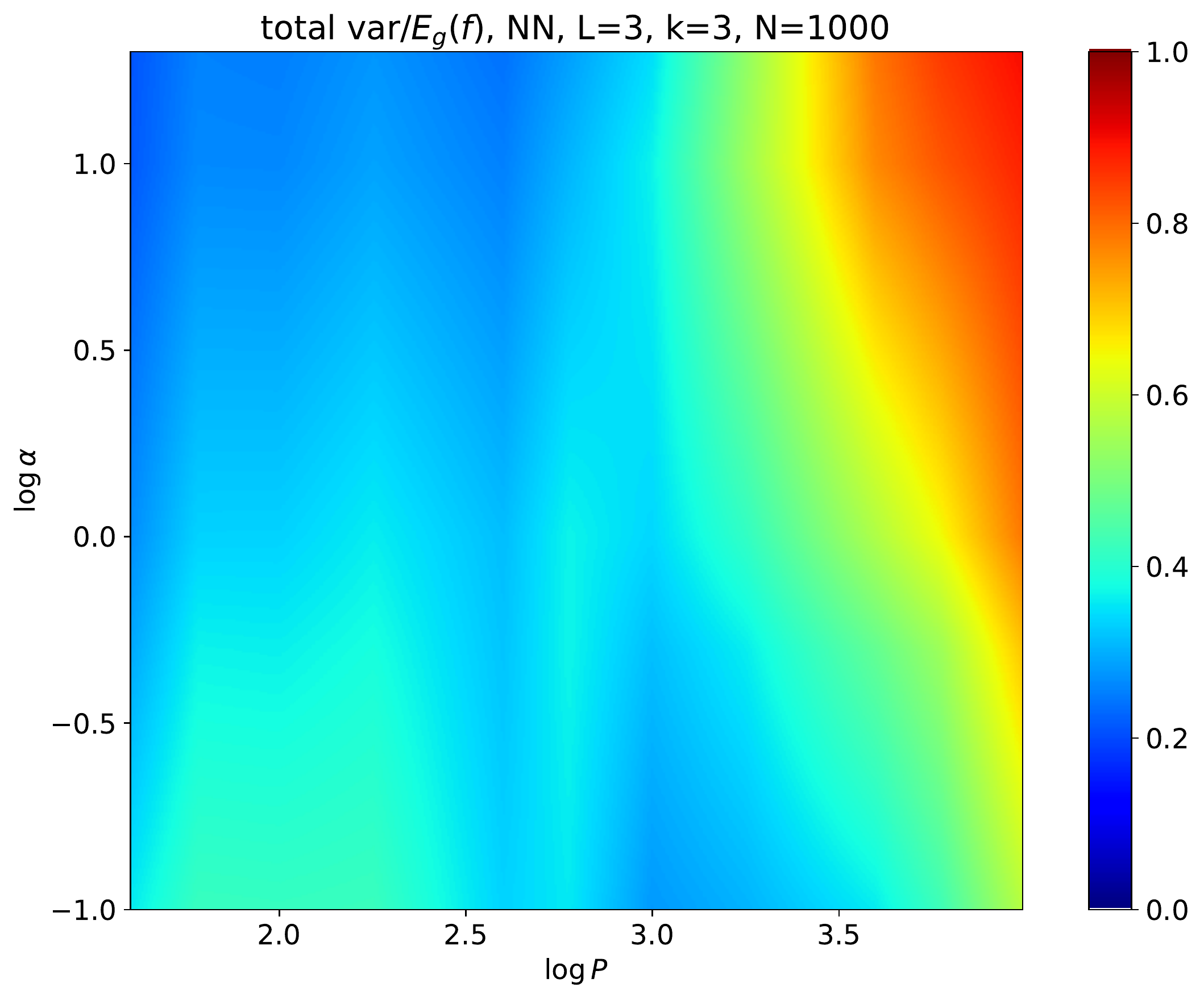}}
    \subfigure[Total variance $k=4$]{\includegraphics[width=0.32\linewidth]{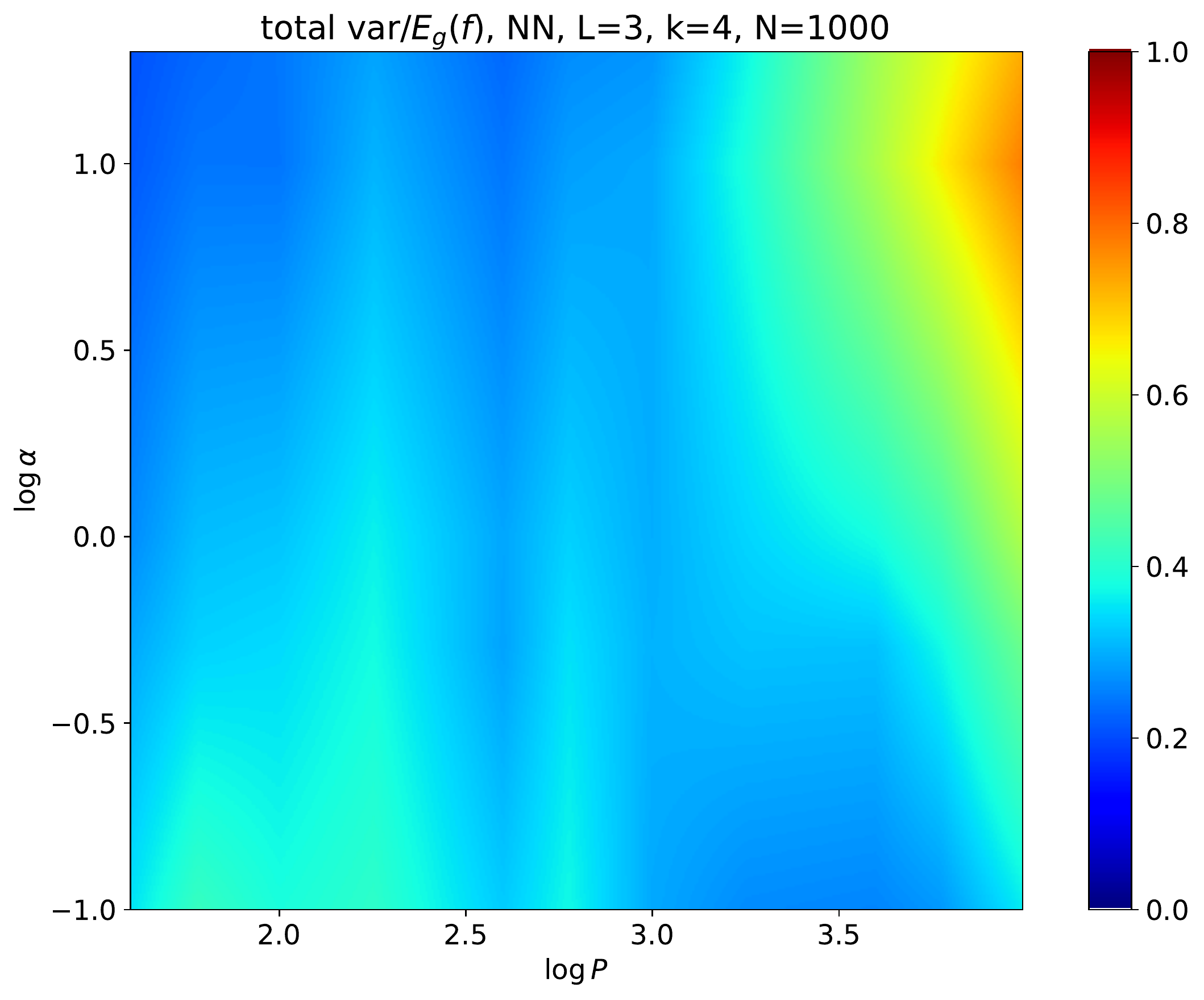}}
    \caption{Phase plots of the fraction of the generalization error due to the initialization variance, the dataset variance, and their combined contribution. The columns correspond to the tasks of polynomial regression for degree $2, 3$ and $4$ polynomials. Neural network has width 1000 and depth 3. Notice that the initialization variance dominates in the large $P$ large $\alpha$ regime}
    \label{fig:fine_grained_variance}
\end{figure}

Using the network simulations, one can show that the bagged predictor does not have substantially lower generalization error in the regimes that we are interested in. This implies that most of the variance driving higher generalization error is due to variance over initializations. In figure \ref{fig:fine_grained_variance}, we make phase plots of the fraction of $E_g$ that arises from variance due to initialization, variance over datasets, and total variance for width 1000. This can be obtained by computing the ensembled predictor, the bagged predictor, and the ensembled-bagged predictor respectively.

\subsection{Relating Ensembled Network Generalization to Infinite Width Generalization}

Making use of the fact that at leading order, the \entkf (either in the rich or lazy regime) of a trained network has $\theta_0$-dependent fluctuations with variance $1/N$, one can write the kernel Gram matrices as 
\begin{equation}
\begin{aligned}
     \,[\bm K_{\theta_0}]_{\mu \nu} &= [\bm K_{\infty}]_{\mu \nu} + \frac{1}{\sqrt{N}} [\bm {\delta K}_{\theta_0}]_{\mu \nu} + O(1/N)\\
     [\bm k_{\theta_0}(\bm x)]_{\mu} &= [\bm k_{\infty}(\bm x)]_{\mu} + \frac{1}{\sqrt{N}} [\bm {\delta k}_{\theta_0}(\bm x)]_{\mu} + O(1/N).
\end{aligned}
\end{equation}
Here, $\bm {\delta K}_{\theta_0}, \bm {\delta k}_{\theta_0}$ are the leading order fluctuations around the infinite width network. Because of how we have written them, their variance is $O(1)$ with respect to $N$. Using perturbation theory \citep{Dyer2020Asymptotics}, one can demonstrate that these leading order terms have mean zero around their infinite-width limit.

The predictor for the \entk (or for a sufficiently large $\alpha$ neural network) for a training set with target labels $\bm y$ is given by:
\begin{equation}
    \begin{aligned}
       f^*(\bm x)_{\theta_0} &=  \bm k_{\theta_0}(\bm x)_{\mu}^\top \bm K_{\theta_0}^{-1} \cdot \bm y \\
       &= f^\infty(\bm x) + \frac{1}{\sqrt N} \bm {\delta k}_{\theta_0}(\bm x)^\top \bm K_{\infty}^{-1} \cdot \bm y - \frac{1}{\sqrt N}  \bm {k}_{\infty}(\bm x)^\top \bm K_{\infty}^{-1} \bm \delta K_{\theta_0} \bm K_{\infty}^{-1} \cdot \bm y + O\left(N^{-1}\right).
    \end{aligned}
\end{equation}
This implies \citep{geiger2020scaling}:
\begin{equation}
    \langle (f^*_{\theta_0} (\bm x) - f^\infty (\bm x))^2 \rangle_{\bm x} = O\left(N^{-1}\right).
\end{equation}
Upon taking the ensemble, because of the mean zero property of the deviations, we get that 
\begin{equation}\label{eq:infinite_pred_deviance}
    \begin{aligned}
       \langle f^*(\bm x)_{\theta_0} \rangle_{\theta_0}
       &= f^\infty(\bm x)  +O\left(N^{-1}\right)\\
       &\Rightarrow \langle( \langle f^*(\bm x)_{\theta_0} \rangle_{\theta_0} - f^\infty(\bm x))^2 \rangle = O\left(N^{-2}\right).
    \end{aligned}
\end{equation}
We can now bound the generalization error of the ensemble of networks in terms of the infinite-width generalization:
\begin{equation}
\begin{aligned}
    \langle ( \langle f^*_{\theta_0}(\bm x) \rangle_{\theta_0}  - y )^2 \rangle_{ \bm x, y} &= \langle ( f^\infty (\bm x) - y )^2 \rangle_{\bm x, y} +  \langle ( \langle f^*_{\theta_0}(\bm x) \rangle_{\theta_0}  - f^\infty(\bm x) )^2 \rangle_{\bm x} \\
    & \quad - 2 \langle (f^\infty(\bm x) - y) (f^\infty(\bm x) - \langle f^*_{\theta_0}(\bm x) \rangle_{\theta_0}) \rangle_{\bm x, y} .
\end{aligned}
\end{equation}
By equation \ref{eq:infinite_pred_deviance}, the second term yields a positive contribution going as $O(N^{-2})$. The last term can be bounded by Cauchy-Schwarz:
\begin{equation}
    \begin{aligned}
         | \langle (f^\infty(\bm x) - y) (f^\infty(\bm x) - \langle f^*_{\theta_0}(\bm x) \rangle_{\theta_0}) \rangle_{\bm x, y} |& \leq \sqrt{ \langle (f^\infty(\bm x) - y)^2 \rangle \langle (f^\infty(\bm x) - \langle f^*_{\theta_0}(\bm x) \rangle_{\theta_0})^2\rangle_{\bm x, y}  }\\
         & = \sqrt{E_g^\infty(P) c_1} / N.
    \end{aligned}
\end{equation}
After we enter the variance limited regime by taking $P>P_{1/2}$ we get $E_g^\infty \leq O(1/N)$ so this last term is bounded by $N^{-3/2}$. 
Consequently, the difference in generalization error between the infinite width NTK and an ensemble of lazy network or \entk predictors is subleading in $1/N$ compared to the generalization gap, which goes as $N^{-1}$.

The same argument can be extended to any predictor that differs from some infinite width limit. In particular \cite{bordelon2022self} show that the fluctuations of the \entkf in any mean field network are asymptotically mean zero with variance $N^{-1}$. The above argument then applies to the predictor obtained by ensembling networks that have learned features. This implies that in the variance limited regime, ensemble averages of feature learning networks have the same generalization as the infinite-width mean field solutions up to a term that decays faster than $N^{-3/2}$. 


\section{Feature Learning}\label{sec:app_feature_learning}

\subsection{Controlling Feature Learning Through Initialization Scale}\label{sec:why_alpha}

Given the feed-forward network defined in equation \ref{eq:network_defn}, one can see that the components of the activations satisfy $h_i^{(\ell)} = O(\sigma h_i)^{(\ell-1)}$ and consequently that the output $h^{(L)}_1 = O(\sigma^L)$. Because of the way the network is parameterized, the changes in the output $\frac{\partial  f}{\partial \theta}$ also scale as $O(\sigma^L)$. This implies that the eNTK at any given time scales as
\begin{equation}
    K_{\theta}(\bm x, \bm x') = \sum_{\theta} \frac{\partial f(\bm x)}{\partial \theta} \frac{\partial f(\bm x')}{\partial \theta} = O(\sigma^{2L}).
\end{equation}
After appropriately rescaling learning rate to $\eta = \sigma^{-2L}$ we get 
\begin{equation}
    \frac{df(\bm x)}{dt} = -\eta \sum_{\mu} K_{\theta}(\bm x, \bm x^\mu) ( f(\bm x^\mu) - y^\mu).
\end{equation}
Under the assumption that $\sigma^L \ll 1$ and $y^\mu = O(1)$ so that the error term is $O(1)$ we get that the output changes in time as $df/dt = O(1)$.

On the other hand, using the chain rule one can show that the features change as a product of the gradient update and the features in the prior layer, yielding the scaling 
\begin{equation}
    \frac{dh^{(\ell)}}{dt}  = \eta \frac{\sigma^L}{\sqrt N} = \frac{1}{\sigma^L \sqrt{N}} = (\alpha \sqrt{N})^{-1}.
\end{equation}
This gives us that the change in the features scales as $(\alpha \sqrt N)^{-1}$ while the change in the output scales as $O(1)$. Thus, for $\alpha \sqrt{N}$ sufficiently small, the features can move dramatically.

\subsection{Output Rescaling without Rescaling Weights}\label{app:alpha_no_weight_rescale}

In the main text, we use the scale $\sigma$ at every layer to change the scale of the output function. This relies on the homogeneity of the activation function so that $W^\ell \to \sigma W^\ell$ for all $\ell$ leads to a rescaling $f \to f \sigma^L$. This would not work for nonhomogenous activations like $\phi(h) = \tanh(h)$. However, following \cite{Chizat2019OnLT, geiger2020scaling}, we note that we can set all weights to be $O_\alpha(1)$ and introduce the $\alpha$ only in the definition of the neural network function
\begin{align}
    f = \frac{\alpha}{\sqrt N} \sum_{i=1}^N w^{L+1}_i \varphi(h^L_i) \ , \ h^{\ell}_i = \frac{1}{\sqrt N} \sum_{j=1}^N W_{ij}^\ell \varphi(h^{\ell-1}_j) \ , \ h_i^1 = \frac{1}{\sqrt D} W_{ij}^1 x_j.
\end{align}
We note that all preactivations $h^\ell$ have scale $O_\alpha(1)$ for any choice of nonlinearity, but that $f = \Theta_\alpha(\alpha)$. Several works have established that the $\alpha \sim \frac{1}{\sqrt N}$ allows feature learning even as the network approaches infinite width \cite{mei2018mean, Yang2020FeatureLI, bordelon2022self}. This is known as the mean field or $\mu$-limit. 



\subsection{Kernel Alignment}\label{sec:kernel_align}

In this section we comment on our choice of kernel alignment metric 
\begin{equation}
    A(\bm K) := \frac{\bm y^\top \bm K \bm y}{\Tr \bm K |\bm y|^2} .
\end{equation}
For kernels that are diagonally dominant, such as those encountered in the experiments, this metric is related to another alignment metric
\begin{equation}
    A_F(\bm K) := \frac{\bm y^\top \bm K \bm y}{|\bm K|_F |\bm y|^2} .
\end{equation}
Here $|\bm K|_F$ is the Frobenius norm of the Gram matrix of the kernel. This metric was extensively used in \citet{baratin2021implicit}. The advantage of the first metric over the second is that one can quickly estimate the denominator of $A(\bm K)$ via Monte Carlo estimation of $\langle \bm u^\top \bm K \bm u \rangle_{\bm u \sim \mathcal N(0, \bm 1)}$.

We use $A(\bm K_f)$ as a measure of feature learning, as we have found that this more finely captures elements of feature learning than other related metrics. We list several metrics we tried that did not work.

One option for a representation-learning metric involves measuring the magnitude of the change between the initial and final kernels, $\bm K_i, \bm K_f$:
\begin{equation}
    \Delta \bm K := |\bm K_f - \bm K_i|_F.
\end{equation}
However, this is more sensitive to the raw parameter change than any task-relevant data. If one instead were to normalize the kernels to be unit norm at the beginning and the end, the modified metric
\begin{equation}
    \Delta \bm K := \left|\frac{\bm K_f}{|\bm K_f|_F} - \frac{\bm K_i}{|\bm K_i|_F}\right|_F.
\end{equation}
This metric however remains remarkably flat over the whole range of $\alpha, P$, as does the centered kernel alignment (CKA) of \cite{cortes2012algorithms}
\begin{equation}
    \mathrm{CKA}(\bm K_i, \bm K_f) = \frac{\Tr[\bm K_i^c \bm K_f^c]}{\sqrt{|\bm K_i^c|_F |\bm K_f^c|_F}}, \quad \bm K^c = \bm C \bm K  \bm C, \quad \bm C = \bm 1 - \frac{1}{P} \vec 1 \, \vec 1^{\; T}.
\end{equation}
Here $C$ is the centering matrix that subtracts off the mean components of the kernel for a $P \times P$ kernel. This alignment metric has been shown to be useful in comparing neural representations \citep{kornblith2019similarity}. For our task, however, because the signal is low-dimensional, only a small set of eigenspaces of the kernel align to this task. As a result, the CKA, which counts all eigenspaces equally, appears to be too coarse to capture the low-dimensional feature learning that is happening.

On the other hand, we find that $A(\bm K_f)$ (with $\bm K_f$ given by the \entkf evaluated on a test set) can very finely detect alignment along the task relevant directions. This produces a clear signal of feature learning at small $\alpha$ and large $P$ as shown in Figure \ref{fig:phase_plot}c. 

 $A(\bm K_f)$ can be related to the centered kernel alignment between the \entkf and the (mean zero) task kernel $\bm y \bm y^\top$, where $\bm y$ is a vector of draws from the population distribution $p(\bm x, y)$.

\subsection{Relationship Between Trained Network and Final Kernel}\label{sec:entkf}

In general, the learned function contains contributions from the instantaneous NTKs at every point in the training. Concretely, following \citet{atanasov2021neural} we have the following formula for the final network predictor $f(x)$
\begin{equation}
f(x) = \int_0^\infty dt\, \bm k(x,t) \cdot \exp\left( - \int_0^t ds \bm K(s) \right) \bm y,    
\end{equation}
where $[\bm k(x,t)]_\mu = K(x,x_\mu, t)$ and $[\bm K(s)]_{\mu \nu} = K(x_\mu, x_\nu, s)$ and $[\bm y]_{\mu} = y_\mu$.  In general there are contributions from earlier kernels $\bm k(x,t)$ for $t < \infty$ and so the function $f$ cannot always be written as a linear combination of the final NTK $K_f$ on training data: $f = \sum_{\mu} \alpha_\mu K_f(x,x_\mu)$. However, as \citet{vyas2022limitations, atanasov2021neural} have shown, the final predictions of the network are often well modeled by regression with the final NTK. We verify this for our task in section \ref{sec:variance}.

\section{Generic Random Feature Model}\label{sec:replica_calculation}

\subsection{Setting up the Problem: Feature Definitions}

For a random kernel, $K(\vx,\vx';\theta)$, we first compute its Mercer decomposition
\begin{align}
    \int d\vx \ p(\vx) K(\vx,\vx';\theta) \phi_k(\vx) = \lambda_k \phi_k(\vx').
\end{align}
From the eigenvalues $\lambda_k$ and eigenfunctions $\phi_k$, we can construct the square root
\begin{align}\label{eq:K_sqrt}
    K^{1/2}(\vx,\vx';\theta) = \sum_k \sqrt{\lambda_k} \phi_k(\vx) \phi_k(\vx').
\end{align}
Lastly, using $K^{1/2}$, we can get a feature map by projecting against a static basis $\{ b_k \}$ giving 
\begin{align}
    \psi_k(\vx) = \int d\vx' p(\vx') K^{1/2}(\vx,\vx';\theta) b_k(\vx').
\end{align}
These features reproduce the kernel so that $K(\vx,\vx';\theta) = \sum_k \psi_k(\vx) \psi_k(\vx')$. This can be observed from the following observation 
\begin{align}
    \psi_k(\vx) &= \sum_\ell \sqrt{\lambda_\ell} \phi_\ell(\vx) U_{\ell k} \ , \ U_{\ell k} = \left< \phi_\ell(\vx) b_k(\vx)\right> 
    \\
    \Rightarrow \sum_k \psi_k(\vx) \psi_k(\vx') &= \sum_{\ell,m} \sqrt{\lambda_\ell \lambda_m} \phi_\ell(\vx) \phi_m(\vx') \sum_{k} U_{\ell k } U_{m k}  
    = \sum_\ell \lambda_\ell \phi_\ell(\vx), \phi_\ell(\vx')
\end{align}
where the last line follows from the orthogonality of $U_{km}$ and recovers $K(\bm x, \bm x'; \theta)$. 

\subsection{Decomposition of Finite Width Features}

We now attempt to characterize the variance in the features over the sample distribution. We will first consider the case of a fixed realization of $\bm\theta_0$ before providing a typical case analysis over random $\bm\theta_0$. For a fixed initialization $\bm\theta_0$ we define the following covariance matrices
\begin{align}
    \bm\Sigma_M = \left< \bm\psi_{M}(\vx) \bm\psi_{M}(\vx)^\top \right> \in \mathbb{R}^{M \times M}.
\end{align}
where $\bm\psi_M$ are the truncated (but deterministic) features induced by the deterministic infinite width kernel. We will mainly be interested in the case where $M \to \infty$ and where the target function can be expressed as the linear combination $y(\vx) = \vw^* \cdot \bm\psi_M(\vx)$ of these features. For example, in the case of our experiments on the sphere, $\bm\psi_M$ could be the spherical harmonic functions. Further, in the $M \to \infty$ limit, we will be able to express the target features $\bm\psi$ as linear combinations of the features $\bm\psi_M$
\begin{align}
    \bm\psi(\vx,\bm\theta_0) = \mA(\bm\theta_0) \bm\psi_{M}(\vx) \ , \ \mA(\bm\theta) \in \mathbb{R}^{N_{\mathcal H} \times M}.
\end{align}
The matrix $\bm A(\bm\theta_0)$ are the coefficients of the decomposition which can vary over initializations. Crucially $\bm A(\bm\theta_0)$ projects to the subspace of dimension $N_{\mathcal H}$ where the finite width features have variance over $\vx$. The population risk for this $\bm\theta_0$ has an irreducible component
\begin{align}
    E_g(\bm\theta_0) &= \left< \left( \vw^* \cdot \bm\psi_M - \vw \cdot \bm\psi \right)^2 \right> \nonumber
    \\
    &\geq \vw^{*\top} \left[ \bm\Sigma_M - \bm\Sigma_M \mA(\bm\theta)^\top \left( \mA(\bm\theta_0) \bm\Sigma_M \mA(\bm\theta_0)^\top \right)^{-1} \mA(\bm\theta_0) \bm\Sigma_M \right]\vw^*.
\end{align}
where the bound is tight for the optimal weights $\vw = \left( \mA(\bm\theta_0) \bm\Sigma_M \mA(\bm\theta_0)^\top \right)^{-1} \mA(\bm\theta_0) \bm\Sigma_M \vw^*$. The irreducible error is determined by a projection matrix which preserves the subspace where the features $\bm\psi(\vx,\bm\theta_0)$ have variance: $\mI - \mA(\bm\theta)^\top \left( \mA(\bm\theta_0) \bm\Sigma_M \mA(\bm\theta_0)^\top \right)^{-1} \mA(\bm\theta_0) \bm\Sigma_M$. In general, this will preserve some fraction of the variance in the target function, but some variance in the target function will not be expressible by linear combinations of the features $\bm\psi(\vx,\bm\theta)$. We expect that random finite width $N$ neural networks will have unexplained variance in the target function on the order $\sim 1/N$. 

\subsection{Gaussian Covariate Model}

Following prior works on learning curves for kernel regression \citep{bordelon_icml_learning_curve,Canatar2021SpectralBA, loureiro_lenka_feature_maps}, we will approximate the learning problem with a Gaussian covariates model with matching second moments. 

The features $\bm\psi_M(\vx)$ will be treated as Gaussian over random draws of datapoints. We will assume centered features. We decompose the features in the orthonormal basis $\vb(\vx)$, which we approximate as a Gaussian vector $\vb \sim \mathcal{N}(0,\mI)$.
\begin{align}
    f &=  \bm\psi(\bm\theta_0) \cdot \vw \ , \ y =  \bar{\bm\psi}_M \cdot \vw^* \nonumber
    \\
    {\psi}_M &= \bm\Sigma_{s}^{1/2} \vb \ , \ \bm\psi(\bm\theta_0) = \mA(\bm\theta_0)^\top \bm\psi_M + \bm\Sigma_{\epsilon}^{1/2}  \bm\epsilon \nonumber
    \\
    \vb &\sim \mathcal{N}(0,\mI) \ , \ \bm\epsilon \sim \mathcal{N}(0,\mI) 
\end{align}
This is a special case of the Gaussian covariate model introduced by \cite{loureiro_lenka_feature_maps} and subsumes the popular two-layer random feature models \citep{mei2022generalization, adlam2020understanding} as a special case. In a subsequent section, we go beyond \cite{loureiro_lenka_feature_maps} by computing typical case learning curves over Gaussian $\mA(\bm\theta_0)$ matrices. In particular, we have for the two layer random feature model in the proportional asymptotic limit $P, N, D \to \infty$ with $P/D = O(1)$ and $P/N = O(1)$ with $\bm\psi(\vx) = \phi( \bm F \vx_\mu)$ for fixed feature matrix $\mF \in \mathbb{R}^{N \times D}$ nonlinearity $\phi$ and $\vx = \vb \sim \mathcal{N}(0, D^{-1} \bm I)$
\begin{align}
    \bm\Sigma_M &= \mI \ , \ \bm\Sigma_\epsilon = c_*^2 \mI \ , \ \mA^\top = c_1 \mF \nonumber
    \\
    c_1 &= \left< z \phi(z) \right>_{z \sim \mathcal{N}(0,1)} \ , \ c_*^2 = \left< \phi(z)^2 \right>_{z \sim \mathcal{N}(0,1)} - c_1^2.
\end{align}
We refer readers to \citet{hu2020universality} for a discussion of this equivalence between random feature regression and this Gaussian covariate model.

\subsection{Replica Calculation of the Learning Curve}

To analyze the typical case performance of kernel regression, we define the following partition function which is dominated 
\begin{align}
    Z[\mathcal D,\bm\theta_0] &= \int d\vw \exp\left( - \frac{\beta}{2\lambda} \sum_{\mu=1}^P [  \vw \cdot \bm\psi_\mu - \vw^* \cdot {\bm\psi}_{M,\mu} ]^2 - \frac{\beta}{2} |\vw|^2 - \frac{J\beta M}{2} E_g(\vw)   \right) \nonumber
    \\
    &E_g(\vw) = \frac{1}{M} |\bm\Sigma_M^{1/2} \vw^* - \bm\Sigma_M^{1/2} \mA(\bm\theta_0) \vw |^2 + \frac{1}{M} \vw^\top \bm\Sigma_{\epsilon} \vw .
\end{align}
For proper normalization, we assume that $\left< \bm\psi_M \bm\psi_M^\top \right> =\frac{1}{M} \bm\Sigma_M$ and $\left< \bm\epsilon \bm\epsilon^\top \right> = \frac{1}{M} \bm\Sigma_{\epsilon}$. We note that in the $\beta \to \infty$ limit, the partition function is dominated by the unique minimizer of the regularized least squares objective \citep{Canatar2021SpectralBA, loureiro_lenka_feature_maps}. Further, for a fixed realization of $\bm\theta_0$ the average generalization error over datasets $\mathcal D$ can be computed by differentiation of the source term $J$
\begin{equation}
\begin{aligned}
    &\frac{2}{\beta M} \frac{\partial }{\partial J}|_{J=0} \left< \ln Z[\mathcal D,\bm\theta_0] \right>_{\mathcal D}\\
    &= \left< \frac{1}{Z} \int d\vw \exp\left( - \frac{\beta}{2\lambda} \sum_{\mu=1}^P [ \vw \cdot \bm\psi_\mu - \vw^* \cdot {\bm\psi}_{M,\mu} ]^2 -\frac{\beta}{2} |\vw|^2 \right)  E_g(\vw)    \right>_{\mathcal D}.
\end{aligned}    
\end{equation}

Thus the $\beta \to \infty$ limit of the above quantity will give the expected generalization error of the risk minimizer. We see the need to average the quantity $\ln Z$ over realizations of datasets $\mathcal D$. For this, we resort to the replica trick $\left< \ln Z \right> = \lim_{n \to 0} \frac{1}{n} \ln \left< Z^n \right>$. We will compute the integer moments $\left< Z^n \right>$ for integer $n$ and then analytically continue the resulting expressions to $n \to 0$ under a symmetry ansatz. The replicated partition function thus has the form
\begin{equation}
\begin{aligned}
    \left< Z^n \right> = \int \prod_{a=1}^n d\vw^a \mathbb{E}_{\{ \vb_\mu ,\bm\epsilon_\mu \}} &\exp\left(- \frac{\beta}{2\lambda} \sum_{\mu=1}^P \sum_{a=1}^n [ \vw^a \cdot \bm\psi_\mu - \vw^* \cdot {\bm\psi}_{M,\mu} ]^2 - \frac{\beta}{2} \sum_{a=1}^n |\vw^a|^2  \right)\\ & \times \exp \left(- \frac{J\beta M}{2} \sum_{a=1}^n E_g(\vw^a)   \right).
\end{aligned}
\end{equation}
We now need to perform the necessary average over the random realizations of data points $\mathcal D = \{\vb_\mu ,\bm\epsilon_\mu \}$. We note that the scalar quantities $h^a_\mu = \vw^a \cdot \bm\psi_\mu - \vw^* \cdot {\bm\psi}_{M,\mu}$ are Gaussian with mean zero and covariance
\begin{align}
    \left< h^a_\mu h^b_\nu \right> &= \delta_{\mu\nu} Q_{ab} \nonumber
    \\
    Q_{ab} &= \frac{1}{M} \left( \mA(\bm\theta_0) \vw^a - \vw^* \right) \bm\Sigma_M \left( \mA(\bm\theta_0) \vw^a - \vw^* \right) +  \frac{1}{M} \vw^a \bm\Sigma_\epsilon \vw^b.
\end{align}
We further see that the generalization error in replica $a$ is $E_{g}(\vw^a) = Q_{aa}$. Performing the Gaussian integral over $\{ h^a_\mu \}$ gives
\begin{align}
    \left< Z^n \right> \propto \int \prod_{a}& d\vw^a \prod_{ab} dQ_{ab} d\hat{Q}_{ab} \exp\left( - \frac{P}{2} \ln\det\left( \lambda\mI + \beta \mQ \right) - \frac{J\beta M}{2} \text{Tr}\mQ - \frac{\beta}{2} \sum_{a} |\vw^a|^2 \right) \nonumber
    \\
    &\exp\left( \frac{1}{2} \sum_{ab} \hat{Q}_{ab}\left( M Q_{ab} -  [\mA(\bm\theta_0) \vw^a - \vw^* ]^\top \bm\Sigma_M [\mA(\bm\theta_0) \vw^a - \vw^*] + \vw^a \bm\Sigma_\epsilon \vw^b \right) \right). \nonumber
\end{align}
We introduced the Lagrange multipliers $\hat{\mQ}$ which enforce the definition of order parameters $\mQ$. We now integrate over $\mW = \text{Vec}\{ \vw^a\}_{a=1}^n$. We let $\bm{\tilde\Sigma}_s =  \mA^\top \bm\Sigma_M \mA$ 
\begin{align}
    \int d\mW &\exp\left( - \frac{1}{2} \mW \left[ \beta\mI + \hat{\mQ} \otimes [\bm{\tilde\Sigma}_s + \bm\Sigma_\epsilon ] \right] \mW   \right) \nonumber
    \\
    &\exp\left(  \mW^\top \left[ \hat{\mQ} \otimes \mI  \right] \left( \bm 1 \otimes \mA^\top \bm{\Sigma}_s \vw^* \right) \right) \nonumber
    \\
    = &\exp\left( \frac{1}{2} \left( \bm 1 \otimes  \mA^\top \bm{\Sigma}_s \vw^* \right)^\top\left[ \hat{\mQ} \otimes \mI   \right] \left[ \beta\mI + \hat{\mQ} \otimes [\bm{\tilde\Sigma}_s +   \bm\Sigma_\epsilon ] \right]^{-1} \left[ \hat{\mQ} \otimes \mI  \right] \left( \bm 1 \otimes \mA^\top \bm{\Sigma}_s \vw^* \right) \right)\nonumber
    \\
    &\exp\left( - \frac{1}{2} \ln\det\left[ \beta\mI + \hat{\mQ} \otimes [\bm{\tilde\Sigma}_s + \bm\Sigma_\epsilon ] \right] \right).
\end{align}
To take the $n\to 0$ limit, we make the replica symmetry ansatz
\begin{align}
    \beta \mQ = q \mI + q_0 \bm 1 \bm 1^\top \ , \ \beta^{-1} \hat{\mQ} = \hat{q} \mI + \hat{q}_0 \bm 1 \bm 1^\top ,
\end{align}
which is well motivated since this is a convex optimization problem. Letting $\alpha = P/N$, we find that under the RS ansatz the replicated partition function has the form
\begin{align}
    \left< Z^n \right> &= \int dq dq_0 d\hat q d\hat{q}_0 \exp\left( \frac{n M}{2} S[q,q_0, \hat q,\hat{q}_0] \right)
    \\
    S &=  q \hat{q} + q_0 \hat{q} + q \hat{q}_0 - \alpha \left[ \ln(\lambda + q) + \frac{q_0}{\lambda + q} \right] \nonumber \\
    &-  \frac{\beta}{M} \vw^* [\hat{q} \bm\Sigma_M] \vw^* + \frac{ \beta}{M} \vw^*  [\hat{q} \tilde{\bm\Sigma}_s ] \mA^\top \mG \mA [\hat{q} \tilde{\bm\Sigma}_s ]  \vw^* \nonumber
    \\
    &- \frac{1}{M} \ln\det \mG - \frac{1}{M} \hat{q}_0 \text{Tr}\mG [ \tilde{\bm\Sigma}_s +  \bm\Sigma_\epsilon]   - J(q+q_0)  \nonumber
    \\
    \mG &= \left( \mI + \hat{q} [\tilde{\bm\Sigma}_s  +  \bm\Sigma_\epsilon] \right)^{-1}.
\end{align}
In a limit where $\alpha = P/M$ is $O(1)$, then this $S$ is intensive $S = O_M(1)$. We can thus appeal to saddle point integration (method of steepest descent) to compute the set of order parameters which have dominant contribution to the free energy. 
\begin{align}
    \left< Z^n \right> &= \int dq dq_0 d\hat q d\hat{q}_0 \exp\left( \frac{n M}{2} S[q,q_0, \hat q,\hat{q}_0] \right) \sim \exp\left( \frac{nM}{2} S[q^*,q_0^*, \hat q^*,\hat{q}_0^*] \right) \nonumber
    \\
    \implies \left< \ln Z \right> &= \frac{M}{2} S[q^*,q_0^*, \hat q^*,\hat{q}_0^*].
\end{align}
The order parameters $q^*,q_0^*, \hat q^*,\hat{q}_0^*$ are defined via the saddle point equations $\frac{\partial S}{\partial q} =\frac{\partial S}{\partial q_0} =\frac{\partial S}{\partial \hat q} =\frac{\partial S}{\partial \hat q_0} = 0$. For our purposes, it suffices to analyze two of these equations
\begin{align}
    \frac{\partial S}{\partial q_0} &= \hat{q} - \frac{\alpha}{\lambda+ q} - J = 0, \nonumber
    \\
    \frac{\partial S}{\partial \hat{q}_0} &= q - \frac{1}{M} \text{Tr} \mG \left[ \bm{\tilde\Sigma}_s + \bm\Sigma_{\epsilon} \right] = 0 .
\end{align}
We can now take the zero temperature ($\beta \to \infty$) limit to solve for the generalization error
\begin{align}
    E_g &= - \frac{\partial }{\partial J}|_{J=0} \lim_{\beta \to \infty} \frac{1}{\beta } F  \nonumber
    \\
    &= \frac{1}{M} \partial_J  \vw^* \left[ \hat{q} \bm\Sigma_M - \hat{q}^2 \bm\Sigma_M \mA^\top \mG  \mA \bm\Sigma_M \right] \vw^*.
\end{align}
We see that we need to compute the $J$ derivatives on $\hat{q}$. We let $\kappa = \lambda + q$ and note
\begin{align}
    \partial_J \hat{q} &= - \alpha \kappa^{-2} \partial_J \kappa + 1 \nonumber
    \\ 
    \partial_J \kappa &= - \partial_J \hat{q} \ \frac{1}{M} \text{Tr} \mG^2\left[ \bm{\tilde\Sigma}_s  +  \bm\Sigma_{\epsilon} \right]^2  = - \left( - \alpha \kappa^{-2} \partial_J \kappa + 1 \right) \frac{1}{M} \text{Tr} \mG^2\left[ \bm{\tilde\Sigma}_s  +  \bm\Sigma_{\epsilon} \right]^2 .
\end{align}
We solve the equation for $\partial_J \kappa$ which gives $\partial_J \kappa = - \frac{\kappa^2}{\alpha} \frac{\gamma}{1-\gamma}$ where $\gamma = \frac{\alpha}{\kappa^2} \frac{1}{M} \text{Tr} \mG^2 \left[ \bm{\tilde\Sigma}_s  + \bm\Sigma_{\epsilon} \right]^2$. With this definition we have $\partial_J \hat{q} = 1 + \frac{\gamma}{1-\gamma} = \frac{1}{1-\gamma}$.
\begin{align}
    E_g &= \frac{1}{1-\gamma} \frac{1}{M} \vw^* \bm\Sigma_M^{1/2} \left[ \mI - 2 \hat{q} {\bm\Sigma}_s^{1/2} \mA^\top \mG \mA {\bm\Sigma}_s^{1/2}  + \hat{q}^2  {\bm\Sigma}_s^{1/2} \mA^\top \mG  \left[ \bm{\tilde\Sigma}_s  +  \bm\Sigma_{\epsilon} \right] \mG \mA {\bm\Sigma}_s^{1/2} \right] \bm\Sigma_M^{1/2} \vw^* \nonumber
    \\
    &= \frac{1}{1-\gamma} \frac{1}{M} \vw^* \bm\Sigma_M^{1/2} \left[ \mI - \hat{q} {\bm\Sigma}_s^{1/2} \mA^\top \mG \mA {\bm\Sigma}_s^{1/2}  - \hat{q}  {\bm\Sigma}_s^{1/2}  \mA^\top \mG^2 \mA  {\bm\Sigma}_s^{1/2} \right] \bm\Sigma_M^{1/2} \vw^*.
\end{align}
This reproduces the derived expression from \cite{loureiro_lenka_feature_maps}. The matching covariance $\tilde{\bm\Sigma}_s = \bm\Sigma_M$ and zero feature-noise limit $\bm\Sigma_\epsilon = 0$ recovers the prior results of \cite{bordelon_icml_learning_curve, Canatar2021SpectralBA, simon2021neural}. In general, this error will asymptote to the the irreducible error
\begin{align}
    \lim_{P \to \infty} E_g = \frac{1}{M} \vw^* \left[ \bm\Sigma_M - {\bm\Sigma}_s \mA^\top \left( \tilde{\bm\Sigma}_s +  \bm\Sigma_{\epsilon} \right)^{-1} \mA {\bm\Sigma}_s   \right] \vw^* .
\end{align}
We see that this recovers the minimal possible error in the $P\to\infty$ limit. The derived learning curves depend on the instance of random initial condition $\bm\theta_0$. To get the average case performance, we take an additional average of this expression over $\bm\theta_0$
\begin{align}\label{eq:avg_eg_over_theta}
    \mathbb{E}_{\bm\theta_0} E_g(\bm\theta_0) = \mathbb{E}_{\theta_0} \frac{1}{1-\gamma} \frac{1}{M} \vw^* \bm\Sigma_M^{1/2} \left[\mI - \hat{q} {\bm\Sigma}_s^{1/2} \mA^\top \mG \mA {\bm\Sigma}_s^{1/2}  - \hat{q}  {\bm\Sigma}_s^{1/2}  \mA^\top \mG^2 \mA  {\bm\Sigma}_s^{1/2} \right] \bm\Sigma_M^{1/2} \vw^*.
\end{align}
This average is complicated since $\gamma, \hat{q}, \mG$ all depend on $\bm\theta_0$. In the next section we go beyond this analysis to try average case analysis for random Gaussian $\mA$.

\subsection{Quenched Average over Gaussian A}\label{app:gauss_A_computation}

In this section we will define a distribution of features which allows an exact asymptotic prediction over random realizations of disorder $\bm\theta_0$ and datasets $\mathcal D$. This is a nontrivial extension of the result of \cite{loureiro_lenka_feature_maps} since the number of necessary saddle point equations to be solved doubles from two to four. However, this more complicated theory allows us to exactly compute the expectation in \eqref{eq:avg_eg_over_theta} under an ansatz for the random matrix $\mA$. We construct our features with
\begin{align}
    \bm\psi | \mA &= \frac{1}{\sqrt N} \mA^\top {\bm\psi}_M + \bm\Sigma_{\epsilon}^{1/2} \bm\epsilon \nonumber \ , \ A_{ij} \sim \mathcal{N}(0, \sigma^2 ).
\end{align}
We will now perform an approximate average over both datasets $\mathcal D$ and realizations of $\bm A$
\begin{equation}
\begin{aligned}
    \left< Z^n \right> = \int \prod_{a=1}^n d\vw^a \mathbb{E}_{\{ \vb_\mu ,\bm\epsilon_\mu, \mA \}} &\exp\left(- \frac{\beta}{2\lambda} \sum_{\mu=1}^P \sum_{a=1}^n [ \vw^a \cdot \bm\psi_\mu - \vw^* \cdot {\bm\psi}_{M,\mu} ]^2 - \frac{\beta}{2} \sum_{a=1}^n |\vw^a|^2 \right) \\ & \times \exp \left(- \frac{J M \beta}{2} \sum_{a=1}^n E_g(\vw^a)   \right).
\end{aligned}    
\end{equation}
As before, we first average over $\vb_\mu, \bm\epsilon_\mu | A$ and define order parameters $Q_{ab}$ as before.
\begin{align}
    \left< Z^n \right> = \int \prod_{a}& d\vw^a \prod_{ab} dQ_{ab} d\hat{Q}_{ab} \exp\left( - \frac{P}{2} \ln\det\left( \lambda\mI + \beta \mQ \right) - \frac{J\beta M}{2} \text{Tr}\mQ - \frac{\beta }{2} \sum_{a} |\vw^a|^2 \right) \nonumber
    \\
    \mathbb{E}_{\{\vg^a\}} &\exp\left( \frac{1}{2} \sum_{ab} \hat{Q}_{ab}\left( M Q_{ab} - [\vg^a - \vw^* ]^\top \bm\Sigma_M [\vg^b - \vw^*] +  \vw^a \bm\Sigma_\epsilon \vw^b \right) \right). \nonumber
\end{align}
where we defined the fields $\vg^a = \frac{1}{\sqrt N} \mA \vw^a$ which are mean zero Gaussian with covariance $\left< \vg^a \vg^{b \top} \right> = V_{ab} \mI$ where $V_{ab} = \frac{\sigma^2}{N} \vw^a \cdot \vw^b$. Performing the Gaussian integral over $\mG = \text{Vec}\{\vg^a\}$, we find
\begin{align}
    &\int\prod_a d\vg^a \exp\left( - \frac{1}{2} \mG \left[ \mI \otimes \mV^{-1} + \bm\Sigma_M \otimes \hat{\mQ} \right] \mG + (\bm\Sigma_M \vw^* \otimes \hat{\mQ} \bm 1) \mG - \frac{1}{2} \ln\det\left( \mI \otimes \mV \right) \right) \nonumber
    \\
    &= \exp\left( \frac{1}{2}(\bm\Sigma_M \vw^* \otimes \hat{\mQ} \bm 1)  \left[ \mI \otimes \mV^{-1} + \bm\Sigma_M \otimes \hat{\mQ} \right]^{-1} (\bm\Sigma_M \vw^* \otimes \hat{\mQ} \bm 1) - \frac{1}{2} \ln\det\left( \mI + \bm\Sigma_M \otimes \hat{\mQ} \mV \right)  \right).
\end{align}
Next, we need to integrate over $\mW = \text{Vec}\{\vw^a\}$ which gives
\begin{align}
    \int d\mW \exp\left( - \frac{1}{2}\mW \left[ \beta \mI + \sigma^2 \mI \otimes \hat{\mV} +  \bm\Sigma_{\epsilon} \otimes \hat{\mQ} \right] \mW \right) = \exp\left( - \frac{1}{2} \ln\det\left[ \beta \mI + \sigma^2 \mI \otimes \hat{\mV} + \bm\Sigma_{\epsilon} \otimes \hat{\mQ} \right]  \right).
\end{align}
Now the replicated partition function has the form
\begin{equation}
\begin{aligned}
    \left<Z^n\right> = &\int d\mQ d\hat{\mQ}  d\mV d\hat{\mV} \exp\left( \frac{M}{2} \text{Tr} [\mQ \hat{\mQ} +  \eta \mV\hat{\mV}] - \frac{J\beta M}{2} \text{Tr}\mQ - \frac{P}{2}\ln\det\left[ \lambda \mI + \beta \mQ \right] \right) 
    \\
    & \times \exp\left( - \frac{1}{2}(\vw^* \otimes \bm 1)^\top [\bm\Sigma_M \otimes \hat{\mQ} ] (\vw^* \otimes \bm 1) \right)\\
    & \times \exp\left( \frac{1}{2} (\bm\Sigma_M \vw^* \otimes \hat{\mQ} \bm 1)^\top  \left[ \mI \otimes \mV^{-1} + \bm\Sigma_M \otimes \hat{\mQ} \right]^{-1}  (\bm\Sigma_M \vw^* \otimes \hat{\mQ} \bm 1)  \right) 
    \\
    &\times \exp\left( - \frac{1}{2} \ln\det\left[ \mI + \bm\Sigma_M \otimes \hat{\mQ} \mV \right]  - \frac{1}{2} \ln\det\left[ \beta \mI + \sigma^2 \mI \otimes \hat{\mV} +  \bm\Sigma_{\epsilon} \otimes \hat{\mQ} \right]  \right) .
\end{aligned}
\end{equation}
Now we make a replica symmetry ansatz on the order parameters $\mQ ,\hat{\mQ} , \mV,\hat{\mV}$
\begin{align}
    &\beta \mQ = q \mI + q_0 \bm 1 \bm 1^\top \ , \ \beta \mV = v \mI + v_0 \bm 1 \bm 1^\top \nonumber
    \\
    &\beta^{-1} \hat{\mQ} = \hat{q} \mI + \hat{q}_0 \bm 1 \bm 1^\top \ , \ \beta^{-1} \hat{\mV} = \hat{v} \mI + \hat{v}_0 \bm 1 \bm 1^\top .
\end{align}
We introduce the shorthand for normalized trace of a matrix $\mG$ as $\text{tr} \ \mG = \frac{1}{M} \text{Tr} \mG$. Under the replica symmetry ansatz, we find the following free energy 
\begin{align}
    \frac{2}{M} \left< \ln Z \right> &= q \hat{q} + q_0 \hat{q} + q \hat{q}_0 + \eta( v \hat{v} + v_0 \hat{v} + v \hat{v}_0) - J ( q+q_0) - \alpha \left[ \ln(\lambda+ q) + \frac{q_0}{\lambda + q} \right] \nonumber
    \\
    &- \frac{\beta}{M} \vw^* [\hat{q} \bm\Sigma_M] \vw^* + \frac{\beta}{M} \vw^* [\hat q \bm\Sigma_M] [ v^{-1} \mI + \hat{q} \bm\Sigma_M]^{-1} [\hat{q} \bm\Sigma_M ]  \nonumber
    \\
    &- \text{tr} \log \left[ \mI + \hat{q} v \bm\Sigma_M \right] - (\hat{q}_0 v + \hat{q} v_0 ) \ \text{tr}[\mI + \hat q v \bm\Sigma_M]^{-1} \bm\Sigma_M \nonumber
    \\
    &-\text{tr} \log\left[ \mI + \sigma^2 \hat{v} \mI +  \bm\Sigma_{\epsilon} \hat{q} \right] - \text{tr}\left[ \mI + \sigma^2 \hat{v} \mI +  \bm\Sigma_{\epsilon} \hat{q} \right]^{-1} \left[ \hat{v}_0 \sigma^2  \mI +  \hat{q}_0 \bm\Sigma_{\epsilon}  \right].
\end{align}
Letting $F = 2 M^{-1} \left< \ln Z \right>$, the saddle point equations read
\begin{align}
    \frac{\partial F}{\partial q_0} &= \hat{q} - \frac{\alpha}{\lambda + q} - J = 0, \nonumber
    \\
    \frac{\partial F}{\partial \hat{q}_0} &= q - v \ \text{tr}[\mI + \hat{q} v \bm\Sigma_M]^{-1}\bm\Sigma_M -  \text{tr}[\mI + \sigma^2 \hat v \mI + \bm\Sigma_{\epsilon} \hat{q} ]^{-1} \bm\Sigma_{\epsilon} =0, \nonumber
    \\
    \frac{\partial F}{\partial v_0 } &= \eta \hat{v} - \hat{q} \  \text{tr}[\mI + \hat{q} v \bm\Sigma_M]^{-1}\bm\Sigma_M = 0, \nonumber
    \\
    \frac{\partial F}{\partial \hat{v}_0 } &= \eta v - \sigma^2 \  \text{tr}[\mI + \sigma^2 \hat v \mI +  \bm\Sigma_{\epsilon} \hat{q} ]^{-1} = 0.
\end{align}
Now the generalization error can be determined from
\begin{align}
    E_g &= - \frac{\partial}{\partial J} \lim_{\beta \to \infty} \frac{2}{\beta M} \left<  \ln Z \right> = \partial_J \frac{1}{M} \vw^* \left[ \hat{q} \bm\Sigma_M - (\hat{q} \bm\Sigma_M)[v^{-1} \mI + \hat{q} \bm\Sigma_M]^{-1} (\hat{q} \bm\Sigma_M)   \right] \vw^*.
\end{align}
We see that it is necessary to compute $\partial_J \hat{q} $ and $\partial_J v$ in order to obtain the final result. For simplicity, we set $\sigma^2 = 1$. The equations for the source derivatives are
\begin{align}
    &\partial_J \hat q = - \frac{\alpha}{(\lambda + q)^2} \partial_J q +  1, \nonumber
    \\
    &\partial_J q = -  \text{tr}[ \mI + v \hat{q} \bm\Sigma_M]^{-2}[ -   \partial_J v \mI + v^2 \partial_J \hat q \bm\Sigma_M ] \bm\Sigma_M - \text{tr}[\mI + \hat{v}\mI + \hat{q} \bm\Sigma_{\epsilon}]^{-2}\bm\Sigma_{\epsilon}[ \partial_J \hat{v} \mI + \partial_J \hat q \bm\Sigma_{\epsilon} ], \nonumber
    \\
    &\eta\partial_J \hat v = - \text{tr}[ \mI + v \hat{q} \bm\Sigma_M ]^{-2}\bm\Sigma_M [ -  \partial_J \hat{q} \mI + \hat{q}^2 \partial_J v \bm\Sigma_M ], \nonumber
    \\
    &\eta \partial_J v = -  \text{tr}[\mI + \hat{v} \mI + \bm\Sigma_{\epsilon} \hat{q}]^{-2}[ \partial_J \hat{v} \mI + \partial_J \hat q \bm\Sigma_{\epsilon} ].
\end{align}

Once the value of the order parameters $(q,\hat q, v,\hat v)$ have been determined, these source derivatives can be obtained by solving a $4 \times 4$ linear system. Examples of these solutions are provided in Figure \ref{fig:gauss_covariate_verify}.



\subsubsection{Asymptotics in Underparameterized Regime}

We can compute the asymptotic ($\alpha \to \infty$) generalization error due to the random projection $\mA$ in the limit of $\bm\Sigma_{\epsilon} = 0$. First, note that if $\hat{v} \to O_\alpha(1)$, then the asymptotic error would be zero. Therefore, we will assume that $\hat{v} \sim a \alpha^c$ for some $a, c > 0$. The saddle point equations give the following asymptotic conditions 
\begin{align}
    \hat{q} \sim \frac{\alpha}{\lambda} \ , \ \eta \sim \hat{q} \ \text{tr}[\hat{v} \mI + \hat q\bm\Sigma_M]^{-1} \bm\Sigma_M \nonumber
    \\
    \implies \eta = \text{tr}[ \lambda a \alpha^{c-1} \bm I + \bm\Sigma_M ]^{-1} \bm\Sigma_M.
\end{align}
For $0 < \eta < 1$, this equation can only be satisfied as $\alpha \to \infty$ if $c = 1$ so that $\hat{v}$ has the same scaling with $\alpha$ as $\hat q$. If $c < 1$ then we could get the equation $\eta = 1$. If $c > 1$, then the equation would give $\eta = 0$. The constant $a$ solves the equation
\begin{align}
    \eta = \text{tr}[ \lambda a \mI + \bm\Sigma_M ]^{-1} \bm\Sigma_M.
\end{align}
Using this fact, our order parameters satisfy the following large $\alpha$ scalings
\begin{align}
    \hat{q} \sim \frac{\alpha}{\lambda} \ , \ q \sim 0 \ , \ \hat{v} \sim a \alpha \ , \ v \sim 0.
\end{align}

The source derivative equations simplify to $\partial_J \hat{q} \sim 1 \ , \ \partial_J q \sim 0$ and
\begin{align}
    \eta \partial_J \hat{v} &\sim  ( \hat{v} \partial_J \hat q +\hat q \partial_J \hat v) \ \text{tr}[\hat v \bm I + \hat q \bm\Sigma_M]^{-1} \bm\Sigma_M - \hat q \hat v \text{tr}[\hat v \mI + \hat q \bm\Sigma]^{-2}[ \partial_J \hat v \bm\Sigma_M + \partial_J \hat q \bm\Sigma_M^2 ] \nonumber
    \\
    &\sim \hat{q}^2 \text{tr}[\hat v \mI + \hat q \bm\Sigma_M]^{-2} \bm\Sigma_M^2  \partial_J \hat{v} + \hat{v}^2 \text{tr} [ \hat{v} \mI + \hat{q} \bm\Sigma_M ]^{-2} \bm\Sigma_M \nonumber
    \\
    \Rightarrow \partial_J \hat v &\sim  \frac{\text{tr}[ \mI + a^{-1} \lambda^{-1} \bm\Sigma_M]^{-2} \bm\Sigma_M }{\eta - \text{tr}[a \lambda \mI + \bm\Sigma_M]^{-2} \bm\Sigma_M^2}.
\end{align}
We note that $\partial_J \hat v$ only depends on the product $a \lambda$ which is an implicit function of $\eta$ and $\bm\Sigma_M$.  The generalization error is $E_g = \frac{1}{M} \partial_J \hat q (1+\hat v) \vw^* [(1+\hat v) \mI + \hat q \bm\Sigma]^{-1} \bm\Sigma_M \vw^*$
\begin{equation}
\begin{aligned}
    E_g &\sim \frac{1}{M} \vw^* \left[  \mI + a^{-1}\lambda^{-1} \bm\Sigma_M  \right]^{-2} \bm\Sigma_M \vw^* \\
    & \quad + \frac{1}{M} \vw^* [\lambda a \mI + \bm\Sigma_M]^{-2} \bm\Sigma_M^2 \vw^* \times \frac{\text{tr}[ \mI + a^{-1} \lambda^{-1} \bm\Sigma_M]^{-2} \bm\Sigma_M }{\eta - \text{tr}[a \lambda \mI + \bm\Sigma_M]^{-2} \bm\Sigma_M^2}.
\end{aligned}    
\end{equation}
We see that in the generic case, the asymptotic error has a nontrivial dependence on the task $\vw^*$ and the correlation structure $\bm\Sigma_M$. To gain more intuition, we will now consider the special case of isotropic features $\bm\Sigma_M = \mI$. In this case, we have $\eta = \frac{1}{1+\lambda a}$ so that $\lambda a = \frac{1-\eta}{\eta}$. This results in the following generalization error
\begin{equation}
\begin{aligned}
    E_g \sim & \frac{1}{M} |\vw^*|^2 \left[ (1-\eta)^2 + \eta^2 \frac{(1-\eta)^2 }{\eta - \eta^2}  \right] \sim \frac{1}{M} |\vw^*|^2 (1-\eta) .
\end{aligned}    
\end{equation}
We see that as $\eta = \frac{N_{\mathcal H}}{M} \to 1$, the asymptotic error converges to zero since all information in the original features is preserved. 

\subsubsection{Simplified Isotropic Feature Noise}
\begin{figure}
    \centering
    \subfigure[Gaussian $\mA$, vary $N_{\mathcal H}$]{\includegraphics[width=0.45\linewidth]{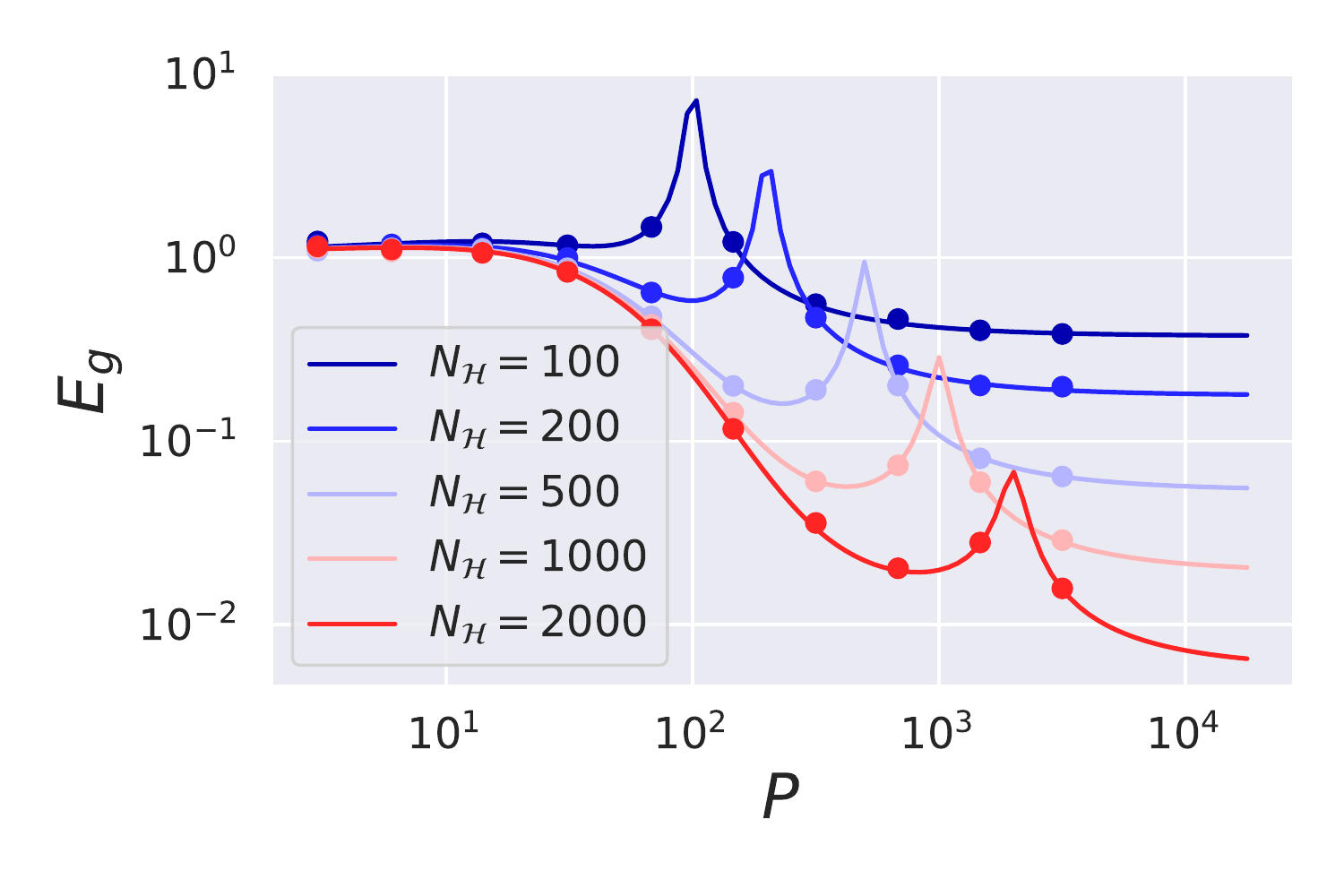}}
    \subfigure[$N_{\mathcal H} = 750$, vary $\sigma^2_{\epsilon}$]{\includegraphics[width=0.45\linewidth]{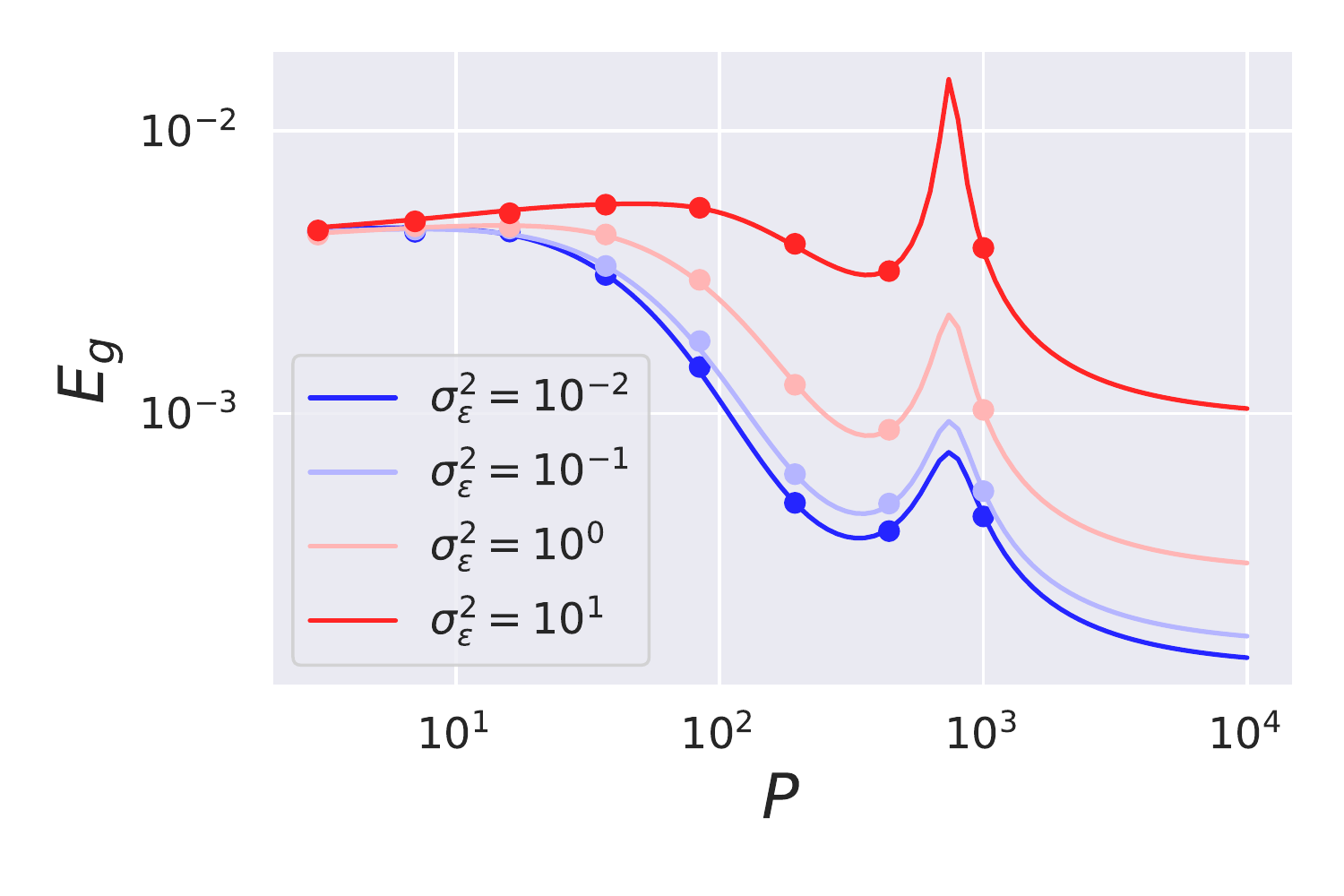}}
    \subfigure[$N_{\mathcal H} = 1000$, Vary $k$]{\includegraphics[width=0.45\linewidth]{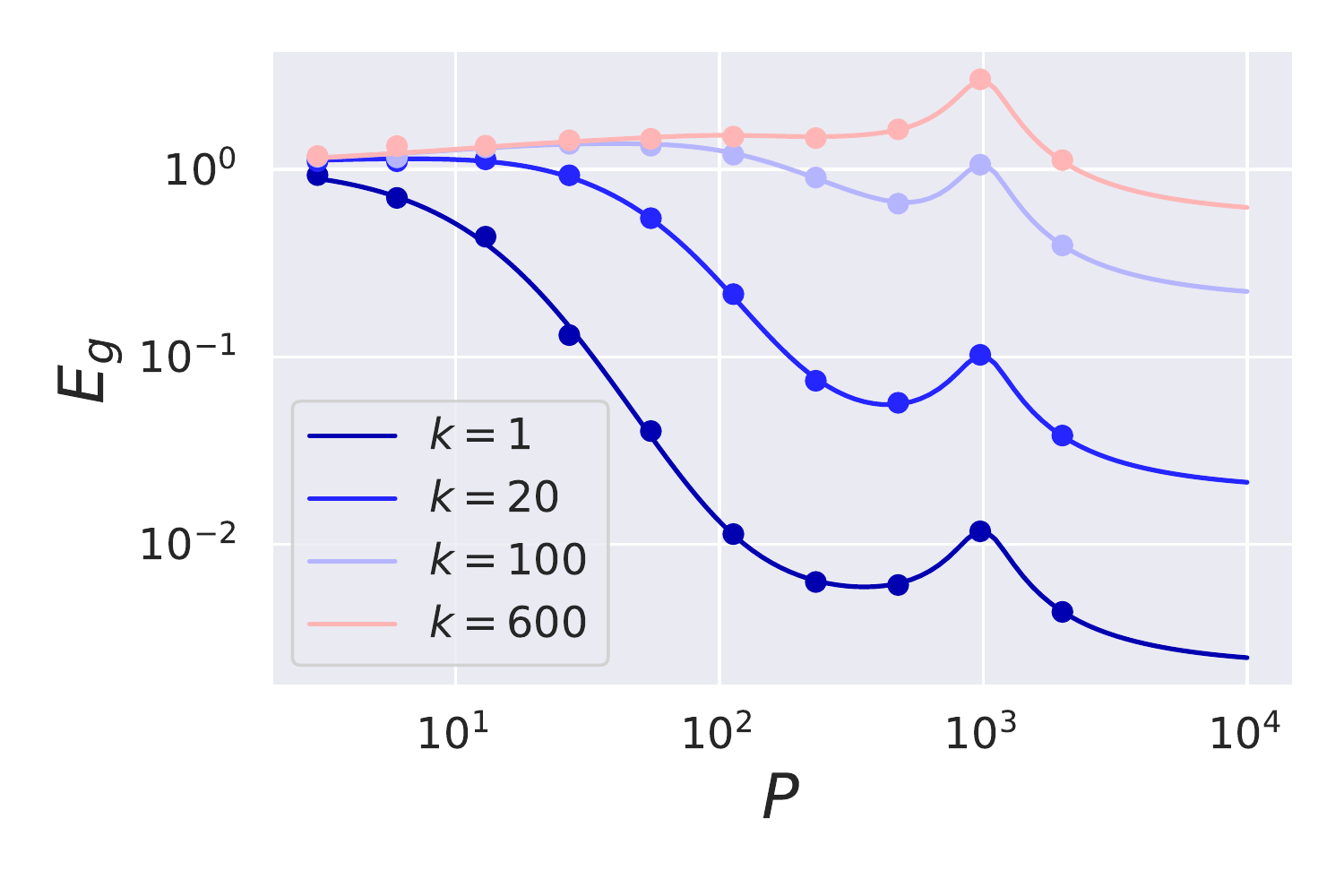}}
    \subfigure[$N_{\mathcal H}=1000$, Vary $\lambda$]{\includegraphics[width=0.45\linewidth]{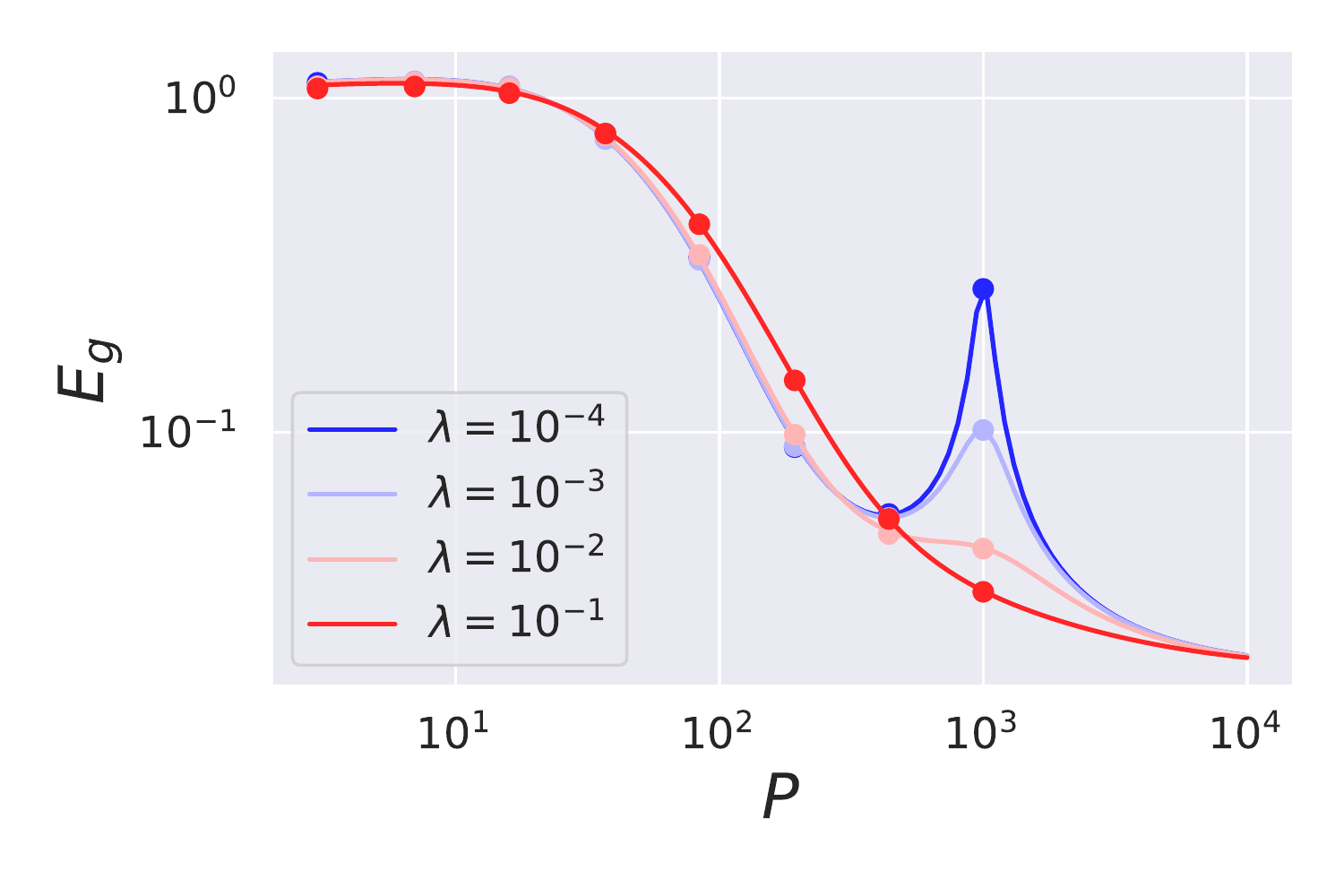}}
    \caption{Verification of Gaussian $\mA$ model. Solid lines are theory and dots are experiments. (a) The effect of changing the student's RKHS dimension $N_{\mathcal H}$. Double descent overfitting peaks occur at $P = N_{\mathcal H}$ (b) The effect of additive noise in the student features $\bm\Sigma_{\epsilon} = \sigma_{\epsilon}^2 \bm\Sigma_M$. (c) Learning curves for fitting the $k$-th eigenfunction. All mode errors exhibit a double descent peak at $P = {N}_{\mathcal H}$ regardless of the task. (d) Regularization can prevent the overfitting peak. 
    }
    \label{fig:gauss_covariate_verify}
\end{figure}

We can simplify the above expressions somewhat in the case where $\sigma^2=1$ and $\bm\Sigma_{\epsilon} =   \sigma^2_{\epsilon} \mI$. In this case, the order parameters become
\begin{align}
    \eta v &= \eta (1+\hat{v} + \sigma^2_{\epsilon} \hat{q})^{-1} \implies v = \frac{1}{1+\hat{v} +  \sigma^2_{\epsilon} \hat{q}} \nonumber
    \\
    \implies \eta\hat{v} &= \hat{q} \  \text{tr}\left[\mI + \frac{\hat q}{1+\hat v +  \sigma^2_{\epsilon} \hat q} \bm\Sigma_M \right]^{-1} \bm\Sigma_M = \hat q(1+\hat v + \sigma^2_{\epsilon} \hat q) \ \text{tr}[ (1+\hat v + \sigma^2_{\epsilon} \hat q)\mI + \hat{q} \bm\Sigma_M ]^{-1} \bm\Sigma_M \nonumber
    \\
    q &= \text{tr}\left[(1+\hat v + \sigma^2_{\epsilon} \hat q) \mI + \hat q \bm\Sigma_M \right]^{-1} \bm\Sigma_M + \frac{\eta \sigma_{\epsilon}^2}{1+\hat v + \sigma^2_{\epsilon} \hat q}.
\end{align}
Letting $\mG = [(1+\hat v + \sigma^2_{\epsilon} \hat q) \mI + \hat q \bm\Sigma_M]^{-1}$, the source derivatives have the form
\begin{align}
    \partial_J \hat{q} &= 1 - \frac{\alpha}{(\lambda+q)^2} \partial_J q   \\
    &= 1 + \frac{\alpha}{(\lambda+q)^2 } \left[ \text{tr}\mG^2 \bm\Sigma[ \partial_J \hat{v} \mI + \partial_J \hat{q} \bm\Sigma_M ] + \frac{\eta \sigma^2_{\epsilon}}{(1+\hat v + \sigma^2_{\epsilon} \hat q )^2} (\partial_J \hat v + \sigma^2_{\epsilon} \partial_J \hat q ) \right], \nonumber
    \\
    \eta \partial_J \hat{v} &= ((1+\hat v + 2\sigma^2_{\epsilon} \hat q)\partial_J \hat{q} + \hat q \partial_J \hat v  ) \text{tr}\mG \bm\Sigma_M\\
    & \quad - \hat{q}(1+\hat v + \sigma^2_{\epsilon} \hat q) \text{tr}\mG^2\bm\Sigma_M [ (\partial_J \hat v + \sigma^2_{\epsilon}\partial_J \hat q) \mI + \partial_J \hat q \bm\Sigma_M ] \nonumber
    \\
    &= (\partial_J \hat{q}) (1+\hat v + \sigma^2_{\epsilon} \hat q)^2  \text{tr}\mG^2 \bm\Sigma + (\partial_J \hat{v} + \sigma^2_{\epsilon} \partial_J \hat q) \hat{q}^2 \text{tr}\mG^2 \bm\Sigma^2.
\end{align}
This is a $2 \times 2$ linear system
\begin{align}\nonumber
    \begin{bmatrix}
    1 - \frac{\alpha}{(\lambda+q)^2} [  \text{tr}\mG^2 \bm\Sigma^2 +  \frac{\eta \sigma^4_{\epsilon}}{(1+\hat v +\sigma^2_{\epsilon} \hat q)^2}]  & - \frac{\alpha}{(\lambda+q)^2}[ \text{tr}\mG^2 \bm\Sigma +  \frac{\eta \sigma^2_{\epsilon}}{(1+\hat v + \sigma^2_{\epsilon} \hat q)^2} ]
    \\
    -(1+\hat v + \sigma^2_{\epsilon}\hat q)^2   \text{tr}\mG^2 \bm\Sigma_M - \sigma^2_{\epsilon} \hat{q}^2  \text{tr}\mG^2 \bm\Sigma_M^2 & \eta - \hat{q}^2  \text{tr}\mG^2 \bm\Sigma^2_s
    \end{bmatrix} 
    \begin{bmatrix}
    \partial_J \hat q
    \\
    \partial_J \hat v
    \end{bmatrix} = \begin{bmatrix}
    1
    \\
    0
    \end{bmatrix}.
\end{align}
For each $\alpha$, we can solve for $\partial_J \hat{q}$ and $\partial_J \hat{v}$ to get the final generalization error with the formula
\begin{align}
    E_g &= \partial_J  \frac{1}{M} \vw^* \left[ (1+\hat v + \sigma^2_{\epsilon}\hat q) \hat q \bm\Sigma \mG  \right] \vw^* \nonumber
    \\
    &= \frac{1}{M} \vw^* [ \partial_J ( \hat{q} + \hat{q} \hat v + \sigma^2_{\epsilon}\hat q^2 ) \bm\Sigma \mG - (1+\hat v + \sigma^2_{\epsilon}\hat q) \hat{q} \bm\Sigma \mG^2 (\partial \hat v \mI + \sigma^2_{\epsilon} \partial \hat q \mI + \partial \hat{q} \bm\Sigma_M )   ] \vw^*.
\end{align}
An example of these solutions can be found in Figure \ref{fig:gauss_covariate_verify}, where we show good agreement between theory and experiment.

\section{ResNet on CIFAR Experiments}\label{sec:resnet}

\begin{figure}[h]
    \centering
    \subfigure[Generalization MSE]{\includegraphics[width=0.45\linewidth]{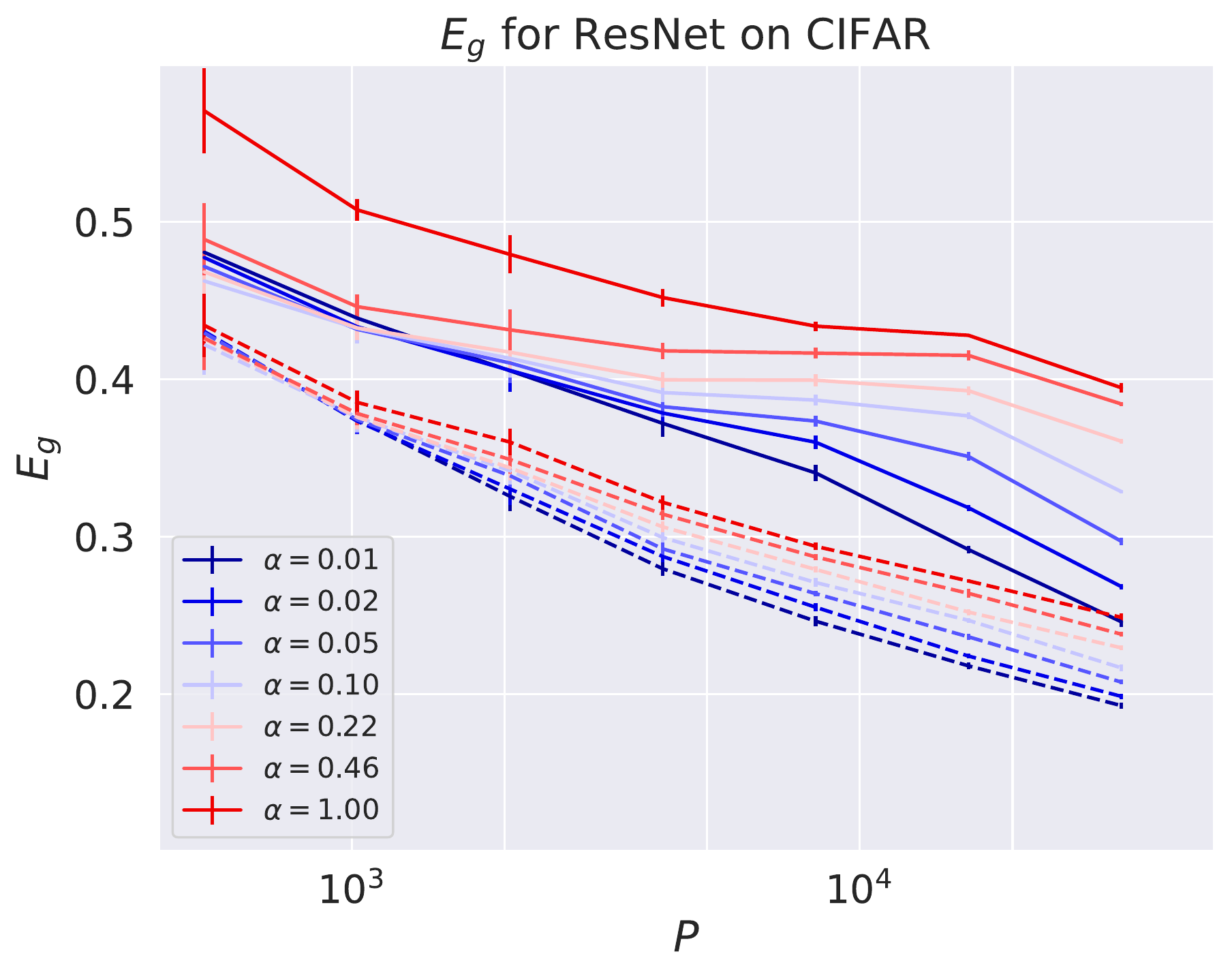}}
    \subfigure[Accuracy]
    {\includegraphics[width=0.45\linewidth]{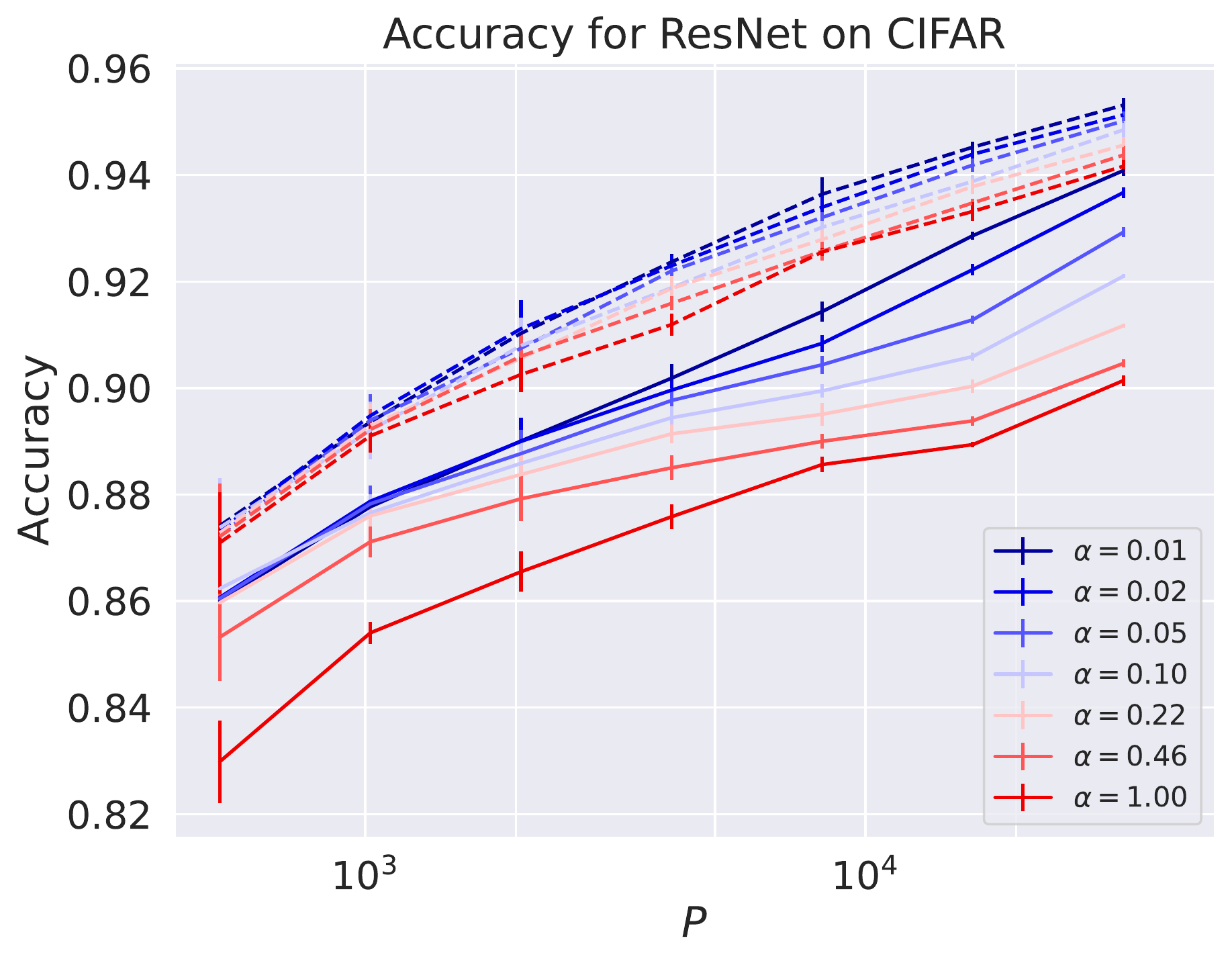}}\\
    \subfigure[var/$E_g$]{\includegraphics[width=0.45\linewidth]{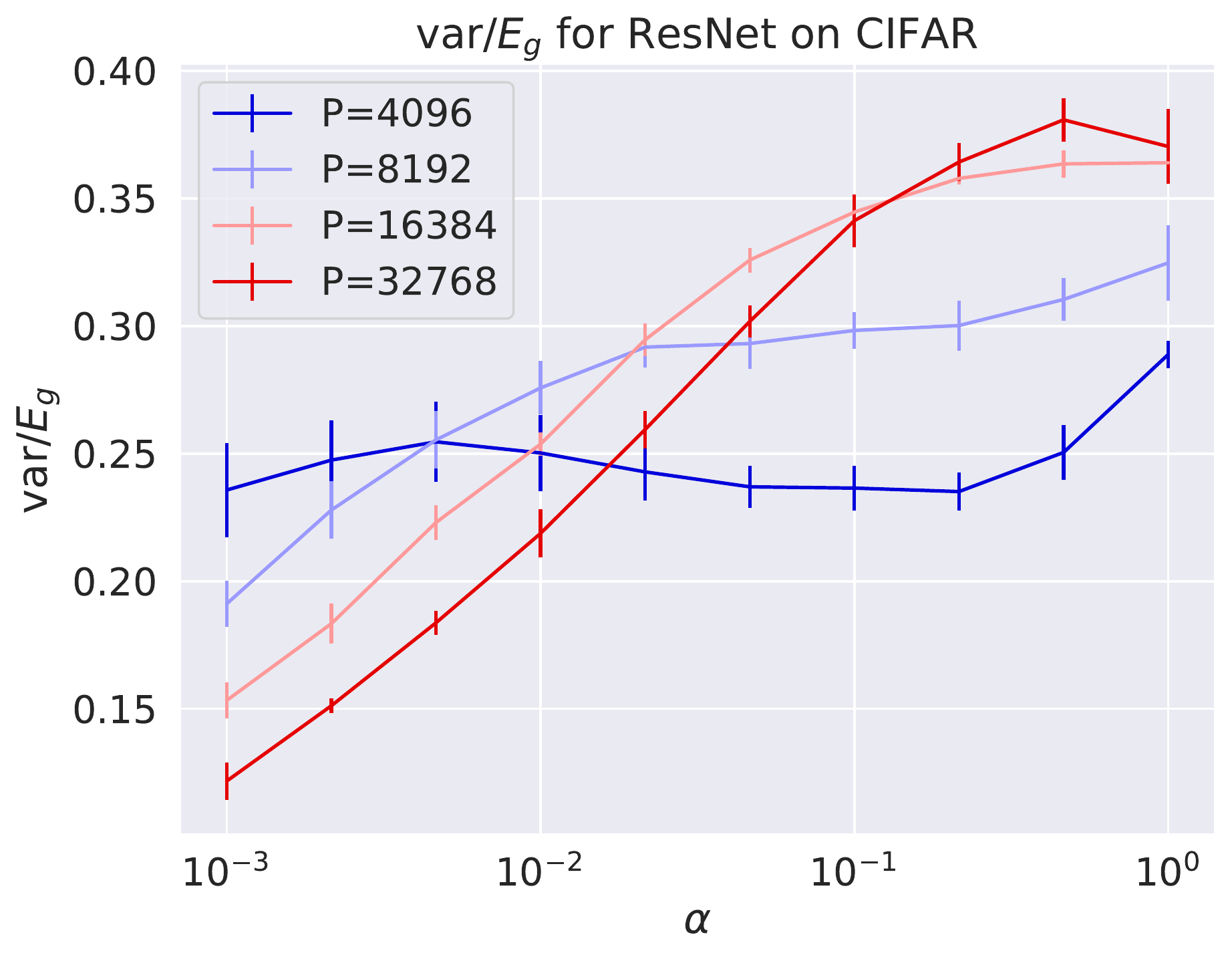}}
    \subfigure[var/$E_g$]{\includegraphics[width=0.45\linewidth]{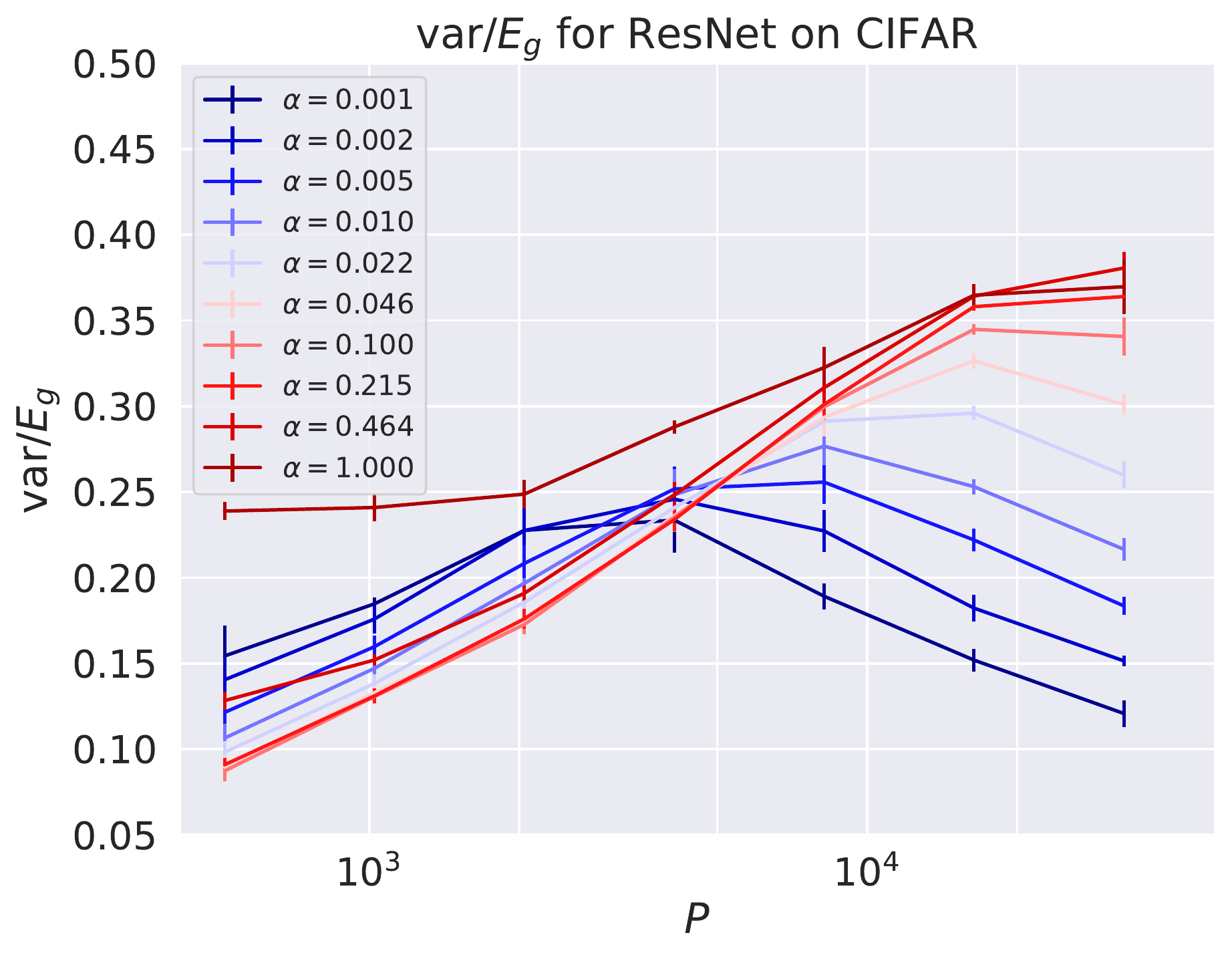}}
    \caption{A Wide ResNet \cite{zagoruyko2017wide} trained on a superclassed CIFAR task comparing animate vs inanimate objects. Each learning curve is averaged over $5$ different samples of the train set, yielding the means and error bars shown in the figures. a) Generalization error $E_g$. The dashed lines are the error of a 20-fold ensemble over different values of $\alpha$. Across all $P$, lazy networks attain worse generalization error. As with the MLP task, the best performing networks are ensembles of rich networks. b) The accuracy also has the same trend: richer networks perform better and ensembling lazy networks helps them more. c) Once $P$ is large enough, lazier networks tend to benefit more from ensembling. d) Very lazy networks transition to variance limited behavior earlier. For ResNets on this task, we see that rich, feature learning networks eventually begin reducing their variance on this task. Further details of the experiment are given in section \ref{sec:CIFAR_expt}.}
    \label{fig:CIFAR_expt}
\end{figure}

\newpage 

\section{Additional Experiments}\label{sec:more_plots}

\begin{figure}[h]
    \centering
    \subfigure[$E_g$ for \entk $k=2$]{\includegraphics[width=0.32\linewidth]{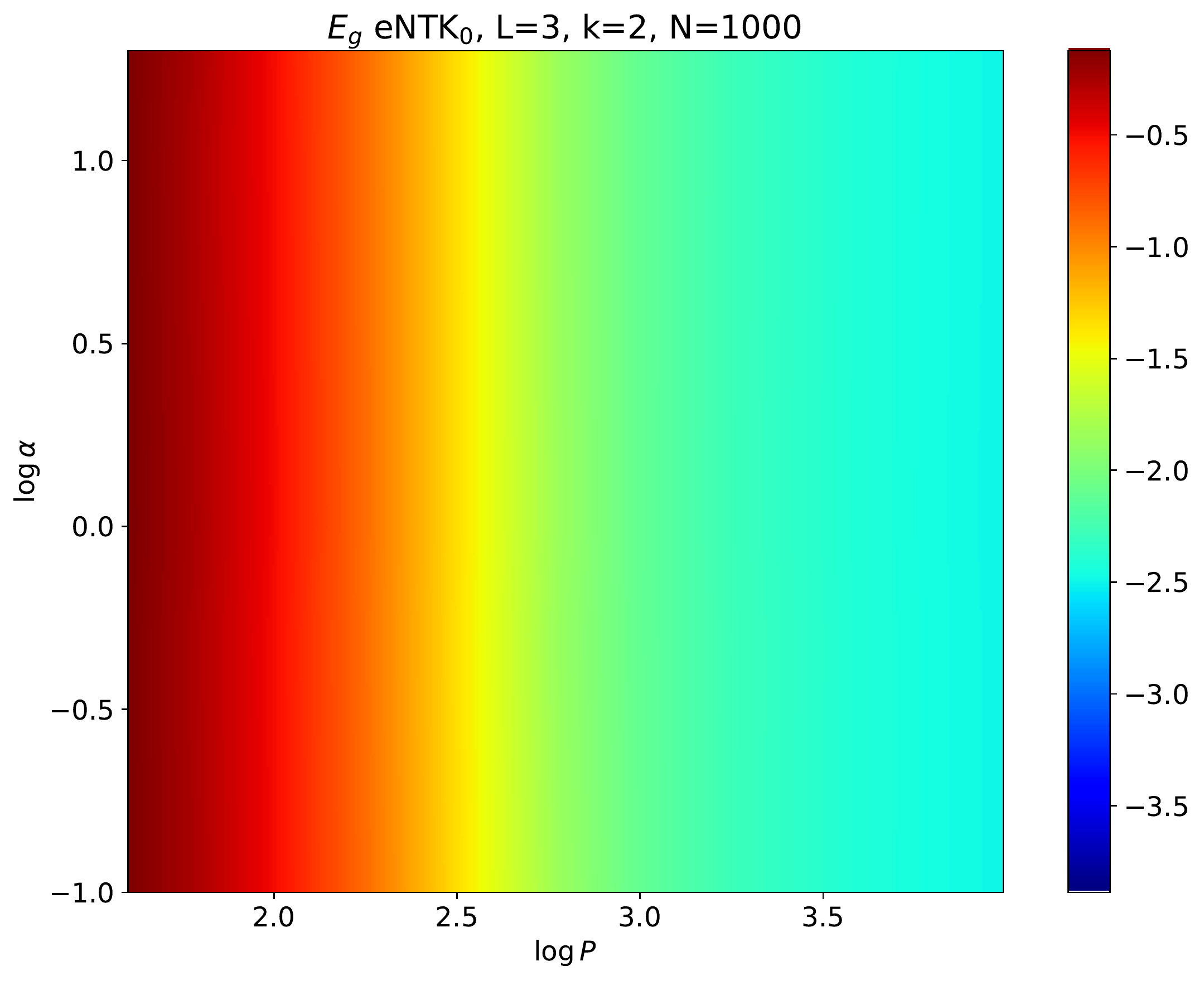}}
    \subfigure[$E_g$ for \entk $k=3$]{\includegraphics[width=0.32\linewidth]{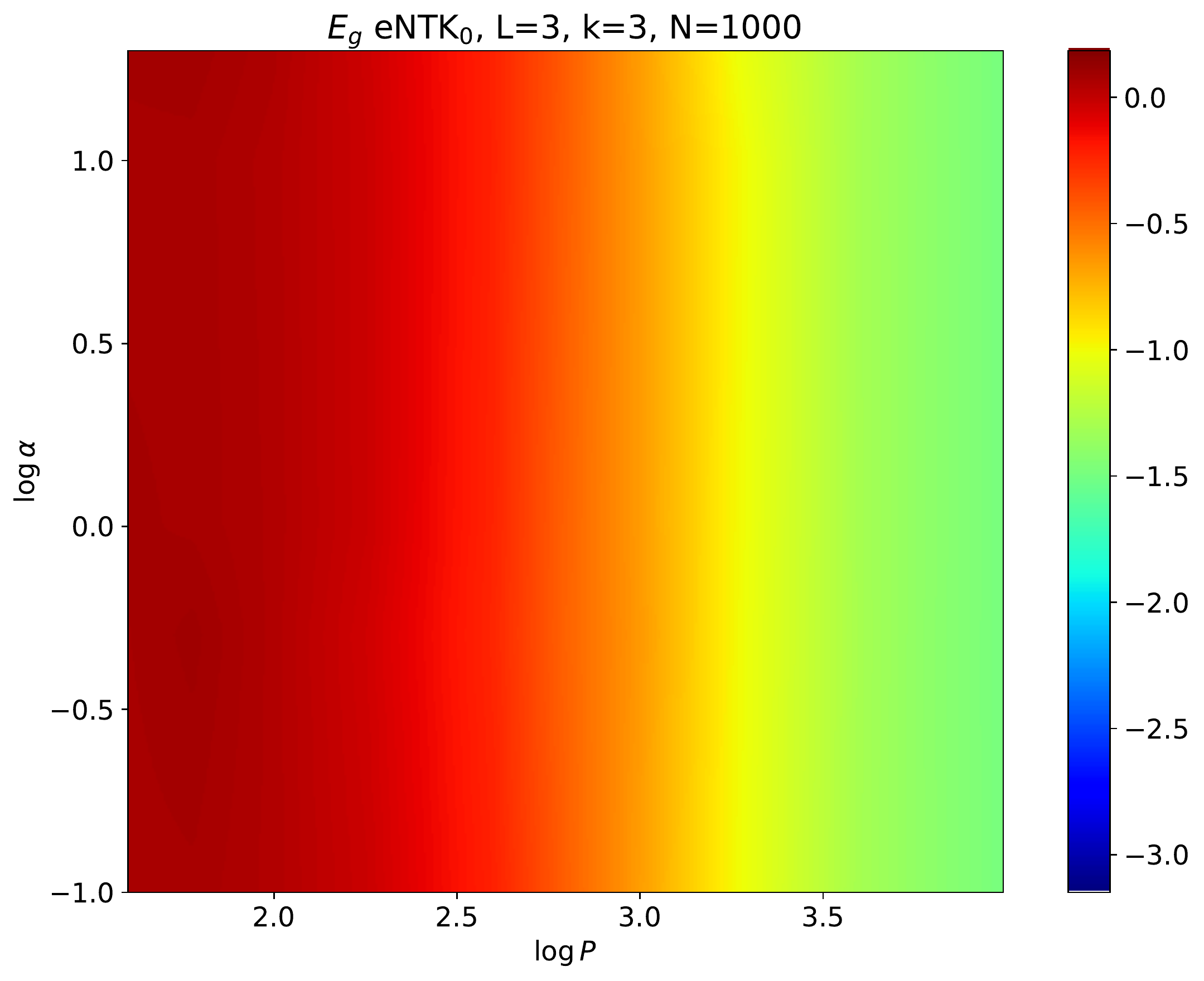}}
    \subfigure[$E_g$ for \entk $k=4$]{\includegraphics[width=0.32\linewidth]{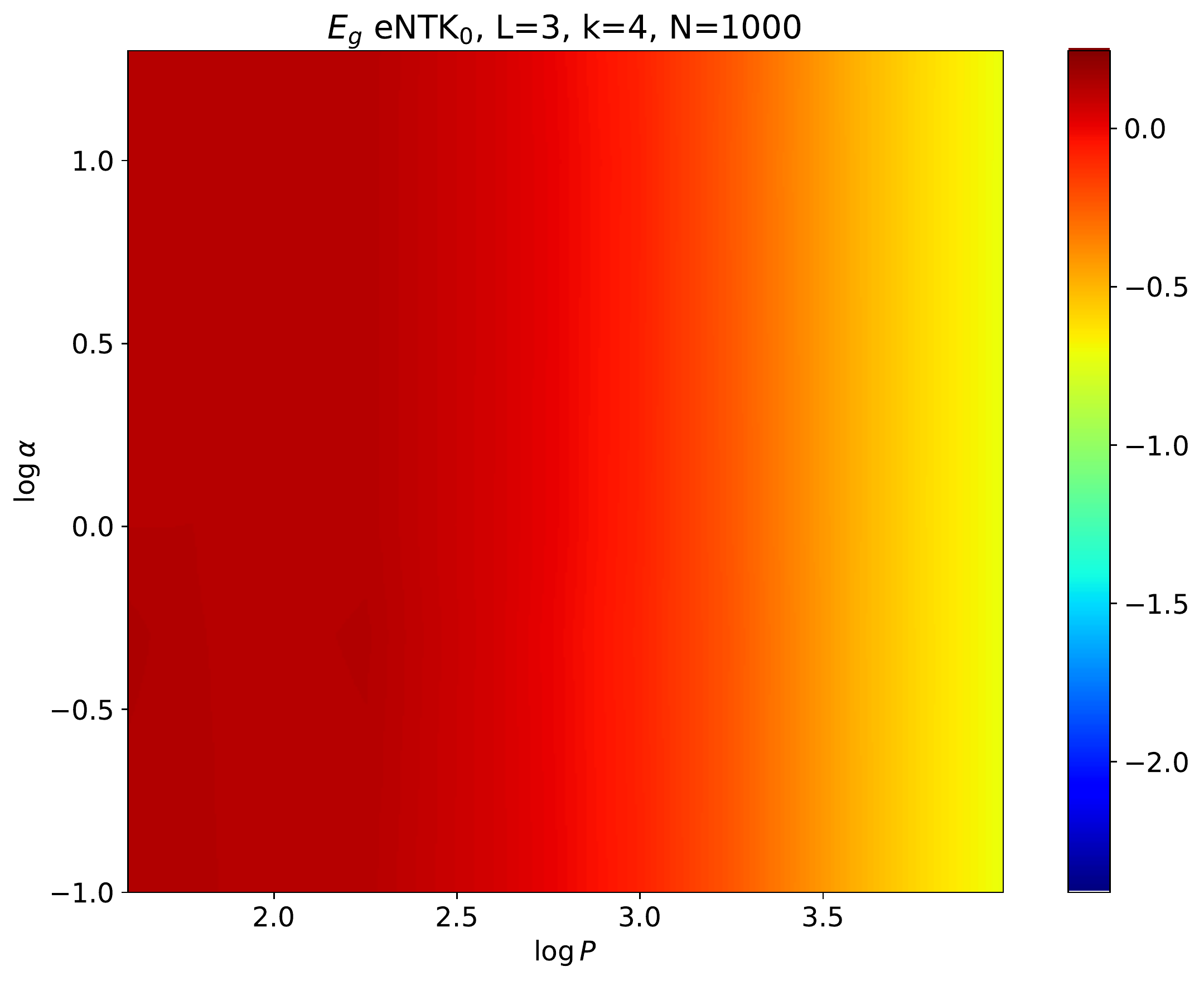}}
    \subfigure[$E_g$ for NN $k=2$]{\includegraphics[width=0.32\linewidth]{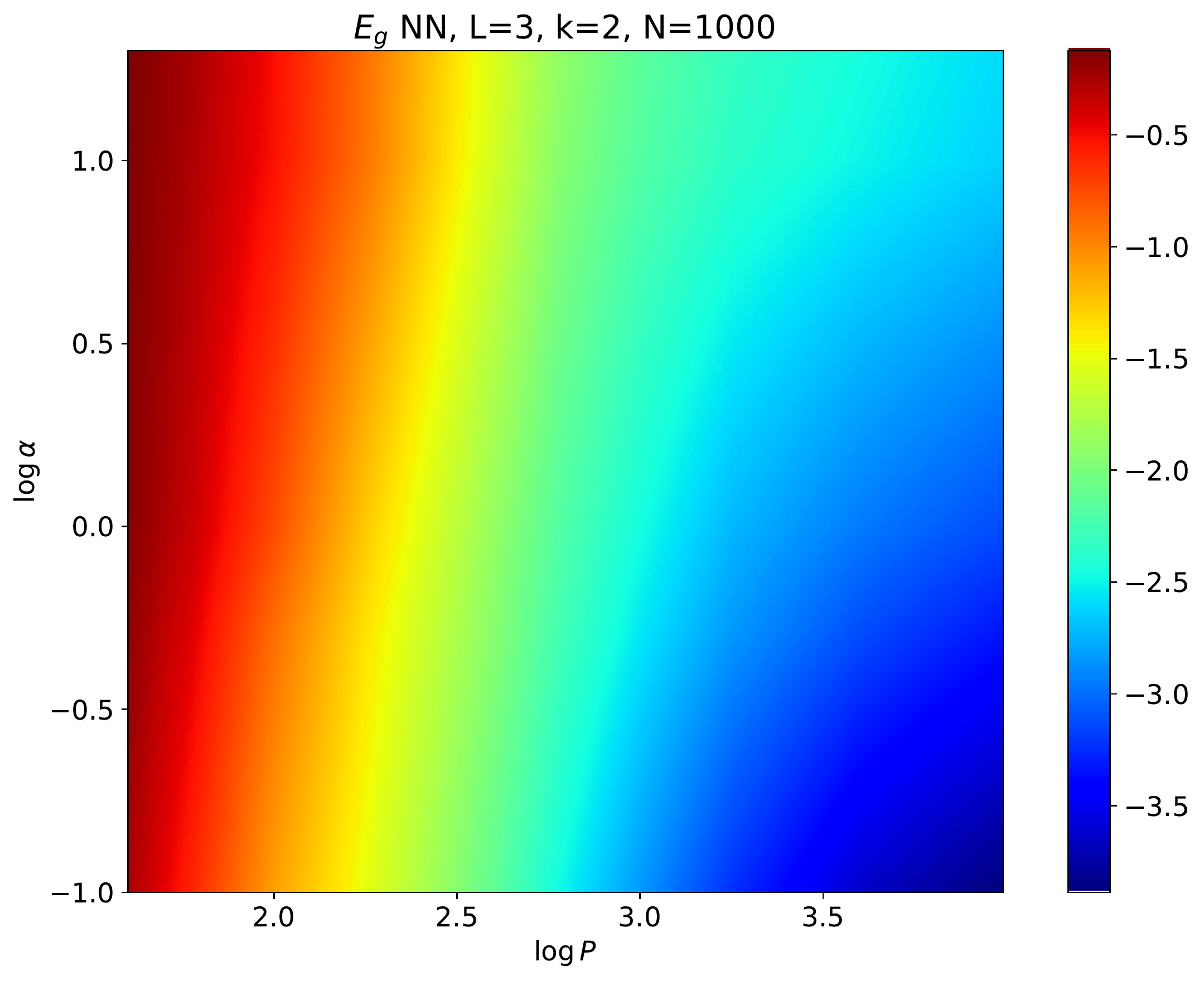}}
    \subfigure[$E_g$ for NN $k=3$]{\includegraphics[width=0.32\linewidth]{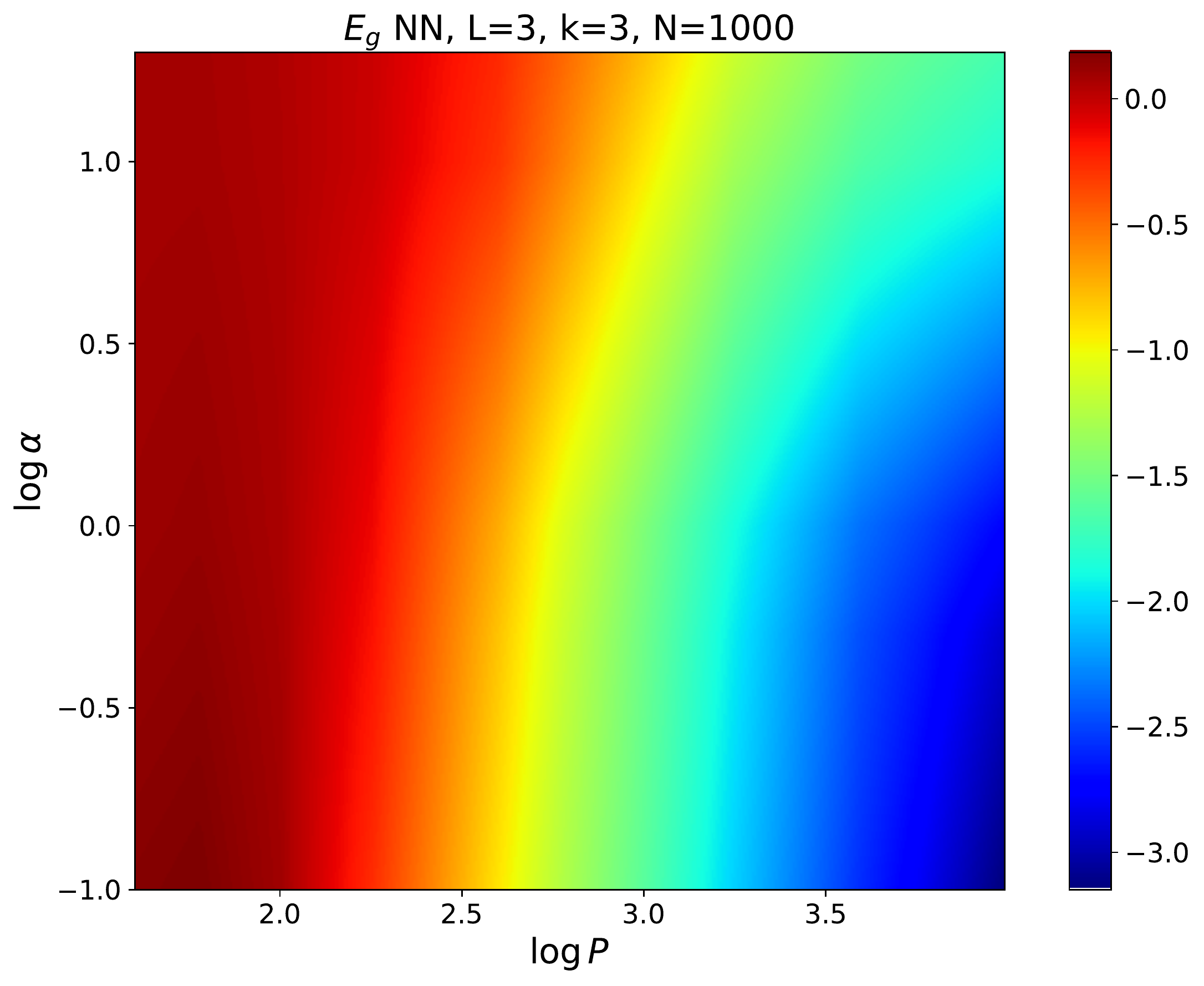}}
    \subfigure[$E_g$ for NN $k=4$]{\includegraphics[width=0.32\linewidth]{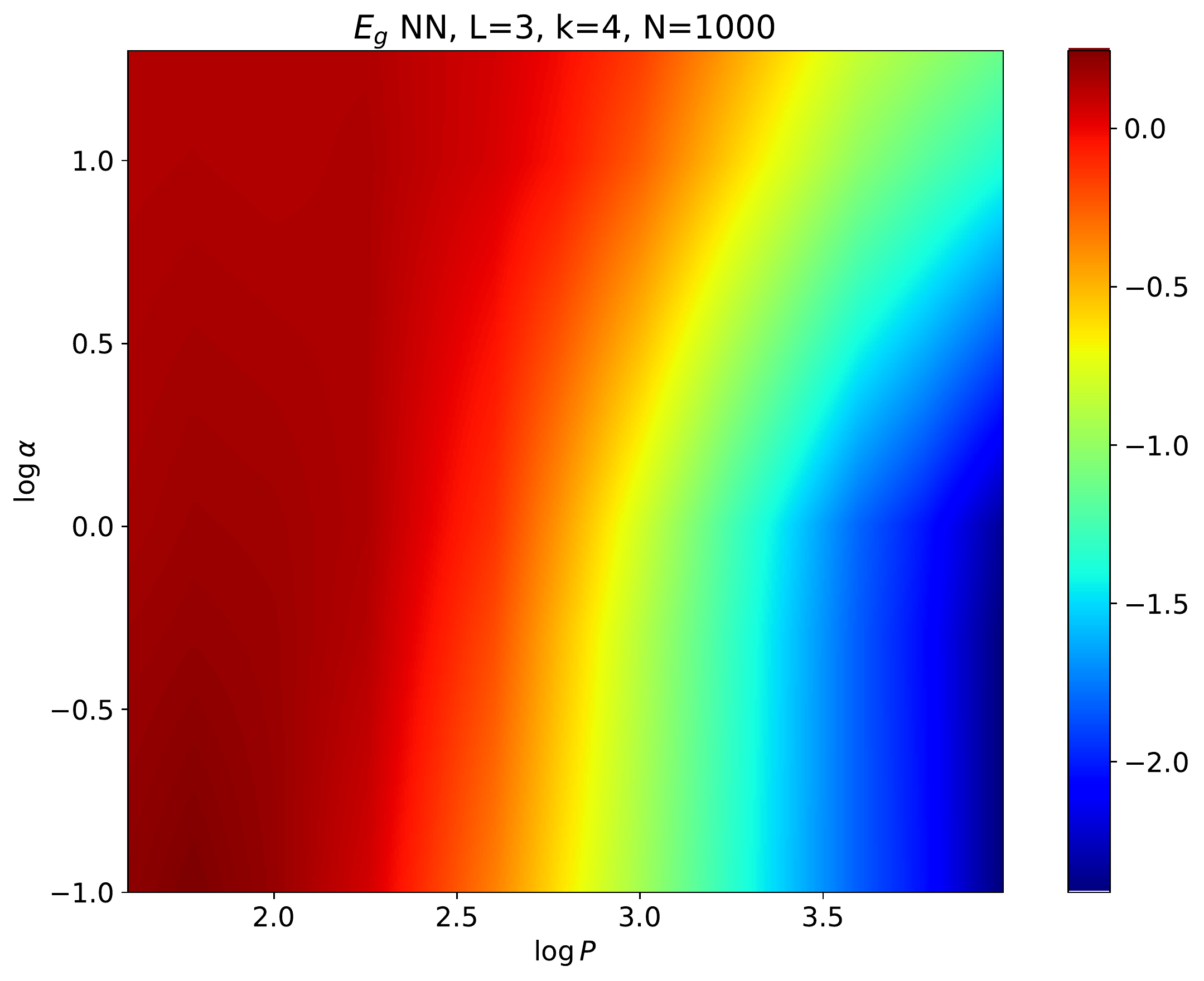}}
    \caption{Phase plots of $\log_{10} E_g$ for initial eNTKs (top) and neural networks (bottom). The large $\alpha$ behavior of the neural network generalization matches the generalization of the corresponding \entk \!. As a sanity check, the \entk generalization error is independent of re-scaling the network initialization because of the homogeneity of the ReLU network output. }
    \label{fig:gen_errs_over_k}
\end{figure}

\begin{figure}[h]
    \centering
    \subfigure[]{\includegraphics[width=0.32\linewidth]{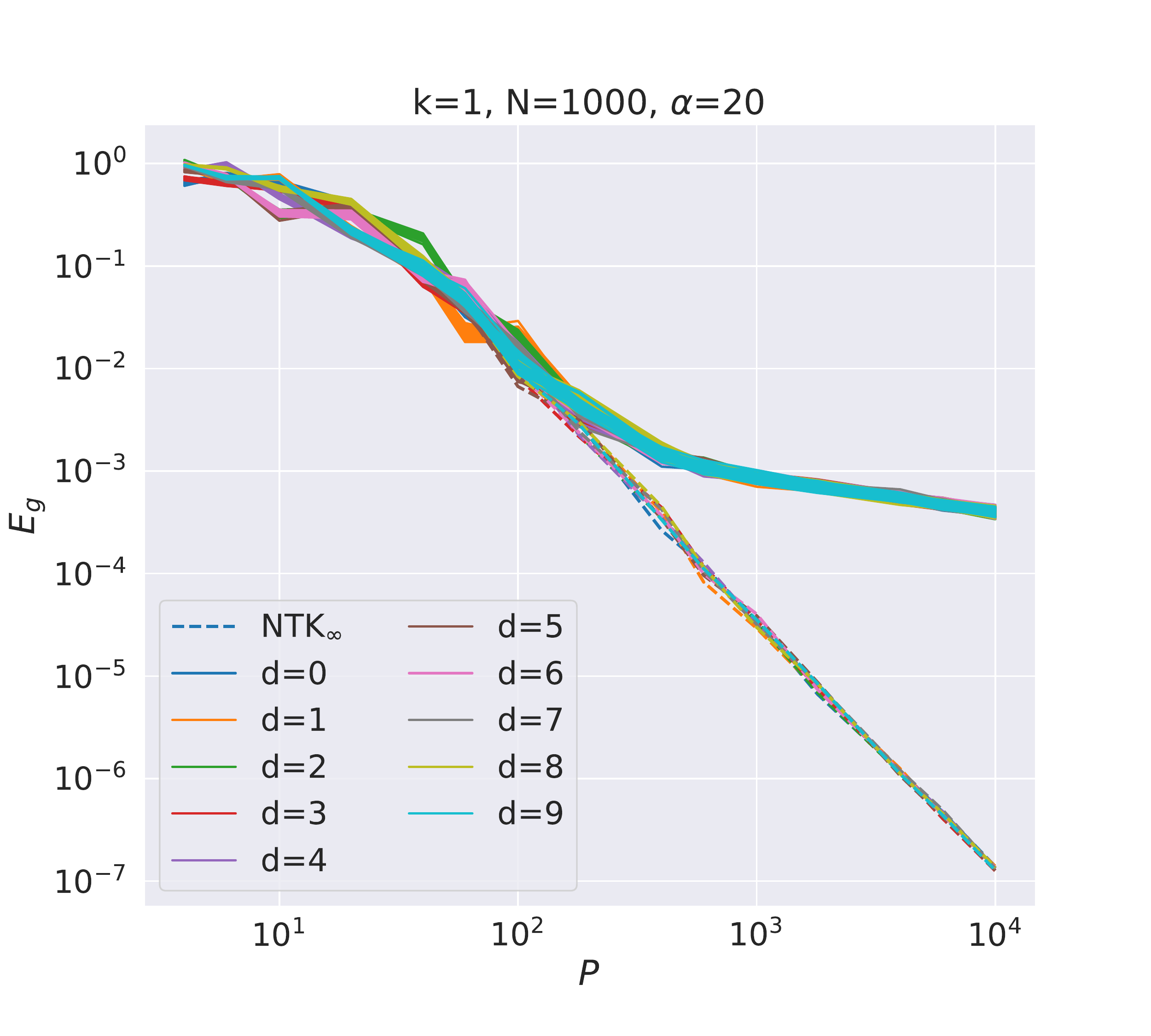}}
    \subfigure[]{\includegraphics[width=0.32\linewidth]{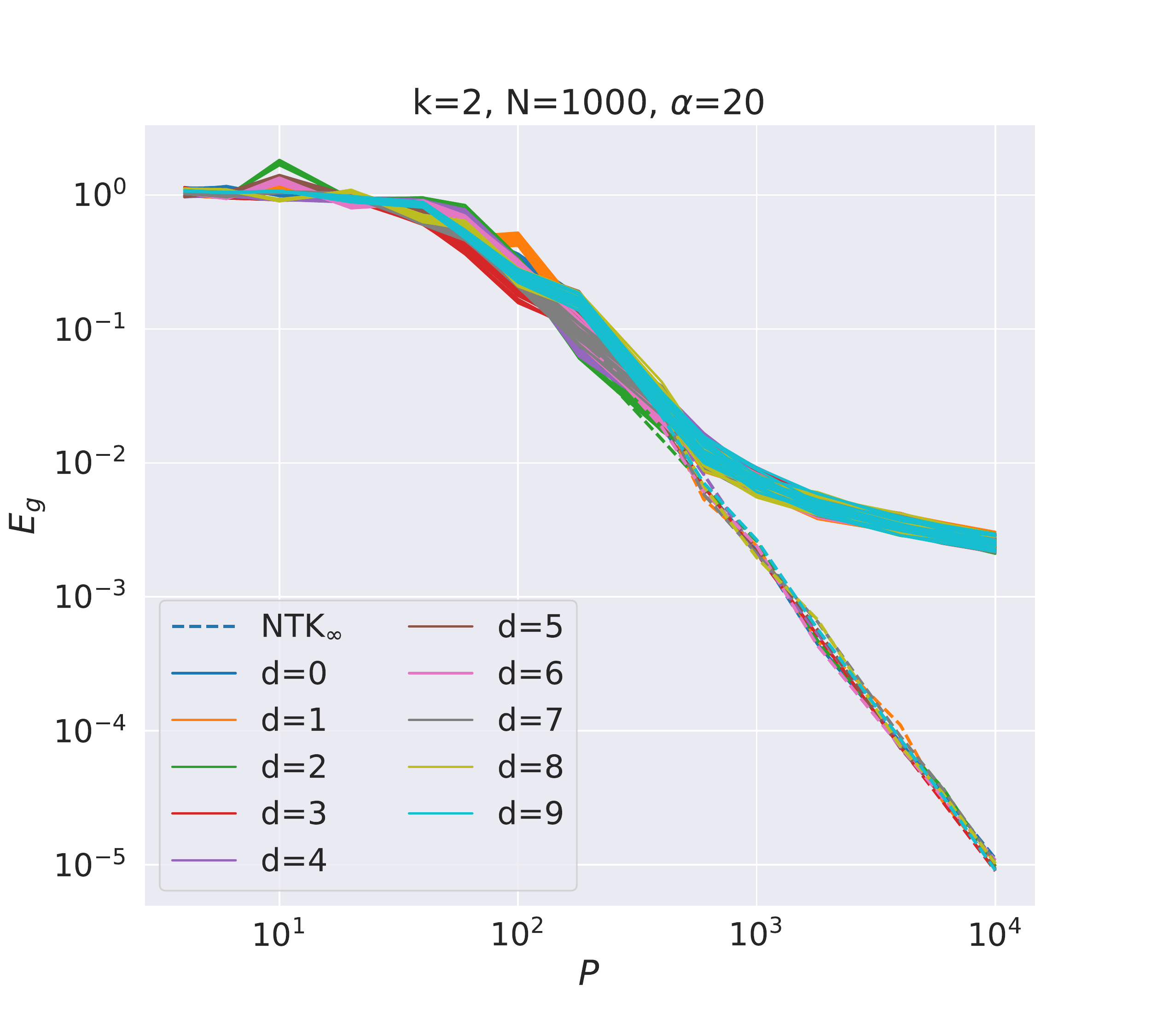}}
    \subfigure[]{\includegraphics[width=0.32\linewidth]{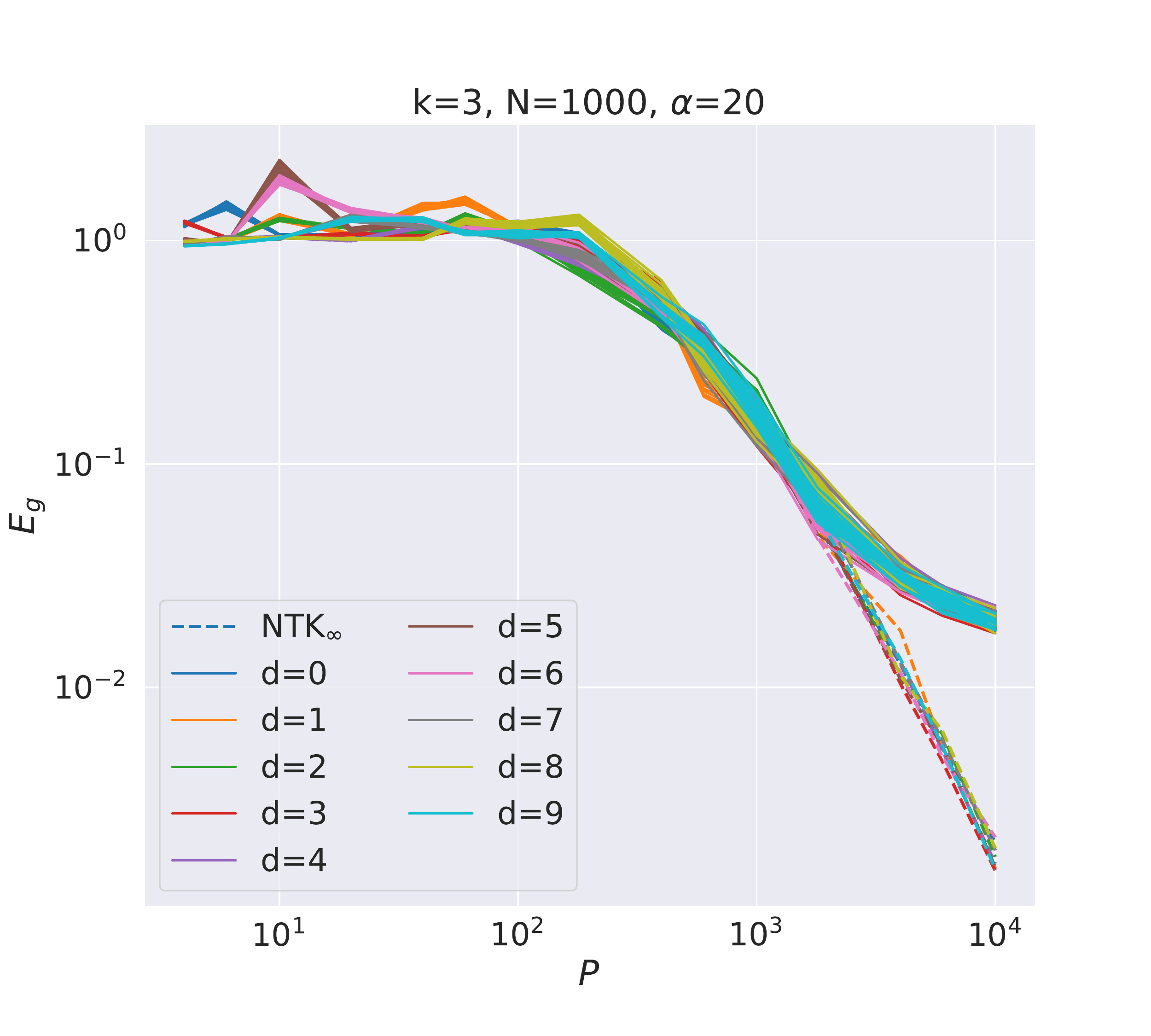}}
    \subfigure[]{\includegraphics[width=0.32\linewidth]{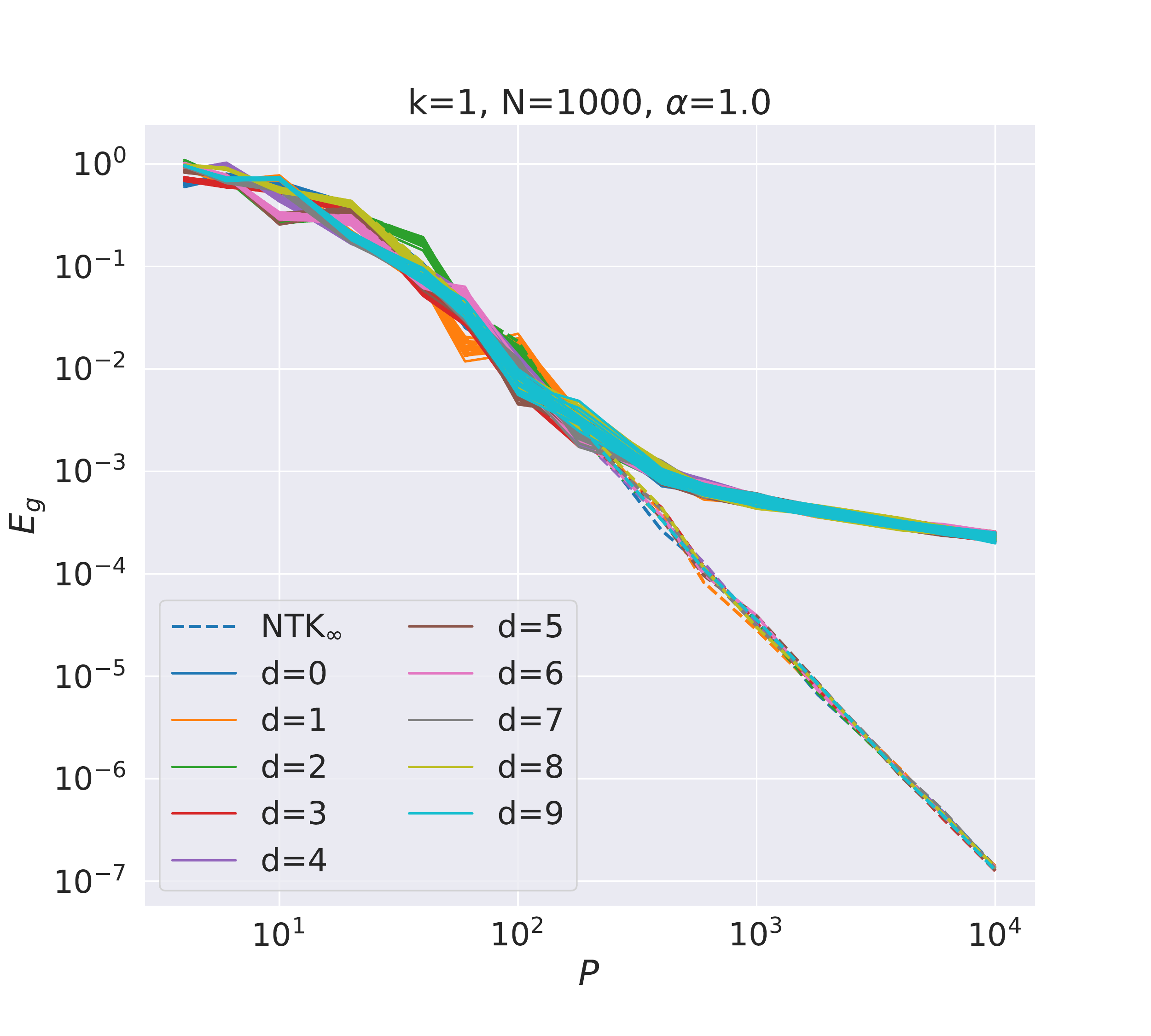}}
    \subfigure[]{\includegraphics[width=0.32\linewidth]{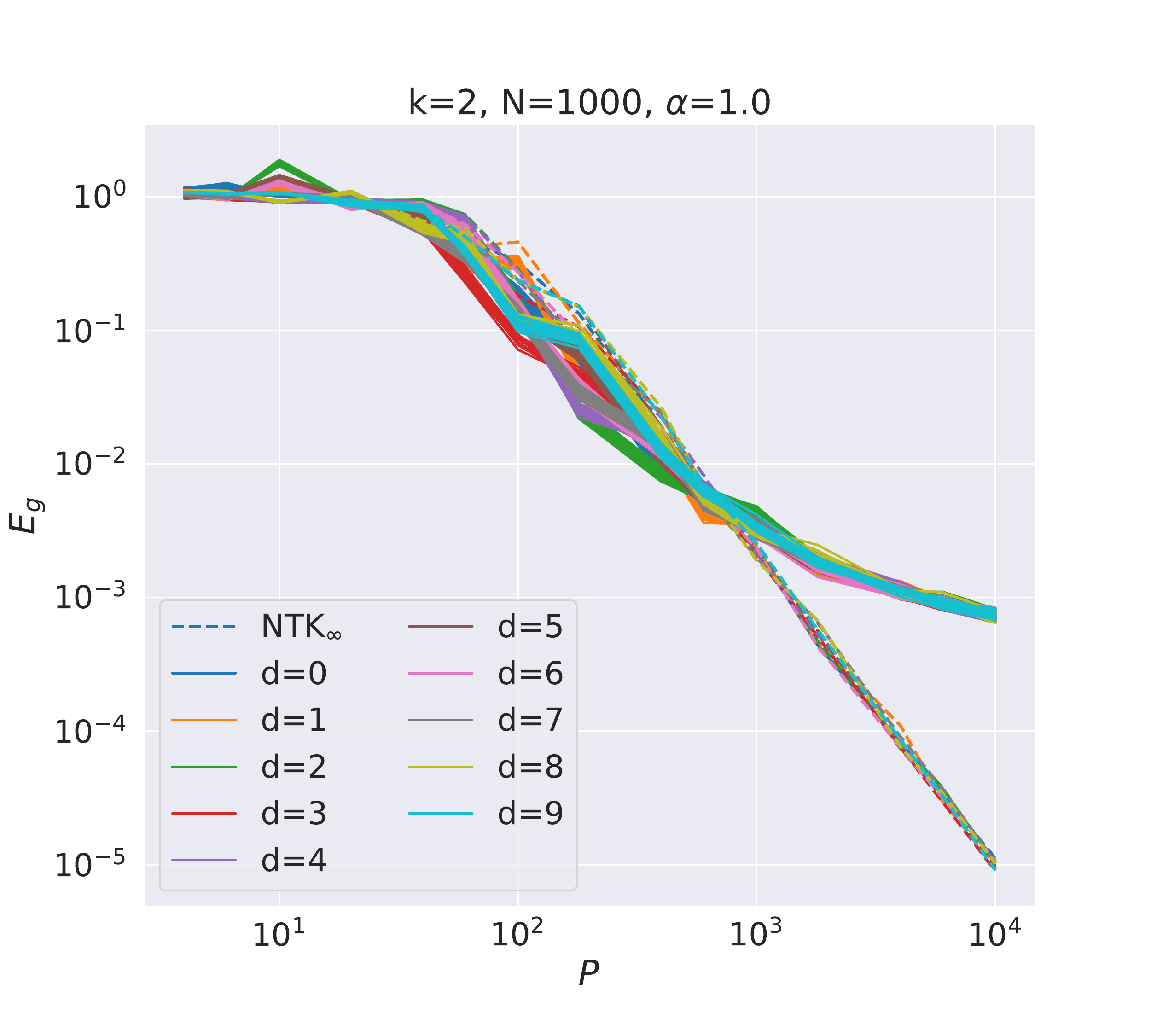}}
    \subfigure[]{\includegraphics[width=0.32\linewidth]{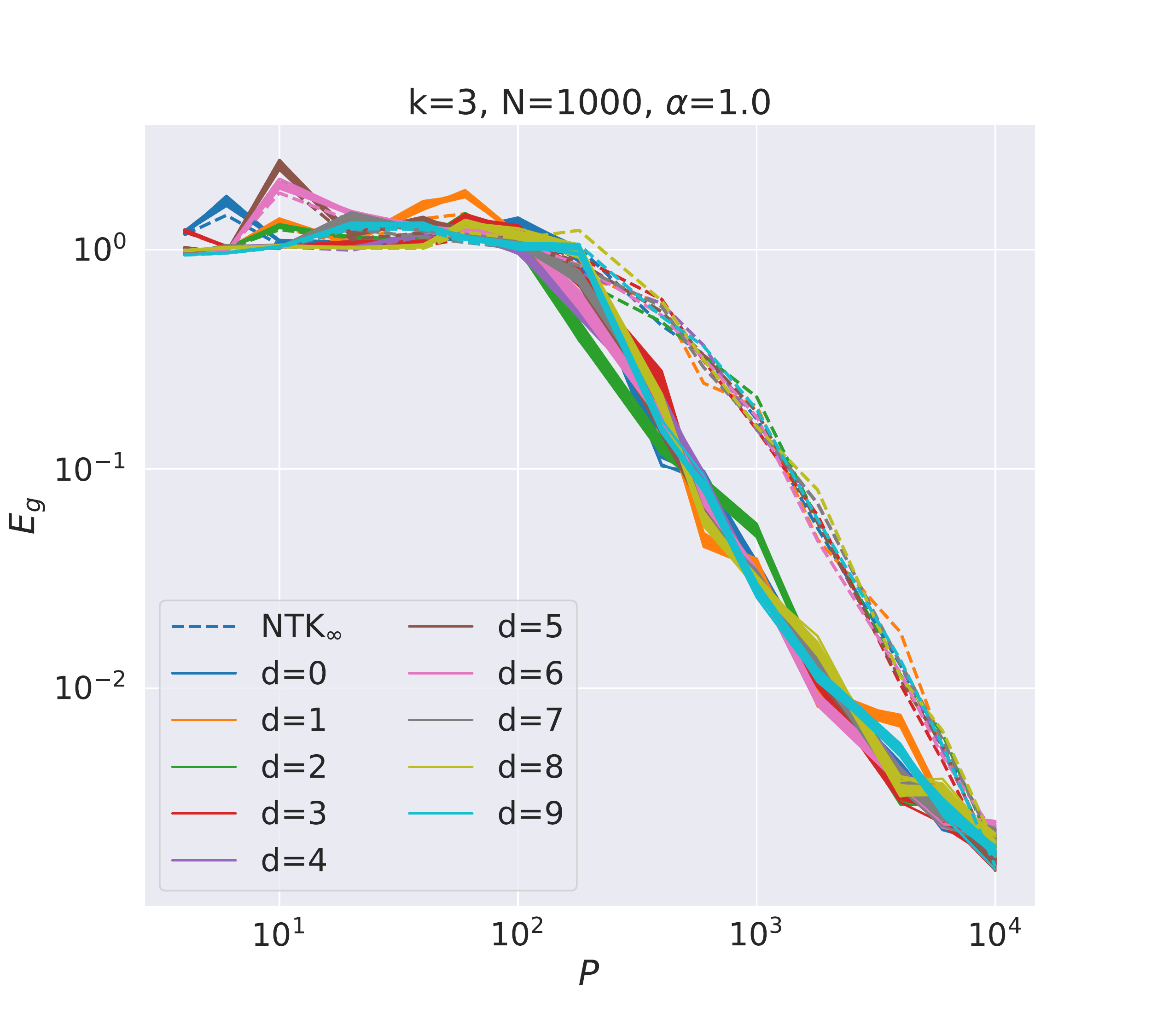}}
    \subfigure[]{\includegraphics[width=0.32\linewidth]{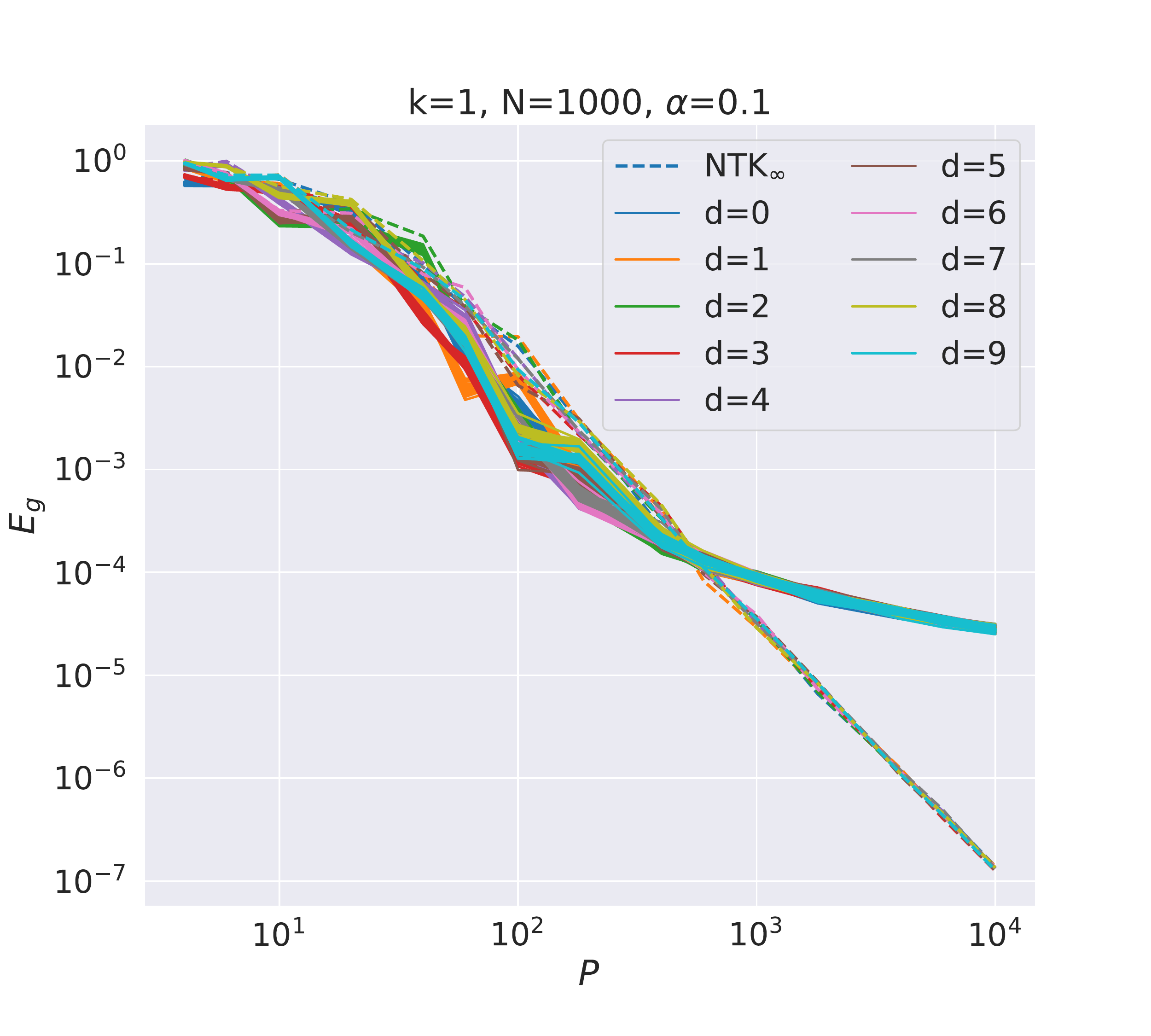}}
    \subfigure[]{\includegraphics[width=0.32\linewidth]{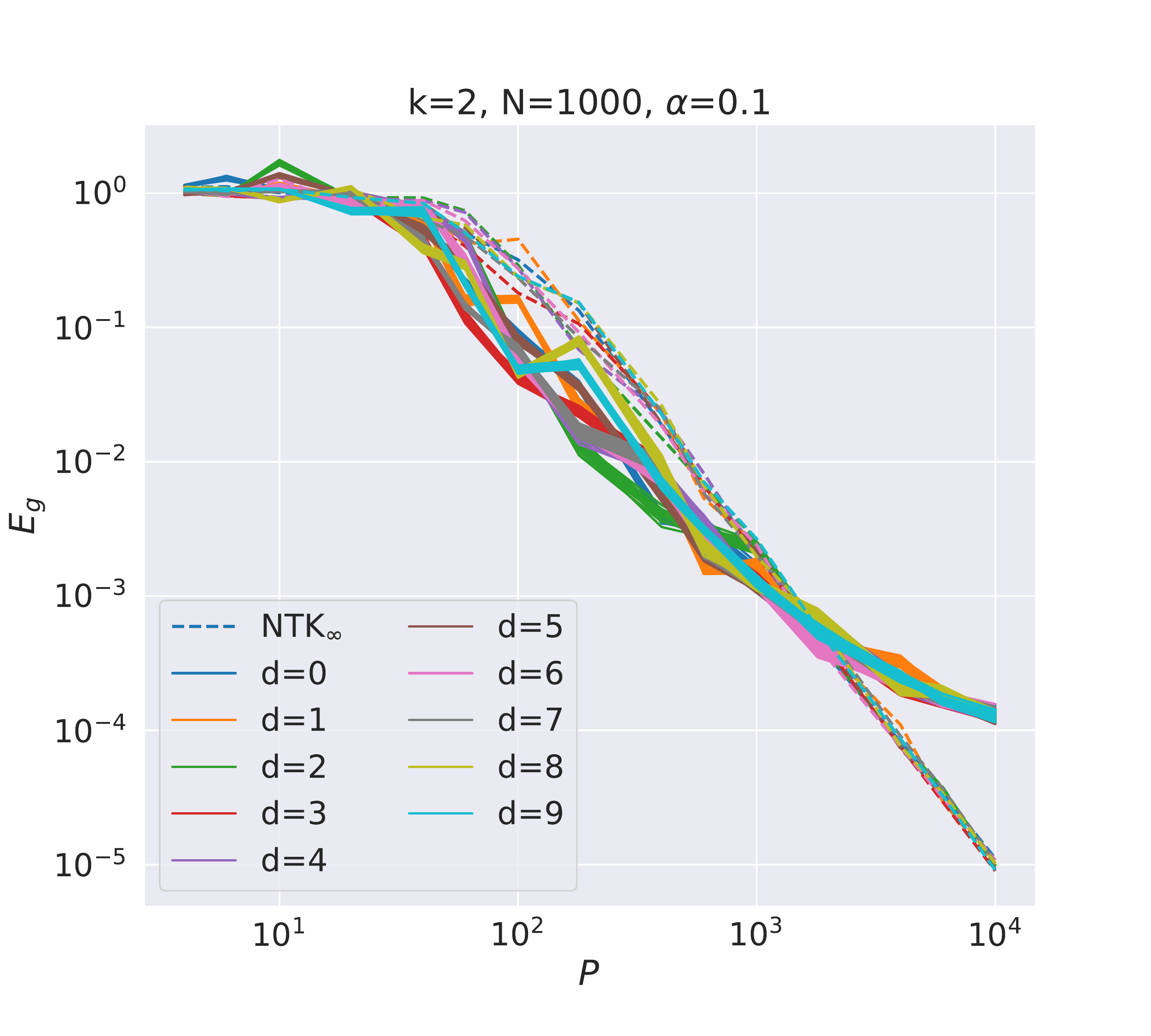}}
    \subfigure[]{\includegraphics[width=0.32\linewidth]{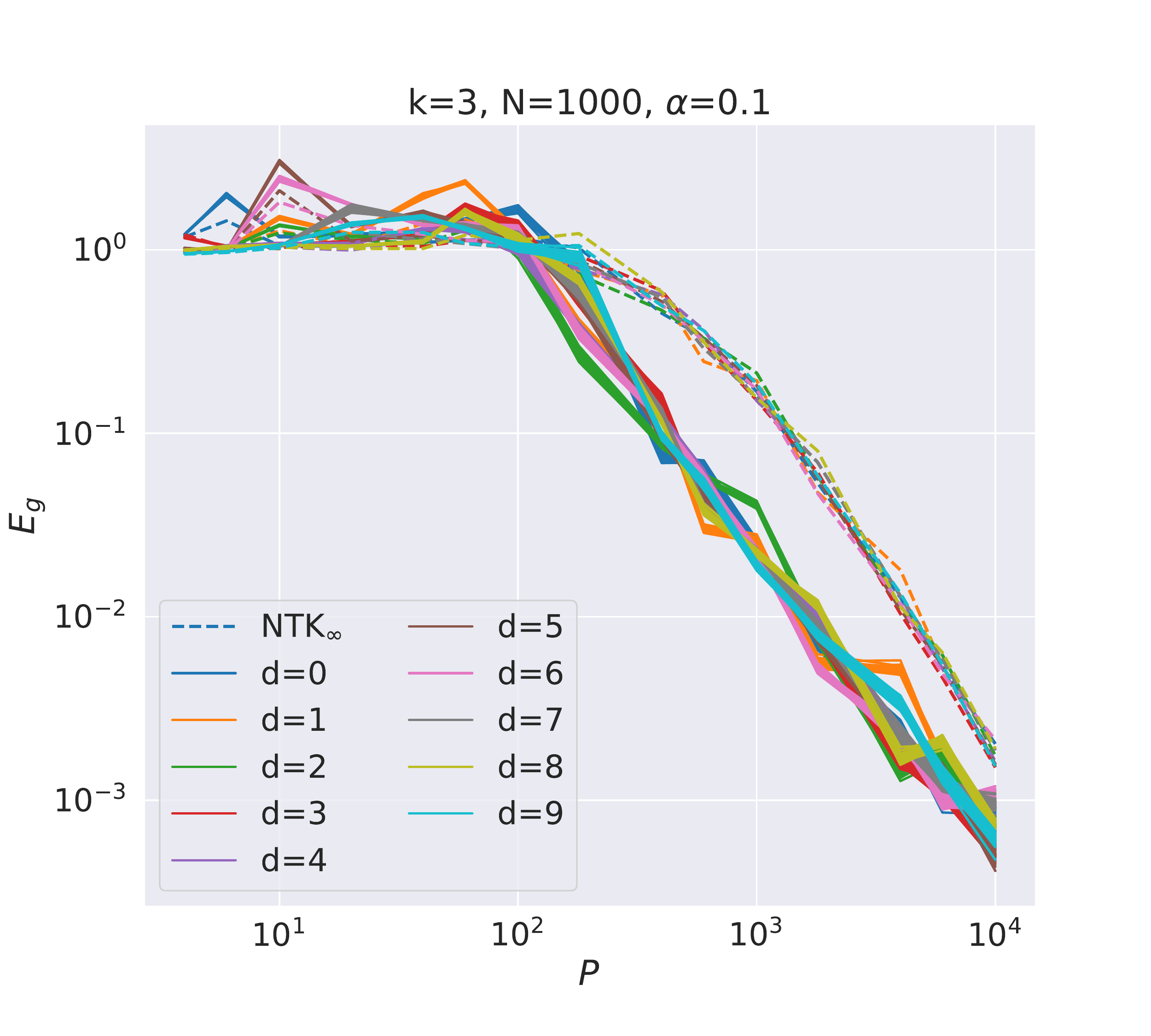}}
    \caption{A fine-grained view of the generalization error across different datasets and ensembles. Solid curves are depth $3$ neural networks, dashed curves are the infinite width NTK (which only has variance over datasets). Each color is a set of networks trained on the same dataset but different initializations. Different colors correspond to different datasets indexed by $d \in \{0, \dots, 9\}$.}
    \label{fig:fine_grained_gen_curves}
\end{figure}

\begin{figure}[h]
    \centering
    \subfigure[$E_g$ for $k=1, P=600$]{\includegraphics[width=0.32\linewidth]{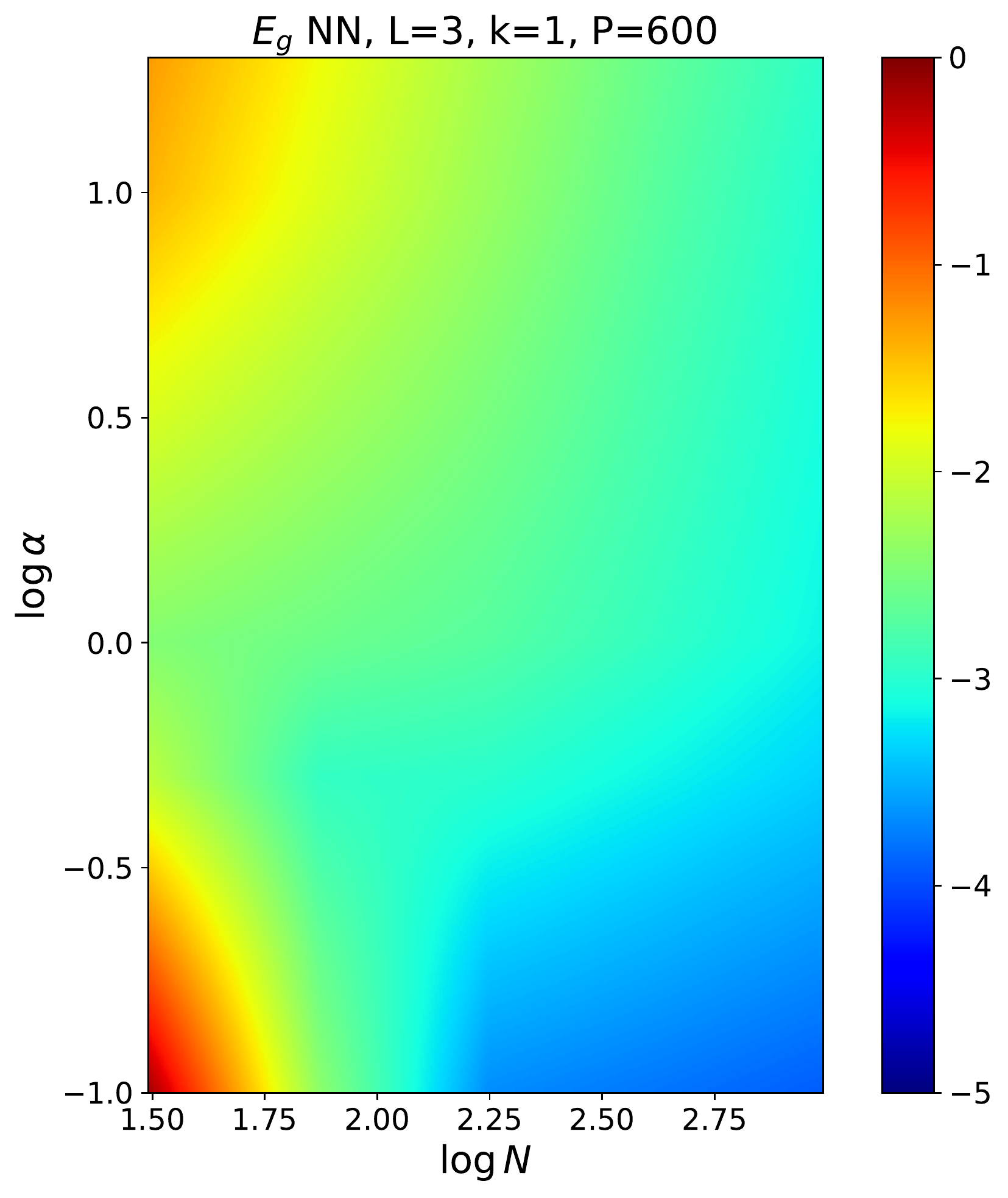}}
    \subfigure[$E_g$ for $k=1, P=10000$]{\includegraphics[width=0.32\linewidth]{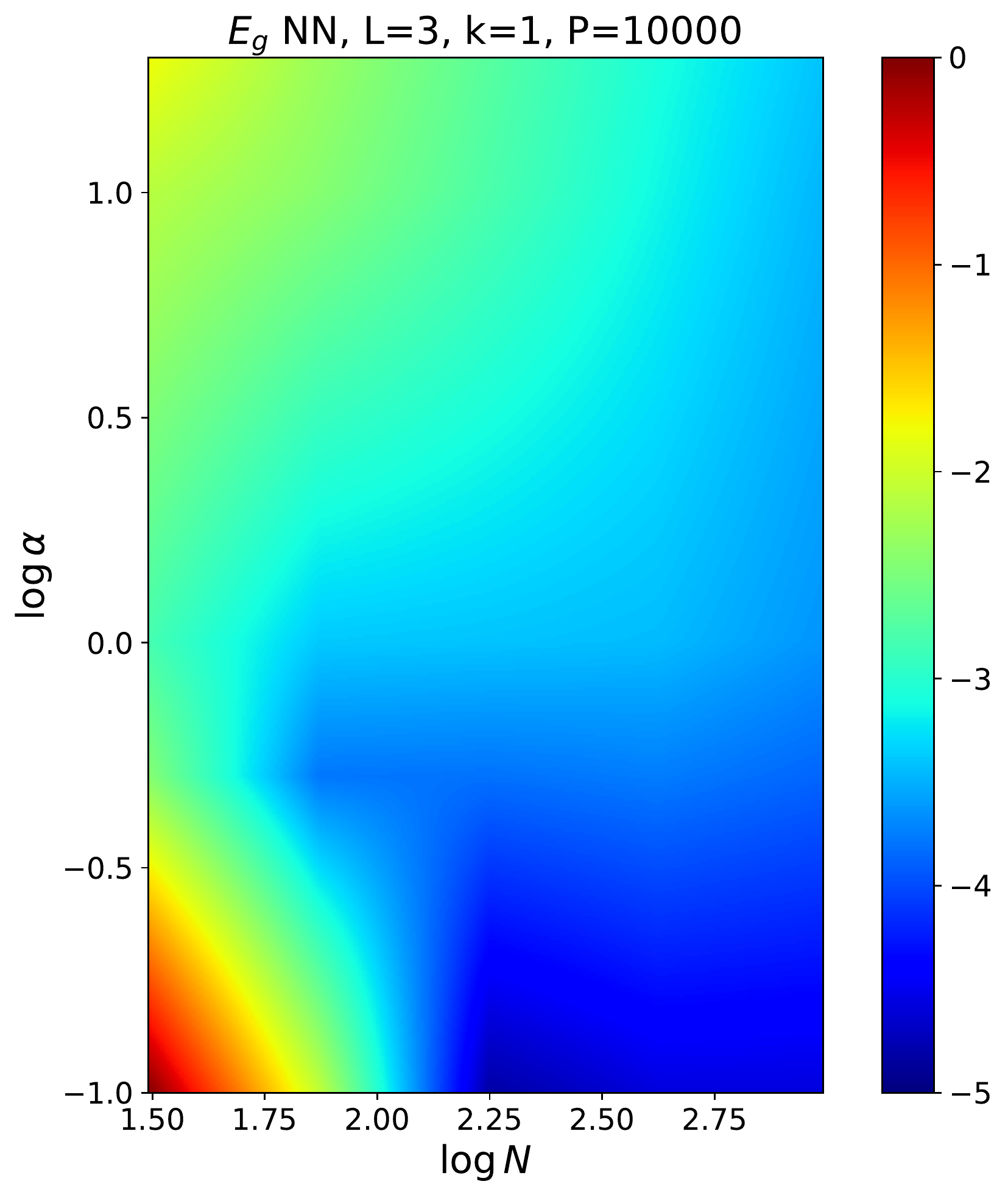}}\\
    \subfigure[$E_g$ for $k=3, P=600$]{\includegraphics[width=0.32\linewidth]{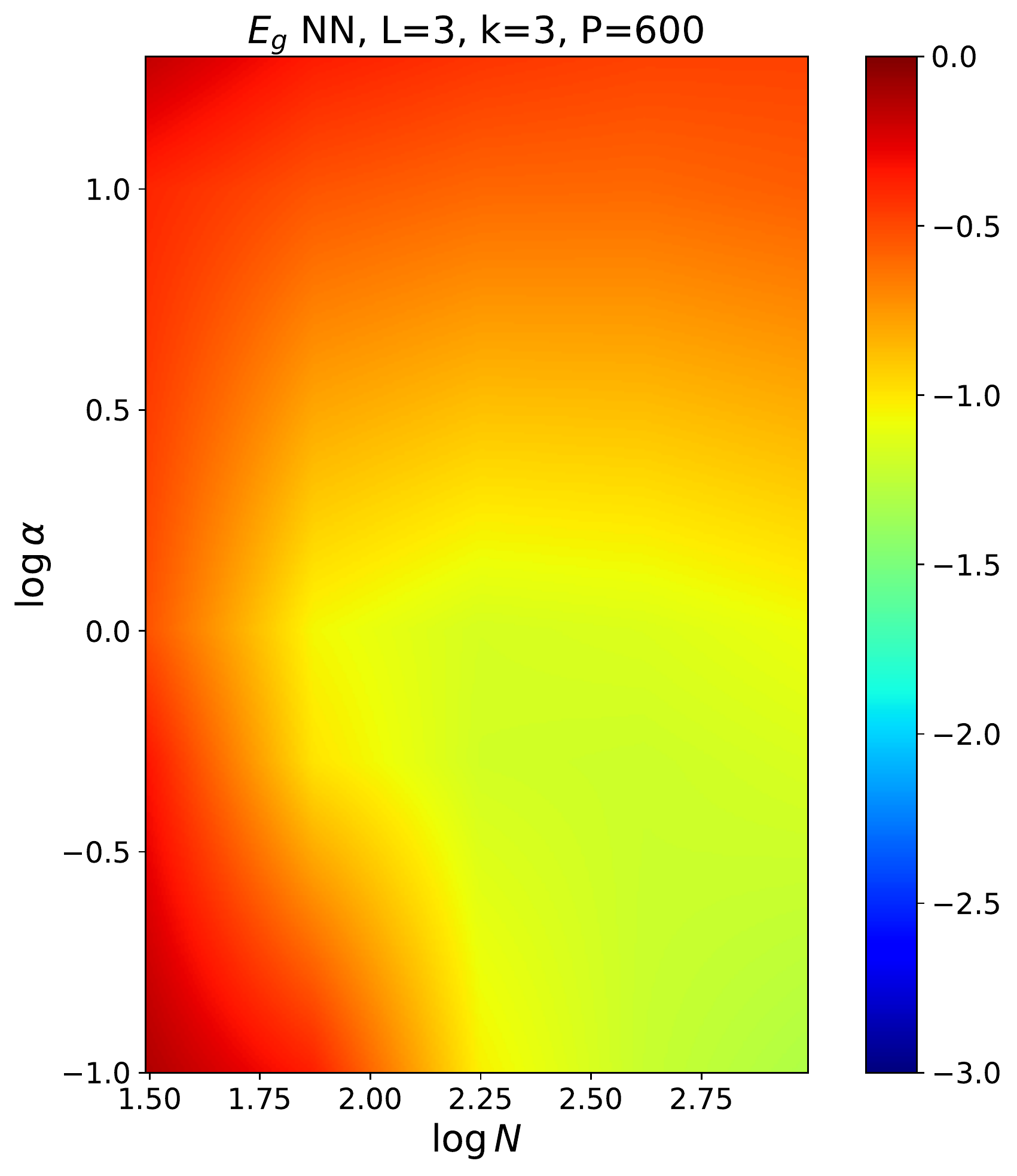}}
    \subfigure[$E_g$ for $k=3, P=10000$]{\includegraphics[width=0.32\linewidth]{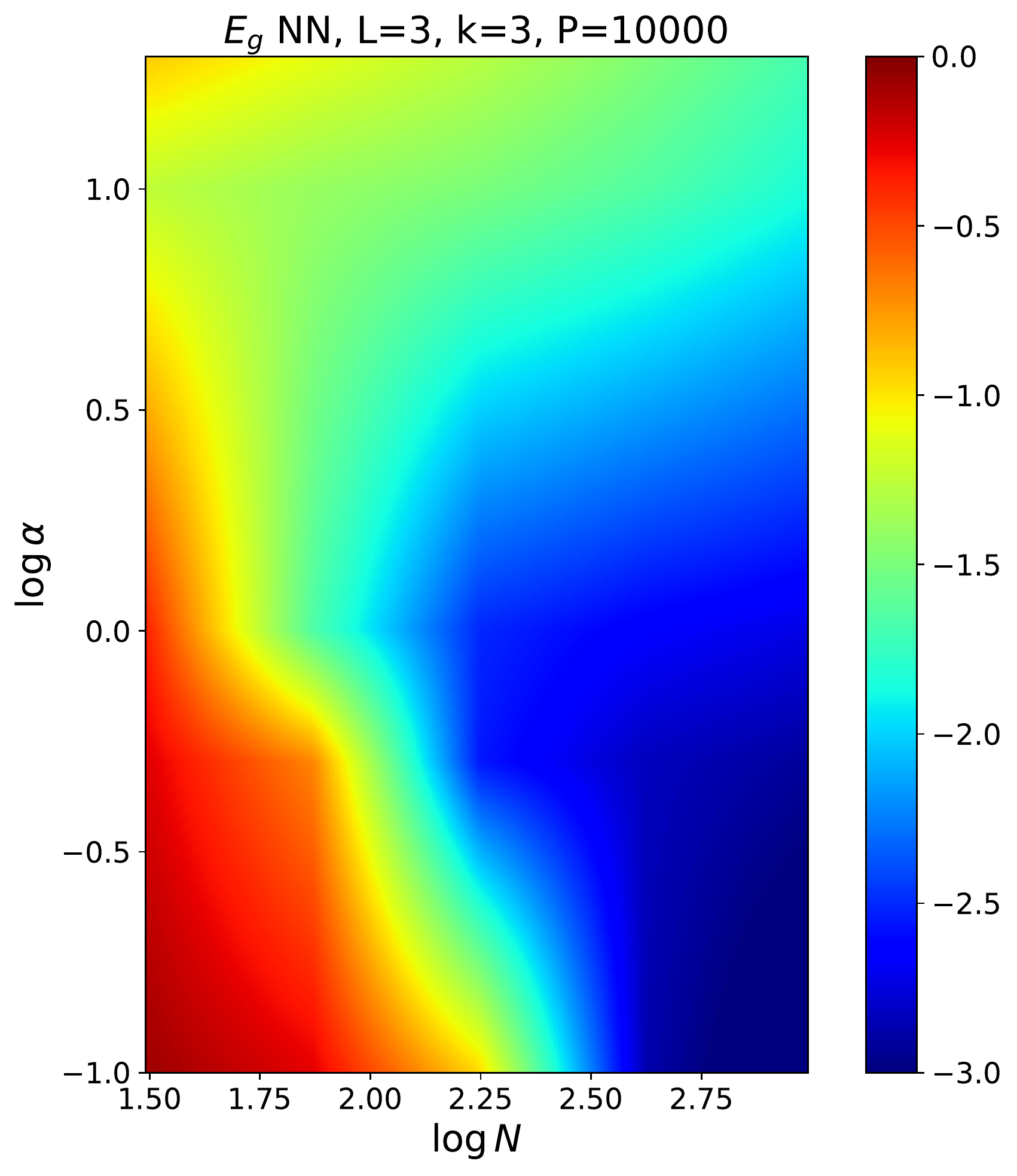}}
    \caption{Phase plots of $\log_{10} E_g$ for neural networks in the $N$-$\alpha$ plane. We plot these at different train set sizes $P$ and different tasks $k$. The colors are fixed to match across networks trained on the same task.}
    \label{fig:N_alpha_plots}
\end{figure}

\begin{figure}[h]
    \centering
    \includegraphics[width=0.4\linewidth]{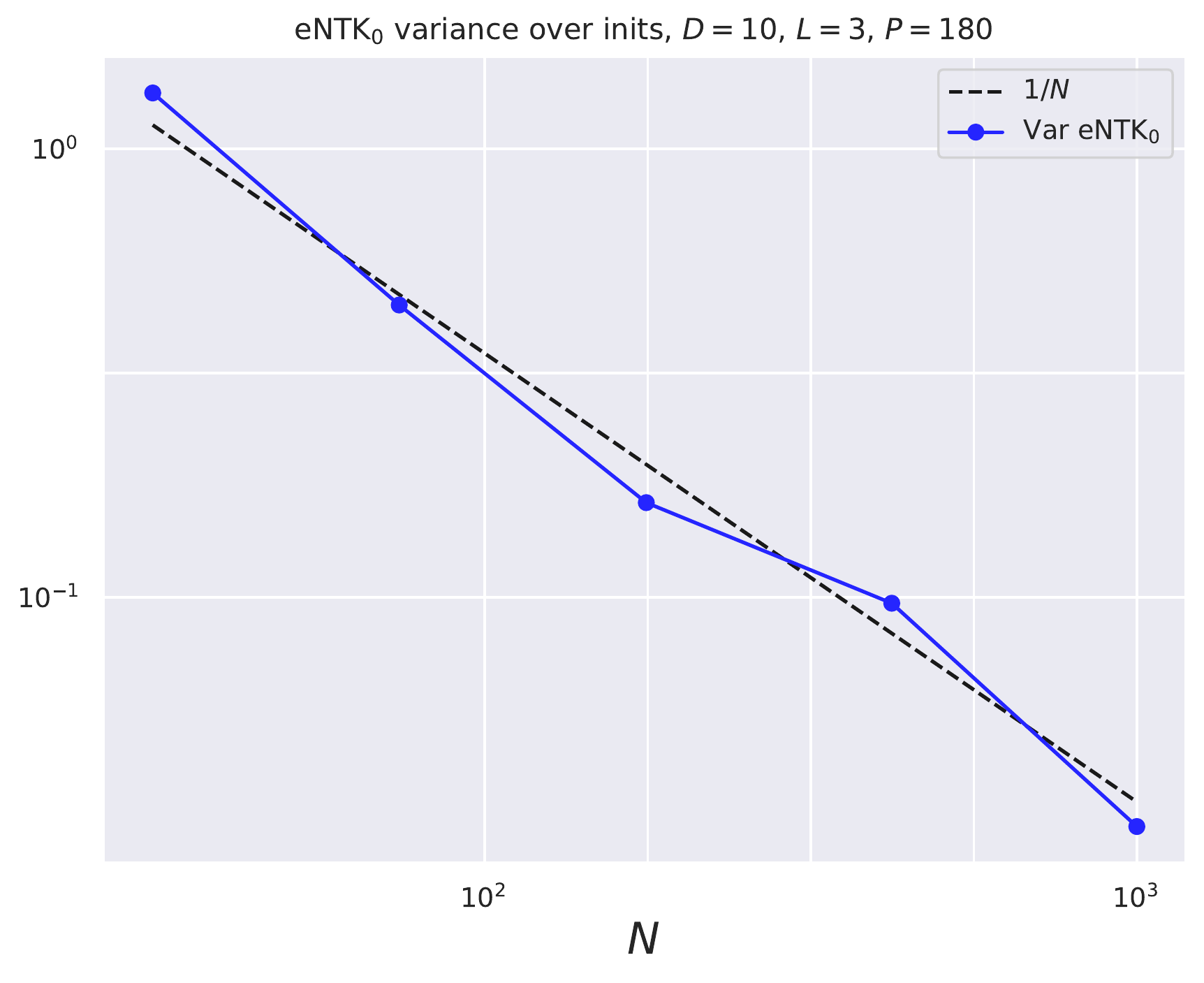}
    \caption{Empirical plot of of the scaling of the variance of the \entk with $N$ with variance taken over 10 initializations and averaged 10 different datasets.}
    \label{fig:var_scaling}
\end{figure}

\begin{figure}[h]
    \centering
    \subfigure[$E_g$ for $L=2$]{\includegraphics[width=0.32\linewidth]{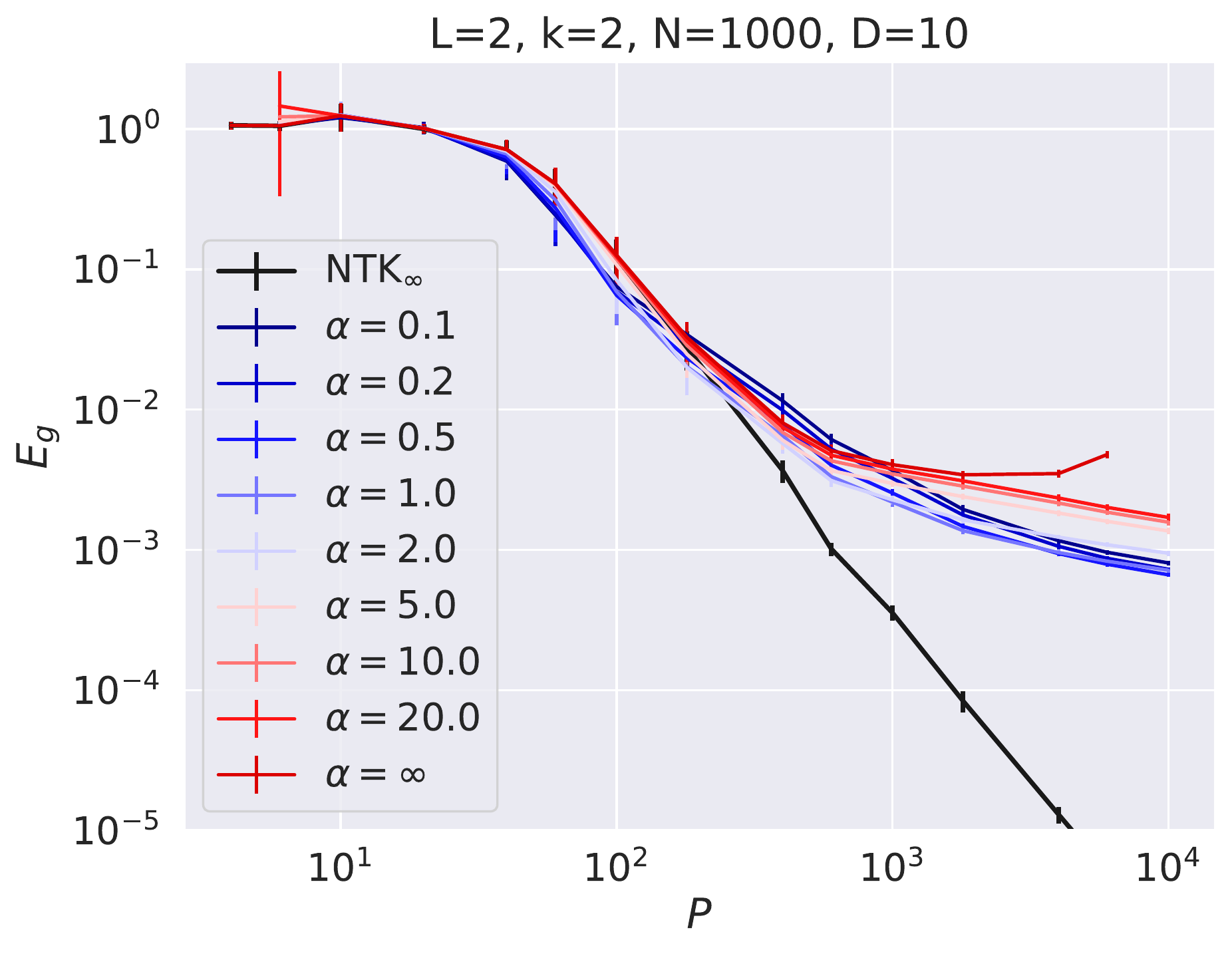}}
    \subfigure[$E_g$ for $L=3$]{\includegraphics[width=0.32\linewidth]{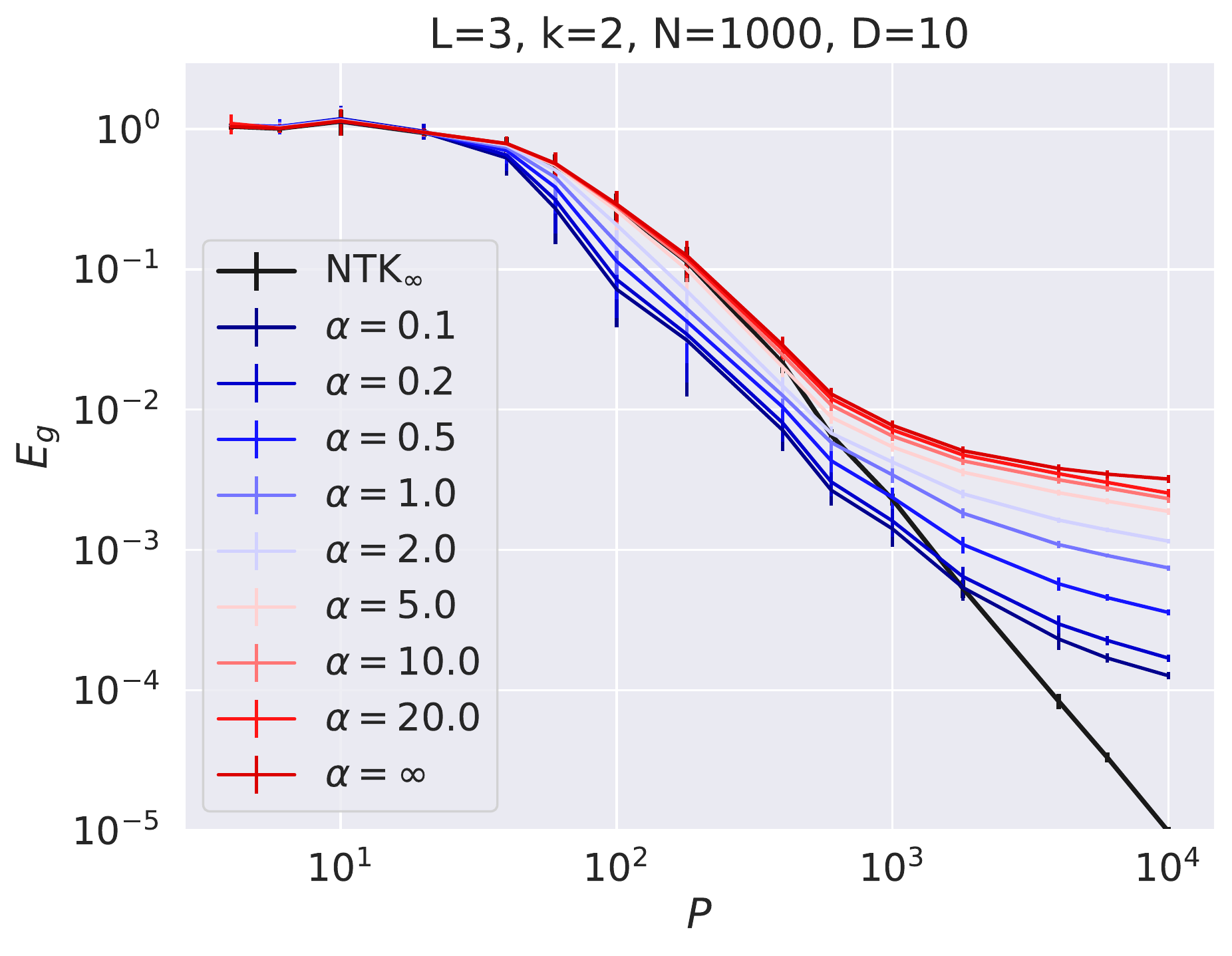}}
    \subfigure[$E_g$ for $L=4$]{\includegraphics[width=0.32\linewidth]{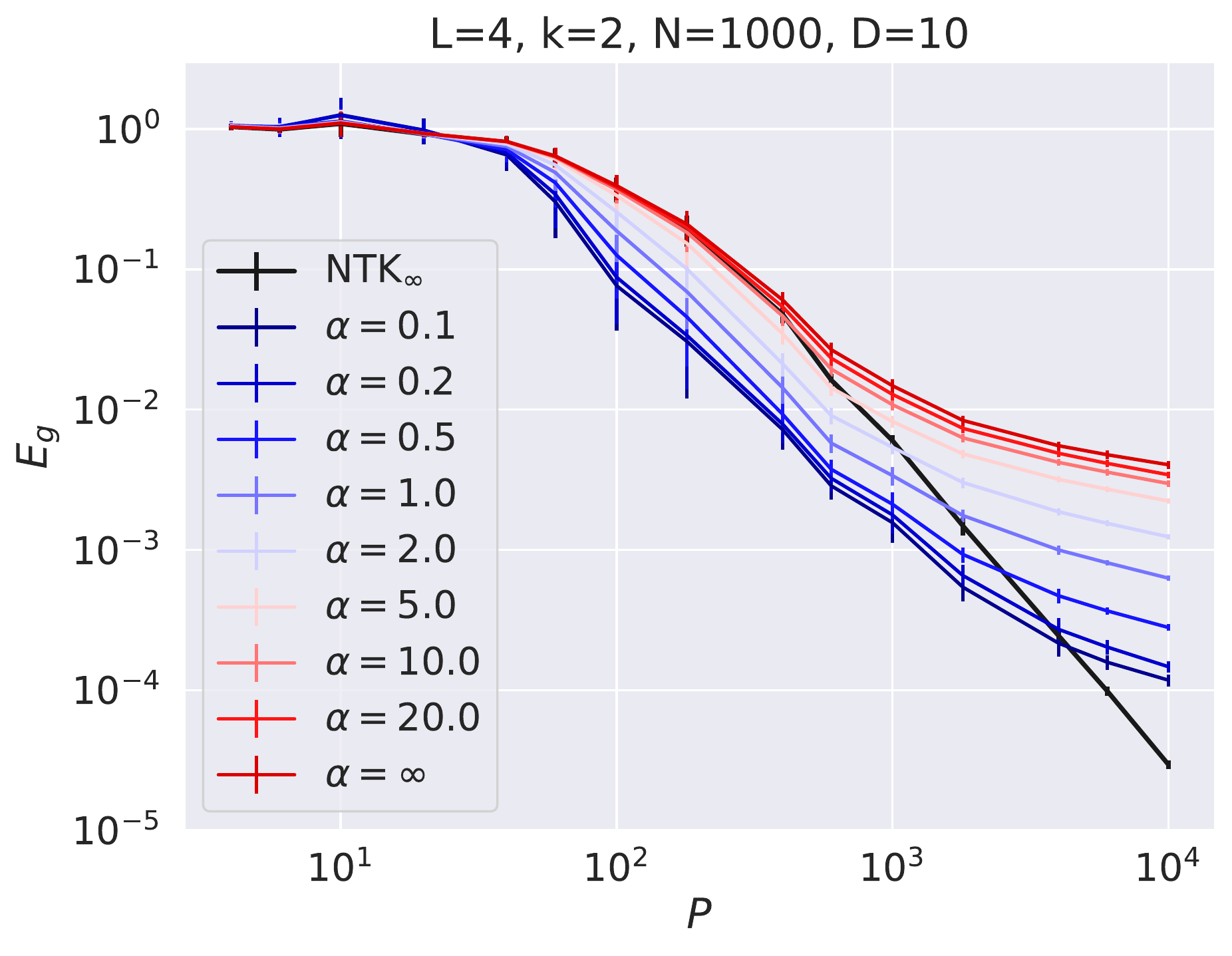}}\\
    \subfigure[Ensembled $E_g$ for $L=2$]
    {\includegraphics[width=0.32\linewidth]{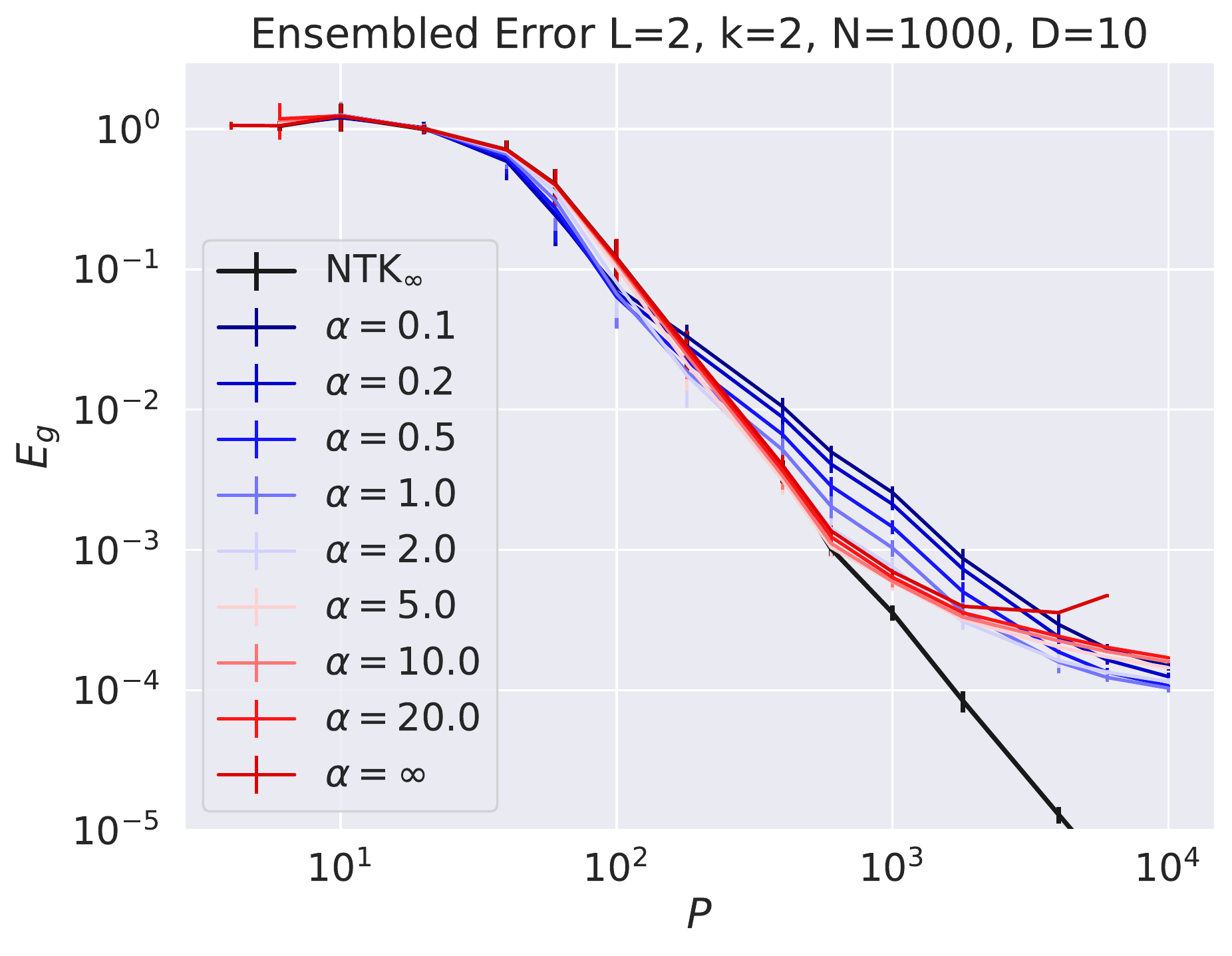}}
    \subfigure[Ensembled $E_g$ for $L=3$]{\includegraphics[width=0.32\linewidth]{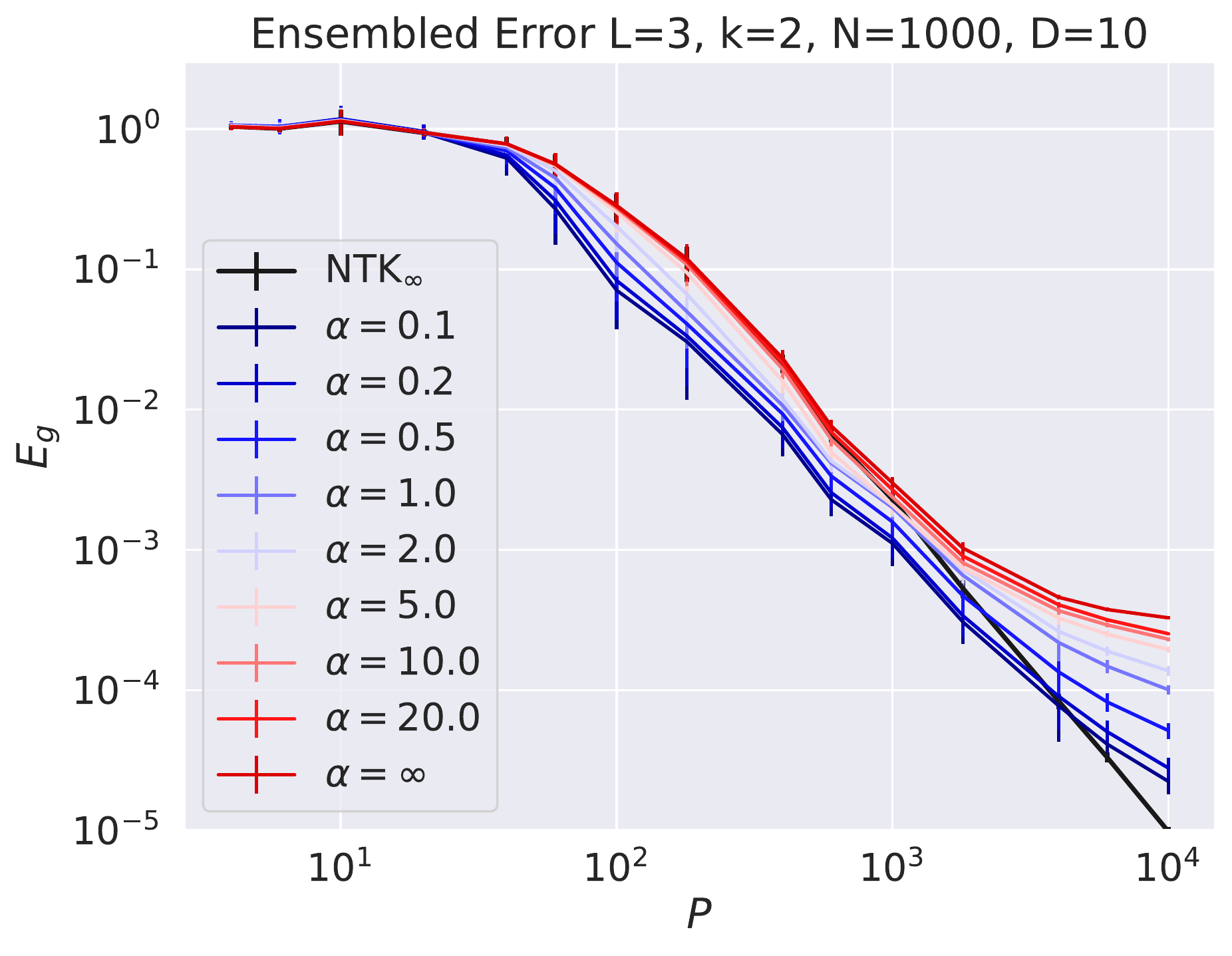}}
    \subfigure[Ensembled $E_g$ for $L=4$]{\includegraphics[width=0.32\linewidth]{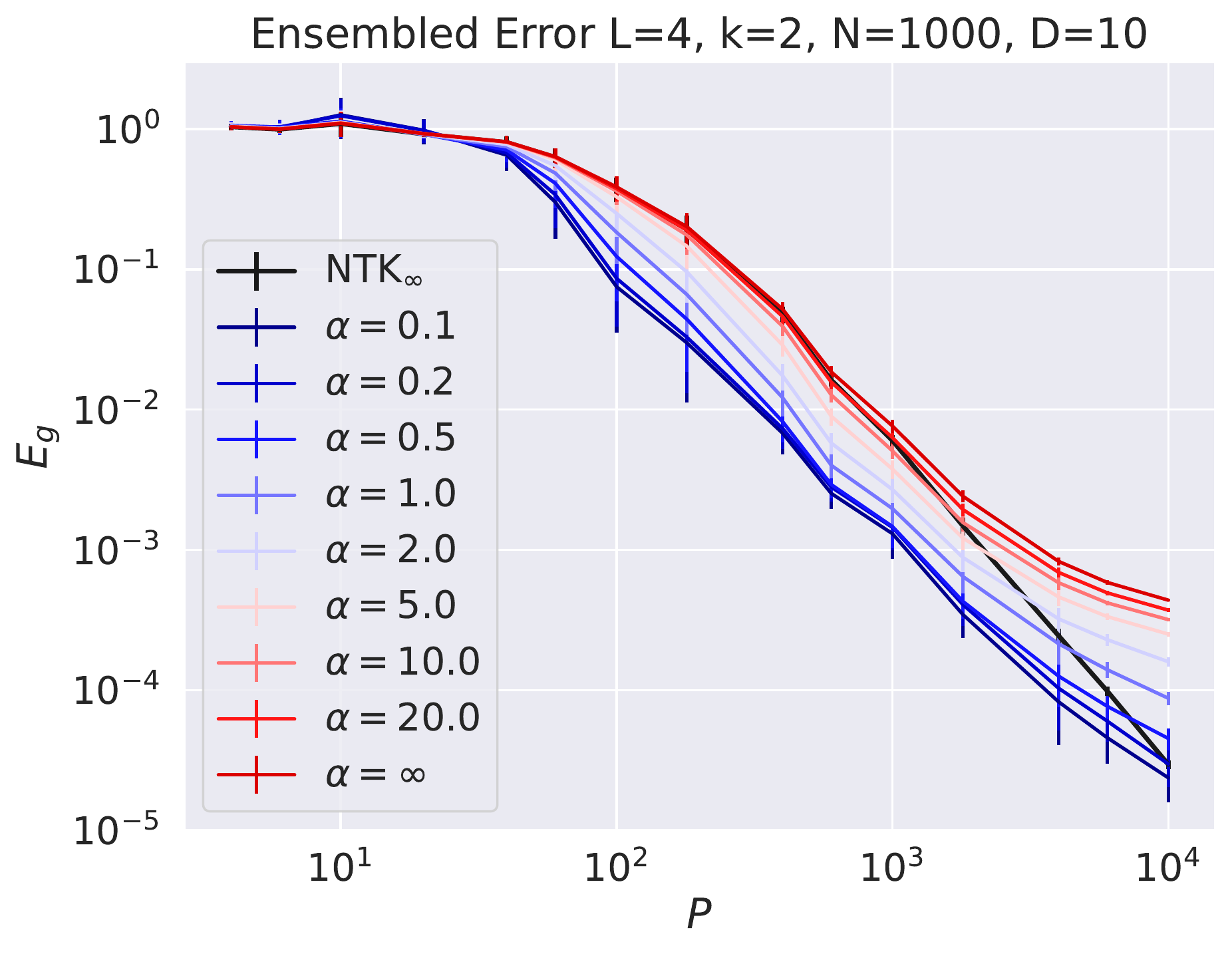}}\\
    \subfigure[$\mathrm{var}/E_g$ for $L=2$]
    {\includegraphics[width=0.32\linewidth]{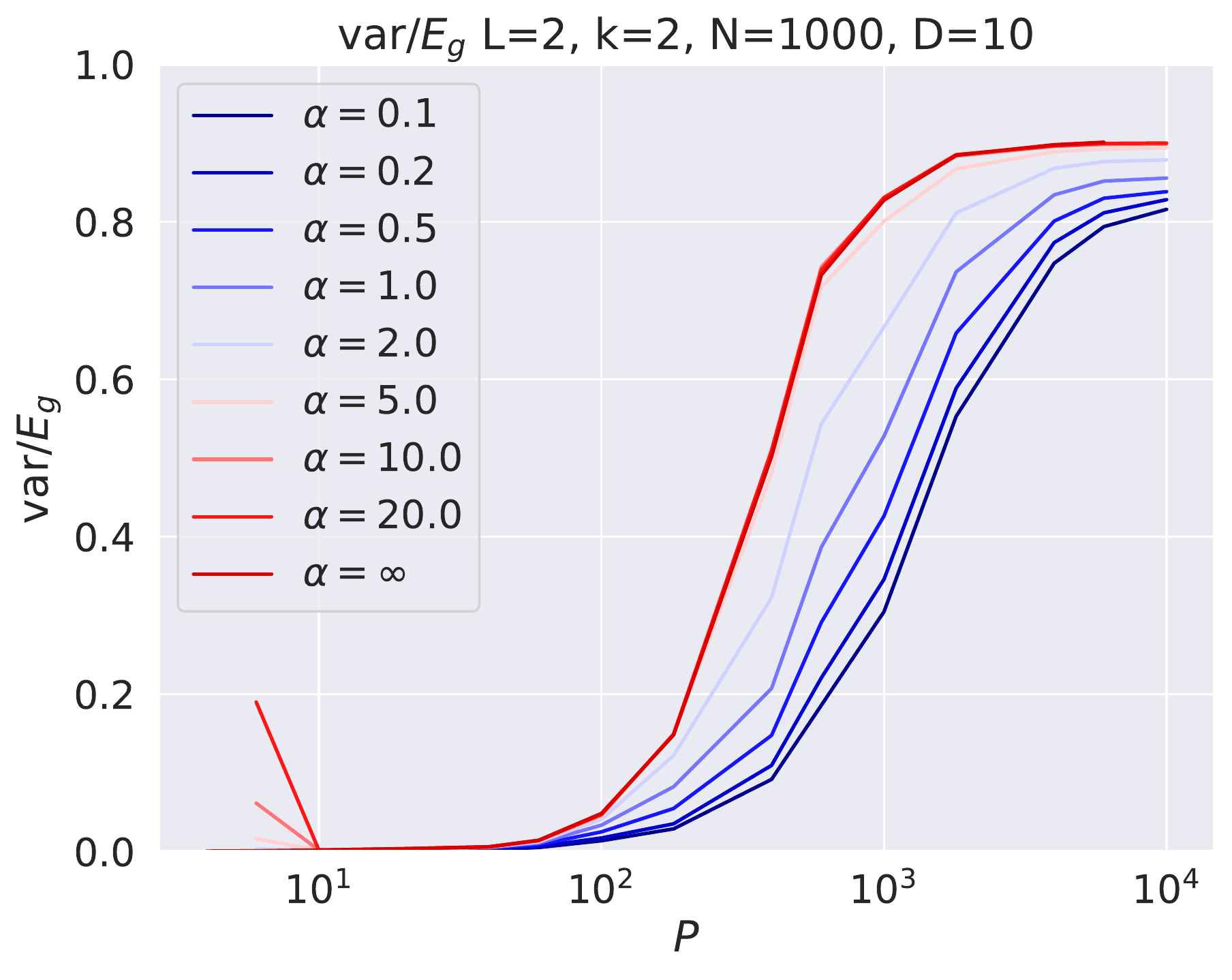}}
    \subfigure[$\mathrm{var}/E_g$ for $L=3$]{\includegraphics[width=0.32\linewidth]{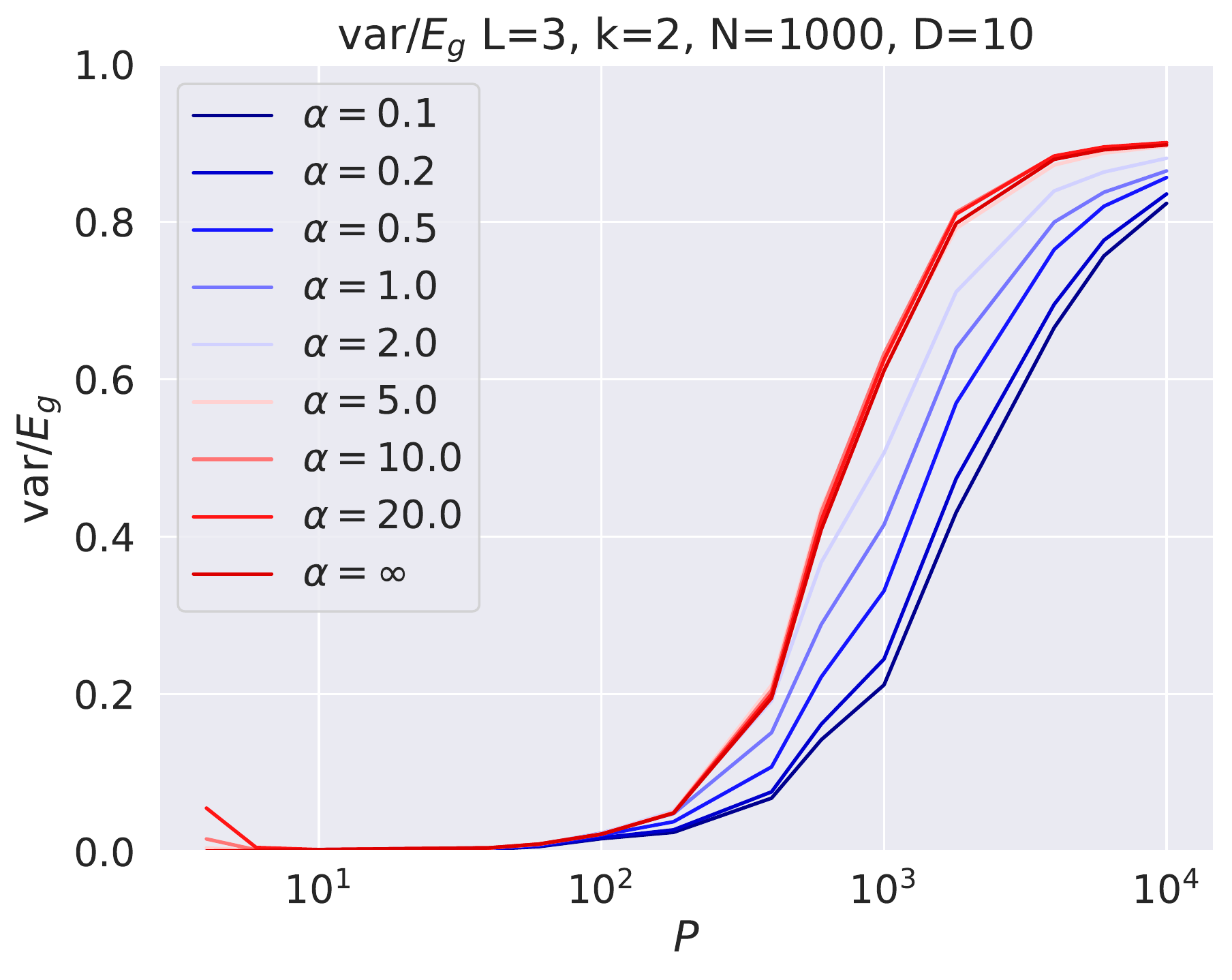}}
    \subfigure[$\mathrm{var}/E_g$ for $L=4$]{\includegraphics[width=0.32\linewidth]{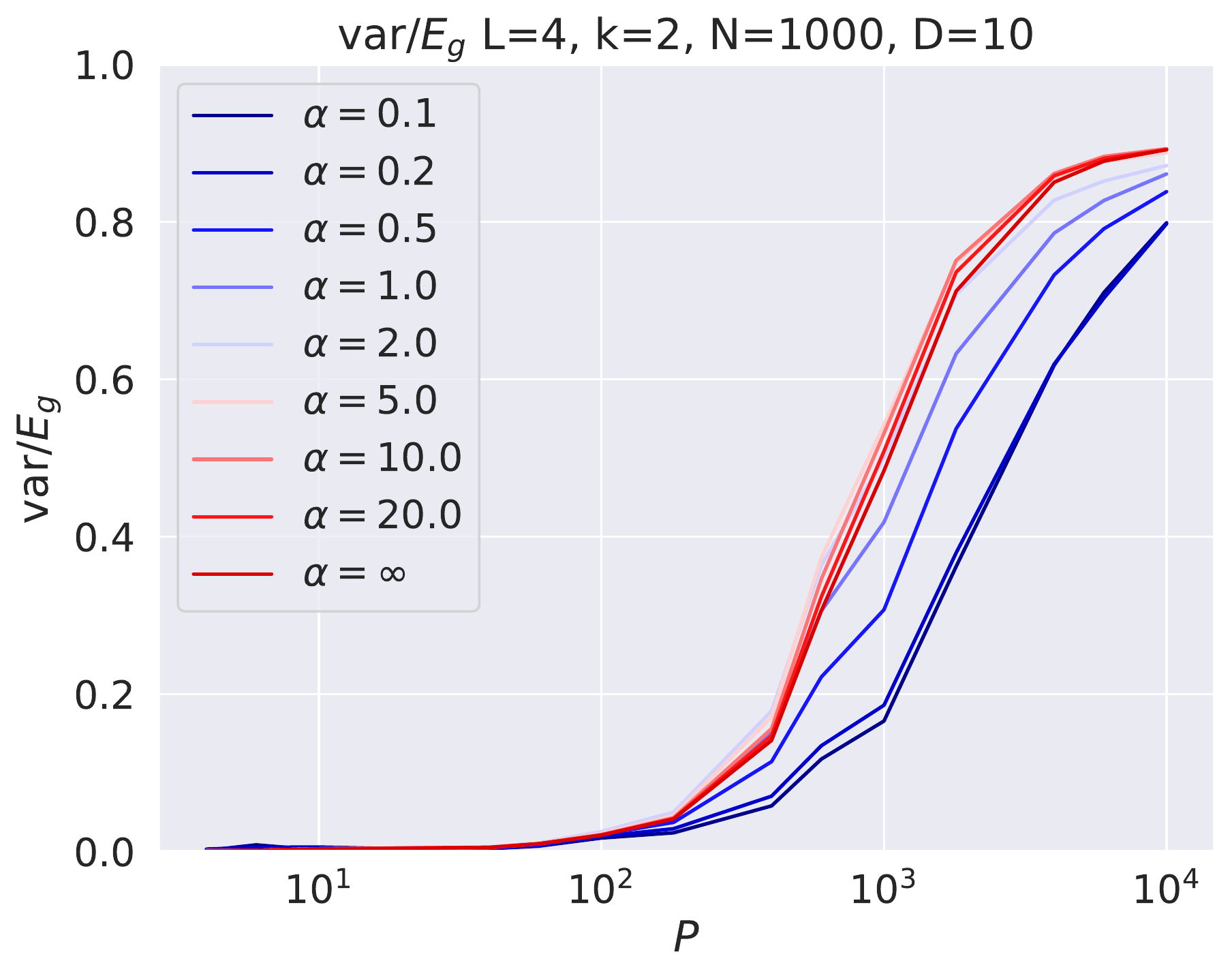}}
    \caption{Sweep over depth $L=\{2,3,4\}$. Deeper networks in the rich regime can more easily outperform the infinite width network for a larger range of $P$. Also, for larger $L$ it is easier to deviate from the lazy regime at a given $\alpha$. By contrast, on this task the shallower \ntk outperforms deeper \ntk \!s.  As before, ensembled lazy networks approach \ntk and the variance rises with $P$.}
    \label{fig:L_sweep}
\end{figure}

\begin{figure}[h]
    \centering
    \subfigure[$E_g$ for $L=2$]{\includegraphics[width=0.32\linewidth]{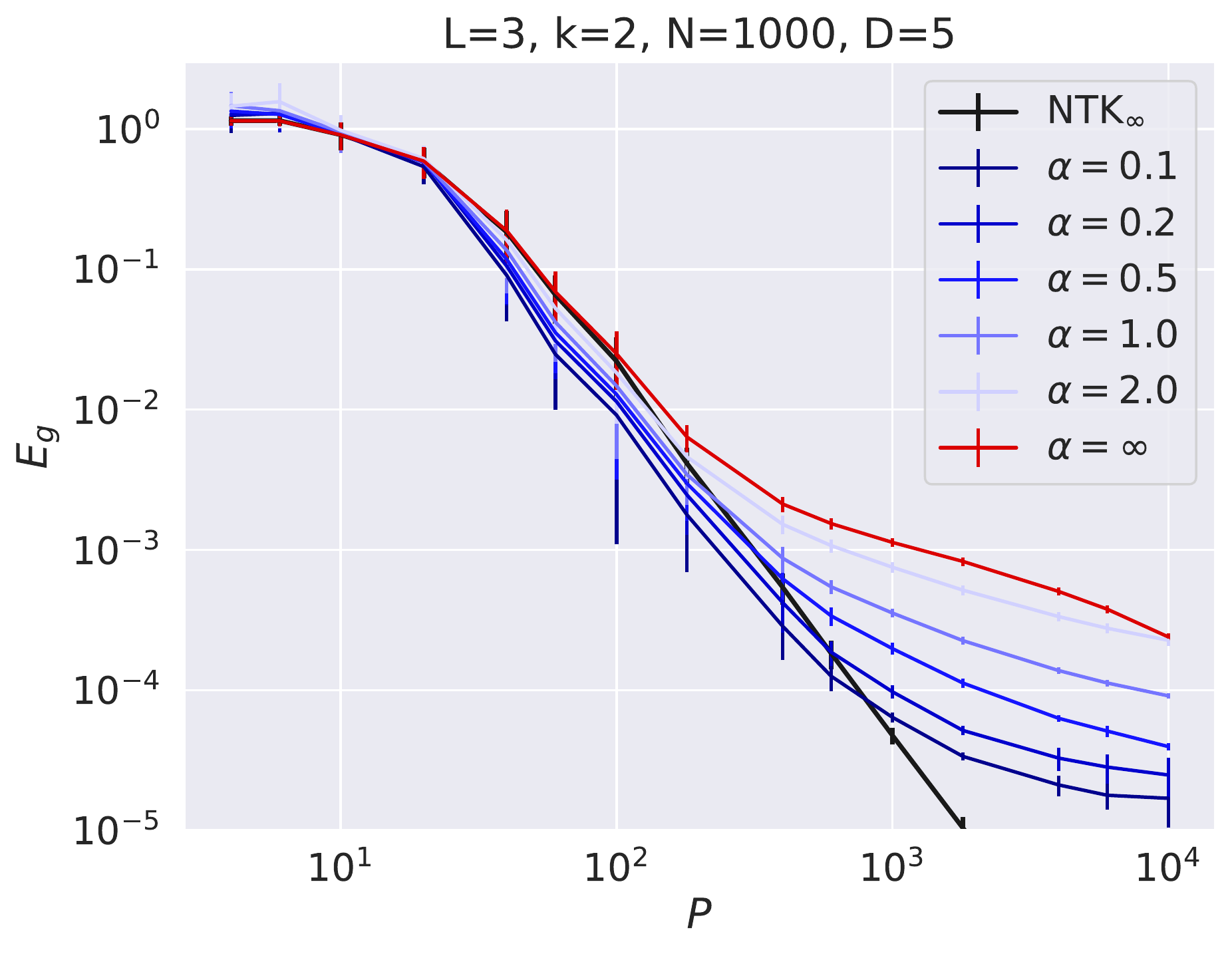}}
    \subfigure[$E_g$ for $L=3$]{\includegraphics[width=0.32\linewidth]{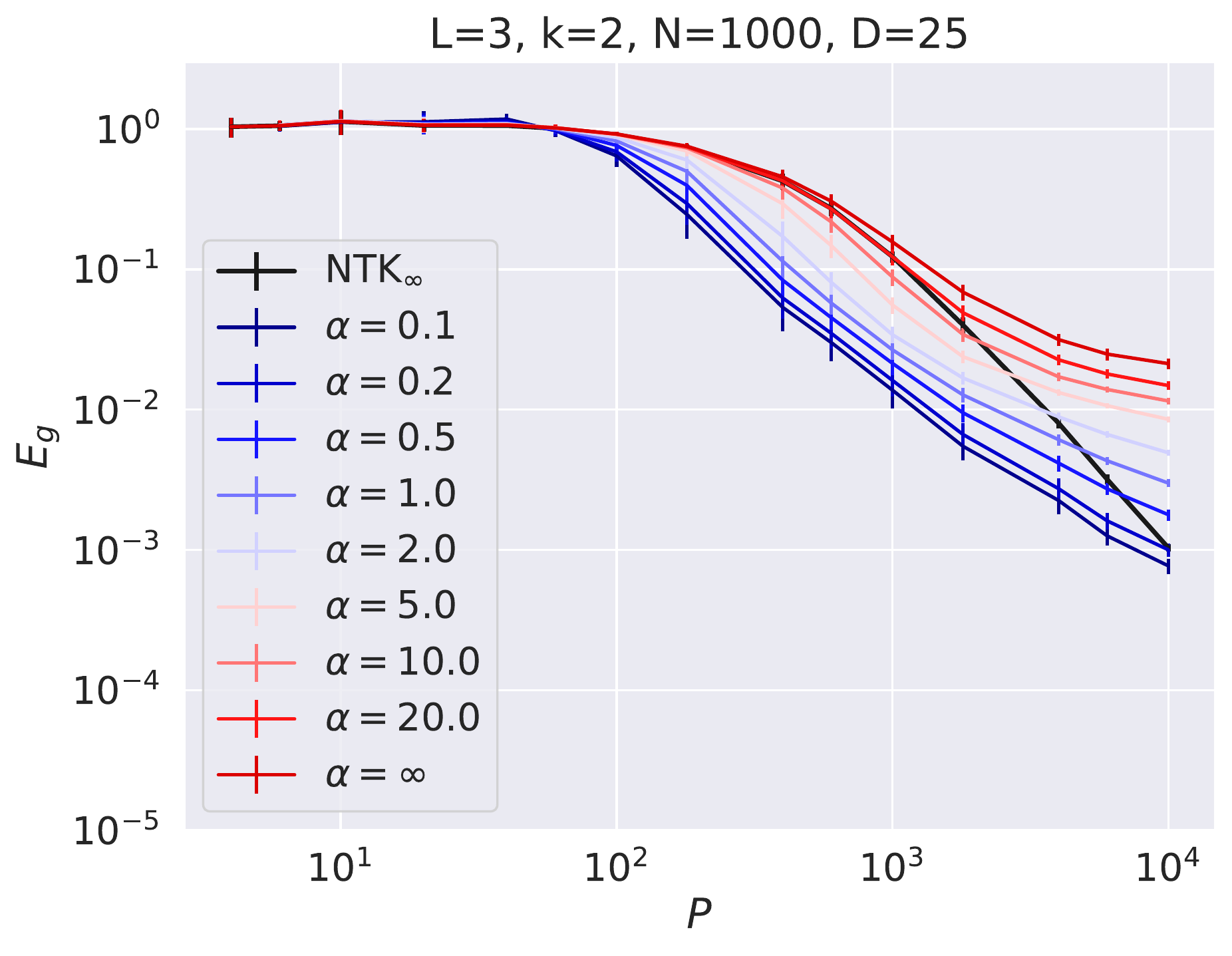}}
    \subfigure[$E_g$ for $L=4$]{\includegraphics[width=0.32\linewidth]{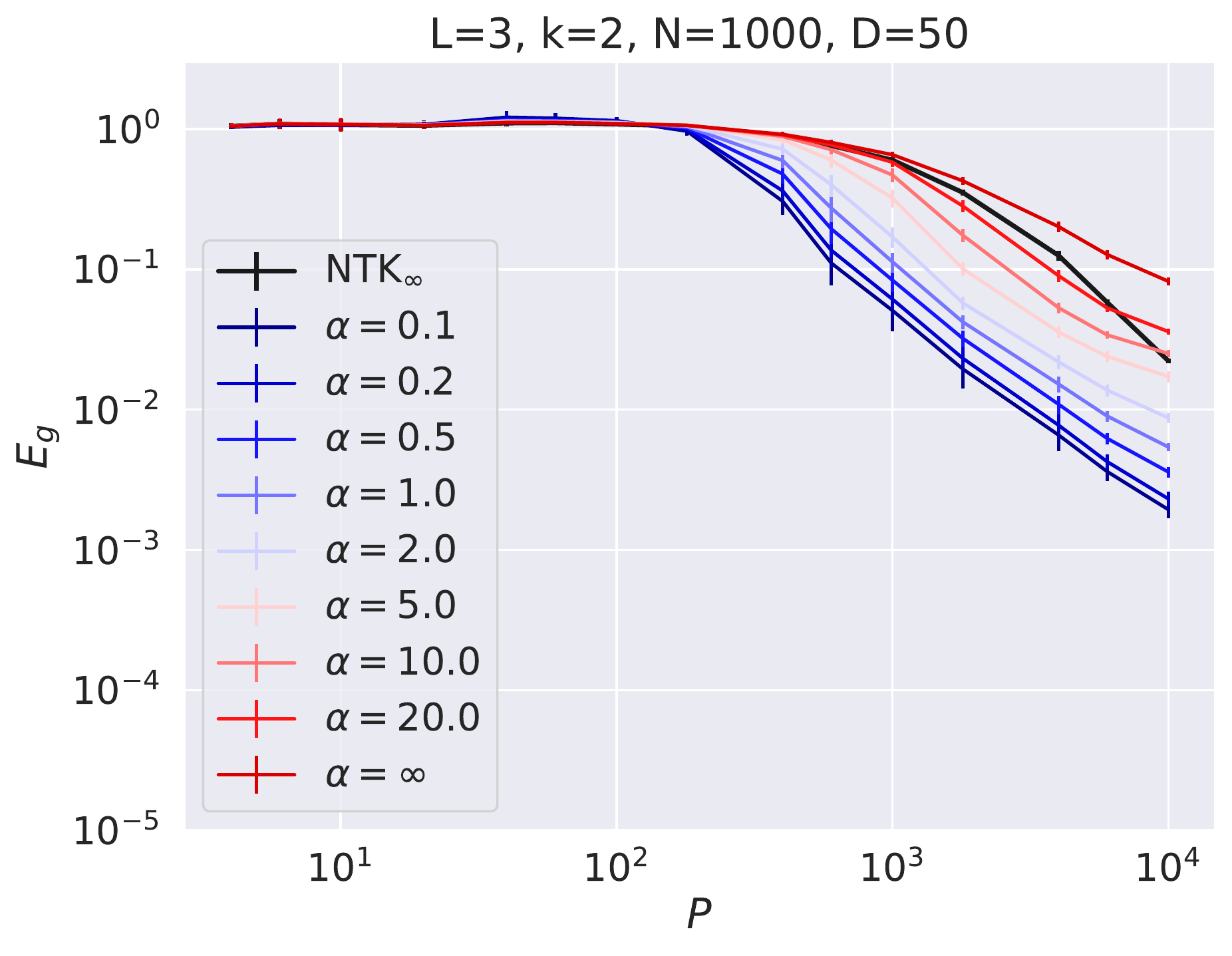}}\\
    \subfigure[Ensembled $E_g$ for $L=2$]
    {\includegraphics[width=0.32\linewidth]{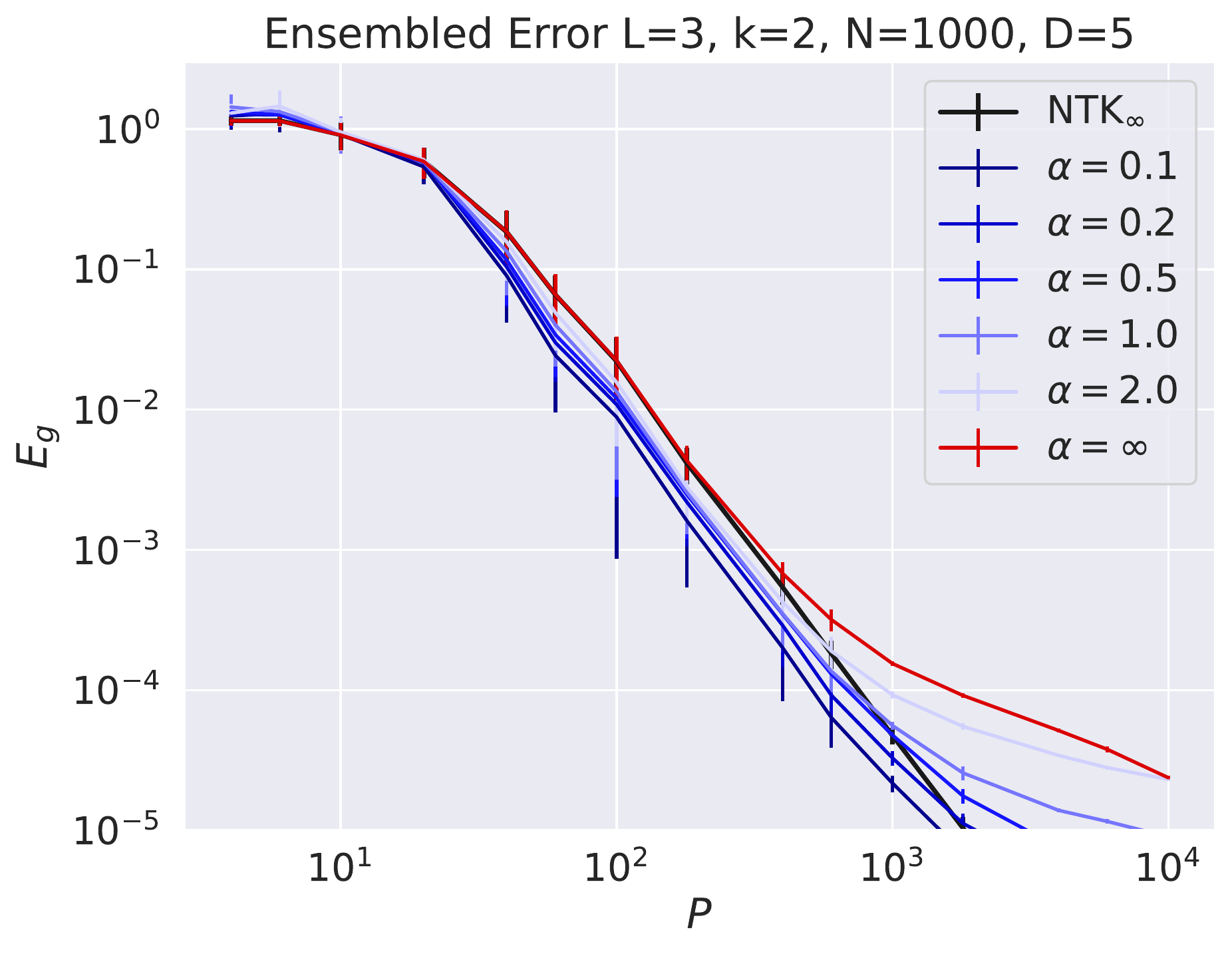}}
    \subfigure[Ensembled $E_g$ for $L=3$]{\includegraphics[width=0.32\linewidth]{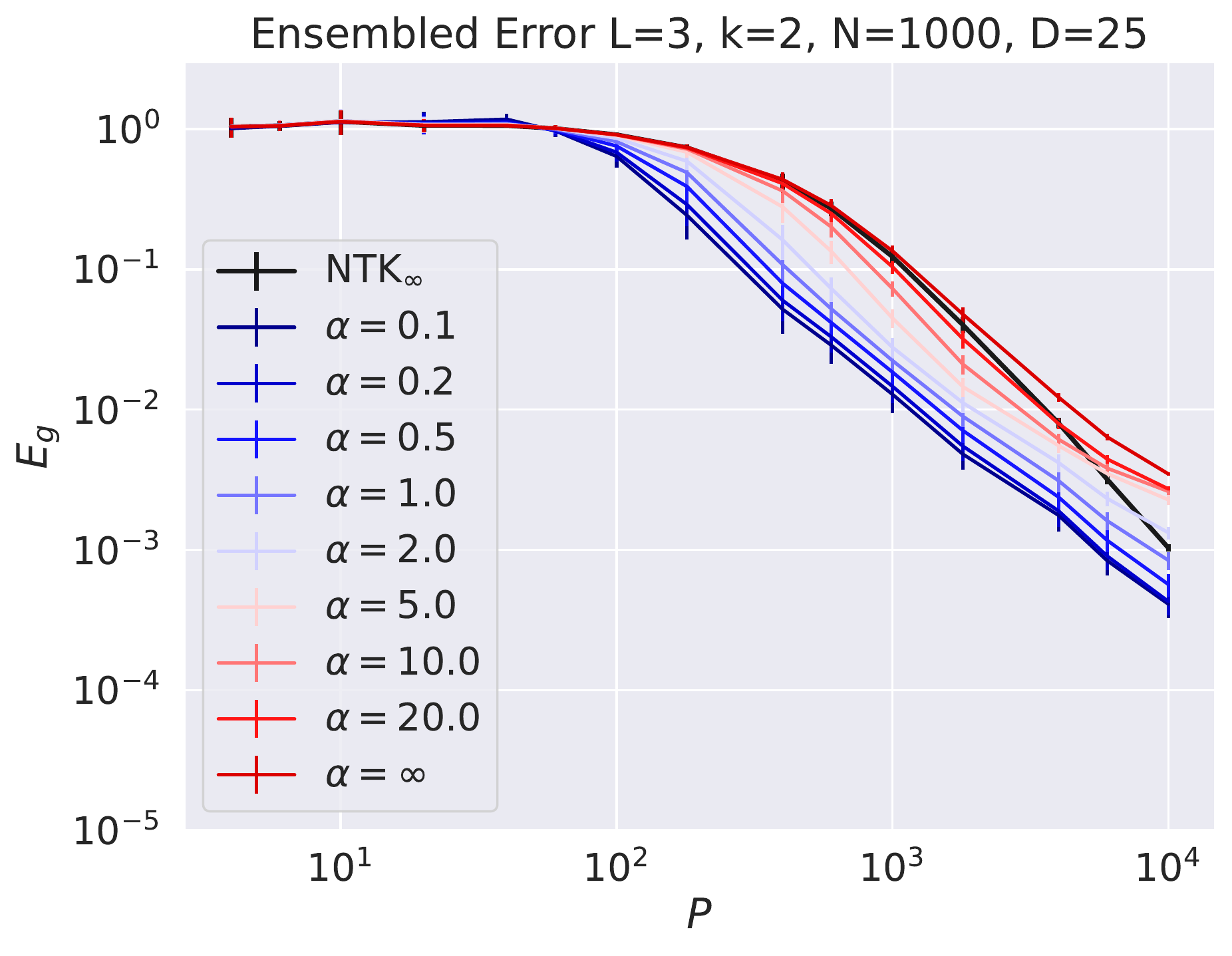}}
    \subfigure[Ensembled $E_g$ for $L=4$]{\includegraphics[width=0.32\linewidth]{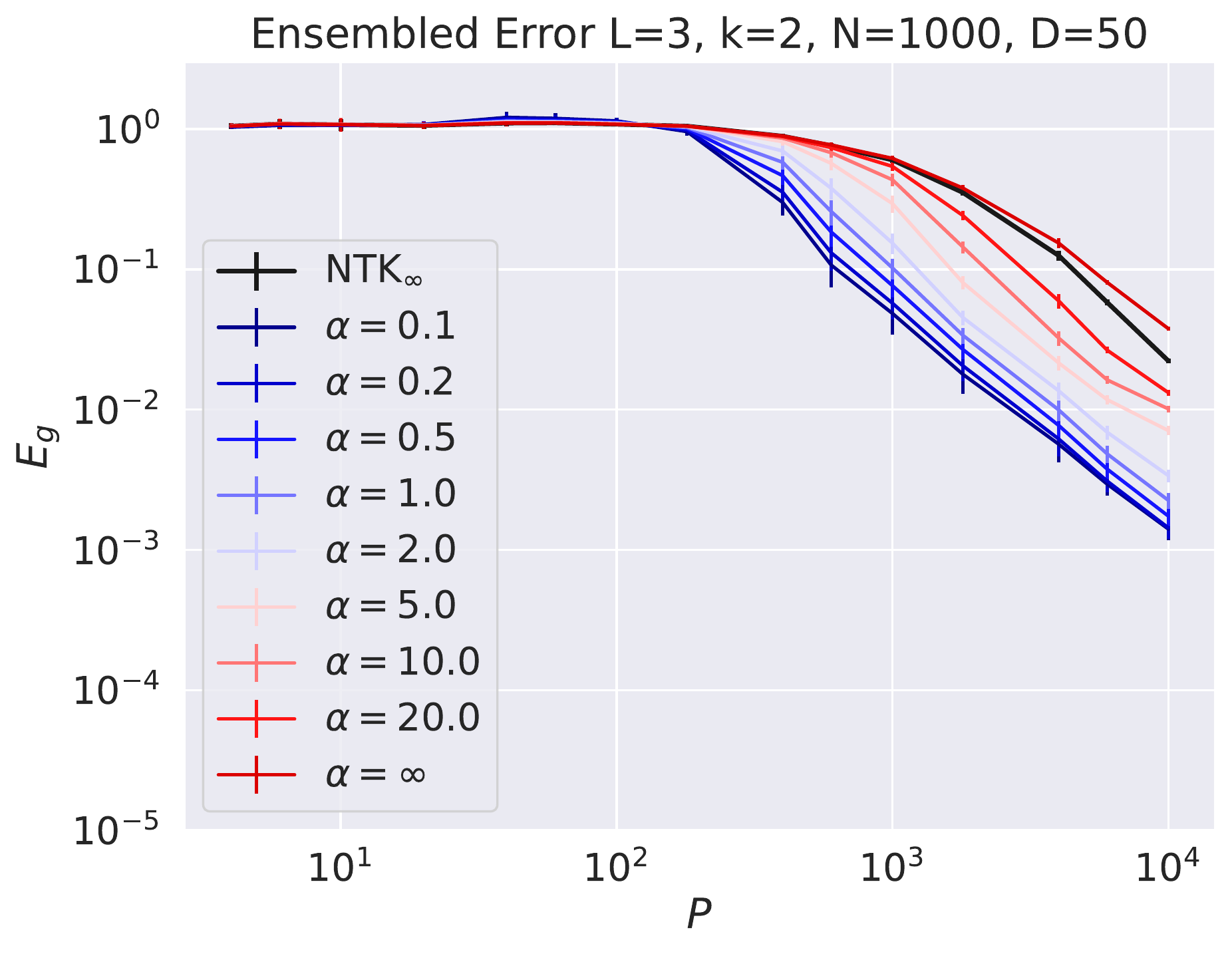}}\\
    \subfigure[$\mathrm{var}/E_g$ for $L=2$]
    {\includegraphics[width=0.32\linewidth]{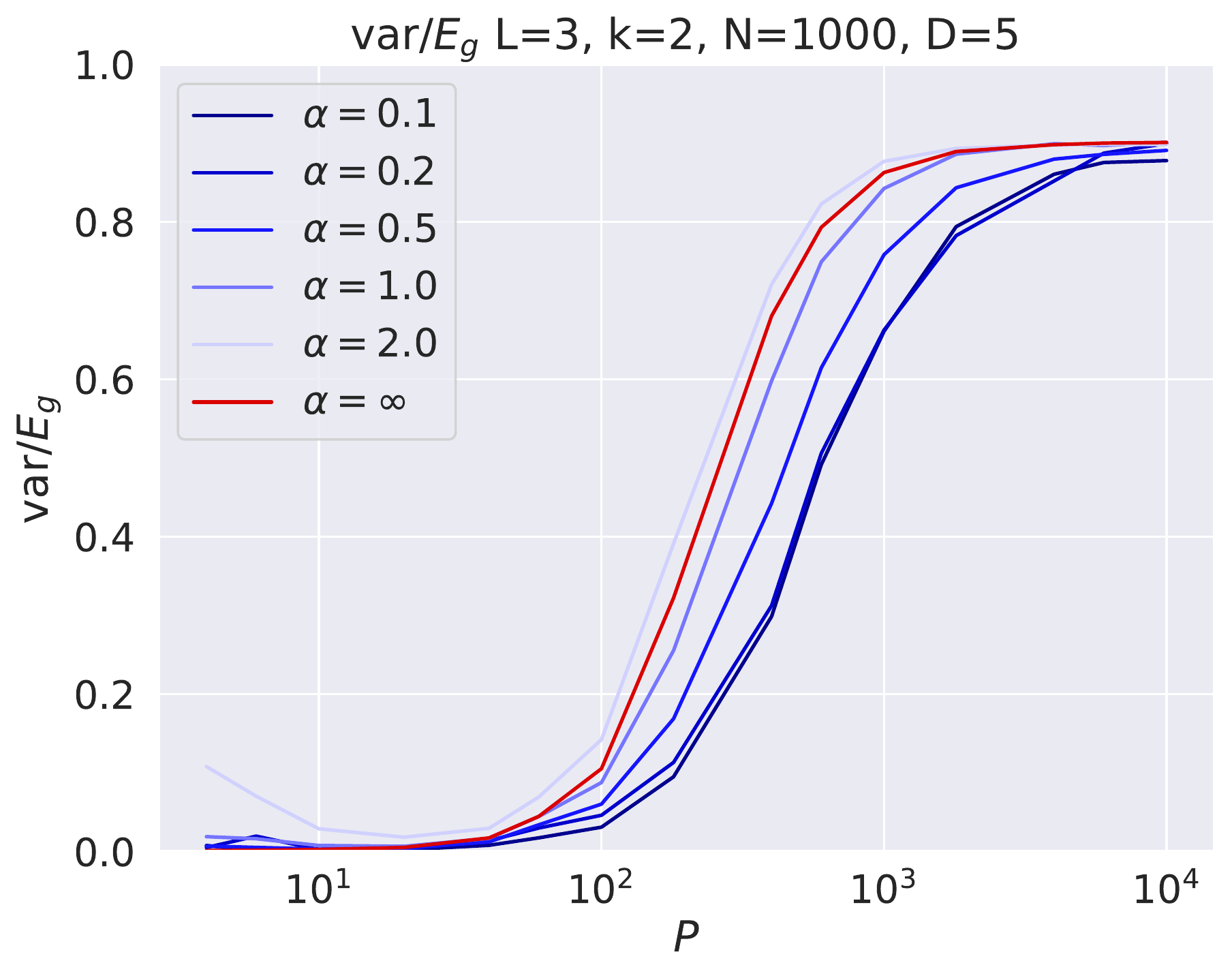}}
    \subfigure[$\mathrm{var}/E_g$ for $L=3$]{\includegraphics[width=0.32\linewidth]{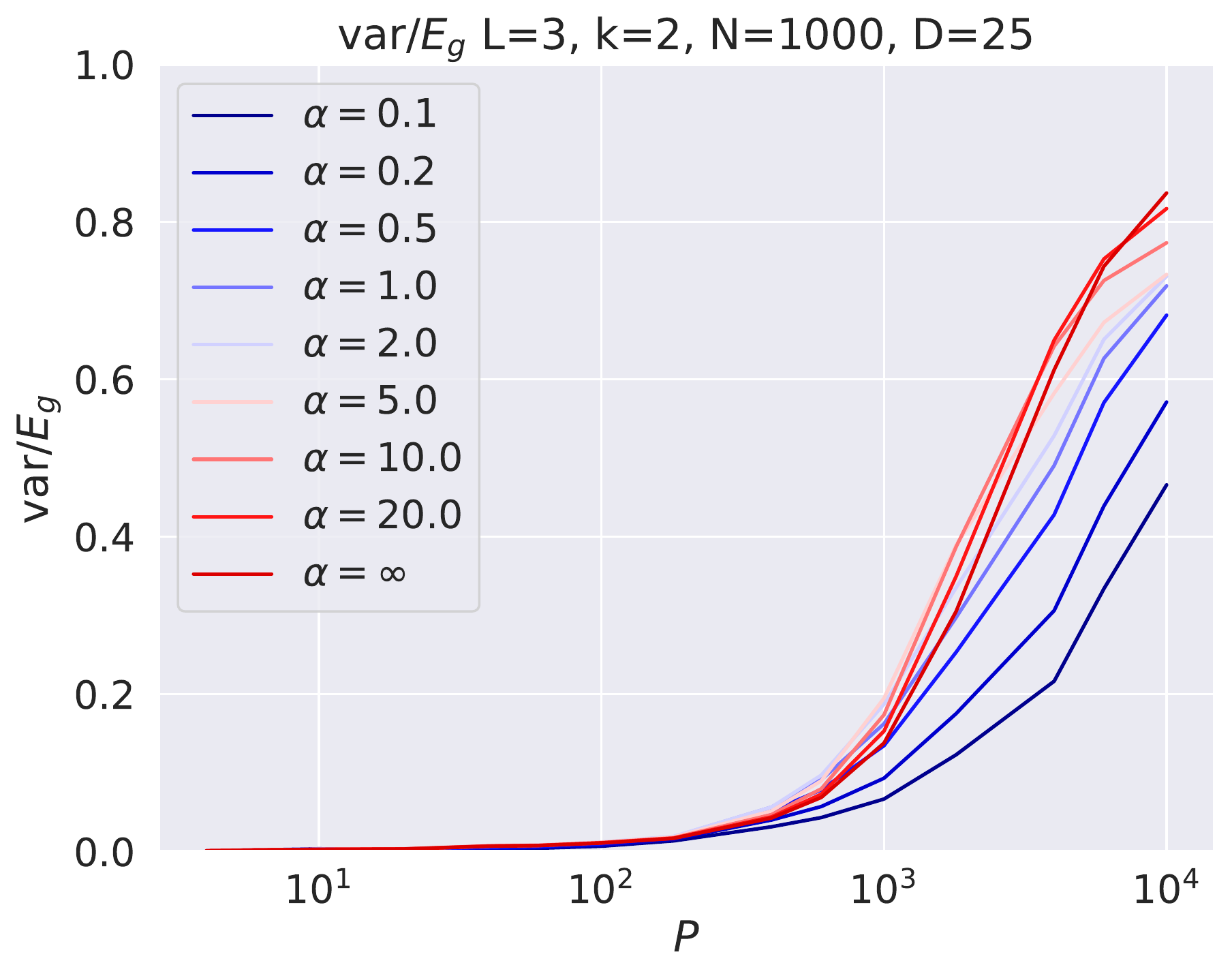}}
    \subfigure[$\mathrm{var}/E_g$ for $L=4$]{\includegraphics[width=0.32\linewidth]{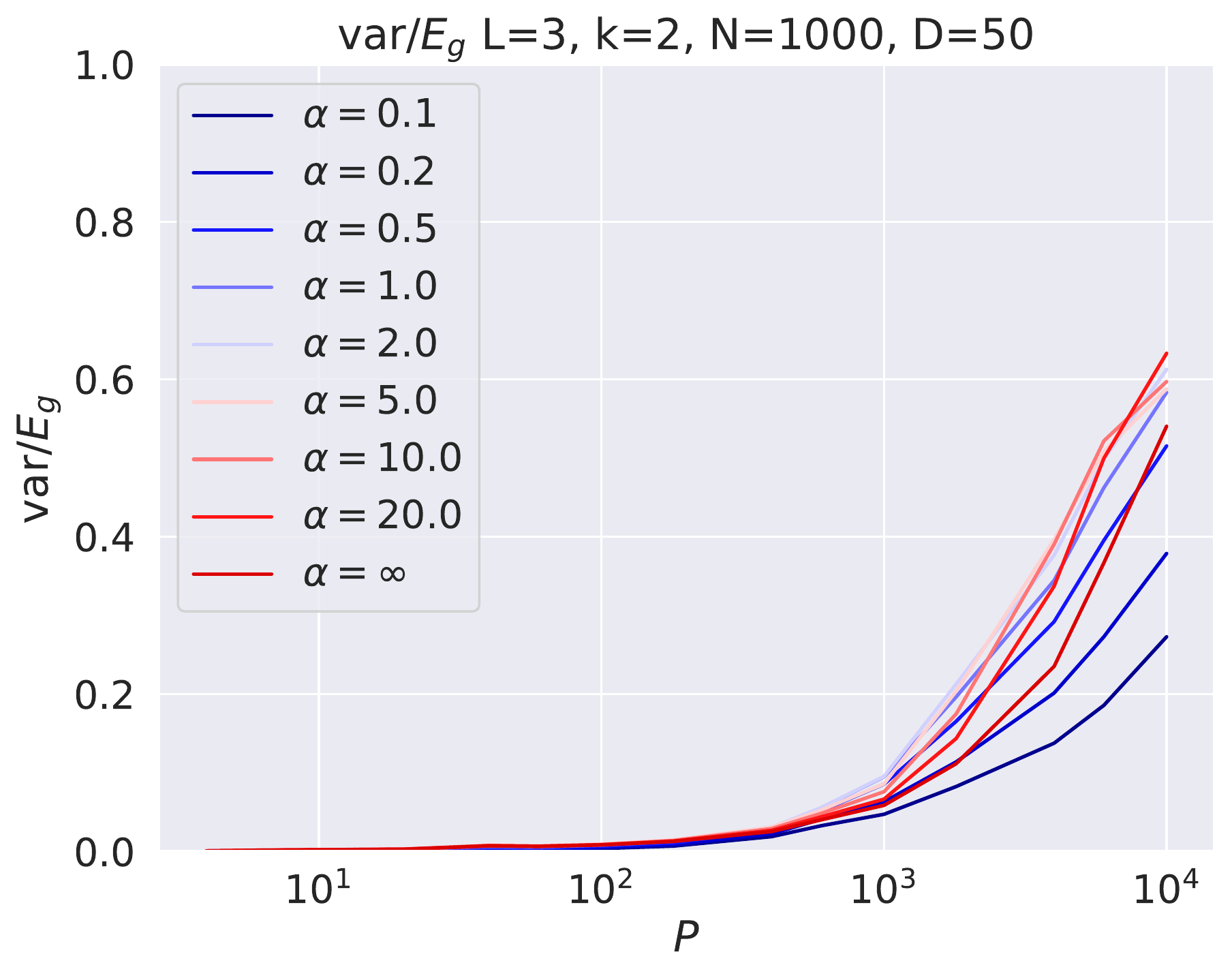}}
    \caption{Sweep over input dimension $D=\{5, 25, 50\}$. At larger input dimensions rich networks can more easily outperform \ntk\!. This is a consequence of the task depending on the low-dimensional projection $\bm \beta \cdot \bm x$.}
    \label{fig:D_sweep}
\end{figure}

\begin{figure}
\centering
    \subfigure[$E_g$ for centered vs uncentered]{
    \includegraphics[width=0.31\linewidth]{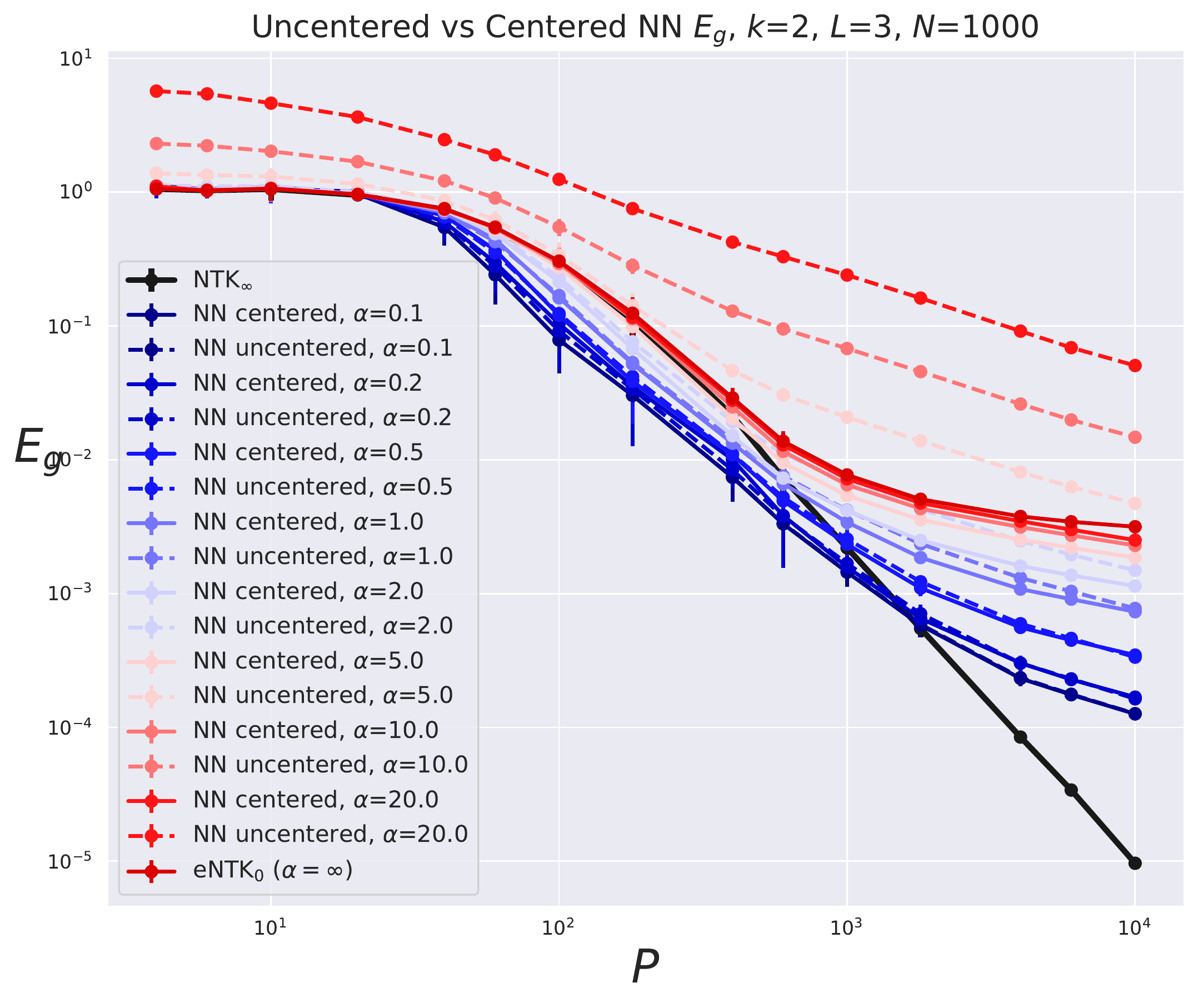}
    }
    \subfigure[Effect of ensembling]{
    \includegraphics[width=0.31\linewidth]{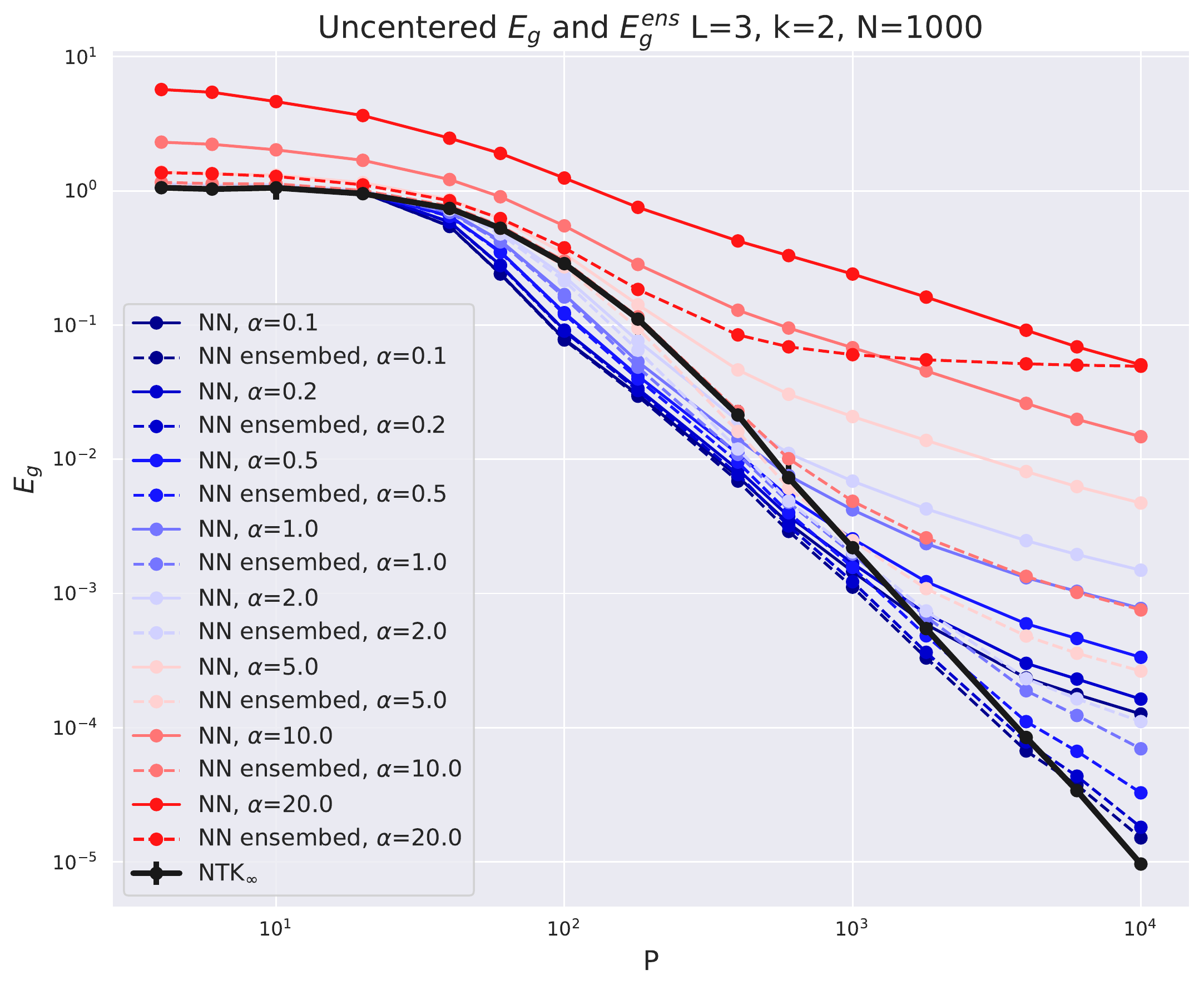}}
    \subfigure[$E_g$ uncentered color plot]{
    \includegraphics[width=0.31\linewidth]{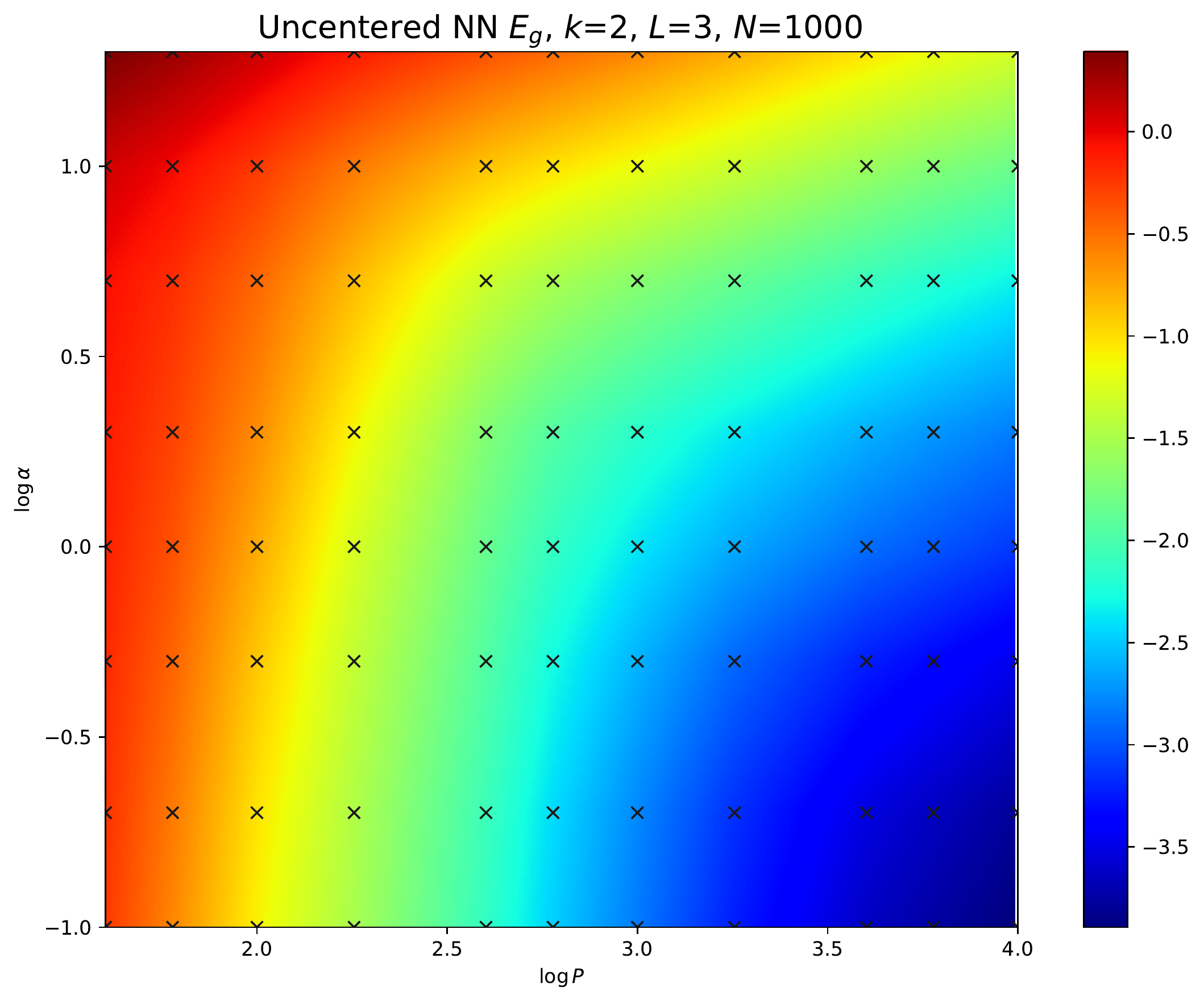}}
    \caption{a) $E_g$ for the centered predictor $\tilde f_\theta(\bm x) - \tilde f_{\theta_0}(\bm x)$ (solid) compared to the generalization of the uncentered predictor $\tilde f_\theta(\bm x)$ (dashed). At small $\alpha$, the difference is negligible, while at large $\alpha$ the uncentered predictor does worse and does not approach \entk. The worse generalization can be understood as $\tilde f_{\theta_0}(\bm x)$ effectively adding an initialization-dependent noise to the target $\bm y$. b) The effect of ensembling becomes less beneficial for uncentered lazy networks. c) Color plot of $E_g$. The lazy regime is different from the \entk generalization (c.f. Figure \ref{fig:gen_errs_over_k}).}
    \label{fig:uncentered}
\end{figure}

\begin{figure}
    \centering
    \subfigure[$E_g$ for mixed mode task]{\includegraphics[width=0.4\linewidth]{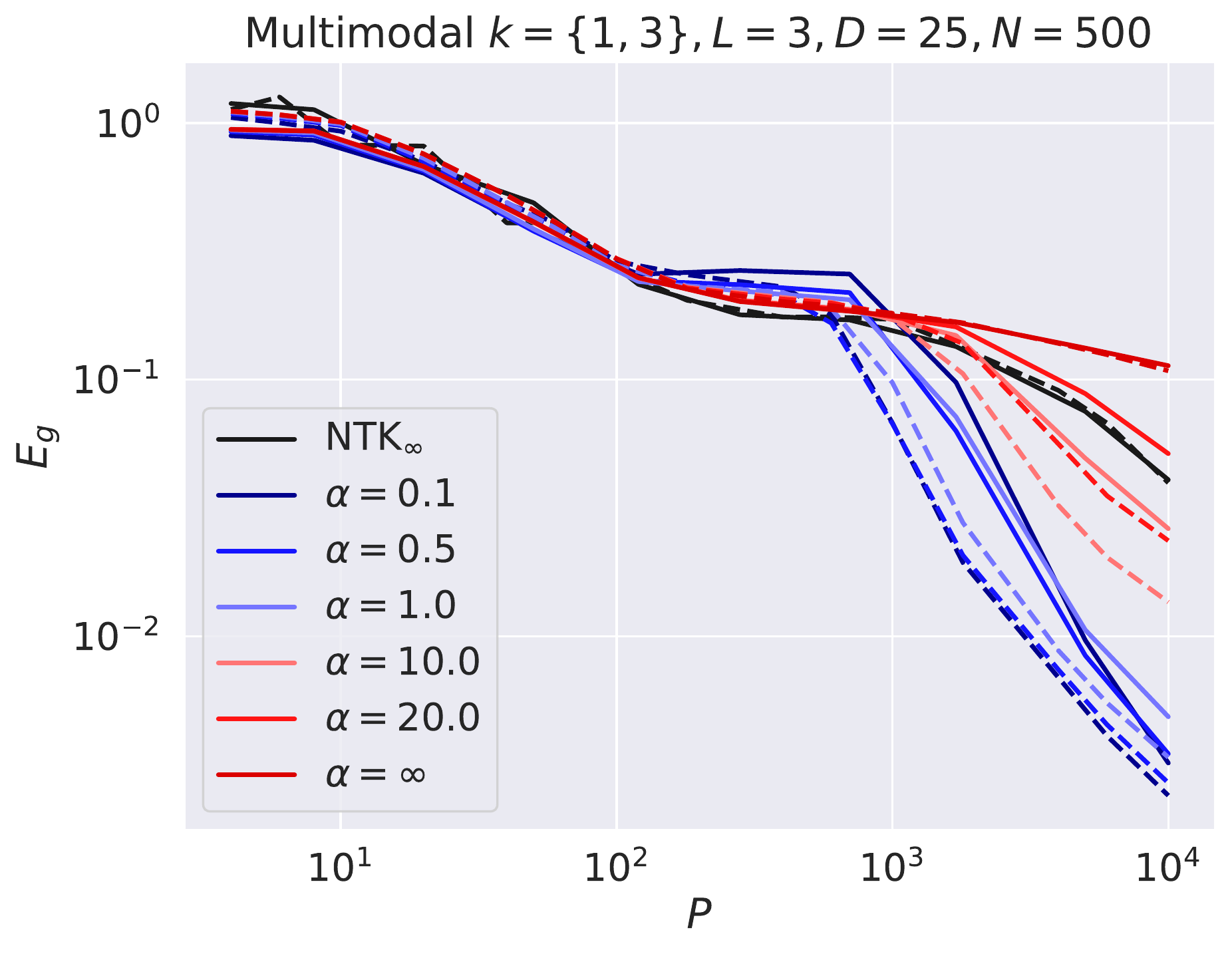}}
    \subfigure[$\mathrm{Var}\, \hat y$ for mixed mode task]{\includegraphics[width=0.4\linewidth]{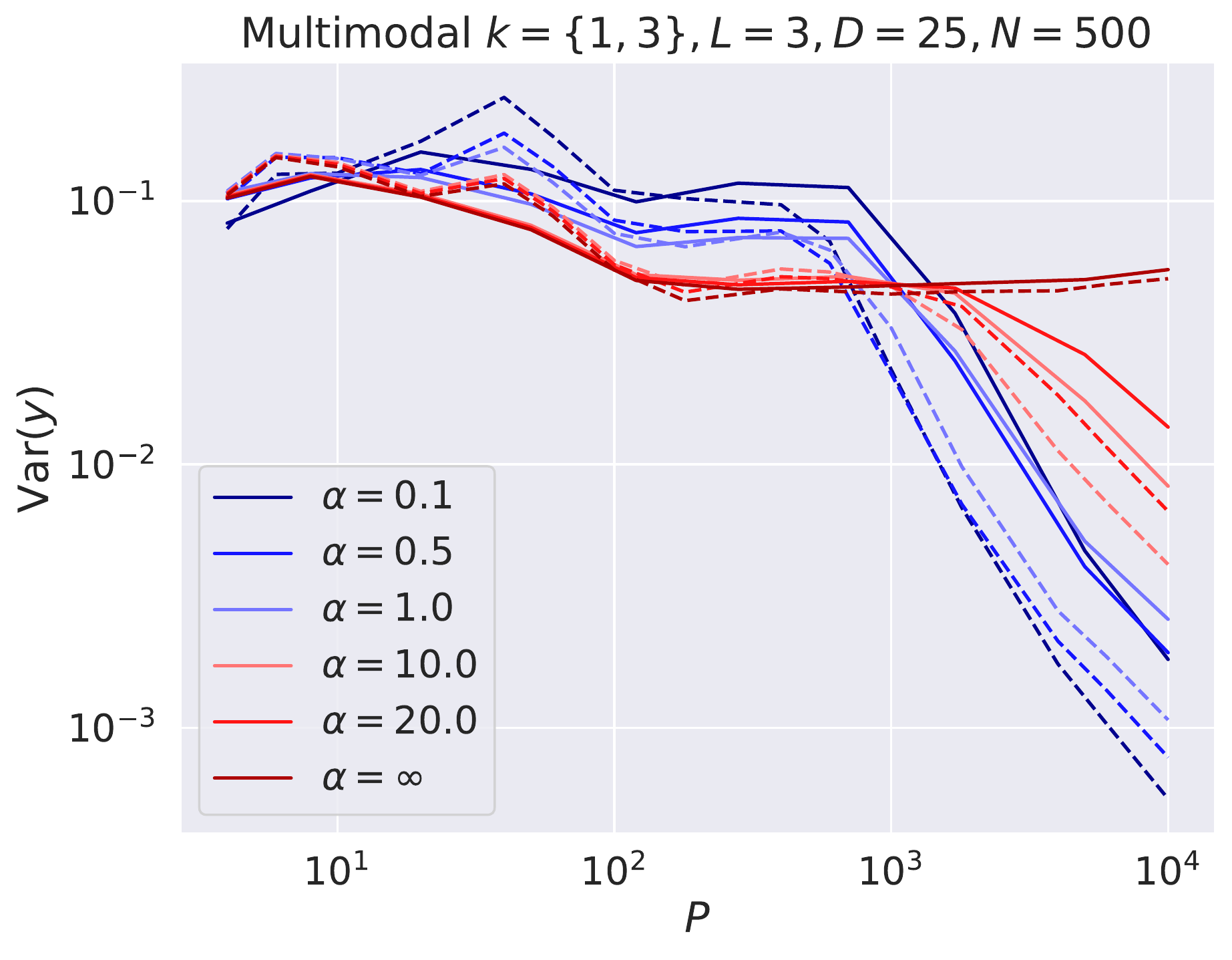}}
    \caption{Width 500 depth 3 MLP learning a $D=25$ mixture of a linear and cubic polynomials. a) Generalization error of \ntk (solid black) and MLP (solid colored lines) on mixed mode task. The dashed lines are convex combinations of  the generalization curves for the pure mode $k=1, k=3$ tasks. For the \ntk, the generalization curves sum to give the mixed mode curve, as observed in \cite{bordelon_icml_learning_curve}. We see that this also holds for the \entk for sufficiently lazy networks, as predicted by the simple renadom feature model considred in section \ref{sec:random_kernel} of  this paper.
    b) The variance curves for the same task. Again, for sufficiently lazy networks the variance is a sum of the variances of the individual pure mode tasks, as predicted by our random feature model.}
    \label{fig:mixed_mode}
\end{figure}

\begin{figure}[h]
    \centering
    \subfigure[$k=3, N=177, \alpha=0.1$]{\includegraphics[width=0.49\linewidth]{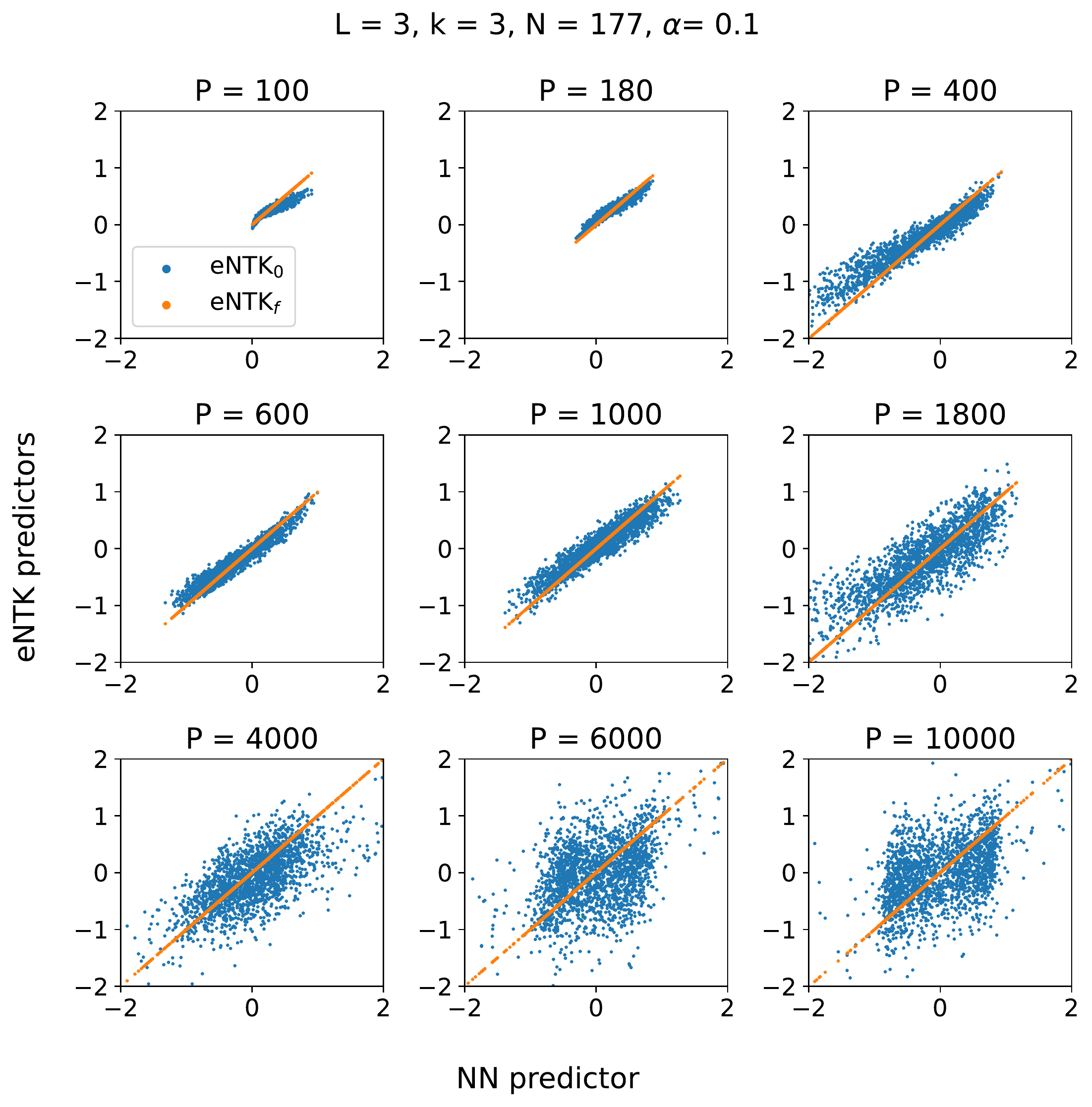}}
    \subfigure[$k=3, N=177, \alpha=20$]{\includegraphics[width=0.49\linewidth]{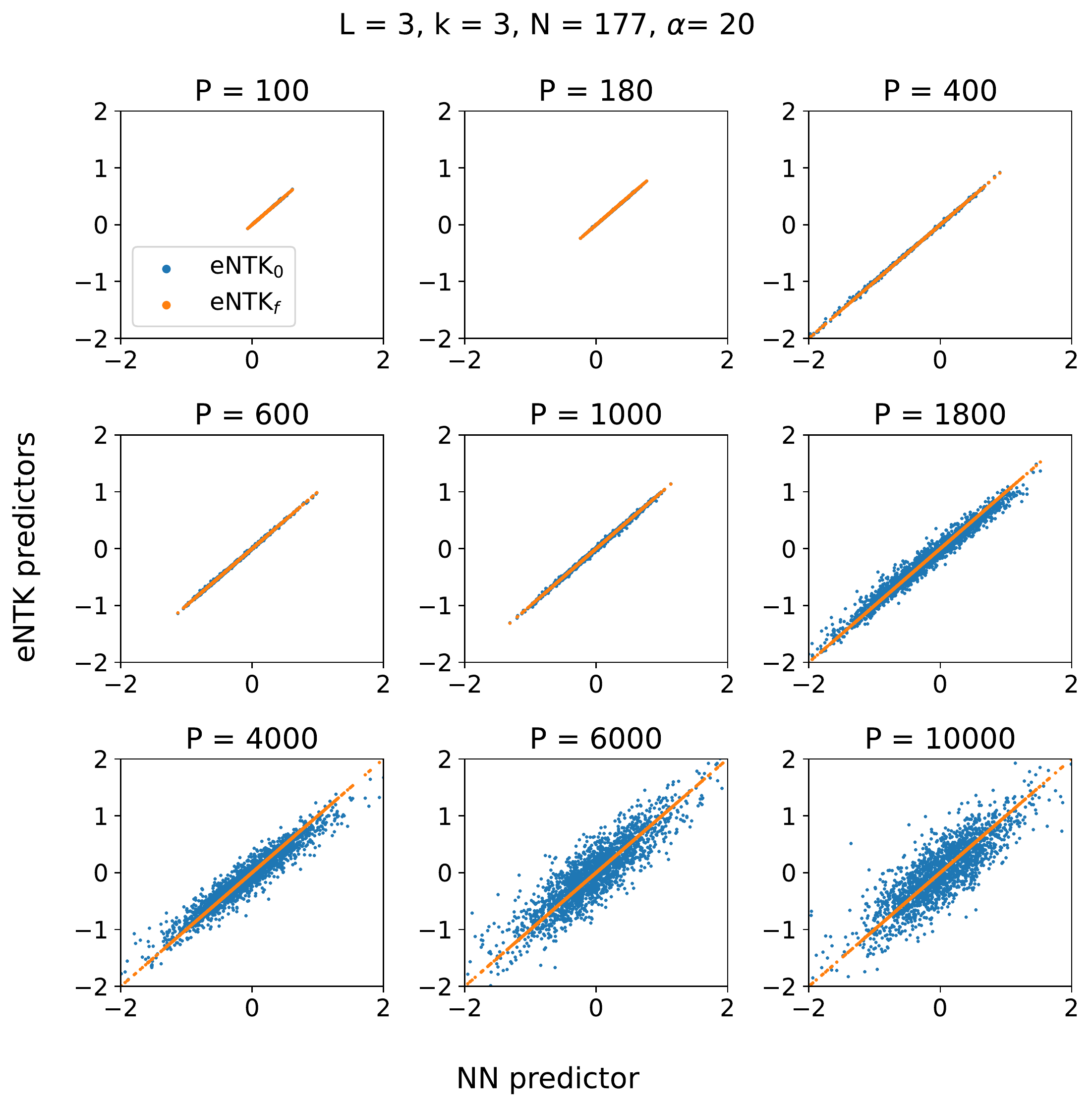}}
    \subfigure[$k=3, N=1000, \alpha=0.1$]{\includegraphics[width=0.49\linewidth]{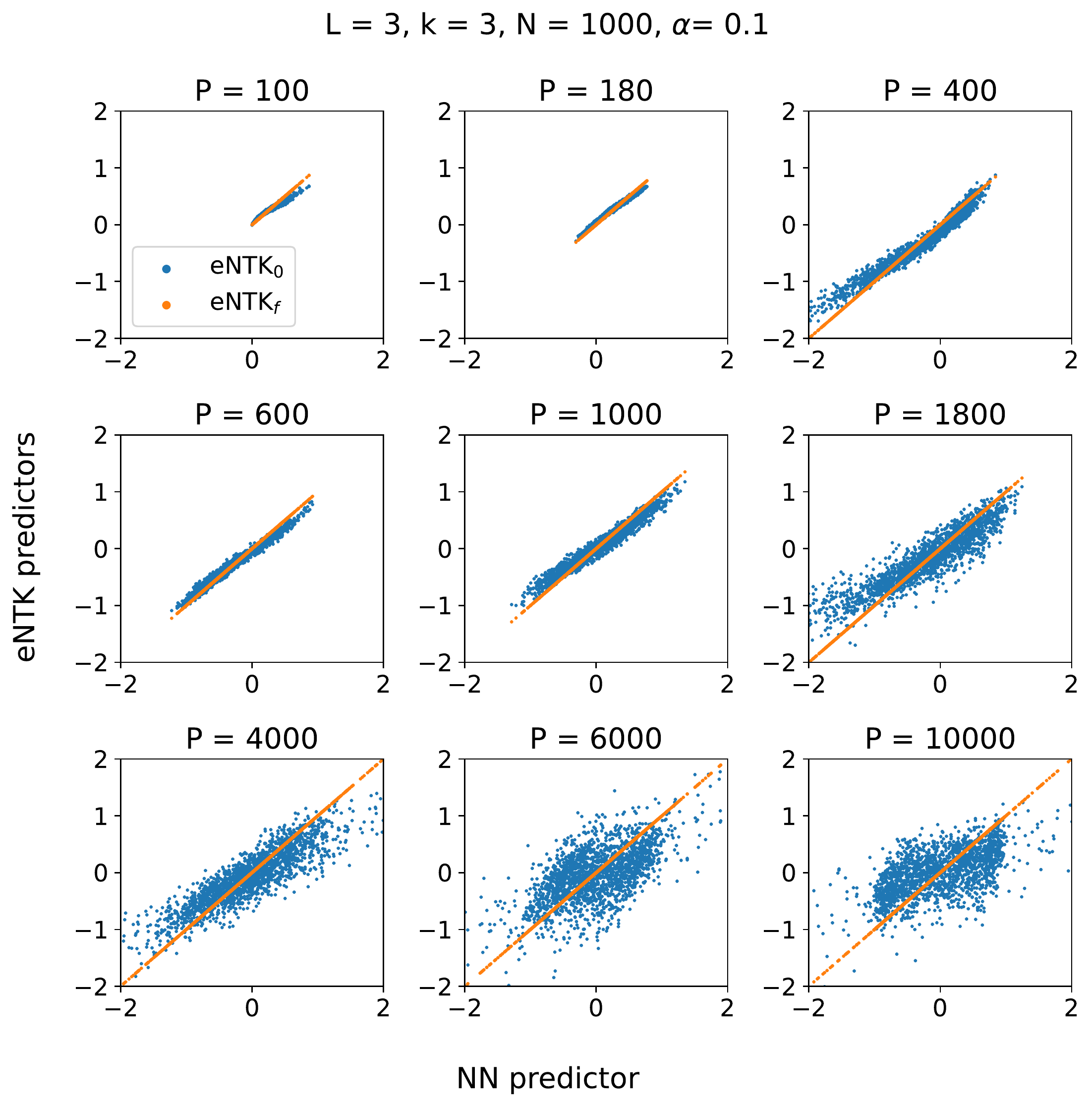}}
    \subfigure[$k=3, N=1000, \alpha=20$]{\includegraphics[width=0.49\linewidth]{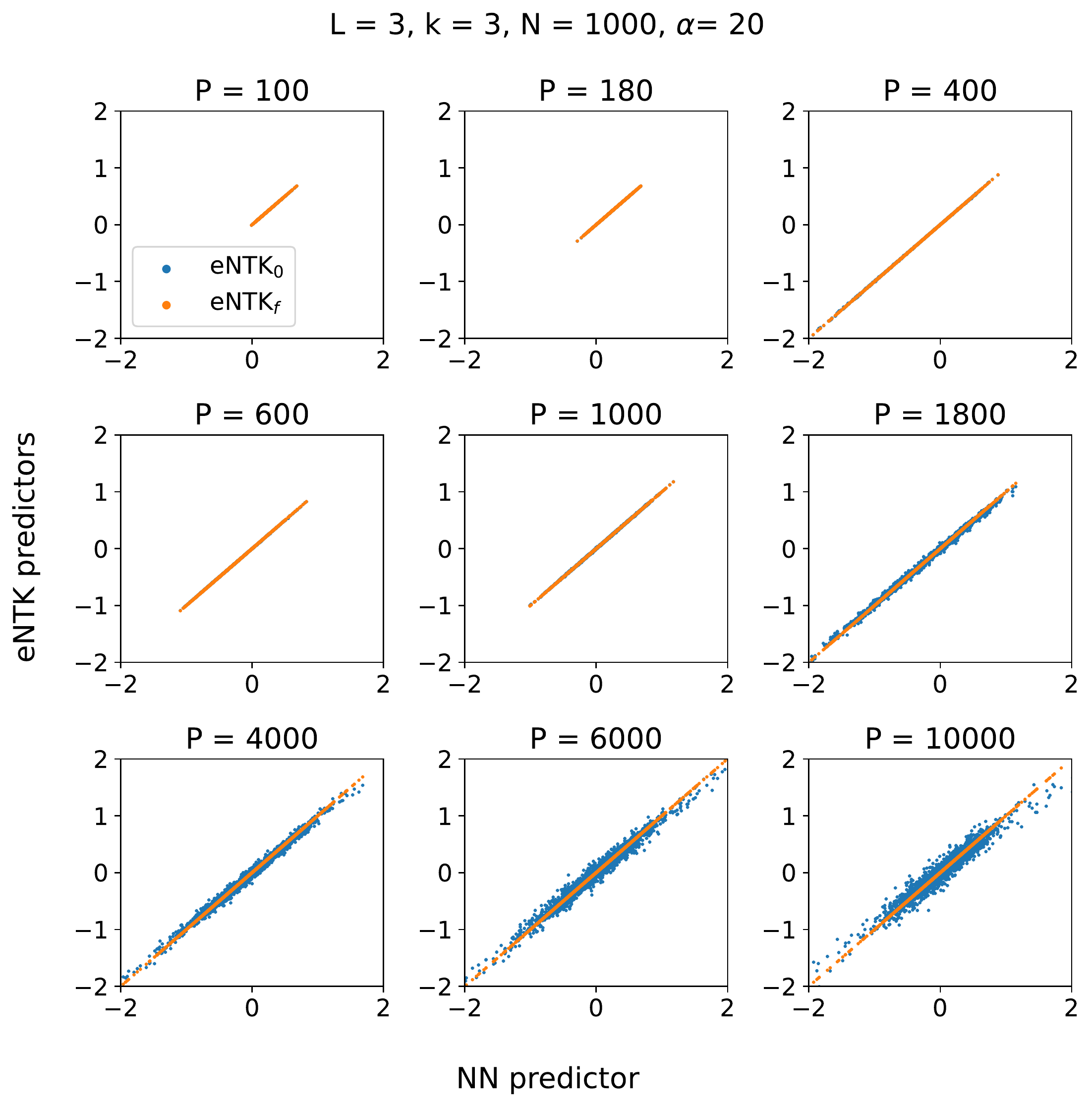}}
\caption{Comparison of the neural network predictor ($x$-axis) to \entk, \entkf (blue, orange respectively, $y$-axis) across several training dataset sizes.  a) $N=177, \alpha=0.1$, b)  $N=177, \alpha=20$, c) $N=1000, \alpha=0.1$ d)  $N=1000, \alpha=20$.  All networks are depth 3. Note how in the rich regime there is a much stronger distinction between the \entk and the neural network. In all regimes, \entkf matches the NN output.}
    \label{fig:pred_comp_k=3}
\end{figure}

\begin{figure}[h]
    \centering
    \subfigure[$k=1, N=177, \alpha=0.1$]{\includegraphics[width=0.49\linewidth]{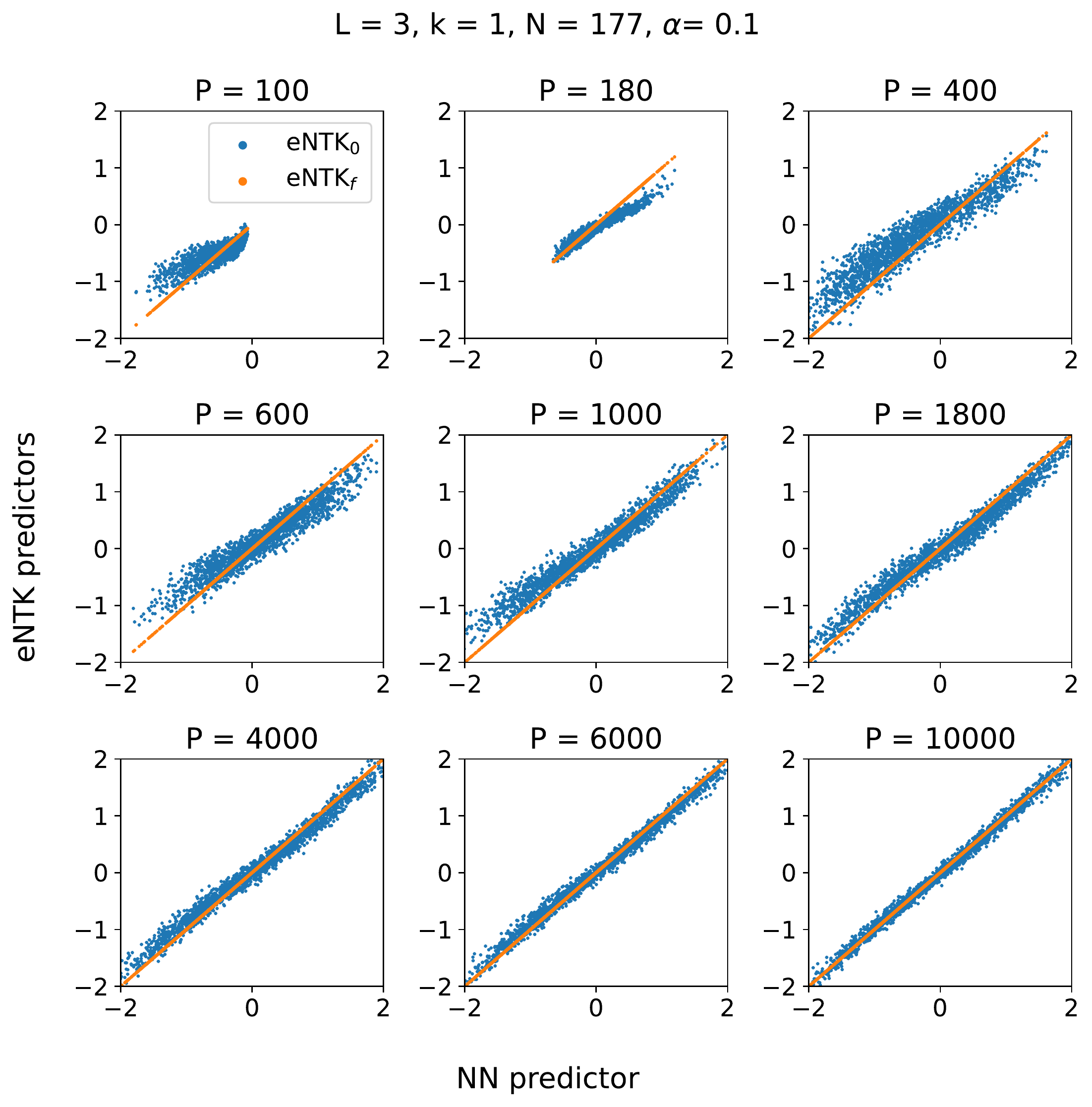}}
    \subfigure[$k=1, N=177, \alpha=20$]{\includegraphics[width=0.49\linewidth]{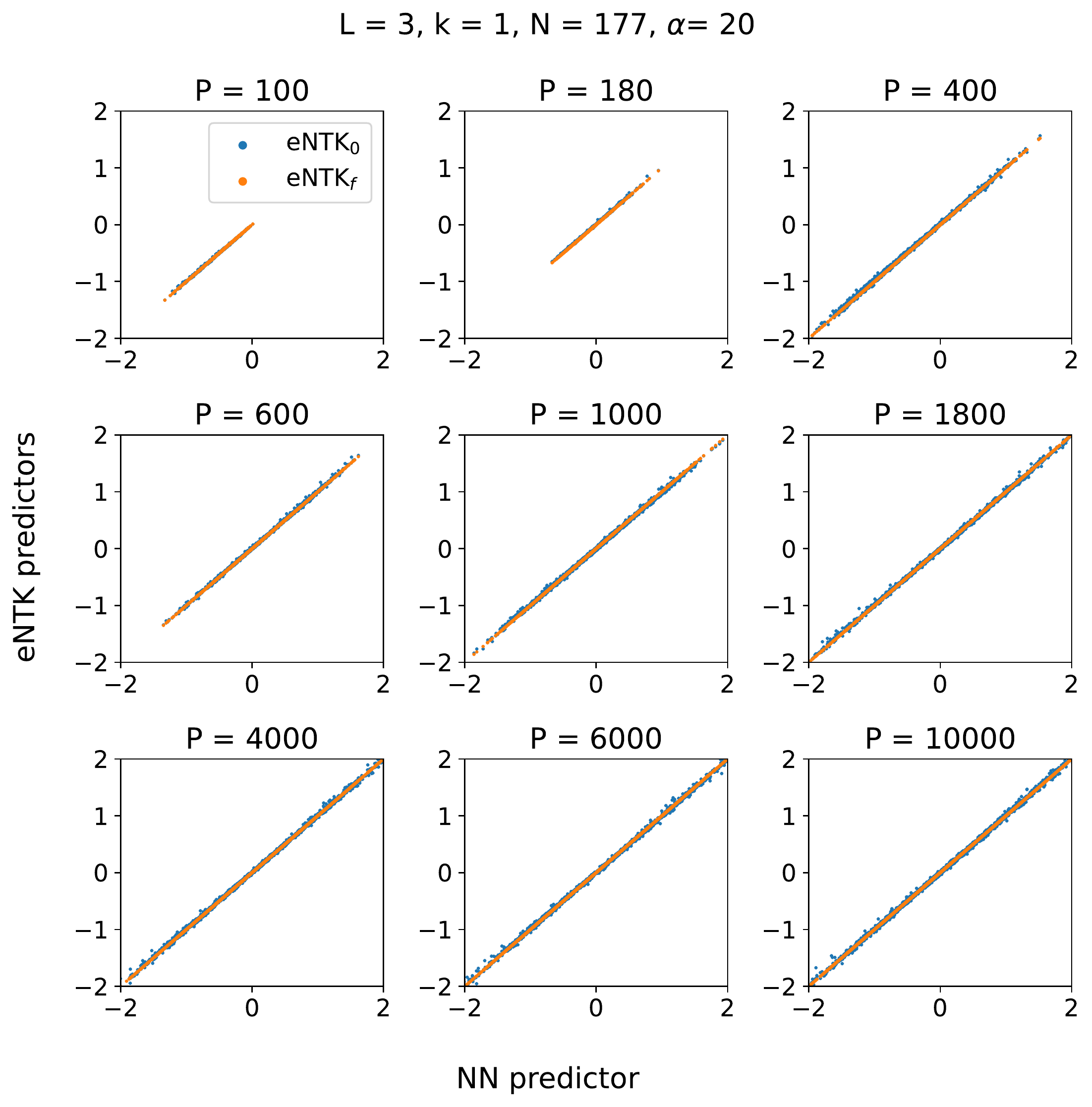}}
    \subfigure[$k=1, N=1000, \alpha=0.1$]{\includegraphics[width=0.49\linewidth]{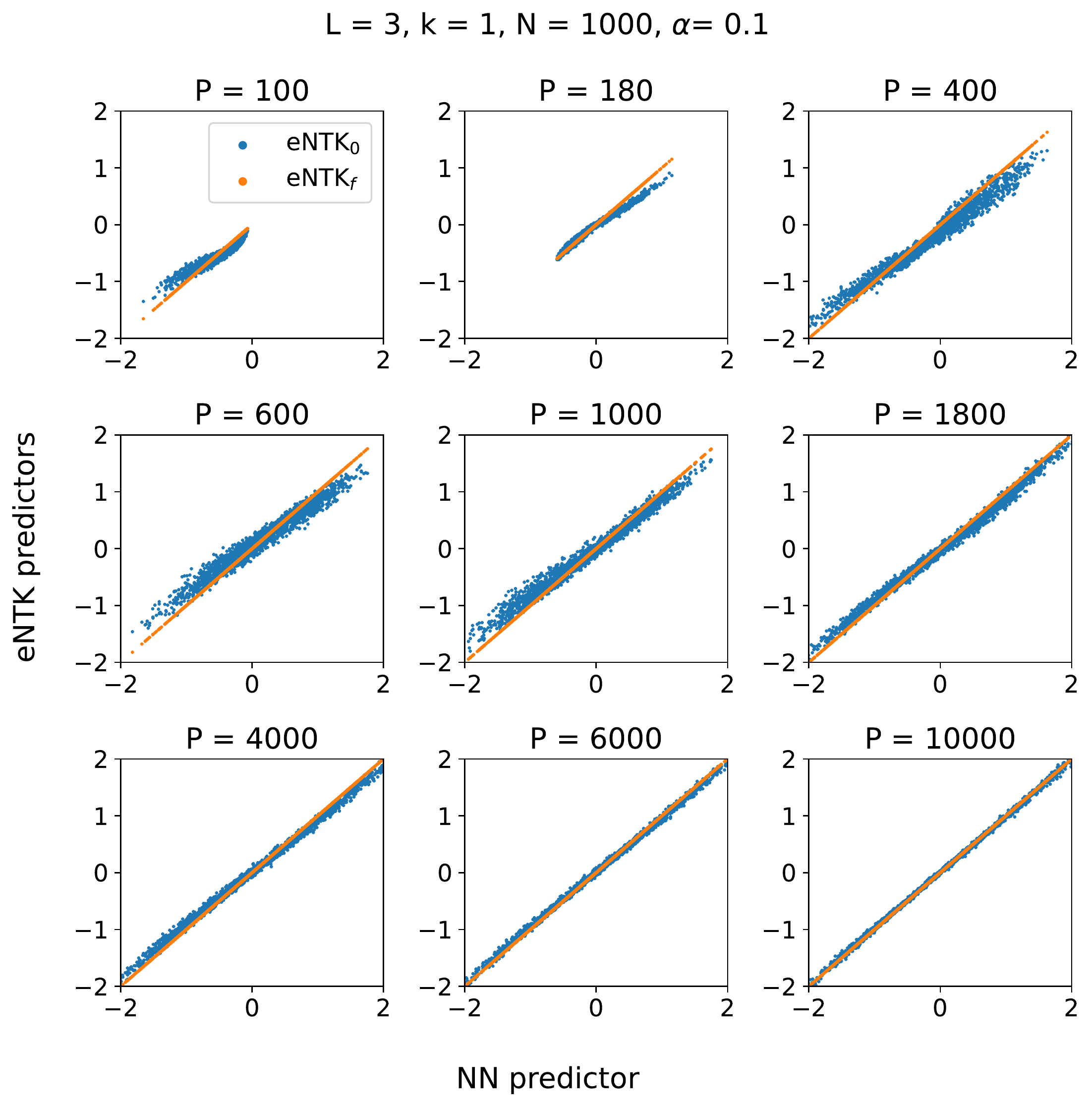}}
    \subfigure[$k=1, N=1000, \alpha=20$]{\includegraphics[width=0.49\linewidth]{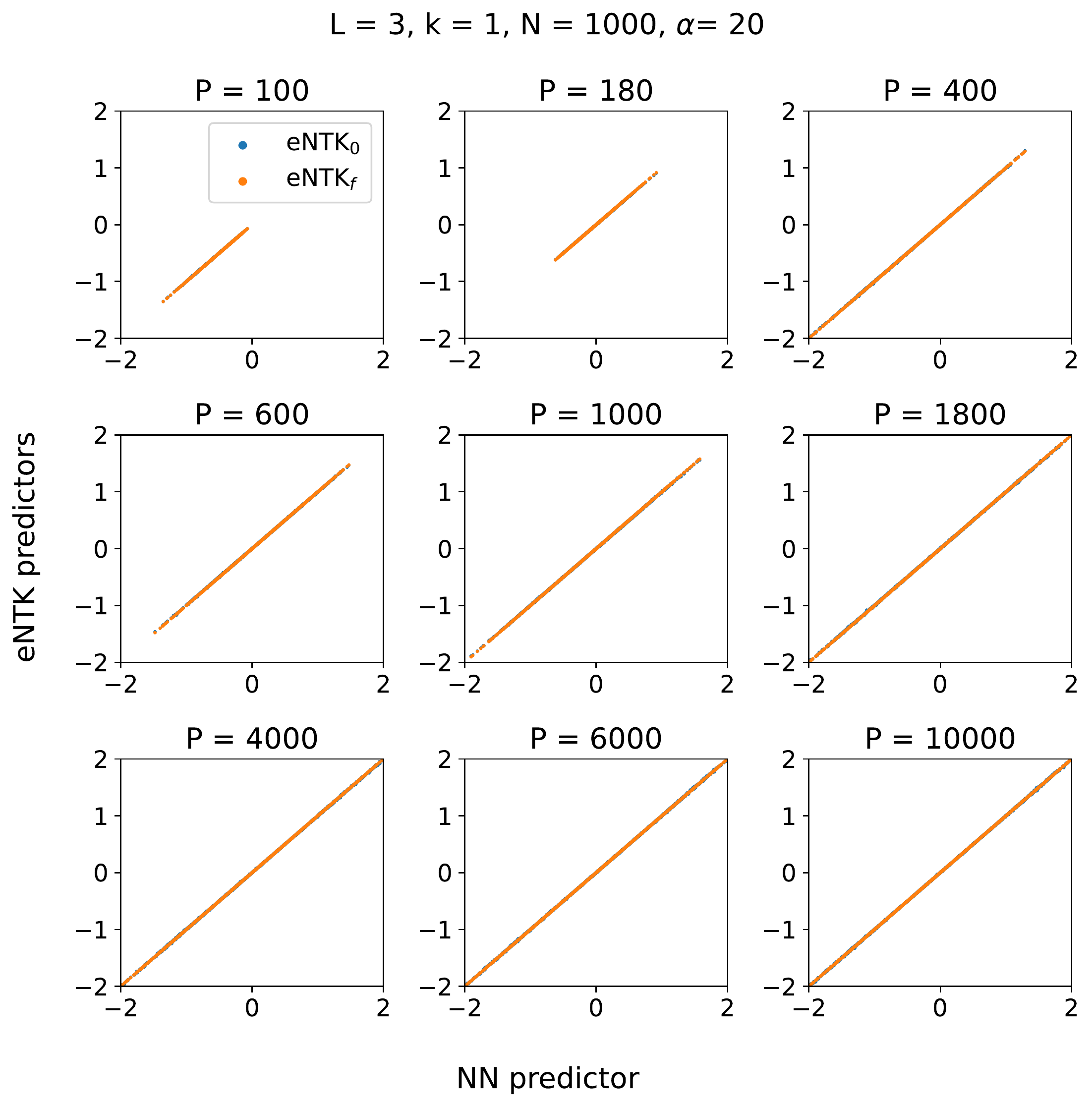}}
    \caption{The same as figure \ref{fig:pred_comp_k=3} but for fitting a linear $k=1$ mode. Because the task is simpler, it doesn't require as large of an $\alpha$ to enter the lazy regime.}
    \label{fig:pred_comp_k=1}
\end{figure}


\end{document}

%% file: math_commands.tex
\usepackage{amsmath,amsfonts,bm}

\def\eqref#1{equation~\ref{#1}}

\def\1{\bm{1}}

\def\vb{{\bm{b}}}

\def\vg{{\bm{g}}}

\def\vw{{\bm{w}}}
\def\vx{{\bm{x}}}
\def\vy{{\bm{y}}}

\def\mA{{\bm{A}}}

\def\mF{{\bm{F}}}
\def\mG{{\bm{G}}}

\def\mI{{\bm{I}}}

\def\mK{{\bm{K}}}

\def\mQ{{\bm{Q}}}

\def\mV{{\bm{V}}}
\def\mW{{\bm{W}}}

\DeclareMathAlphabet{\mathsfit}{\encodingdefault}{\sfdefault}{m}{sl}
\SetMathAlphabet{\mathsfit}{bold}{\encodingdefault}{\sfdefault}{bx}{n}

\DeclareMathOperator{\Tr}{Tr}

%% file: iclr2023_conference.bbl
\begin{thebibliography}{46}
\providecommand{\natexlab}[1]{#1}
\providecommand{\url}[1]{\texttt{#1}}
\expandafter\ifx\csname urlstyle\endcsname\relax
  \providecommand{\doi}[1]{doi: #1}\else
  \providecommand{\doi}{doi: \begingroup \urlstyle{rm}\Url}\fi

\bibitem[Adlam \& Pennington(2020{\natexlab{a}})Adlam and
  Pennington]{adlam2020neural}
Ben Adlam and Jeffrey Pennington.
\newblock The neural tangent kernel in high dimensions: Triple descent and a
  multi-scale theory of generalization.
\newblock In \emph{International Conference on Machine Learning}, pp.\  74--84.
  PMLR, 2020{\natexlab{a}}.

\bibitem[Adlam \& Pennington(2020{\natexlab{b}})Adlam and
  Pennington]{adlam2020understanding}
Ben Adlam and Jeffrey Pennington.
\newblock Understanding double descent requires a fine-grained bias-variance
  decomposition.
\newblock \emph{Advances in neural information processing systems},
  33:\penalty0 11022--11032, 2020{\natexlab{b}}.

\bibitem[Atanasov et~al.(2021)Atanasov, Bordelon, and
  Pehlevan]{atanasov2021neural}
Alexander Atanasov, Blake Bordelon, and Cengiz Pehlevan.
\newblock Neural networks as kernel learners: The silent alignment effect.
\newblock In \emph{International Conference on Learning Representations}, 2021.

\bibitem[Ba et~al.(2022)Ba, Erdogdu, Suzuki, Wang, Wu, and Yang]{ba2022high}
Jimmy Ba, Murat~A Erdogdu, Taiji Suzuki, Zhichao Wang, Denny Wu, and Greg Yang.
\newblock High-dimensional asymptotics of feature learning: How one gradient
  step improves the representation.
\newblock \emph{arXiv preprint arXiv:2205.01445}, 2022.

\bibitem[Bahri et~al.(2021)Bahri, Dyer, Kaplan, Lee, and
  Sharma]{bahri2021scaling}
Yasaman Bahri, Ethan Dyer, Jared Kaplan, Jaehoon Lee, and Utkarsh Sharma.
\newblock Explaining neural scaling laws, 2021.
\newblock URL \url{https://arxiv.org/abs/2102.06701}.

\bibitem[Baratin et~al.(2021)Baratin, George, Laurent, Hjelm, Lajoie, Vincent,
  and Lacoste-Julien]{baratin2021implicit}
Aristide Baratin, Thomas George, C{\'e}sar Laurent, R~Devon Hjelm, Guillaume
  Lajoie, Pascal Vincent, and Simon Lacoste-Julien.
\newblock Implicit regularization via neural feature alignment.
\newblock In \emph{International Conference on Artificial Intelligence and
  Statistics}, pp.\  2269--2277. PMLR, 2021.

\bibitem[Bordelon \& Pehlevan(2022)Bordelon and Pehlevan]{bordelon2022self}
Blake Bordelon and Cengiz Pehlevan.
\newblock Self-consistent dynamical field theory of kernel evolution in wide
  neural networks.
\newblock \emph{arXiv preprint arXiv:2205.09653}, 2022.

\bibitem[Bordelon et~al.(2020)Bordelon, Canatar, and
  Pehlevan]{bordelon_icml_learning_curve}
Blake Bordelon, Abdulkadir Canatar, and Cengiz Pehlevan.
\newblock Spectrum dependent learning curves in kernel regression and wide
  neural networks.
\newblock In Hal~Daumé III and Aarti Singh (eds.), \emph{Proceedings of the
  37th International Conference on Machine Learning}, volume 119 of
  \emph{Proceedings of Machine Learning Research}, pp.\  1024--1034. PMLR,
  13--18 Jul 2020.
\newblock URL \url{https://proceedings.mlr.press/v119/bordelon20a.html}.

\bibitem[Bradbury et~al.(2018)Bradbury, Frostig, Hawkins, Johnson, Leary,
  Maclaurin, Necula, Paszke, VanderPlas, Wanderman-Milne,
  et~al.]{bradbury2018jax}
James Bradbury, Roy Frostig, Peter Hawkins, Matthew~James Johnson, Chris Leary,
  Dougal Maclaurin, George Necula, Adam Paszke, Jake VanderPlas, Skye
  Wanderman-Milne, et~al.
\newblock Jax: composable transformations of python+ numpy programs.
\newblock \emph{Version 0.2}, 5:\penalty0 14--24, 2018.

\bibitem[Canatar et~al.(2021)Canatar, Bordelon, and
  Pehlevan]{Canatar2021SpectralBA}
Abdulkadir Canatar, Blake Bordelon, and Cengiz Pehlevan.
\newblock Spectral bias and task-model alignment explain generalization in
  kernel regression and infinitely wide neural networks.
\newblock \emph{Nature Communications}, 12, 2021.

\bibitem[Chizat et~al.(2019)Chizat, Oyallon, and Bach]{Chizat2019OnLT}
L{\'e}na{\"i}c Chizat, Edouard Oyallon, and Francis~R. Bach.
\newblock On lazy training in differentiable programming.
\newblock In \emph{NeurIPS}, 2019.

\bibitem[Cortes et~al.(2012)Cortes, Mohri, and
  Rostamizadeh]{cortes2012algorithms}
Corinna Cortes, Mehryar Mohri, and Afshin Rostamizadeh.
\newblock Algorithms for learning kernels based on centered alignment.
\newblock \emph{The Journal of Machine Learning Research}, 13\penalty0
  (1):\penalty0 795--828, 2012.

\bibitem[d'Ascoli et~al.(2020)d'Ascoli, Sagun, and Biroli]{d2020triple}
St{\'e}phane d'Ascoli, Levent Sagun, and Giulio Biroli.
\newblock Triple descent and the two kinds of overfitting: Where \& why do they
  appear?
\newblock \emph{Advances in Neural Information Processing Systems},
  33:\penalty0 3058--3069, 2020.

\bibitem[Dhifallah \& Lu(2020)Dhifallah and Lu]{dhifallah2020precise}
Oussama Dhifallah and Yue~M Lu.
\newblock A precise performance analysis of learning with random features.
\newblock \emph{arXiv preprint arXiv:2008.11904}, 2020.

\bibitem[Dyer \& Gur-Ari(2020)Dyer and Gur-Ari]{Dyer2020Asymptotics}
Ethan Dyer and Guy Gur-Ari.
\newblock Asymptotics of wide networks from feynman diagrams.
\newblock In \emph{International Conference on Learning Representations}, 2020.
\newblock URL \url{https://openreview.net/forum?id=S1gFvANKDS}.

\bibitem[d’Ascoli et~al.(2020)d’Ascoli, Refinetti, Biroli, and
  Krzakala]{d2020double}
St{\'e}phane d’Ascoli, Maria Refinetti, Giulio Biroli, and Florent Krzakala.
\newblock Double trouble in double descent: Bias and variance (s) in the lazy
  regime.
\newblock In \emph{International Conference on Machine Learning}, pp.\
  2280--2290. PMLR, 2020.

\bibitem[Fort et~al.(2020)Fort, Dziugaite, Paul, Kharaghani, Roy, and
  Ganguli]{fort2020deep}
Stanislav Fort, Gintare~Karolina Dziugaite, Mansheej Paul, Sepideh Kharaghani,
  Daniel~M Roy, and Surya Ganguli.
\newblock Deep learning versus kernel learning: an empirical study of loss
  landscape geometry and the time evolution of the neural tangent kernel.
\newblock In H.~Larochelle, M.~Ranzato, R.~Hadsell, M.~F. Balcan, and H.~Lin
  (eds.), \emph{Advances in Neural Information Processing Systems}, volume~33,
  pp.\  5850--5861. Curran Associates, Inc., 2020.
\newblock URL
  \url{https://proceedings.neurips.cc/paper/2020/file/405075699f065e43581f27d67bb68478-Paper.pdf}.

\bibitem[Geiger et~al.(2020{\natexlab{a}})Geiger, Jacot, Spigler, Gabriel,
  Sagun, d’Ascoli, Biroli, Hongler, and Wyart]{geiger2020scaling}
Mario Geiger, Arthur Jacot, Stefano Spigler, Franck Gabriel, Levent Sagun,
  St{\'e}phane d’Ascoli, Giulio Biroli, Cl{\'e}ment Hongler, and Matthieu
  Wyart.
\newblock Scaling description of generalization with number of parameters in
  deep learning.
\newblock \emph{Journal of Statistical Mechanics: Theory and Experiment},
  2020\penalty0 (2):\penalty0 023401, 2020{\natexlab{a}}.

\bibitem[Geiger et~al.(2020{\natexlab{b}})Geiger, Spigler, Jacot, and
  Wyart]{geiger2020disentangling}
Mario Geiger, Stefano Spigler, Arthur Jacot, and Matthieu Wyart.
\newblock Disentangling feature and lazy training in deep neural networks.
\newblock \emph{Journal of Statistical Mechanics: Theory and Experiment},
  2020\penalty0 (11):\penalty0 113301, 2020{\natexlab{b}}.

\bibitem[Gerace et~al.(2020)Gerace, Loureiro, Krzakala, M{\'e}zard, and
  Zdeborov{\'a}]{gerace2020generalisation}
Federica Gerace, Bruno Loureiro, Florent Krzakala, Marc M{\'e}zard, and Lenka
  Zdeborov{\'a}.
\newblock Generalisation error in learning with random features and the hidden
  manifold model.
\newblock In \emph{International Conference on Machine Learning}, pp.\
  3452--3462. PMLR, 2020.

\bibitem[Ghorbani et~al.(2020)Ghorbani, Mei, Misiakiewicz, and
  Montanari]{ghorbani_outperform}
Behrooz Ghorbani, Song Mei, Theodor Misiakiewicz, and Andrea Montanari.
\newblock When do neural networks outperform kernel methods?
\newblock In \emph{NeurIPS}, 2020.
\newblock URL
  \url{https://proceedings.neurips.cc/paper/2020/hash/a9df2255ad642b923d95503b9a7958d8-Abstract.html}.

\bibitem[Hanin \& Nica(2019)Hanin and Nica]{hanin2019finite}
Boris Hanin and Mihai Nica.
\newblock Finite depth and width corrections to the neural tangent kernel.
\newblock In \emph{International Conference on Learning Representations}, 2019.

\bibitem[Hoffmann et~al.(2022)Hoffmann, Borgeaud, Mensch, Buchatskaya, Cai,
  Rutherford, Casas, Hendricks, Welbl, Clark, et~al.]{hoffmann2022training}
Jordan Hoffmann, Sebastian Borgeaud, Arthur Mensch, Elena Buchatskaya, Trevor
  Cai, Eliza Rutherford, Diego de~Las Casas, Lisa~Anne Hendricks, Johannes
  Welbl, Aidan Clark, et~al.
\newblock Training compute-optimal large language models.
\newblock \emph{arXiv preprint arXiv:2203.15556}, 2022.

\bibitem[Hu \& Lu(2020)Hu and Lu]{hu2020universality}
Hong Hu and Yue~M Lu.
\newblock Universality laws for high-dimensional learning with random features.
\newblock \emph{arXiv preprint arXiv:2009.07669}, 2020.

\bibitem[Jacot et~al.(2018)Jacot, Gabriel, and Hongler]{Jacot2018NeuralTK}
Arthur Jacot, Franck Gabriel, and Cl{\'e}ment Hongler.
\newblock Neural tangent kernel: convergence and generalization in neural
  networks (invited paper).
\newblock \emph{Proceedings of the 53rd Annual ACM SIGACT Symposium on Theory
  of Computing}, 2018.

\bibitem[Kingma \& Ba(2014)Kingma and Ba]{kingma2014adam}
Diederik~P Kingma and Jimmy Ba.
\newblock Adam: A method for stochastic optimization.
\newblock \emph{arXiv preprint arXiv:1412.6980}, 2014.

\bibitem[Kornblith et~al.(2019)Kornblith, Norouzi, Lee, and
  Hinton]{kornblith2019similarity}
Simon Kornblith, Mohammad Norouzi, Honglak Lee, and Geoffrey Hinton.
\newblock Similarity of neural network representations revisited.
\newblock In \emph{International Conference on Machine Learning}, pp.\
  3519--3529. PMLR, 2019.

\bibitem[Lee et~al.(2019)Lee, Xiao, Schoenholz, Bahri, Novak, Sohl-Dickstein,
  and Sohl-Dickstein]{Lee2019WideNN}
Jaehoon Lee, Lechao Xiao, Samuel~S. Schoenholz, Yasaman Bahri, Roman Novak,
  Jascha Sohl-Dickstein, and Jascha Sohl-Dickstein.
\newblock Wide neural networks of any depth evolve as linear models under
  gradient descent.
\newblock \emph{ArXiv}, abs/1902.06720, 2019.

\bibitem[Long(2021)]{long2021after}
Philip~M. Long.
\newblock Properties of the after kernel.
\newblock \emph{CoRR}, abs/2105.10585, 2021.
\newblock URL \url{https://arxiv.org/abs/2105.10585}.

\bibitem[Loureiro et~al.(2021)Loureiro, Gerbelot, Cui, Goldt, Krzakala,
  Mézard, and Zdeborová]{loureiro_lenka_feature_maps}
Bruno Loureiro, Cédric Gerbelot, Hugo Cui, Sebastian Goldt, Florent Krzakala,
  Marc Mézard, and Lenka Zdeborová.
\newblock Capturing the learning curves of generic features maps for realistic
  data sets with a teacher-student model.
\newblock \emph{CoRR}, abs/2102.08127, 2021.
\newblock URL \url{https://arxiv.org/abs/2102.08127}.

\bibitem[Maloney et~al.(2022)Maloney, Roberts, and
  Sully]{roberts_random_feature}
Alexander Maloney, Daniel~A. Roberts, and James Sully.
\newblock A solvable model of neural scaling laws, 2022.
\newblock URL \url{https://arxiv.org/abs/2210.16859}.

\bibitem[Mei \& Montanari(2022)Mei and Montanari]{mei2022generalization}
Song Mei and Andrea Montanari.
\newblock The generalization error of random features regression: Precise
  asymptotics and the double descent curve.
\newblock \emph{Communications on Pure and Applied Mathematics}, 75\penalty0
  (4):\penalty0 667--766, 2022.

\bibitem[Mei et~al.(2018)Mei, Montanari, and Nguyen]{mei2018mean}
Song Mei, Andrea Montanari, and Phan-Minh Nguyen.
\newblock A mean field view of the landscape of two-layer neural networks.
\newblock \emph{Proceedings of the National Academy of Sciences}, 115\penalty0
  (33):\penalty0 E7665--E7671, 2018.

\bibitem[Novak et~al.(2020)Novak, Xiao, Hron, Lee, Alemi, Sohl-Dickstein, and
  Schoenholz]{neuraltangents2020}
Roman Novak, Lechao Xiao, Jiri Hron, Jaehoon Lee, Alexander~A. Alemi, Jascha
  Sohl-Dickstein, and Samuel~S. Schoenholz.
\newblock Neural tangents: Fast and easy infinite neural networks in python.
\newblock In \emph{International Conference on Learning Representations}, 2020.
\newblock URL \url{https://github.com/google/neural-tangents}.

\bibitem[Ortiz-Jim{\'e}nez et~al.(2021)Ortiz-Jim{\'e}nez, Moosavi-Dezfooli, and
  Frossard]{ortiz2021can}
Guillermo Ortiz-Jim{\'e}nez, Seyed-Mohsen Moosavi-Dezfooli, and Pascal
  Frossard.
\newblock What can linearized neural networks actually say about
  generalization?
\newblock \emph{Advances in Neural Information Processing Systems},
  34:\penalty0 8998--9010, 2021.

\bibitem[Paccolat et~al.(2021{\natexlab{a}})Paccolat, Petrini, Geiger, Tyloo,
  and Wyart]{Paccolat_2021}
Jonas Paccolat, Leonardo Petrini, Mario Geiger, Kevin Tyloo, and Matthieu
  Wyart.
\newblock Geometric compression of invariant manifolds in neural networks.
\newblock \emph{Journal of Statistical Mechanics: Theory and Experiment},
  2021\penalty0 (4):\penalty0 044001, apr 2021{\natexlab{a}}.
\newblock \doi{10.1088/1742-5468/abf1f3}.
\newblock URL \url{https://doi.org/10.1088/1742-5468/abf1f3}.

\bibitem[Paccolat et~al.(2021{\natexlab{b}})Paccolat, Petrini, Geiger, Tyloo,
  and Wyart]{paccolat2021geometric}
Jonas Paccolat, Leonardo Petrini, Mario Geiger, Kevin Tyloo, and Matthieu
  Wyart.
\newblock Geometric compression of invariant manifolds in neural networks.
\newblock \emph{Journal of Statistical Mechanics: Theory and Experiment},
  2021\penalty0 (4):\penalty0 044001, 2021{\natexlab{b}}.

\bibitem[Roberts et~al.(2021)Roberts, Yaida, and Hanin]{roberts2021principles}
Daniel~A. Roberts, Sho Yaida, and Boris Hanin.
\newblock The principles of deep learning theory, 2021.

\bibitem[Simon et~al.(2021)Simon, Dickens, and DeWeese]{simon2021neural}
James~B. Simon, Madeline Dickens, and Michael~R. DeWeese.
\newblock Neural tangent kernel eigenvalues accurately predict generalization,
  2021.

\bibitem[Spigler et~al.(2020)Spigler, Geiger, and Wyart]{spigler2020asymptotic}
Stefano Spigler, Mario Geiger, and Matthieu Wyart.
\newblock Asymptotic learning curves of kernel methods: empirical data versus
  teacher--student paradigm.
\newblock \emph{Journal of Statistical Mechanics: Theory and Experiment},
  2020\penalty0 (12):\penalty0 124001, 2020.

\bibitem[Tan \& Le(2019)Tan and Le]{tan2019efficientnet}
Mingxing Tan and Quoc Le.
\newblock Efficientnet: Rethinking model scaling for convolutional neural
  networks.
\newblock In \emph{International conference on machine learning}, pp.\
  6105--6114. PMLR, 2019.

\bibitem[Vyas et~al.(2022)Vyas, Bansal, and Nakkiran]{vyas2022limitations}
Nikhil Vyas, Yamini Bansal, and Preetum Nakkiran.
\newblock Limitations of the ntk for understanding generalization in deep
  learning.
\newblock \emph{arXiv preprint arXiv:2206.10012}, 2022.

\bibitem[Wei et~al.(2022)Wei, Hu, and Steinhardt]{wei2022more}
Alexander Wei, Wei Hu, and Jacob Steinhardt.
\newblock More than a toy: Random matrix models predict how real-world neural
  representations generalize.
\newblock \emph{arXiv preprint arXiv:2203.06176}, 2022.

\bibitem[Yang \& Hu(2020)Yang and Hu]{Yang2020FeatureLI}
Greg Yang and Edward~J. Hu.
\newblock Feature learning in infinite-width neural networks.
\newblock \emph{ArXiv}, abs/2011.14522, 2020.

\bibitem[Zagoruyko \& Komodakis(2017)Zagoruyko and
  Komodakis]{zagoruyko2017wide}
Sergey Zagoruyko and Nikos Komodakis.
\newblock Wide residual networks, 2017.

\bibitem[Zavatone-Veth et~al.(2022)Zavatone-Veth, Tong, and
  Pehlevan]{zavatone2022contrasting}
Jacob~A Zavatone-Veth, William~L Tong, and Cengiz Pehlevan.
\newblock Contrasting random and learned features in deep bayesian linear
  regression.
\newblock \emph{arXiv preprint arXiv:2203.00573}, 2022.

\end{thebibliography}
